\documentclass[11pt, oneside]{mnthesis}
\usepackage{bigstrut}
\usepackage{epsfig,epic,eepic,units}
\usepackage{hyperref}
\usepackage{url}
\usepackage{longtable}
\usepackage{mathrsfs}
\usepackage{multirow}
\usepackage{bigstrut}
\usepackage{amssymb}
\usepackage{graphicx}
\usepackage{amsmath}

\usepackage{array}
\usepackage{enumerate}
\usepackage{multirow} % tables
\usepackage{subfigure}
\usepackage{float}
\usepackage{amsfonts}
\newcolumntype{L}[1]{>{\raggedright\let\newline\\\arraybackslash\hspace{0pt}}m{#1}}
\newcolumntype{C}[1]{>{\centering\let\newline\\\arraybackslash\hspace{0pt}}m{#1}}
\newcolumntype{R}[1]{>{\raggedleft\let\newline\\\arraybackslash\hspace{0pt}}m{#1}}

\newcommand{\ie}{\textit{i.e.}}
\newcommand{\eg}{\textit{e.g.}}
\newcommand{\etal}{\textit{et al.}} 

\usepackage{appendix}
\usepackage{algorithm}  
\usepackage{algpseudocode}

\newcommand*\rot{\rotatebox{90}}
\DeclareMathOperator*{\argmax}{arg\,max}
\DeclareMathOperator*{\argmin}{arg\,min}

\usepackage{lscape} 

\usepackage{makecell}

\begin{document}

%%%%%%%%%%%%%%%%%%%%%%%%%%%%%%%%%%%%%%%%%%%%%%%%%%%%%%%%%%%%%%%%%%%%%%%%%%%%%%%%
% title.tex - Set up the beginning of thesis.
%%%%%%%%%%%%%%%%%%%%%%%%%%%%%%%%%%%%%%%%%%%%%%%%%%%%%%%%%%%%%%%%%%%%%%%%%%%%%%%%
% For a  PhD give the command \phd. Default is masters
%\degree (normally Doctor of Philosophy or Master of Science)
%\initials (normally Ph.D. or M.S.)
%\ms % use if for a Master of Science thesis
\phd % use if for a Ph.D. dissertation

\title{\bf Machine Vision for Improved Human-Robot Cooperation in Adverse Underwater Conditions}
\author{Md Jahidul Islam}
\campus{University of Minnesota} 
\program{CS} 
\degree{DOCTOR OF PHILOSOPHY}
\director{Junaed Sattar} 

% Optionally specify the month and year.
\submissionmonth{May} % defaults to current month.
\submissionyear{2021} % defaults to current year.

%Comment out below on final copy
\abstract{
Visually-guided underwater robots are deployed alongside human divers for cooperative exploration, inspection, and monitoring tasks in numerous shallow-water and coastal-water applications. 
The most essential capability of such companion robots is to visually interpret their surroundings and assist the divers during various stages of an underwater mission. Despite recent technological advancements, the existing systems and solutions for real-time visual perception are greatly affected by marine artifacts such as poor visibility, lighting variation, and the scarcity of salient features. 
The difficulties are exacerbated by a host of non-linear image distortions caused by the vulnerabilities of underwater light propagation (\eg, wavelength-dependent attenuation, absorption, and scattering).
In this dissertation, we present a set of novel and improved visual perception solutions to address these challenges for effective underwater human-robot cooperation. 
The research outcomes entail novel design and efficient implementation of the underlying vision and learning-based algorithms with extensive field experimental validations and real-time feasibility analyses for single-board deployments.

The dissertation is organized into three parts. The first part focuses on developing practical solutions for autonomous underwater vehicles (AUVs) to accompany human divers during an underwater mission. These include robust vision-based modules that enable AUVs to understand human swimming motion, hand gesture, and body pose for following and interacting with them while maintaining smooth spatiotemporal coordination. A series of closed-water and open-water field experiments demonstrate the utility and effectiveness of our proposed perception algorithms for underwater human-robot cooperation. We also identify and quantify their performance variability over a diverse set of operating constraints in adverse visual conditions. The second part of this dissertation is devoted to designing efficient techniques to overcome the effects of poor visibility and optical distortions in underwater imagery by restoring their perceptual and statistical qualities. We further demonstrate the practical feasibility of these techniques as pre-processors in the autonomy pipeline of visually-guided AUVs. Finally, the third part of this dissertation develops methodologies for high-level decision-making such as modeling spatial attention for fast visual search, learning to identify when image enhancement and super-resolution modules are necessary for a detailed perception, etc. We demonstrate that these methodologies facilitate up to $45\%$ faster processing of the on-board visual perception modules and enable AUVs to make intelligent navigational and operational decisions, particularly in autonomous exploratory tasks.

In summary, this dissertation delineates our attempts to address the environmental and operational challenges of real-time machine vision for underwater human-robot cooperation. Aiming at a variety of important applications, we develop robust and efficient modules for AUVs to \emph{\textbf{follow and interact}} with companion divers by accurately perceiving their surroundings while relying on noisy visual sensing alone.
Moreover, our proposed perception solutions enable visually-guided robots to \textit{\textbf{see better}} in noisy conditions, and \textit{\textbf{do better}} with limited computational resources and real-time constraints. In addition to advancing the state-of-the-art, the proposed methodologies and systems take us one step closer toward bridging the gap between theory and practice for improved human-robot cooperation in the wild.   

}
\words{486}    % number of words in the abstract
\copyrightpage % Do you want copyright protection?
\acknowledgements{First and foremost, I would like to express my sincere gratitude and appreciation toward my academic adviser, Prof. Junaed Sattar. I am thankful for his support, encouragement, and guidance during my journey as a Ph.D. student. I am forever indebted for his confidence vested in me and for giving me the freedom to explore new (and often crazy) ideas throughout my research career. As one of the first members in his group, I had the chance to experience the early days and every stage of the growth of a research lab into its full form. This remarkable experience and everything that comes with it will certainly help me in shaping my professional career in the near future.

I am also grateful to my dissertation committee members: Prof. Eric Van Wyk, Prof. Volkan Isler, and Prof. Timothy Kowalewski. I am thankful for their constant support and advice throughout the WPE/OPE, dissertation proposal examination, and my preparation for the final defense. Their insightful comments and suggestions have immensely enriched this dissertation in countless ways. I absolutely needed their constructive inputs and feedback, particularly during the early stages of my research.

I thank my co-authors and collaborators: Jiawei Mo, Cameron Fabbri, Michael Fulton, Jungseok Hong, Chelsey Edge, Karin de Langis, and Sadman Sakib Enan; they helped me shape and solidify many important ideas in numerous projects. The graduate and undergraduate students whom I worked with and mentored: Marc Ho, Youya Xia, Peigen Luo, Ruobing Wang, Christopher Morse, Yuyang Xiao, Muntaqim Mehtaz, and Tanmay Agarwal - have brought tremendous positive energy in me and kept me motivated every day. I also thank Kimberly Barthelemy, Andrea Walker, Elliott Imhoff, Julian Lagman, and Marcus Oh for their assistance in proofreading documents, collecting data, annotating images, and preparing media files. It was a pleasure to work with them; I would have never been able to conduct so many field experiments and complete all these projects without their active participation. 

I am grateful to Prof. Gregory Dudek and Bellairs Research Institute of Barbados for providing us with the facilities for extensive ocean trials. Albeit through brief interactions, I learned a lot from him about practicing a healthy academic culture and having dynamic organizational skills. I am also thankful to have interacted with Prof. Ioannis Rekleitis and Prof. Alberto Quattrini Li, and their amazing group of students at the Bellairs. I acknowledge our colleagues and all participants of the 2019 and 2018 marine robotics ocean trials at the Bellairs for helping us design and conduct various experiments. It was a great experience to work alongside them and learn numerous practical hacks and challenging aspects of \textit{field robotics} - which one can barely learn from books. In particular, I learned immensely from Prof. Florian Shkurti;  I am thankful for his inspirations and directions that helped me pass through some difficult and confusing times. I look forward to the opportunity to work and collaborate with these amazing people and stellar researchers again in the future.

I have been very fortunate to come across some outstanding faculty members and prominent academics at the University of Minnesota. I am grateful to Prof. Stergios Roumeliotis for giving me the first hands-on exposure to robotics; I am forever amazed and inspired by his work ethic and discipline. I am thankful to Prof. Anand Tripathi, Prof. Yousef Saad, Prof. Jarvis Haupt, and Prof. Daniel Boley for their guidance, direction, and encouragement in my early days - I truly needed those. I am also grateful to Prof. Arindam Banerjee, Prof. Volkan Isler, and Prof. Hyun Soo Park for teaching me the most important and useful topics of advanced machine learning, robot perception, and 3D computer vision; they have also welcomed me in their reading groups and wrote recommendation letters for me on several occasions. I was also incredibly lucky to get two outstanding mentors in the industry: Dr. Guruprasad Somasundaram at 3M and Dr. Ravishankar Sivalingam at Qualcomm; I learned a lot from them during my respective internships. I am also thankful to Srikrishnan Srinivasan, who is like a brother to me and helped me immensely in my career planning. 

I am grateful to the Digital Technology Center (DTC) for awarding me the ADC Graduate Fellowship in the $2015$-$16$ academic year. I gratefully acknowledge the generous supports from the University of Minnesota's MNDrive initiative and the Minnesota Robotics Institute (MnRI) Seed Grant as well. Several research projects in this dissertation were funded by the National Science Foundation (NSF) grant numbers IIS-$1845364$ and IIS-$1637875$. I am also thankful to the University of Minnesota graduate school for awarding me the prestigious Doctoral Dissertation Fellowship (DDF) in the $2019$-$20$ academic year toward my Ph.D. dissertation.

Finally, I would like to express my gratitude and appreciation towards my loving family and friends. My parents, sisters, close friends, and cricket buddies have always supported me and kept me going through the darkest of times. They encouraged me in successes, always been there for me during failures, and extended their support in every possible way - I owe my academic career to them.
}
\dedication{To my father, Md Harunur Rashid, whom I lost during the second year of my Ph.D. And to my superwoman mother, Taslima Akter, who is the source of all my inspiration.
}

% Use a special preface
%\extra{\input{preface}}

% The \beforepreface command actually causes insertion of the title, 
% abstract, signature, and copyright pages into the new document.
\beforepreface 

% Define the text to go before the table of contents
\figurespage
\tablespage

% The \afterpreface command actually causes insertion of the
% contents, list of figures, etc. into the new document.
\afterpreface            
%%%%%%%%%%%%%%%%%%%%%%%%%%%%%%%%%%%%%%%%%%%%%%%%%%%%%%%%%%%%%%%%%%%%%%%%%%%%%%%%

\chapter{Introduction}\label{intro}
\begin{figure}[t]
\centering
    \subfigure[Reef exploration, monitoring, surveying, sampling, and data collection~\cite{shkurti2012multi,beijbom2012automated,koreitem2020one,modasshir2020enhancing,hoegh2007coral}.]{
        \includegraphics[width=0.98\textwidth]{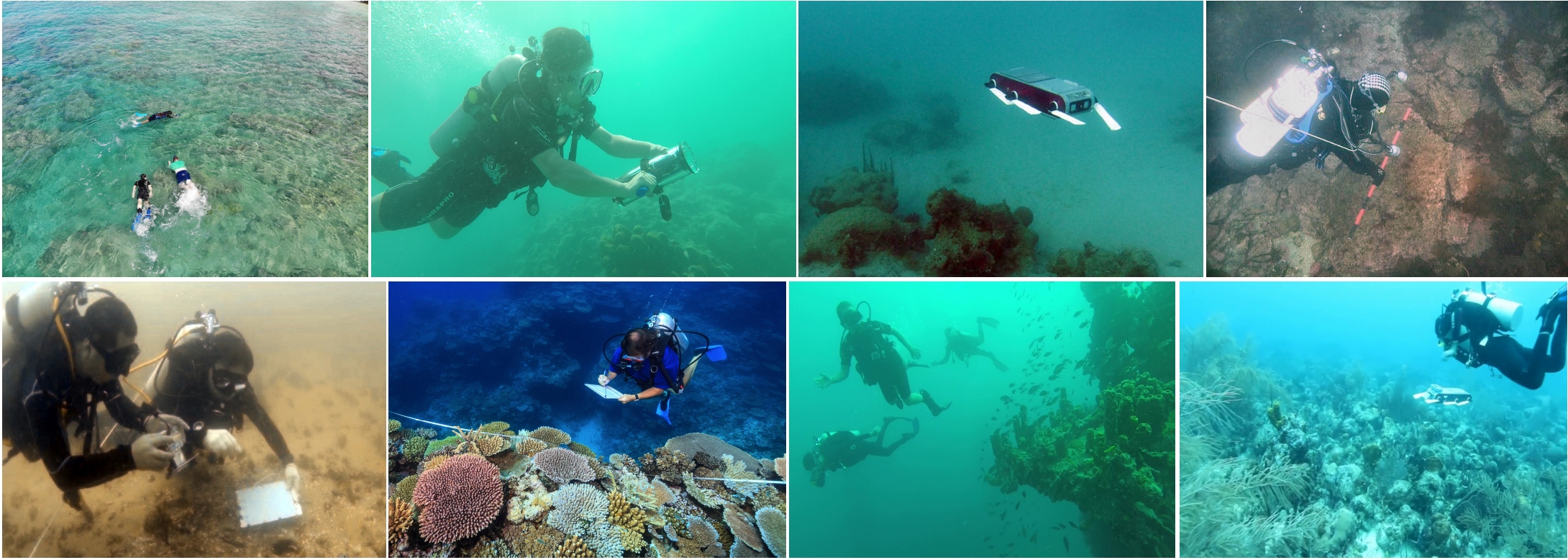}
        \label{fig:1.1a}
    }%
    
    \subfigure[Collaborative tracking, shipwreck inspection, and subsea mapping~\cite{WoodsHole,kim2013real,rekleitis2001multi,ROVHercules}.]{
        \includegraphics[width=0.98\textwidth]{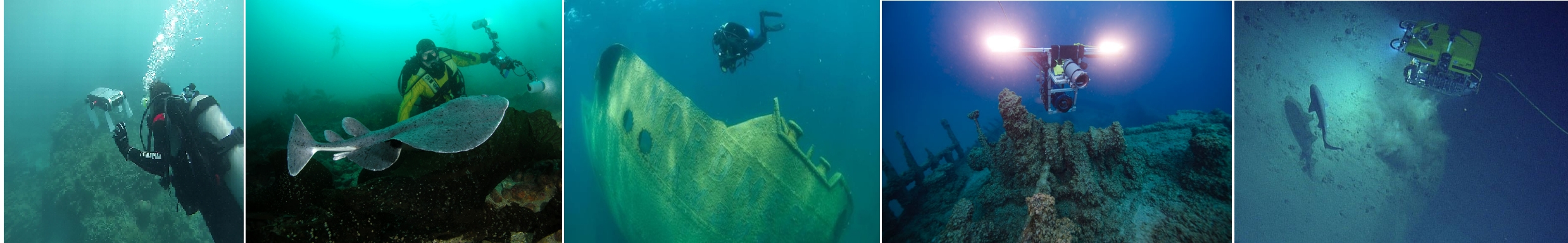}
        \label{fig:1.1c}
    }%
\caption{A few snapshots showing the use of visually-guided AUVs, ROVs, and high-end imaging sensors in oceanic explorations and  human-robot cooperative tasks.}
\label{fig:1.1}
\end{figure}

Underwater robotics is a domain of increasing importance with existing and emerging applications ranging from monitoring, inspection, and surveillance to data collection, surveying, and bathymetric mapping. In particular, visually-guided AUVs (Autonomous Underwater Vehicles) and ROVs (Remotely Operated Vehicles) are widely used in important coastal-water and shallow-water applications such as the monitoring of coral reefs and marine species migration~\cite{dunbabin2006visual,alonso2019coralseg,shkurti2012multi}, the inspection of submarine cables and wreckage~\cite{bingham2010robotic,carreras2018sparus}, underwater scene analysis~\cite{lee2012vision,schechner2005recovery}, seabed mapping~\cite{kennedy2019unknown,paull2018probabilistic}, and more (see Figure~\ref{fig:1.1}). Since truly autonomous navigation is still an open problem, underwater missions are often deployed with a team of human divers and robots that cooperatively perform a set of common tasks. The human divers typically lead the mission and interact with the robots during task execution. Such human-in-the-loop guidance for autonomous and semi-autonomous robots simplifies the mission planning~\cite{islam2018understanding,sattar2008enabling} and significantly reduces the associated operational risks and computational overhead.

In human-robot collaborative settings, underwater robots typically rely on vision for exteroceptive perception. A practical alternative is to use acoustic sensors such as sonars and hydrophones. However, they are mainly used for deep-water target tracking~\cite{mandic2016underwater,demarco2013sonar} or bathymetric mapping~\cite{kennedy2019unknown} as they are not suitable for interactive applications. Acoustic sensors also face challenges in coastal waters due to scattering and reverberation. Additionally, their usage is often limited by government regulations on the sound level in marine environments~\cite{islam2018person}. On the other hand, cameras are passive sensors, \ie, they do not emit energy and hence are not intrusive to the marine ecosystem~\cite{slabbekoorn2010noisy}. Due to these compelling reasons, visual sensing is more feasible and generally preferred for shallow-water and coastal-water applications.

The existing literature and technological solutions for underwater visual perception have progressed on an unprecedented scale over the last decade. In particular, the advent of deep learning and powerful single-board computers have revolutionized the visual perception capabilities of autonomous robots and systems. Despite the recent advancements, the design of robust and efficient perception pipelines for visually-guided AUVs remains an active research problem. The existing systems and state-of-the-art (SOTA)  perception solutions are faced with several domain-specific practicalities. First, visual sensing and estimation are inherently challenging underwater due to limited visibility and a wide range of chromatic distortions caused by the waterbody-specific degraded optical characteristics~\cite{akkaynak2018revised,islam2019fast}. Consequently, ensuring consistent and reliable performance for visual detection and tracking, servoing, and detailed scene understanding are particularly hard problems for underwater robots. Additionally, real-time operating requirements and single-board computational constraints make it a notoriously challenging undertaking to ensure robust yet efficient perception
performance. In this dissertation, we delineate our attempts to address these challenges by designing novel and improved perception pipelines for visually-guided underwater robots. Our contributions include intuitive design, efficient implementation, and field experimental validation of the underlying vision and learning-based algorithms. Over several important application scenarios, we demonstrate the feasibility and effectiveness of our proposed solutions from the perspective of real-time robot vision, human-robot cooperation, and computational imaging technologies.

\section{Research Objectives and Scope}
The research focus of this dissertation abides in the context of underwater exploration, inspection, and monitoring by human-robot teams. In these applications, visually-guided AUVs and human divers cooperatively perform a set of complex operations. In coral reef inspection, for instance, tasks for a robot include following and interacting with its companion divers throughout the mission, maintaining spatial coordination during navigation, identifying interesting features/species of coral reefs, collecting samples, recording events, and other pre-defined atomic tasks. Therefore, understanding human (swimming) motion and instructions, robust detection or tracking of specific objects of interest, and accurate interpretation of the visual scenes are the primary capabilities for a robot. Our objective is to enable these visual perception capabilities while providing intelligent solutions to deal with the unique challenges posed by the unpredictable and noisy underwater environment. Specifically, we address the practicalities that can be grouped into two categories: $i$) challenges of robust underwater visual perception, and $ii$) domain-specific real-time operational complexities.

\subsection{Challenges of Robust Underwater Visual Perception}\label{chal_un_vision}
Underwater imagery suffers from a host of irregular non-linear distortions due to the unique characteristics of underwater light propagation~\cite{akkaynak2018revised} such as wavelength-dependent attenuation, absorption, and scattering. These artifacts depend on the specific optical properties of a waterbody, distance of light sources, salinity, and many other factors. Only some of these aspects can be modeled by physics-based approximations~\cite{berman2018underwater,bryson2016true}, that too with significant computational overhead. Moreover, accurate physics-based modeling of the underlying optical distortions requires information such as the scene depth and water-quality measures~\cite{akkaynak2019sea}, which are not always available in practical applications. Consequently, despite often using high-end cameras, underwater robots have to deal with low-contrast and color-degraded images that lack salient visual features. These artifacts severely affect the performance of perception tasks such as detection, tracking, visual servoing, and scene parsing.

For a given application, adopting data-driven approaches are generally more feasible for combating noisy visual data. These approaches approximate the underlying optical distortion functions by learning \emph{perceptual image enhancement} from a large amount of data. In recent years, various deep learning-based models~\cite{fabbri2018enhancing,yu2018underwater,liu2019underwater} have demonstrated remarkable success in restoring the perceptual and statistical qualities of the distorted underwater images for improved visual perception. However, the generalizability of these models is rather limited because only small-scale and often synthetically distorted images are used for paired learning aiming at a specific application. It is also practically impossible to accumulate large-scale underwater data that capture the whole spectrum of natural image distortions over various setups and visibility conditions in all waterbodies (of five open-ocean spectra and nine coastal spectra based on Jerlov classiﬁcation~\cite{solonenko2015inherent}). Besides, the analogous SOTA perceptual enhancement models for atmospheric imagery~\cite{ignatov2017dslr,chen2018deep} are computationally too demanding for real-time deployments on embedded devices.  Moreover, \emph{transfer learning} from these models pre-trained on terrestrial data is not very beneficial as the visual content of underwater imagery is entirely different in terms of object categories and background waterbody patterns~\cite{islam2019fast}. Consequently, the existing data-driven systems and methodologies are not extendable to fast generalizable solutions that can be used as image pre-processing filters in the autonomy pipeline of visually-guided underwater robots.
% https://www.oceanopticsbook.info/view/overview-optical-oceanography/classification-schemes

\subsection{Domain-specific Real-time Operational Complexities}
There are significant operational complexities involved in effective human-robot cooperation in unstructured, unpredictable, feature-deprived, and GPS-denied underwater environments. The underlying challenges and practicalities are twofold. First, unless high-end surface sensory supports such as the ultra-short base-lines (USBLs) are available, robots rely on their companion divers for important navigational decisions. Hence, a simple and natural human-to-robot communication system with minimal cognitive overhead is paramount for operational success. Secondly, Wi-Fi or radio communication is severely degraded underwater~\cite{dudek2008sensor}, which strictly requires standalone on-board processing for all computations components in real-time. 

Such adverse operating conditions call for two major features of visual perception algorithms: robustness to noisy visual data and real-time performance on single-board robotic platforms. These become extremely difficult in complex tasks such as detailed scene understanding and fast visual search. Therefore, the robot needs capabilities to infer high-resolution scene interpretation from noisy low-resolution measurements and intelligently model visual attention to facilitate efficient single-board computation, which are not explored in-depth in the literature.

\section{Research Contributions}
In this dissertation, we address the inherent challenges of real-time underwater machine vision to ensure effective human-robot cooperation in shallow-water and coastal-water applications. Our research finds novel and improved perception solutions for visually-guided AUVs to \emph{\textbf{follow and interact}} with companion divers while only relying on the low-powered passive sensing. Additionally, our proposed solutions enable underwater robots to \emph{\textbf{see better}} in noisy visual conditions and \emph{\textbf{do better}} with limited computational resources and real-time constraints. We provide a brief overview of these aspects and highlight our research contributions in the following sections.

\subsection{Follow \& Interact: Enabling Features of a Companion Robot}
A significant segment of our research efforts is dedicated to developing the key capabilities of companion robots in underwater human-robot cooperation. In a broader sense, we first identify and benchmark various perception, planning, and interaction modules for autonomous person-following by ground, underwater, and aerial robots. We investigate the operational constraints and experimentally validate the optimal design choices for effective human-robot cooperation in each medium of operation~\cite{islam2018person}. For underwater applications, in particular, the capabilities for an AUV entail visually detecting and following its companion diver, understanding instructions, maintaining spatial coordination with other communicating robots, and performing various tasks autonomously.

\begin{figure}[ht]
\centering
    \subfigure[We adopt two classes of algorithms for diver detection and tracking: (i) spatio-temporal tracking by MDPM~\cite{islam2017mixed} based on spatial-domain and frequency-domain motion signatures in $1$-$2$ Hz bands; (ii) Tracking by appearance-based deep visual detection (bounding box detection~\cite{islam2018towards}, body pose detection~\cite{cao2017realtime}, and pixel-level semantic masking~\cite{islam2020suim}).]{
        \includegraphics[width=0.98\textwidth]{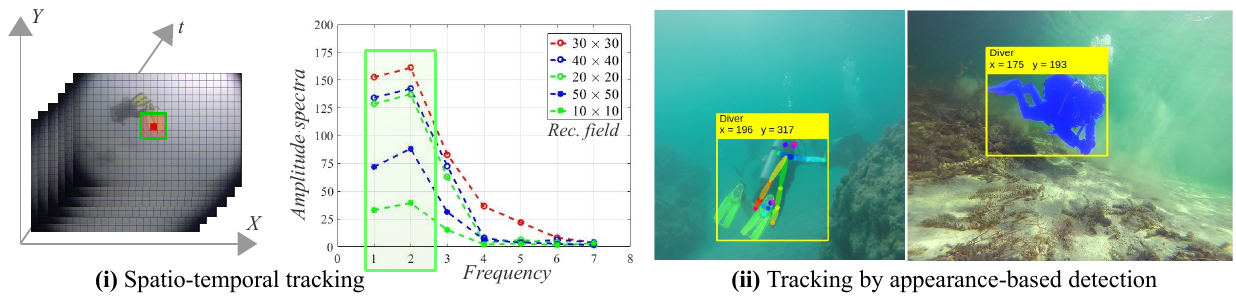}
        \label{fig:1.2a}
    }%
    
    \subfigure[Diver detection by our deep CNN-based model~\cite{islam2018towards} in various scenarios.]{
        \includegraphics[width=0.98\textwidth]{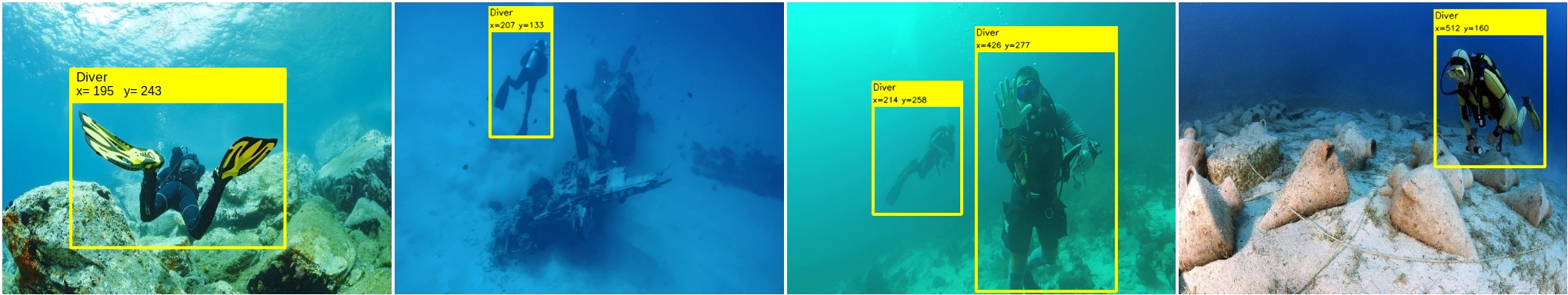}
        \label{fig:1.2b}
    }%
\caption{A few demonstrations of our proposed perception modules for autonomous diver-following in cooperative tasks; more details are provided in Chapter~\ref{diver_following}.}
\label{fig:1.2}
\end{figure}

We present two diver-following modules driven by $i$) fast spatio-temporal motion tracking; and $ii$) deep visual learning. The first one, \textbf{Mixed-Domain Periodic Motion (MDPM)} tracking~\cite{islam2017mixed}, is a classical approach that uses both spatial domain and frequency domain features to track a diver's swimming motion. Specifically, we formulate a spatio-temporal optimization problem to track the flippers' oscillating motion of a diver along a sequence of non-overlapping image regions over time. We then deploy Hidden Markov Model (HMM)-based search-space reduction, followed by frequency-domain filtering to find the optimal motion direction (see Figure~\ref{fig:1.2a}). MDPM tracker relaxes the directional constraint of the existing frequency-domain trackers~\cite{Sattar09RSS} and demonstrates superior performance in tracking arbitrary swimming motions. However, it does not take advantage of important features such as divers' appearance/wearables and its performance is affected by their swimming trajectories (\ie, straight-on or sideways). We address these limitations and improve general tracking performance by leveraging the power of various \textbf{Deep Diver Detection (DDD)} models~\cite{islam2018understanding}. Subsequently, we design an architecturally simple \textbf{Convolutional Neural Network (CNN)-based model}~\cite{islam2018towards} that is significantly faster than SOTA object detection models (\eg, Faster R-CNN~\cite{renNIPS15fasterrcnn}, SSD~\cite{liu2016ssd}, YOLO~\cite{redmon2016yolo9000}), yet provides comparable detection performance. Each building block of the proposed model is fine-tuned to balance the trade-off between robustness and efficiency for a single-board setting under real-time constraints. With the integration of a standard visual servo controller~\cite{espiau1992new}, we validate its performance and general applicability for diver following through numerous field experiments in pools and oceans; a few sample demonstrations are shown in Figure~\ref{fig:1.2b}.

\begin{figure}[t]
\centering
    \subfigure[A scenario where the OpenPose~\cite{cao2017realtime}-based key-points provide more reliable correspondences than standard SIFT-based features due to a lack of salient features and natural landmarks in the scene.]{
        \includegraphics[width=0.49\textwidth]{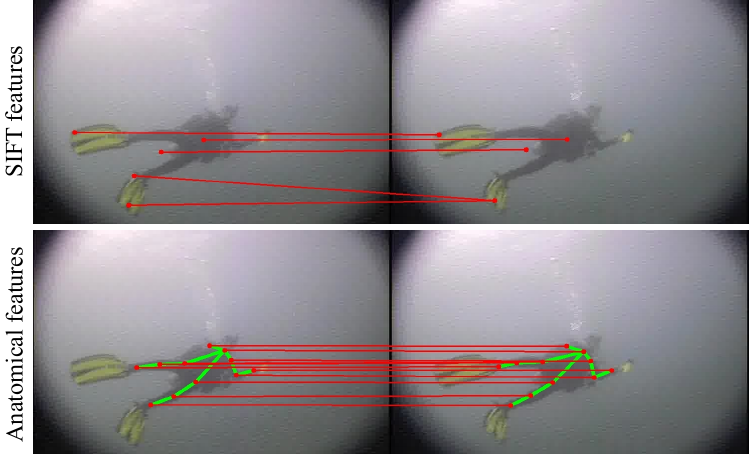}
        \label{fig:1.3a}
    }~
    \subfigure[The use of AR-Tags and hand gestures for diver-to-robot communication~\cite{dudek2007visual} is shown on the left and a snapshot of dynamic robot programming via RoboChatGest~\cite{islam2018dynamic} is shown on the right.]{
        \includegraphics[width=0.48\textwidth]{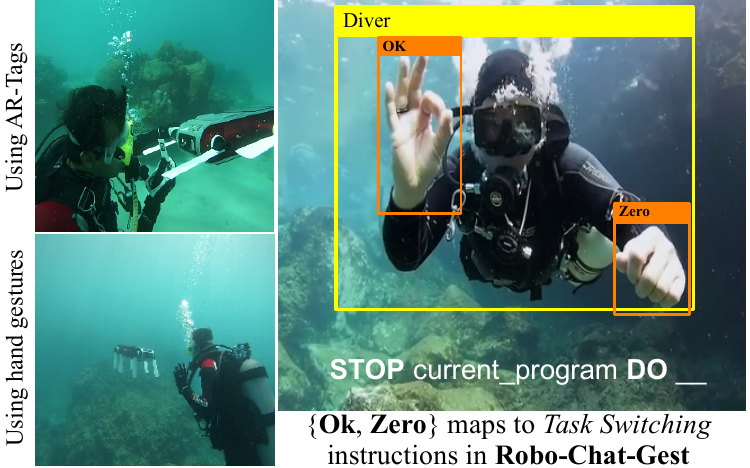}
        \label{fig:1.3b}
    }%
\caption{A few illustrations showing (a) the motivation of using human body pose-based key-points for relative pose estimation in our proposed system~\cite{islam2019robot} (more in Chapter~\ref{r2r_pose}); and (b) a setup of using hand gesture-based instructions for diver-to-robot communication via RoboChatGest~\cite{islam2018dynamic,islam2018understanding} (more in Chapter~\ref{hrc}).}
\label{fig:1.3}
\end{figure}

We also develop a \textbf{robot-to-robot relative pose estimation} method~\cite{islam2019robot} that only uses human body poses to combat the lack of salient visible features in underwater environments (see Figure~\ref{fig:1.3a}). Specifically, we propose a method to determine the 3D relative pose of pairs of communicating robots by regressing human pose (\ie, OpenPose~\cite{cao2017realtime})-based key-points as correspondences. To ensure the perspective geometric validity of pose estimation, we design a person re-identification module and a key-point refinement algorithm for fast body-pose association by exploiting local structural properties in image-space with multi-view constraints. We provide a fast implementation of the proposed system and evaluate its end-to-end performance over several terrestrial and underwater scenarios. The experimental results suggest that it can be useful for \textit{implicit} interactions~\cite{sattar2012thesis,islam2018person} in maintaining spatial coordination by cooperative robots, particularly amidst a lack of natural landmarks in underwater applications.

For \textit{explicit} interaction, we introduce \textbf{RoboChatGest}~\cite{islam2018dynamic}, a hand gesture-based language for real-time robot programming and parameter reconfiguration. RoboChatGest improves upon an existing fiducial tag-based system named RoboChat~\cite{dudek2007visual}, with its syntactic simplicity and computational efficiency. It also reduces the cognitive load on divers by eliminating the need for carrying and manipulating tags to program complex language rules during an underwater mission. The proposed end-to-end system comprises of $i$) a set of intuitive hand-gestures to instruction mapping rules, $ii$) a deep CNN-based hand gesture recognizer, and $iii$) a finite-state machine-based instruction decoder. We also provide the option to use SOTA deep visual detectors (\eg, SSD~\cite{liu2016ssd}, Faster R-CNN~\cite{renNIPS15fasterrcnn}) for more reliable hand gesture recognition. We evaluated the usability benefits of RoboChatGest through user interaction studies and validated its end-to-end performance by extensive field experiments in real-world settings~\cite{islam2018understanding}. A typical setup is shown in Figure~\ref{fig:1.3b}; it is now used as the de facto interaction system on our Aqua MinneBot AUV~\cite{dudek2007aqua} and LoCO AUV~\cite{LoCOAUV} for diver-to-robot communication, with the tag-based system used only as backup.

\subsection{See Better: Conquering Adverse Visual Conditions}
\begin{figure}[t]
\centering
    \subfigure[A few instances of perceptual image enhancement by FUnIE-GAN~\cite{islam2019fast} are shown; its adversarial training is driven by an objective function that compensates image quality based on color, local texture, global image content, and style information.]{
        \includegraphics[width=0.98\textwidth]{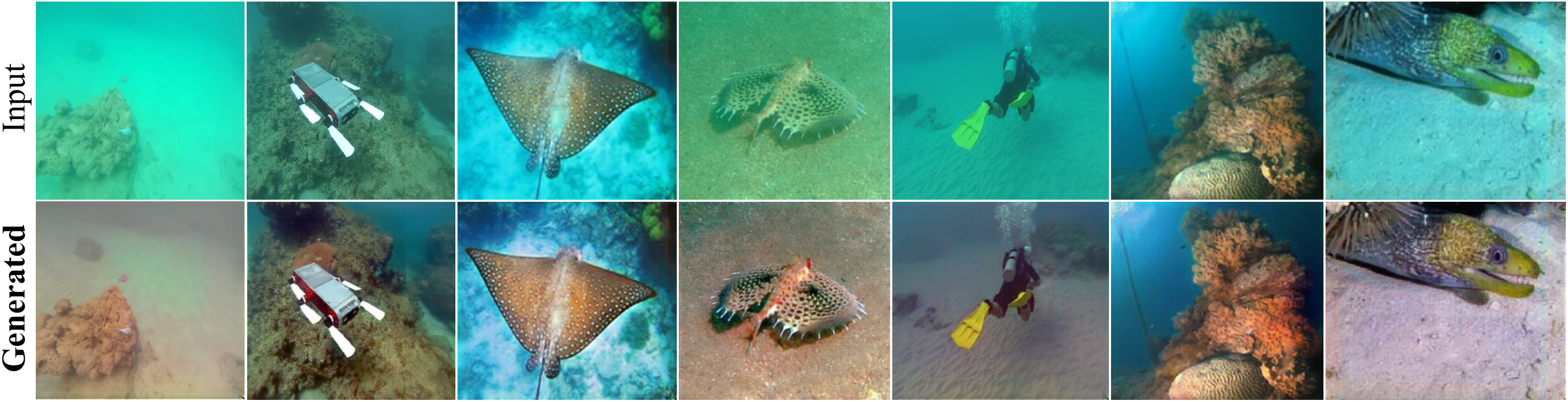}
        \label{fig:1.4a}
    }%
    
    \subfigure[Two examples showing improved performance of standard models for underwater object detection~\cite{islam2018towards}, human pose estimation~\cite{cao2017realtime}, and saliency prediction~\cite{wang2018salient} on FUnIE-GAN-enhanced images.]{
        \includegraphics[width=0.98\textwidth]{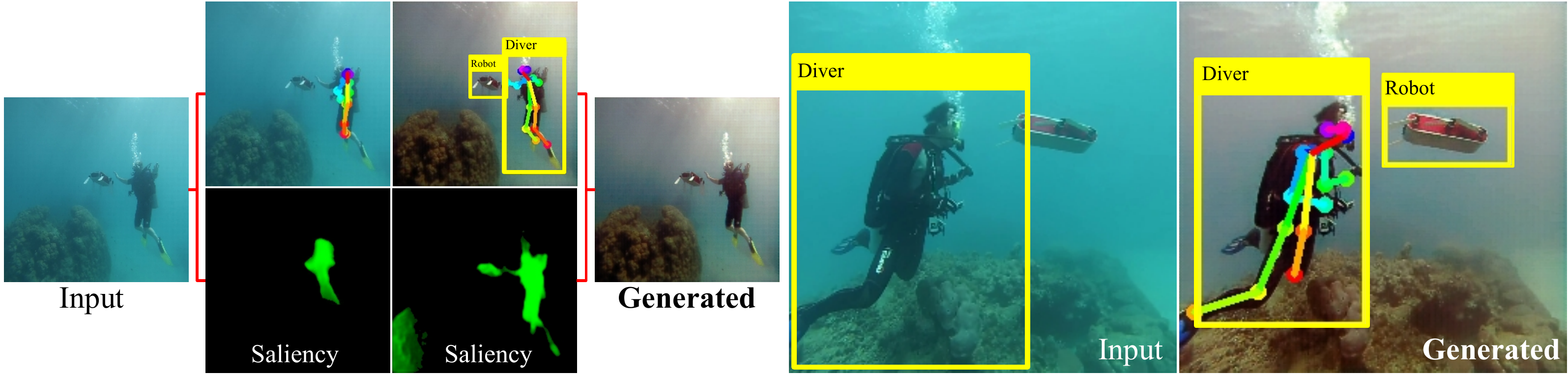}
        \label{fig:1.4b}
    }%
\caption{A few demonstrations of underwater image enhancement
by our proposed FUnIE-GAN model~\cite{islam2019fast} and its practical feasibility for real-time robot vision; more methodological details and experimental results are provided in Chapter~\ref{en_sr}.}
\label{fig:1.4}
\end{figure}

In Section~\ref{chal_un_vision}, we mentioned various artifacts for underwater image distortions and discussed how they affect the visual perception performance of AUVs. Our research on image enhancement has led to the design of robust techniques to alleviate these problems by restoring the perceptual and statistical qualities of distorted underwater images in real-time (see Figure~\ref{fig:1.4a}). In particular, we devise a general-purpose model for  \textbf{Fast Underwater Image Enhancement using a Generative Adversarial Network (GAN): FUnIE-GAN}~\cite{islam2019fast}, which can learn perceptual enhancement from both paired and unpaired data. It is a fully-convolutional conditional GAN-based model, which offers over $48$ FPS inference on NVIDIA\texttrademark{}~AGX Xavier and over $25$ FPS on NVIDIA\texttrademark{}~Jetson TX2, in addition to providing SOTA enhancement performance. Such speeds on single-board devices make it ideal for real-time robotic deployments to combat adverse visual conditions. We empirically validated that the enhanced images provide improved performance for standard perception tasks such as underwater object detection~\cite{islam2018towards} (up to $14$\%), human body-pose estimation~\cite{cao2017realtime} (up to $28$\%), and class-agnostic saliency prediction~\cite{wang2018salient} (up to $28$\%); a few qualitative examples are illustrated in Figure~\ref{fig:1.4}. We also release the \textbf{EUVP dataset}, which we collected and configured for FUnIE-GAN training; it is the first large-scale dataset to facilitate both paired and unpaired learning of underwater image enhancement.

We also design efficient single image super-resolution (SISR) modules that allow visually-guided AUVs to \textit{zoom in} interesting image regions for detailed perception. Specifically, we designed \textbf{Deep Residual Multiplier (DRM)} modules~\cite{islam2019srdrm}, which can be used in both generative (SRDRM) and adversarial (SRDRM-GAN) training pipelines for $2\times$, $4\times$, and $8\times$ SISR of underwater imagery. Such zoom-in capabilities on low-resolution (LR) image region of interests (RoIs) are particularly useful for detailed scene understanding in surveying distant coral reefs or seabed~\cite{girdhar2014autonomous,hoegh2007coral,shkurti2012multi}. However, if the LR image patches suffer from noise and optical distortions, those get amplified by SISR, resulting in high-resolution (HR) yet uninformative RoIs.

\begin{figure}[t]
\centering
    \subfigure[Deep SESR jointly learns saliency prediction and SESR on a shared hierarchical feature space: the predicted salient foreground pixels (shown as green intensity values on the top left corners) in turn guides the network to learn global contrast enhancement, in addition to color and sharpness restoration.]{
        \includegraphics[width=0.98\textwidth]{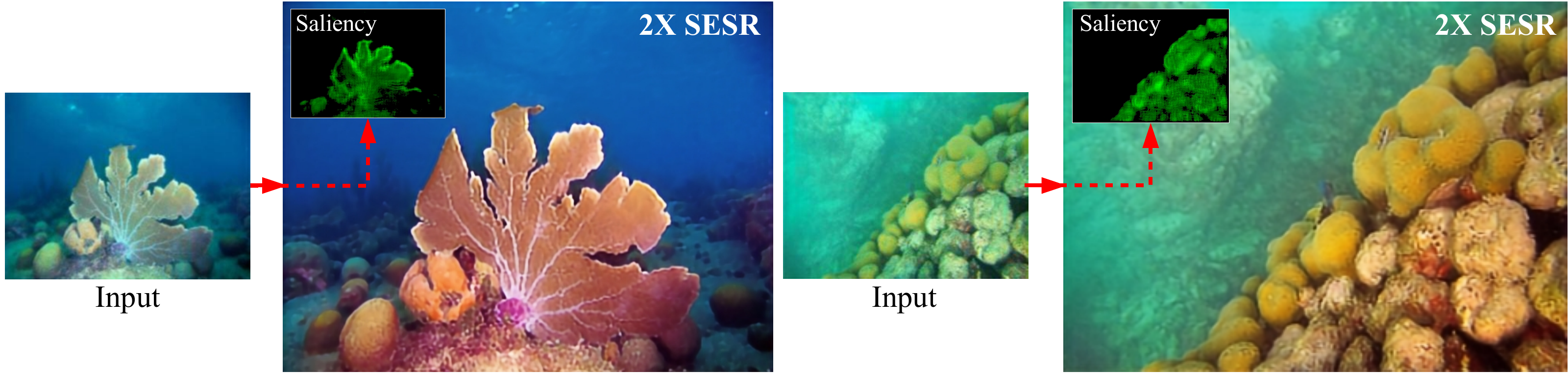}
        \label{fig:1.5a}
    }%
    
    \subfigure[Deep SESR rectifies hue and restores color, contrast, and sharpness at up to $4\times$ higher resolutions.]{
        \includegraphics[width=0.98\textwidth]{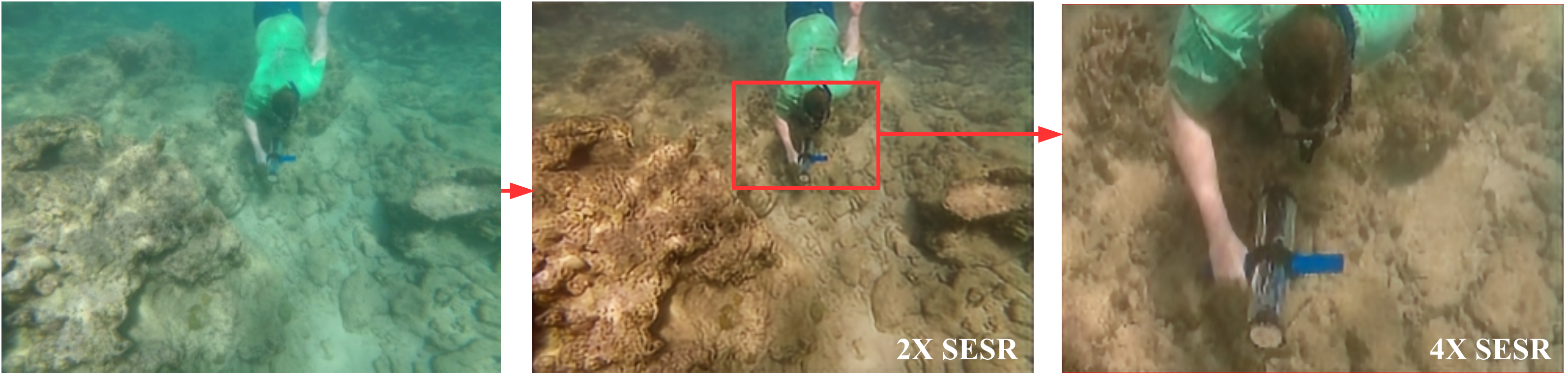}
        \label{fig:1.5b}
    }%
\caption{A few illustrations of perceptually enhanced HR image generation and saliency prediction by our proposed Deep SESR model~\cite{islam2020sesr}. The enhanced images restore color, contrast, and sharpness at $2\times$/$3\times$/$4\times$ spatial scales to facilitate improved underwater visual perception; more details are provided in Chapter~\ref{sesr}.}
\label{fig:1.5}
\end{figure}

To address this practicality, we introduce a new research problem: \textbf{Simultaneous Enhancement \& Super-Resolution (SESR)}, and design an efficient solution for underwater imagery. Our proposed solution, \textbf{Deep SESR}~\cite{islam2020sesr}, is a residual-in-residual network-based model that learns to restore perceptual image qualities for up to $4\times$ higher spatial resolution. We supervise its learning by formulating a multi-modal objective function to address the chrominance-specific underwater color degradation, lack of image sharpness, and loss in high-level feature representation. Moreover, we configure the network to jointly learn saliency prediction and SESR on a shared feature space for accurate contrast recovery on foreground pixels (see Figure~\ref{fig:1.5a}). Over a series of qualitative and quantitative experiments, we demonstrate that Deep SESR outperforms the SOTA solutions for underwater image enhancement and super-resolution. Additionally, it offers over $10$ FPS inference on NVIDIA\texttrademark{}~AGX Xavier, which is significantly faster than any combinations of existing enhancement and super-resolution models for underwater robot vision. As shown in Figure~\ref{fig:1.5}, Deep SESR generates hue rectified, perceptually enhanced, and sharpness restored HR images from noisy LR measurements. In addition to the model, we release \textbf{UFO-120}, the first dataset to facilitate large-scale SESR learning; we also provide several application-specific design choices and training configurations for the underlying SESR problem.

%we also demonstrate the feasibility of SESR in terrestrial imagery for further research on this problem.

\subsection{Do Better: Meeting On-board Computational Constraints}
In addition to ensuring robust performances of the proposed visual perception modules, one key aspect of our research is to analyze their practical feasibility and develop efficient implementations for robotic deployments. We find application-specific design choices for their algorithms~\cite{islam2019fast,islam2020sesr,islam2020svam} and provide effective solutions to deal with the practicalities. While we extensively validate these solutions by oceanic field experiments, we also design capabilities that facilitate intelligent decision-making for AUVs in order to meet the on-board computational requirements in real-world application scenarios. %(\eg, how and when to use these modules)

\begin{figure}[t]
\centering
    \subfigure[SVAM-Net model identifies salient image regions for visual attention modeling by AUVs. The decoupled light model, \ie, SVAM-Net\textsuperscript{Light}, generates abstract saliency maps (shown in green intensity and red object contours) from an early branch, ensuring fast processing on single-board platforms.]{
        \includegraphics[width=0.98\textwidth]{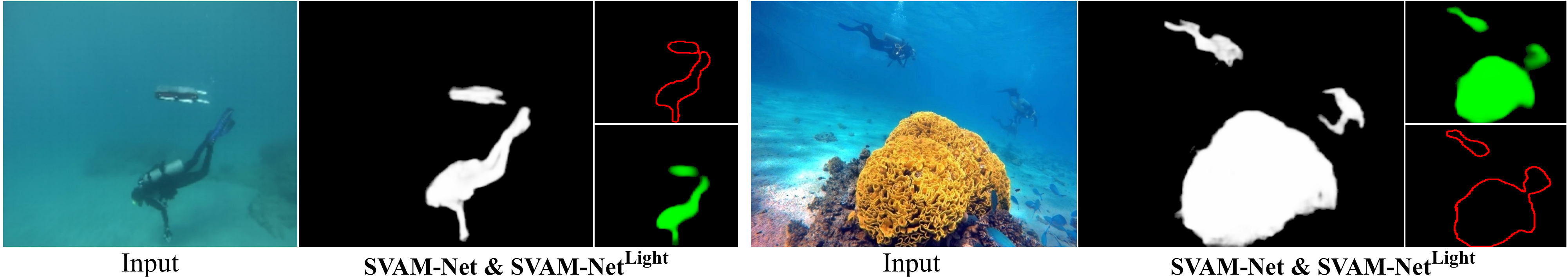}
        \label{fig:1.6a}
    }%
    
    \subfigure[A use case of SVAM-Net\textsuperscript{Light} (denoted by \textit{\textbf{G}}) is shown for high-resolution image enhancement. On the left: \textit{\textbf{G}}-generated saliency maps are used for salient RoI pooling, and then FUnIE-GAN (denoted by \textit{\textbf{F}}) is applied on all $256\times256$ patches; the total processing time is $131$ ms. As shown on the right, it would take $250$ ms to enhance the image at $1024\times768$ resolution (on NVIDIA\texttrademark{}~AGX Xavier).]{
        \includegraphics[width=0.98\textwidth]{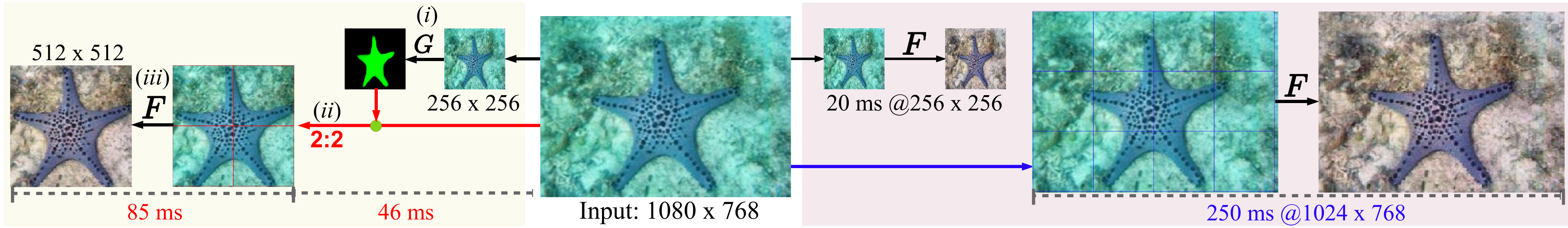}
        \label{fig:1.6b}
    }%
\caption{A few demonstrations of saliency-guided visual attention modeling by our proposed SVAM-Net model~\cite{islam2020svam} and its benefits for salient RoI enhancement are shown; the methodological details and more experimental analyses are provided in Chapter~\ref{svam}.}
\label{fig:1.6}
\end{figure}

To this end, our work on \textbf{Saliency-guided Visual Attention Modeling (SVAM)} enables AUVs to identify interesting and salient objects in images to make fast operational decisions. Our proposed model, \textbf{SVAM-Net}~\cite{islam2020svam}, integrates deep visual features at various scales and semantics for accurate Salient Object Detection (SOD) in natural underwater images. It jointly accommodates bottom-up and top-down learning within two separate branches of the network while sharing the same encoding layers. We design dedicated Spatial Attention Modules (SAMs) along these learning pathways to exploit the coarse-level and top-level semantic features for SOD at multiple stages of abstractions. In the deeper top-down pipeline, we attach a Residual Refinement Module (RRM) for fine-grained saliency estimation that contributes to SOTA performance on benchmark underwater SOD datasets, with better generalization performance on challenging test cases than existing approaches. Besides, the bottom-up pipeline extracts semantically rich features from early encoding layers for an abstract saliency prediction at a significantly faster rate. We denote this decoupled lighter branch as \textbf{SVAM-Net\textsuperscript{Light}}; it offers $21.77$ FPS inference on NVIDIA\texttrademark{}~AGX Xavier, which is equivalent to over $86$ FPS run-time on a GTX 1080 GPU. As shown in Figure~\ref{fig:1.6a}, the SVAM-Net\textsuperscript{Light}-generated saliency maps segment interesting image regions by accurately discarding the background (waterbody) pixels. It facilitates effective visual attention modeling for fast on-board computation in various perception tasks such as image enhancement, super-resolution, scene parsing, and visual search.

A particular use case of SVAM-Net\textsuperscript{Light} for high-resolution image enhancement is demonstrated in Figure~\ref{fig:1.6b}. Here, we consider FUnIE-GAN~\cite{islam2019fast}, the fastest available image enhancement model with an input reception resolution of $256\times256$; it takes $20$ ms processing time on NVIDIA\texttrademark{}~AGX Xavier. Hence, it requires $250$ ms time to enhance and combine all patches of a $1080\times768$ input image. Alternatively, SVAM-Net\textsuperscript{Light}-generated saliency maps (which costs $46$ ms) can be used to perform  `{salient RoI enhancement}' in a total of $131$ ms time. Therefore, it saves $45\%$ processing time even though the salient RoI occupies more than half the image regions. In our experiments, we found up to $75\%$ and $67\%$ faster computation for RoI enhancement and RoI super-resolution tasks, respectively. We also demonstrated that in a broader sense, SVAM-Net provides a general-purpose solution to the `where to look' problem, which is essential for uninformed visual search and object localization in autonomous exploratory tasks by visually-guided underwater robots.

In Chapter~\ref{do_better}, we discuss several other aspects for platform-specific model adaption and investigate a few intriguing research questions. In particular, we discuss the operational utility of a standalone SISR module (\eg, SRDRM or SRDRM-GAN)~\cite{islam2019srdrm} for underwater imagery. We also explored the research problem of class-aware saliency prediction and semantic Segmentation of Underwater IMagery (SUIM)~\cite{islam2020suim}. For a case study with eight object categories, our proposed SUIM-Net model shows promising results in semantic segmentation of underwater scenes. Furthermore, we design a computationally light module for assessing underwater image quality by introducing an \textit{entangled discriminator} within the adversarial training pipeline of GAN-based image enhancement models. This light module enables AUVs to use the relatively heavier image enhancement filters more efficiently and only when the visual quality is poor.          

%which are important for the operational success of visually-guided AUVs in human-robot cooperative applications.

%

\section{Research Publications, Code, and Data}
\subsection{Peer-reviewed Publications}
\begin{itemize}
    \item The MDPM tracker is published at the 2017 IEEE International Conference on Robotics and Automation (ICRA)~\cite{islam2017mixed}, while the CNN-based diver following module and its computational aspects appeared at the IEEE Robotics and Automation Letters (RA-L) journal~\cite{islam2018towards} (in vol. 4, no. 1, 2018). 
    \item The RoboChatGest system is published in the proceedings of ICRA-2018 and its extended version with detailed feasibility analyses appeared at the Journal of Field Robotics (JFR)~\cite{islam2018understanding} (in vol. 36, no. 5, 2018). 
    \item A comprehensive survey paper outlining our investigations on autonomous person-following robots for ground, underwater, and aerial domains appeared at the International Journal of Robotics Research (IJRR)~\cite{islam2018person} (in vol. 38, no. 14, 2019).
    \item The proposed robot-to-robot relative pose estimation method is accepted for publication at the Autonomous Robots (AuRo)~\cite{islam2019robot} journal (to appear).
    \item The FUnIE-GAN model for paired and unpaired underwater image enhancement and its experimental analyses are at published at the IEEE RA-L journal~\cite{islam2019fast} (in vol. 5, no. 2, 2020). The SRDRM and SRDRM-GAN models for underwater image super-resolution are published in the proceedings of ICRA-2020~\cite{islam2019srdrm}. 
    \item The Deep SESR model for simultaneous enhancement \& super-resolution of underwater imagery with its experimental validations are published in the proceedings of the Robotics: Science and Systems (RSS)-2020 conference~\cite{islam2020sesr}. 
    \item The SUIM-Net model for class-aware saliency prediction and semantic segmentation of underwater imagery is published at the 2020 IEEE/RSJ International Conference on Intelligent Robots and Systems (IROS)~\cite{islam2020suim}. 
    \item The SVAM-Net models for saliency-guided visual attention modeling and relevant research findings are under review at the IEEE Transactions on Pattern Analysis and Machine Intelligence (T-PAMI) journal~\cite{islam2020svam} (as of April 2021).

\end{itemize}

\subsection{Shared Code Repositories and Datasets}
Our research contributions are featured as state-of-the-art benchmarks in forums such as Papers-With-Code (\url{https://paperswithcode.com}), with more than $2100$ downloads over the last two years. The research outcomes, \textit{i.e.}, packages, software, and data are released to the broader academic community at \url{http://irvlab.cs.umn.edu/resources}.

\section{Outline of the Manuscript}
The rest of this document is organized as follows. Chapter~\ref{diver_following} presents the autonomous diver following modules (\ie, MDPM tracker~\cite{islam2017mixed} and DDD~\cite{islam2018towards}); it includes elaborate discussions on how we balance the robustness-efficiency trade-off for deep visual diver detection and tracking. Then, Chapter~\ref{r2r_pose} presents a robot-to-robot relative pose estimation method using mutually visible humans' body-pose~\cite{islam2019robot}, while Chapter~\ref{hrc} demonstrates the operational feasibility of our proposed human-to-robot communication system: RoboChatGest~\cite{islam2018dynamic,islam2018understanding}. Chapter~\ref{en_sr} presents the design and implementation details of the FUnIE-GAN model~\cite{islam2019fast}, and validates its utility as a fast underwater image enhancement filter for improved visual perception. Subsequently, we introduce the SESR problem and present the proposed Deep SESR model~\cite{islam2020sesr} in Chapter~\ref{sesr}. Next, Chapter~\ref{svam} reveals how the SVAM-Net~\cite{islam2020svam} model solves the \textit{`where to look'} problem and facilitates general-purpose visual attention modeling. Then, in Chapter~\ref{do_better}, we discuss several platform-specific practices for problem/model adaptation and demonstrate the benefits of our proposed solutions in meeting the real-time operating constraints. Finally, we summarize the research outcomes, offer concluding remarks, and highlight several prospective directions for future research in Chapter~\ref{con}.

\chapter{Balancing Robustness and Efficiency in Diver Following}\label{diver_following}
\begin{figure}[ht]
\centering
    \subfigure[An underwater robot is following a diver.]{
        \includegraphics[width=0.44\textwidth]{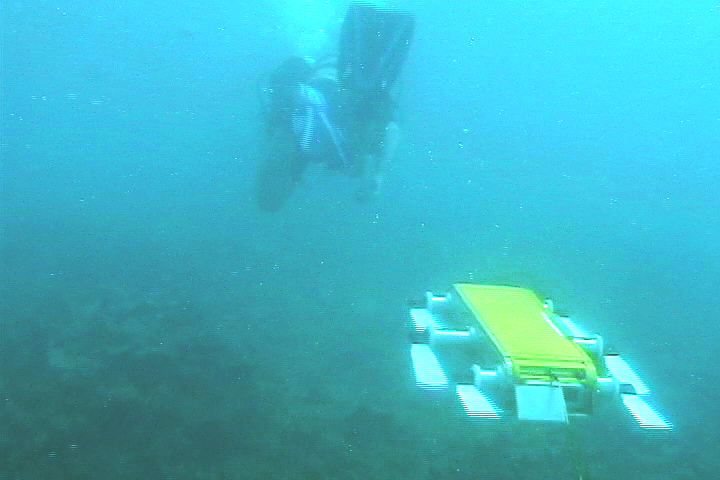}
        \label{fig:2.1a}
    }
    \subfigure[A diver seen from the robot's camera.]{
        \includegraphics[width=0.405\textwidth]{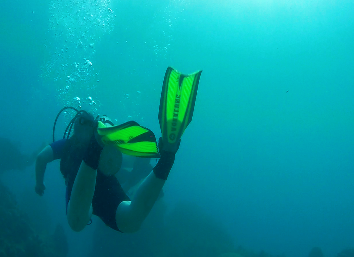}
        \label{fig:2.1b}
    }
\caption{A typical diver-following scenario during a reef exploration task.}
\label{fig:2.1}
\end{figure}

% introduce the problem
In the last chapter, we introduced various operational setups and applications for underwater human-robot cooperative missions. We discussed how a team of human divers and autonomous robots cooperatively perform a common task, while the robots follow and interact with their companion divers at certain stages of the mission~\cite{islam2018person,sattar2007your}. Although following a diver is not the primary objective in these applications, it significantly reduces operational overhead by eliminating the need for complex mission planning \textit{a priori}. Without sacrificing the generality, we consider a single-robot single-diver setting where an Autonomous Underwater Vehicle (AUV) follows its companion diver during cooperative tasks; a sample scenario is illustrated in Figure~\ref{fig:2.1}.

%\subref{fig:2.1a}, \subref{fig:2.1b}

For visually-guided underwater robots, the computational challenges lie in robust visual diver detection, fast on-board tracking, and the generation of smooth motion trajectories for autonomous following. With the focus on designing improved perception solutions, we identify several issues and limitations of existing model-based and model-free approaches (which we discuss elaborately in Section~\ref{ddd_related}). In particular, divers' appearances and motion signatures to a follower robot vary greatly based on their swimming styles, choices of wearables, and relative orientations in the six-degrees-of-freedom (6-DOF) environment. Due to this immense variability, classical model-based detection and tracking algorithms fail to achieve good generalization performance~\cite{sattar2006performance, islam2018person}. The deep visual tracking-by-detection approaches can overcome these challenges by learning complex appearance-based models from large-scale and comprehensive data~\cite{shkurti2017underwater}. They are also considerably more robust to noise and image distortions compared to model-free algorithms, which are prone to target drift in noisy visual conditions. However, the deep visual models are often computationally demanding, hence require meticulous design and efficient implementations for single-board deployments.

In this chapter, we explore the design and development of two efficient algorithms for autonomous diver following. The first algorithm,
named \textbf{Mixed-Domain Periodic Motion (MDPM)} tracker~\cite{islam2017mixed}, combines spatial and frequency domain features to track a diver's swimming motion through image sequences over time. Specifically, we formulate a sptaio-temporal optimization problem to track the flippers' oscillating motion of a diver along a sequence of non-overlapping image regions. We then deploy Hidden Markov Model (HMM)-based search-space reduction, followed by frequency-domain filtering to find the optimal motion direction. The HMM-based pruning step ensures efficient computation by avoiding large search spaces, whereas the frequency-domain detection allows accurate detection of the diver's motion direction. More details of the algorithmic formulation are provided in Section~\ref{mdpm}.

By further consideration of the inherent challenges for underwater visual tracking in diverse real-world settings, we formulate the desired capabilities of a generic diver-following algorithm: \textit{i}) invariant to the color of divers' appearance/wearables, \textit{ii}) invariant to divers' relative motion and orientation, \textit{iii}) robust to noise and image distortions, and \textit{iv}) reasonably efficient for real-time deployments. We attempt to accommodate these capabilities in the second algorithm, by designing an architecturally simple \textbf{Convolutional Neural Network (CNN)-based model} for diver detection. Each building block of the model is fine-tuned to balance the trade-off between robustness and efficiency for a single-board setting under real-time constraints. This trade-off is extensively analyzed and compared with the performance of several state-of-the-art (SOTA) deep visual object detection models as well. The underlying methodologies and design choices are presented in Section~\ref{ddd}.

Finally, we validate the effectiveness of both these models through several field experiments in open-water and closed-water environments, \textit{i.e.}, in oceans and pools, respectively. We demonstrate the real-time tracking performances and analyze their operational feasibility in Section~\ref{servo}.

\section{Background and Related Work}\label{ddd_related}
\begin{figure}[h]
\vspace{4mm}
\centering
    \includegraphics[width=0.65\linewidth]{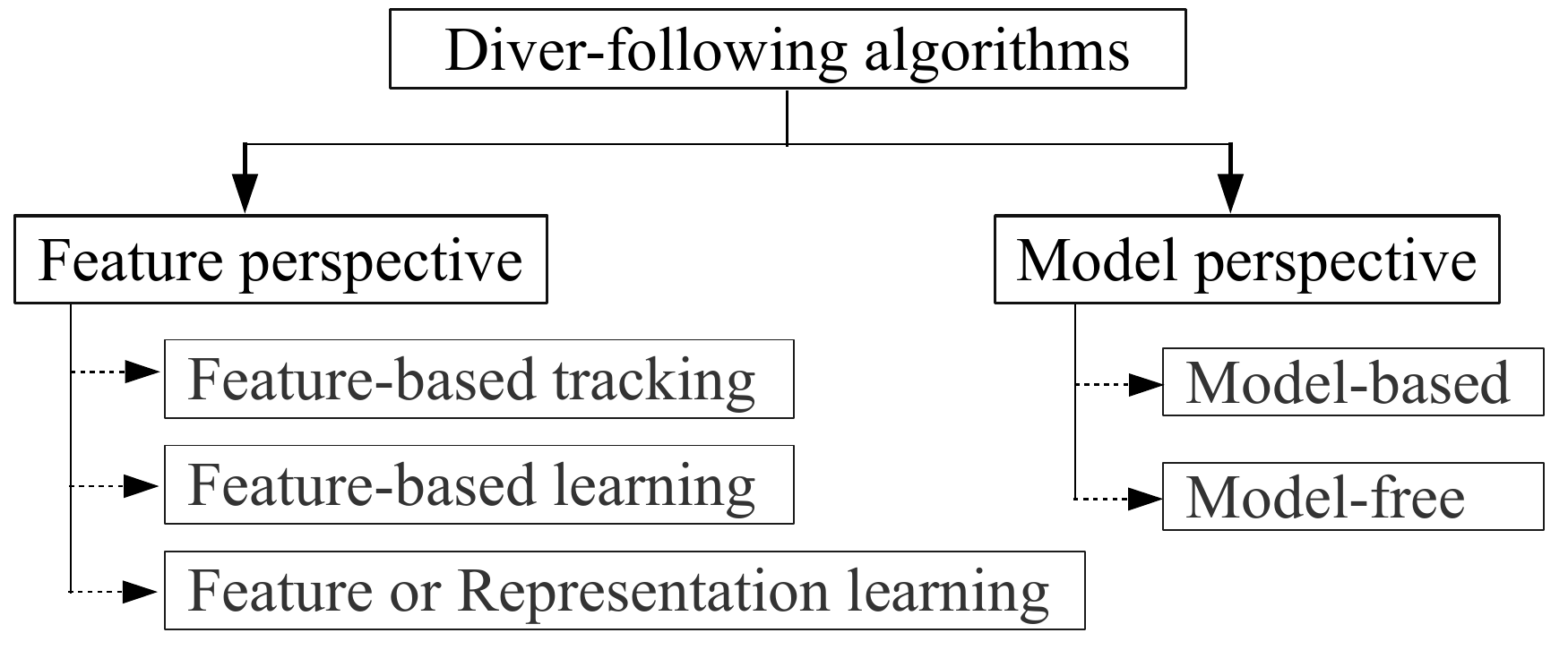}
\caption{An algorithmic categorization of the visual perception techniques used for autonomous diver following.}
\label{fig:perc}
\end{figure}

A categorization of the visual perception techniques that are commonly used for autonomous diver following is illustrated in Figure~\ref{fig:perc}. Based on algorithmic usage of the input features, these techniques can be classified as feature-based tracking, feature-based learning, or representation learning algorithms. On the other hand, they can be categorized as model-based or model-free techniques based on whether or not any prior knowledge about the appearance or motion of the diver is used for tracking. Our discussion is organized based on the feature perspective since it is more relevant to the design of our proposed algorithms. Nevertheless, various aspects of using divers' appearance and motion models are also included in the discussion.

%Robust underwater visual perception is generally  challenging due to marine artifacts such as poor visibility, variations in illumination, suspended particles, etc. Color distortion and scarcity of salient visual features make it harder to accurately detect and follow a diver in arbitrary directions. Moreover, divers' appearances to the robot vary greatly based on their swimming styles, choices of wearables, and relative orientations with respect to the robot. Challenging operating conditions require two major characteristics of a perception algorithm: robustness to noisy visual data and fast run-time with limited on-board resources. 
%Consequently, state-of-the-art approaches focus more on robustness and fast running-time than on the accuracy of perception. 

%\subsection{Feature Perspective}
Due to the operational simplicity and fast run-time, simple feature-based trackers~\cite{sattar2006performance,sattar2005visual} are often practical choices for autonomous diver following. For instance, color-based trackers localize a diver in the image space by performing binary image thresholding based on the color of their flippers or wearables. The binary image is then refined to track the centroid of the target (color blob features) by using standard algorithms such as mean-shift~\cite{comaniciu2002mean} or particle filters~\cite{gustafsson2002particle}. Optical flow~\cite{inoue1992robot}-based methods are also utilized for tracking a diver's motion from one image-frame to another~\cite{islam2018person}. Although these techniques provide reasonable tracking performance, they are prone to \emph{target drift} which is caused by the accumulation of detection errors over time, particularly in noisy visual conditions such as underwater. The tracking instability can be considerably reduced by incorporating a motion model based on human swimming cues in the frequency domain. Sattar~\etal~\cite{Sattar09RSS} showed that intensity variations in the spatio-temporal domain caused by a diver's swimming gait tend to generate high-energy responses in the $1$-$2$ Hz frequency range. This inherent periodicity can be used as a cue for robust tracking in noisy visual conditions. We generalize this idea by fusing multi-domain features to track arbitrary swimming directions of a diver; we elaborately discuss this in Section~\ref{mdpm}.

Another class of approaches uses machine learning techniques to approximate the true underlying function that relates the input features to the exact location of a diver in image space. In particular, the ensemble learning methods such as Adaptive Boosting (AdaBoost) is widely used for reliable diver tracking~\cite{sattar2009robust}. AdaBoost learns a \emph{strong} tracker from a large pool of \emph{weak} feature-based trackers that identify simple cues pertaining to divers' presence in the image. A family of such ensemble methods has been investigated as they are known to be computationally inexpensive yet highly accurate in practice. However, these methods often fail to achieve good generalization performance beyond the training data~\cite{islam2018understanding}. 
On the other hand, although Histogram of Oriented Gradients (HOG) features are used to train Support Vector Machines (SVMs) for robust human detection~\cite{dalal2005histograms} in numerous person-following systems, their applicability for diver following has been limited~\cite{islam2018person}. 
This is mostly because the divers' non-upright body-parts are only partially visible to a robot's camera (from behind or sideways), which is not ideal for HOG-based feature computation.

In recent times, the CNN-based deep visual models have been applied effectively in diver-following applications~\cite{islam2018understanding,shkurti2017underwater}. These models learn a hierarchical feature representation in image space, which significantly improves the generalization performance compared to using hand-crafted features. Once trained with sufficient data, they are quite robust to occlusion, noise, and outliers. Despite robust performance, the applicability of deep visual models in real-time robotic systems is often limited due to their slow run-time on embedded devices. Hence, the trained models are typically quantized and/or pruned to get faster inference which considerably limits their accuracy. We investigate this performance trade-off for several SOTA object detectors, and 
subsequently design a CNN-based model that offers robust detection performance in addition to ensuring that the real-time operating constraints are met; we present this model and relevant discussions in Section~\ref{ddd}.

%\subsection{Model Perspective}
%In model-free algorithms, no prior information about the target (\eg, diver's motion model, color of wearables, etc.) is used for tracking. These algorithms are initialized arbitrarily and then iteratively learn to track the target in a semi-supervised fashion~\cite{yu2008online}. TLD (tracking-learning-detection)  trackers~\cite{kalal2012tracking} and optical flow-based trackers~\cite{islam2018person} are the most commonly used model-free algorithms for general object tracking. The TLD trackers train a detector using positive and negative feedback that are obtained from image-based features. 
%In contrast, 

%On the other hand, model-based algorithms use prior knowledge about the divers' motion and appearances in order to formulate a model in the input feature-space. Iterative search methods are then applied to find the target model in the feature-space~\cite{sattar2007your}. Machine learning techniques are also widely used to learn the diver-specific features~\cite{islam2018person,sattar2009robust} and predict the target location in the feature-space. Performance of the model-free algorithms depend on comprehensiveness of the model descriptors and the underlying input feature-space. Hence, they require careful design and thorough training processes to ensure good tracking performance.

\section{Mixed-Domain Periodic Motion (MDPM) Tracker}\label{mdpm}
\begin{figure}[h]
\vspace{2mm}
 \centering
    \includegraphics[width=0.98\linewidth]{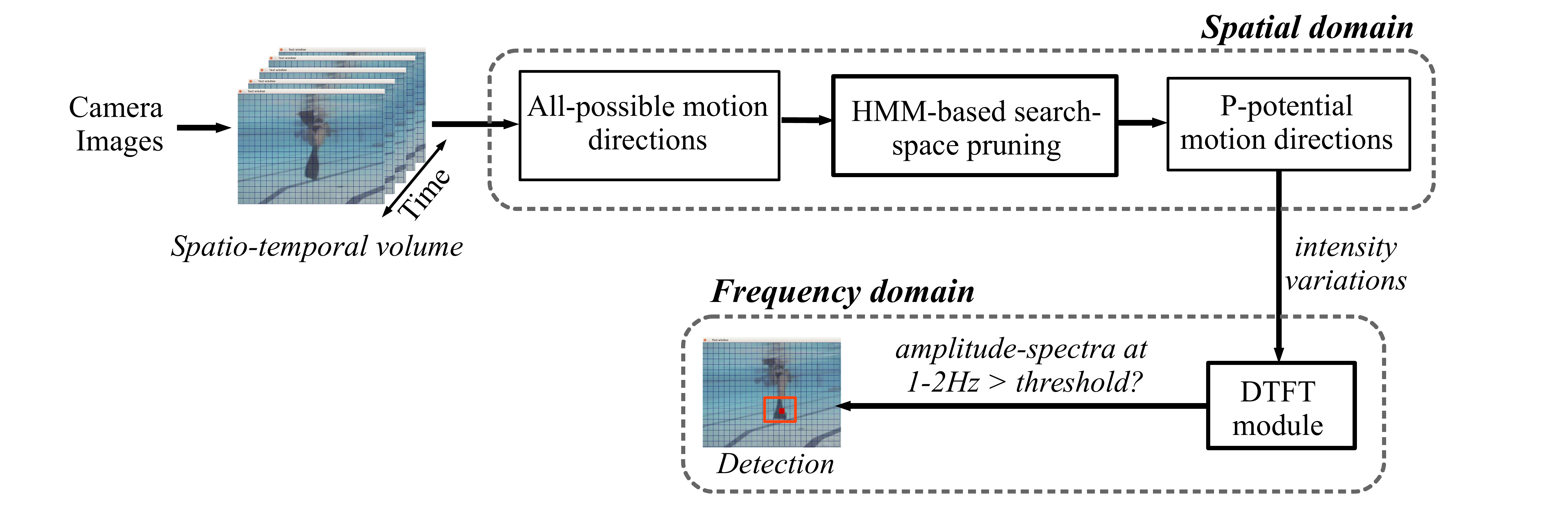}%
 \caption{Outline of the proposed MPDM tracker~\cite{islam2017mixed}.}
 \label{fig:mdpm}
\end{figure}

The proposed MDPM tracker~\cite{islam2017mixed} uses both spatial domain and frequency domain features to track human swimming motion in spatio-temporal volume. As illustrated in Figure~\ref{fig:mdpm}, the motion direction of a diver is modeled as a sequence of non-overlapping image regions over time, and it is quantified by the corresponding vector of intensity values. A HMM-based pruning method exploits these intensity values to track a set of promising motion directions. The potentially optimal motion directions are then validated based on their frequency domain signatures~\cite{Sattar09RSS}, \ie, high energy responses in the $1$-$2$ Hz frequency bands.

\begin{figure}[ht]
\centering
\includegraphics[width=0.7\linewidth]{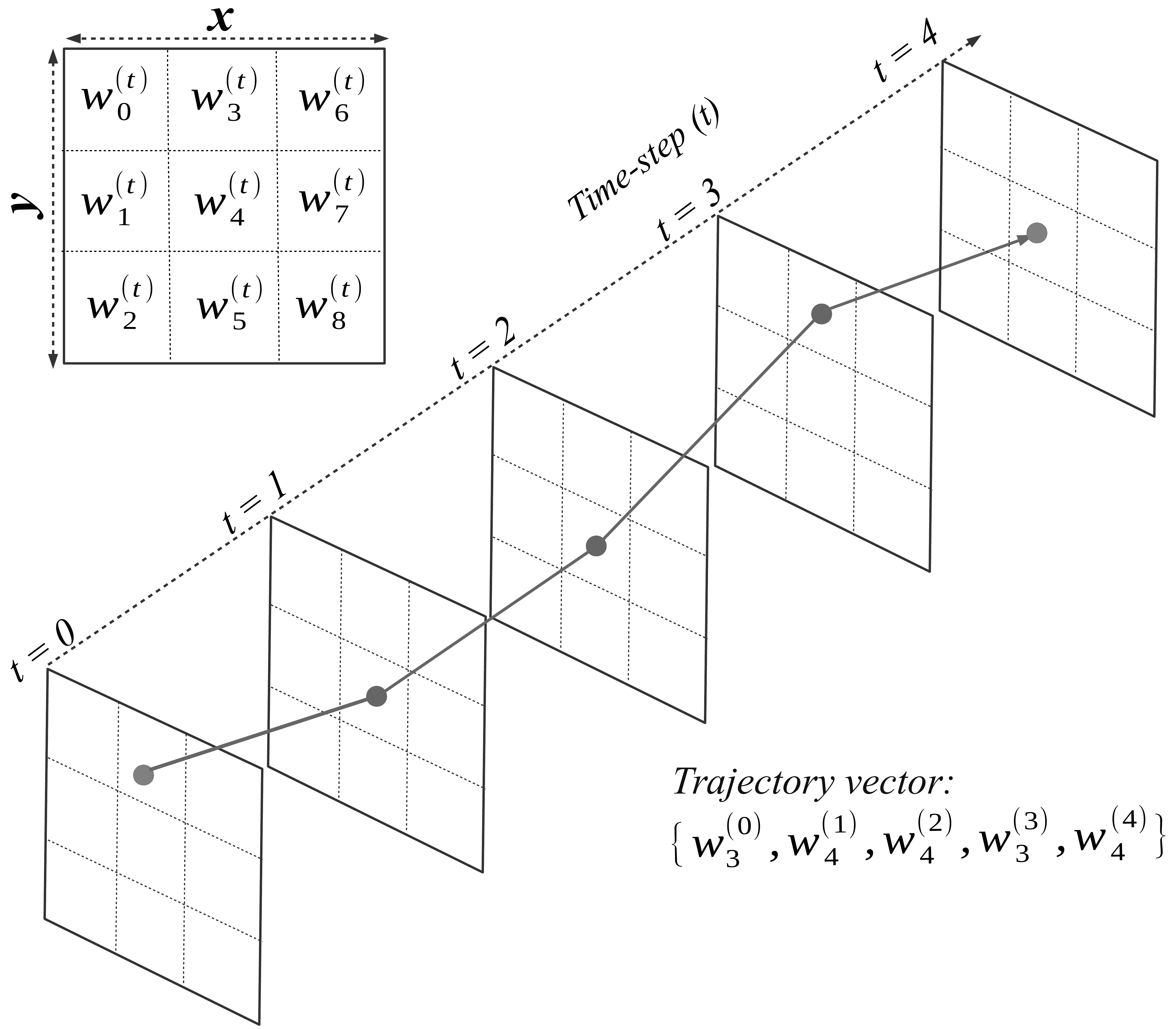}
\caption{A simple scenario with the image space divided into $M$=$9$ windows is illustrated; one particular motion direction is shown, where the corresponding trajectory vector for $T$=$5$ time-steps is $v$ $=$ $\{w^{(0)}_{3}, w^{(1)}_{4}, w^{(2)}_{4}, w^{(3)}_{3}, w^{(4)}_{4}\}$.}
\label{window}
\end{figure}

\subsection{Modeling Motion Directions}
First, the image frame at time-step $t$ is divided into a set of $M$ non-overlapping rectangular windows labeled as $w^{(t)}_0, w^{(t)}_1, \dots, w^{(t)}_{M-1}$. As demonstrated in Figure \ref{window}, the motion directions are quantified as vectors of the form $v$ $=$ $\{ w^{(0)}_{i}, w^{(1)}_{i}, \dots, w^{(t)}_{i}, \dots, w^{(T-1)}_{i}\}$. Here, $T$ stands for the \emph{slide-size} and $w^{(t)}_{i}$ denotes one particular window on the $t^{th}$ frame with $i \in [0,M-1]$ and $t \in [0,T-1]$. We call $v$ the \emph{trajectory vector}. Now, let $x_v$ denote the \emph{intensity vector}, \ie, the sequence of Gaussian-filtered averaged intensity values corresponding to the trajectory vector $v$. 
We interpret this sequence of $T$ numbers in $x_v$ as values of a discrete aperiodic function defined on $t$. This interpretation allows us to take the Discrete-Time Fourier Transform (DTFT) of $x_v$ and get a $T$-periodic sequence of complex numbers which we denote by $X_{v}$. The values of $X_{v}$ represents the discrete frequency components of $x_v$ in the frequency domain. The standard equations~\cite{Oppenheim96} that relate the spatial and frequency domains through a Fourier Transform are:
{\footnotesize
\begin{align}
X_{v} [k] &= \sum_{t=0}^{T-1}{x_v[t] e^{-j 2 \pi t k / N  }}, \qquad (k \in [0, N-1]); \text{ and}     \\
x_v[t] &= \frac{1}{N} \sum_{k=0}^{N-1}{X_{v} [k] e^{j 2 \pi t k / N  } }. 
\label{DFT}
\end{align}
}

As mentioned earlier, we try to capture the periodic motion of the diver in $x_v$ by keeping track of the variations of intensity values along $v$. Then, we take the DTFT of $x_v$ to inspect its energy responses for the discrete frequency components. The flippers of a human diver typically oscillate at $1$-$2$ Hz frequencies \cite{Sattar09RSS}. Hence, our goal is to find the motion direction $v$ for which the corresponding intensity vector $x_v$ produces maximum amplitude-spectra within $1$-$2$ Hz in its frequency domain ($X_{v}$). Therefore, if $\digamma(v)$ is the function that performs DTFT on $x_v$ to generate $X_{v}$ and subsequently finds its energy responses in $1$-$2$ Hz bands, we can formulate the following optimization problem by predicting the motion direction of a diver as:
\begin{equation}
v^* = \argmax_{v}{\digamma(v)}. 
\label{Opti}
\end{equation}

The search-space under consideration for this optimization is of size $M^T$, as there are $M^T$ different trajectory vectors over $T$ time-steps. Performing $O(M^T)$ computations for a single detection is computationally too expensive for real-time implementation. Besides, a large portion of all possible motion directions is irrelevant due to the limited body movement capabilities of human divers. Consequently, we adopt a search-space pruning mechanism to eliminate these infeasible solutions. 

\begin{figure}[h]
\vspace{2mm}
\centering
\includegraphics [width=0.52\linewidth]{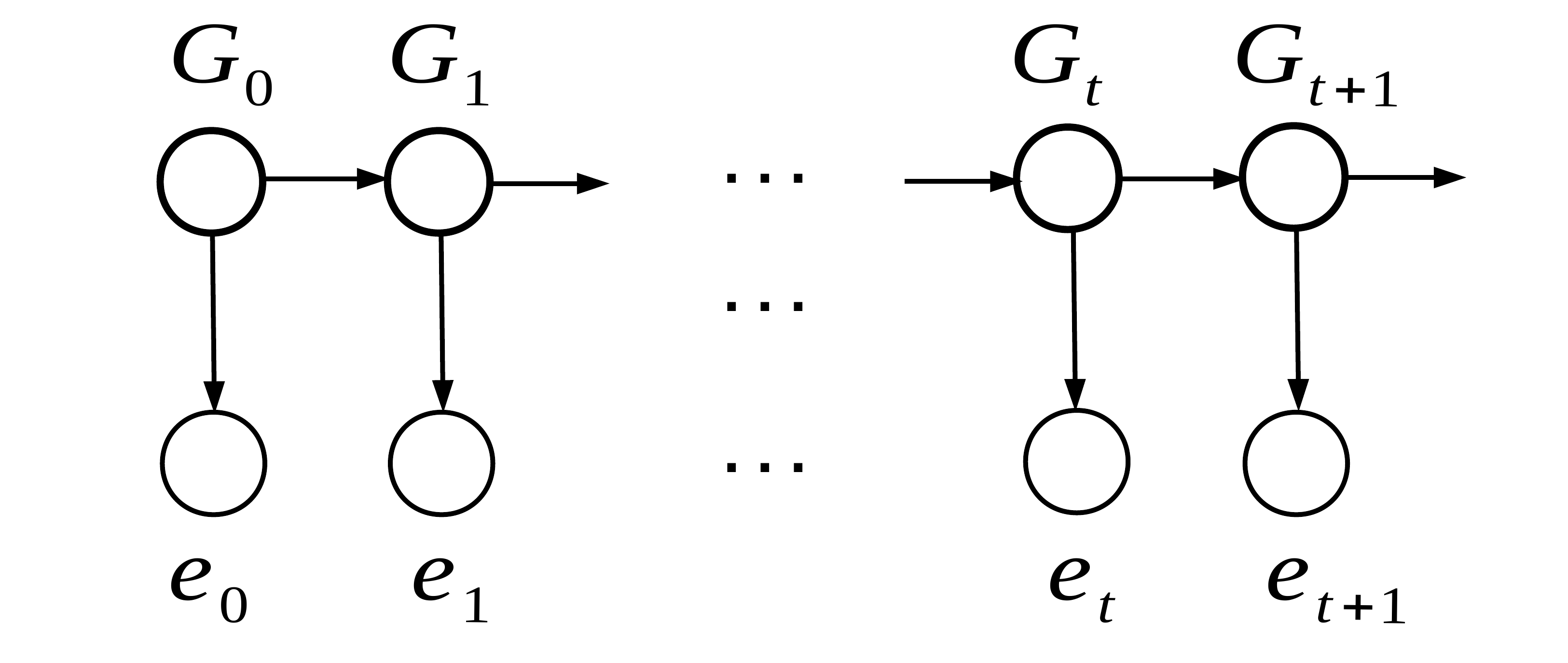}
\caption{An HMM-based representation: the observed states $e_t$ are \emph{evidence vectors} containing intensity values for $w^{(t)}_{i}$, $i \in [0, M-1]$, whereas the hidden states $G_t$ represent the probabilities that $w^{(t)}_{i}$ contains a diver's body-parts or flippers.}
\label{HMM}
\end{figure}

\subsection{HMM-based Search-space Pruning}
We have discussed that the periodic variations of intensity values (transformed into the
frequency domain) contain identifiable signatures for a potential swimming direction of the diver. 
Besides, the raw intensity values of an image region can suggest whether it contains the diver's body-parts/flippers or only waterbody background. 
In particular, we exploit the prior knowledge of the color discrepancies between the diver's flippers and waterbody to set an intensity range $R$ such that the probability of diver's flippers being present in a window $w^{(t)}_{i}$ can be defined as: 
\begin{equation}
\footnotesize
P\{ G_{t} = w^{(t)}_{i} | e_t \} \propto {1}/{Dist(I(w^{(t)}_{i}), R)}.
\end{equation}
Here, $e_t$ is the \emph{evidence vector} that contains intensity values for window $w^{(t)}_{i}$, whereas $Dist(I(w^{(t)}_{i}), R)$ measures the numeric distance between $R$ and the intensity of $w^{(t)}_{i}$. As depicted in Figure \ref{HMM}, we define a HMM representation by considering $G_t$ as the \emph{hidden} state (as we want to predict which windows contain the diver's flippers) and $e_t$ as the \emph{observed} state at time-step $t$. Additionally, we consider it unlikely that the diver's flippers will move too far away from a given window in a single time-step. Based on these assumptions, we define the following Markovian transition probabilities:
\begin{equation}
\footnotesize
\begin{gathered}
P\Big\{G_{t+1}=w^{(t+1)}_{i} \Big| G_{0}=w^{(0)}_{i}, G_{1}=w^{(1)}_{i}, \dots, G_{t}=w^{(t)}_{i}  \Big\} \\
\hspace{15mm} = P\Big\{ G_{t+1} = w^{(t+1)}_{i} \Big| G_{t} = w^{(t)}_{i}  \Big\}  \propto {1}/{Dist(w^{(t+1)}_{i}, w^{(t)}_{i} )};
\end{gathered}
\label{Eq:pair}
\end{equation} 
\begin{equation}
\footnotesize
P\Big\{e_{t} \Big| G_{t}=w^{(t)}_{i}  \Big\} =
   \begin{cases}
       1-\epsilon &\quad\text{if } I(w^{(t)}_{i}) \in R  \\
        \epsilon &\quad\text{otherwise.} \ 
     \end{cases}
 \label{Eq:evidence}
\end{equation} 

In our implementation, we take $\epsilon=0.1$, and adopt an intensity-based range $R$ to define $P\{ G_{t} = w^{(t)}_{i} | e_t \}$; color-based ranges (in RGB-space or HSV-space) can also be adopted instead. One advantage of using an intensity-based range is that the intensity values are already available in the trajectory vector, hence no additional computation is needed. We use this setup to predict \emph{the most likely sequence of states} ($G_{0}, \dots G_{T-1}$) that leads to a given state $G_{T}$. In terms of the parameters and notations mentioned above, this is defined as:
\begin{equation}
\footnotesize
\mu ^*(T) = \argmax_{w^{(0)}_{i}, \dots, w^{(T-1)}_{i}}{ P \Big\{ G_{0}=w^{(0)}_{i}, \dots, G_{T}=w^{(T)}_{i} \Big| e_{0}, \dots, e_{T}  \Big\}.} 
\end{equation} 

Now, using the properties of the Bayesian chain rule and Markovian transition~\cite{rabiner1989tutorial}, a recursive formulation of $\mu ^*(T)$ can be obtained as follows:
\begin{equation}
\footnotesize
\begin{aligned}
\mu ^*(T) &= P \Big\{ e_T \Big| G_{T}=w^{(T)}_{i} \Big\}  \times { \argmax_{w^{(T-1)}_{i}}{ \Big( P\Big\{ G_{T}=w^{(T)}_{i} \Big| G_{T-1}=w^{(T-1)}_{i}  \Big\} } } \times { \mu ^*(T-1)  \Big)  }.
\end{aligned}
\label{Recur}
\end{equation}
The derivation is provided in Appendix~\ref{ApenA1}. Using this recursive definition of $\mu ^*(T)$, we can efficiently keep track of the most likely sequence of states over $T$ time-steps, which is essentially the desired trajectory vector. However, a pool of such trajectory vectors is needed so that we can avoid outliers and false positives by inspecting their frequency responses. Hence, we choose the $p$ most likely sequences of states $\mu^{*}(T,p)$, where $p$ is the \emph{pool-size}. Finally, we rewrite the problem definition in Equation \ref{Opti} as follows:
\begin{equation}
v^* = \argmax_{\mu \in \mu^{*}(T,p)}{\digamma(\mu)}. 
\label{FinalOpti}
\end{equation}

\subsection{Frequency Domain Validation}
We now summarize the procedure for finding $v^*$ at each detection cycle. First, we find the $p$ most potential motion directions (\ie, trajectory vectors) through the HMM-based pruning mechanism discussed above. We do this efficiently by using the notion of \emph{dynamic programming}; it requires $\mathcal{O}(M^2)$ operations to update the dynamic table of probabilities. Once the potential trajectory vectors are found, we perform DTFT to observe their frequency domain responses. The trajectory vector producing the highest amplitude-spectra at $1$-$2$ Hz frequencies is selected as the optimal solution. DTFT can be performed very efficiently as well; for instance, the run-time of a Fast Fourier Transform algorithm is $\mathcal{O}(T\times logT)$. Therefore, we need only $\mathcal{O}(p \times T\times logT)$ operations for inspecting all potential trajectory vectors. Additionally, the approximated location of the diver is readily available in the solution, hence no additional computation is required for bounding-box (BBox) localization.

\subsection{Experimental Setup and Evaluation}
The MDPM tracker has three hyper-parameters: the slide-size ($T$), the window size, and the amplitude threshold ($\delta$) in the frequency domain. We empirically determine their values through extensive simulations on video footage of diver following~\cite{islam2017mixed}. We found that $T$$=$$15$ and a window size of $30$$\times$$30$ work well in practice; also, we set the frequency threshold $\delta$$=$$75$. Once the bootstrapping is done with the first $T$ frames, mixed-domain detection is performed at every frame onward in a sliding-window fashion. At each detection, the tracker estimates $p$ trajectory vectors that represent a set of potential motion directions in spatio-temporal volume. If a motion direction produces amplitude-spectra more than $\delta$, it is reported as a positive detection, and the diver's flippers are subsequently located in the image frame. 

\begin{figure}[ht]
\centering
    \subfigure[For a closed-water experiment, the swimming trajectory of a diver is visualized using a surface-plot; it is prepared by projecting the detected trajectory vectors to the spatio-temporal volume.]{
        \includegraphics[width=0.49\textwidth]{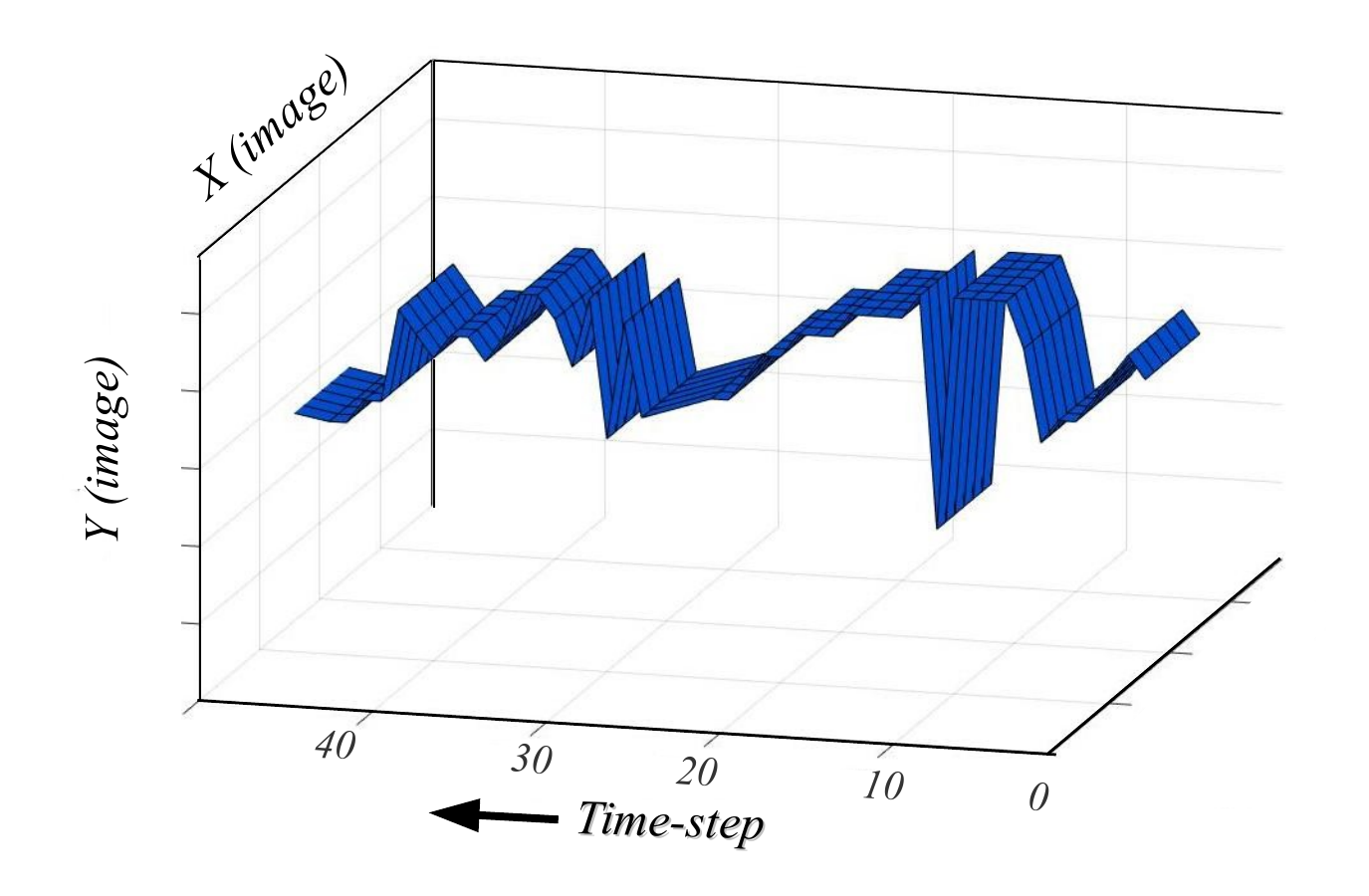}%
        \label{fig:mdpm_a}
    }\hspace{2mm}
    \subfigure[Corresponding frequency domain signatures are shown; each dotted line represents the amplitude spectra for a single detection in low frequency bands.]{
        \includegraphics[width=0.44\textwidth]{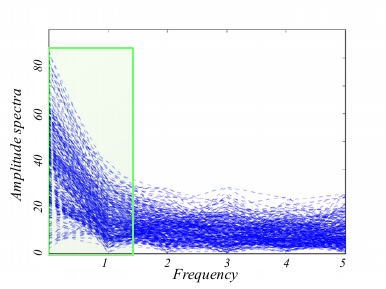}%
        \label{fig:mdpm_b}
    }
    
    \subfigure[Detection of a diver's flippers in different scenarios: swimming straight-on away from the robot and swimming sideways (both in closed-water and open-water conditions).]{
        \includegraphics[width=0.99\textwidth]{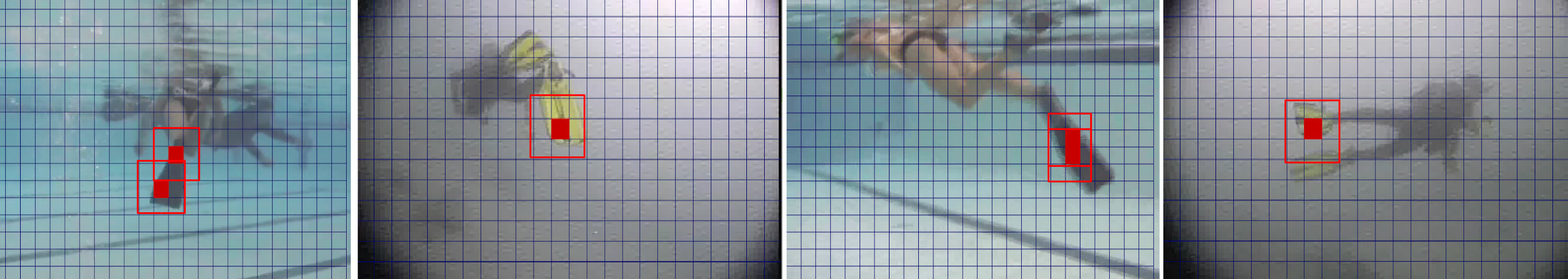}%
        \label{fig:mdpm_c}
    }
\caption{Experimental validations of the MDPM tracker.}
\label{fig:mdpm_res}
\end{figure}
%\subref{fig:2.1a}, \subref{fig:2.1b}

Figure~\ref{fig:mdpm_res} demonstrates how the MDPM tracker combines spatial domain features and frequency domain motion cues for effective detection of a diver's flippers. It keeps track of the diver's motion direction through a sequence of $30$$\times$$30$$\times$$15$ slices in the spatio-temporal volume. The corresponding surface through the image space over time mimics the actual motion direction of the diver, which validates its effectiveness. 
Furthermore, Table~\ref{MDPM_com} quantifies the performance of MDPM tracker in terms of correct detection and missed/wrong detection for various experimental cases. On an average, it achieves a positive detection accuracy of $84.2$-$91.7\%$, which suggests that it provides $8$-$9$ positive detection of a diver per second (considering a frame-rate of $10$\,FPS). We have found this detection rate sufficient for successfully following a diver in practice.

\begin{table}[ht]
\caption{Detection accuracy of MDPM tracker in various swimming conditions.}

\vspace{2mm}
\scriptsize
\centering
  \begin{tabular}{m{5.6cm}||c|c|c|c}
  \Xhline{2\arrayrulewidth}
    \multirow{2}{*}{Cases} &
      \multicolumn{2}{c|}{Closed Water} &
      \multicolumn{2}{c}{Open Water} \\
      \cline{2-5}
      & Straight-on & Sideways & Straight-on & Sideways \\  
      \Xhline{2\arrayrulewidth}
    Correct detection: true positives on target image windows, true negatives on the rest  & \textbf{$647$  ($91.7\%$)} & \textbf{$463$ ($87.3\%$)} & \textbf{$294$ ($85.2\%$)} & \textbf{$240$ ($84.2\%$) } \\
    \hline
    Missed detection: false negatives on target image windows  & $46$ ($6.5\%$) & $57$ ($10.8\%$) & $38$ ($11\%$) & $43$ ($15\%$)  \\
    \hline
    Wrong detection: false positives on non-target image windows & $12$ ($1.8\%$) & $10$ ($1.9\%$) & $13$ ($3.8\%$) & $2$ ($0.8\%$)  \\ 
    \Xhline{2\arrayrulewidth}
  \end{tabular}
\label{MDPM_com} 
\end{table}

\section{Deep Diver Detection (DDD)}\label{ddd}
A major limitation of the MDPM tracker is that it does not model the appearance of a diver; it only detects the periodic signals pertaining to their flippers' motion. Besides, its performance is affected by the swimming trajectory (\ie, straight-on or sideways), the color of divers' wearables, etc. 
We try to address these issues and ensure robust detection performance by designing a CNN-based model for single diver detection. We also investigate the performance and feasibility of several SOTA deep object detection models~\cite{tfzoo} for multi-diver tracking in quasi-real-time.

\begin{figure}[ht]
\centering
    \subfigure[The base model for single diver detection.]{\includegraphics [width=0.68\linewidth]{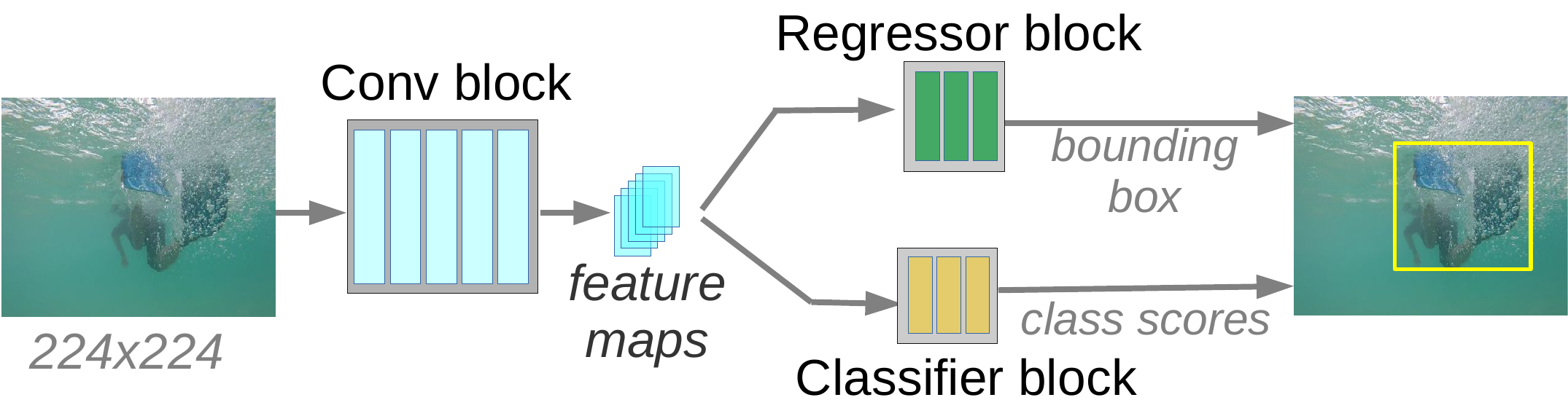}
        \label{dr_detect_single}
    }
    \vspace{2mm}
    
    \subfigure[Allowing detection of multiple divers by using a region selector named Edge-box~\cite{zitnick2014edge}.]{\includegraphics [width=0.68\linewidth]{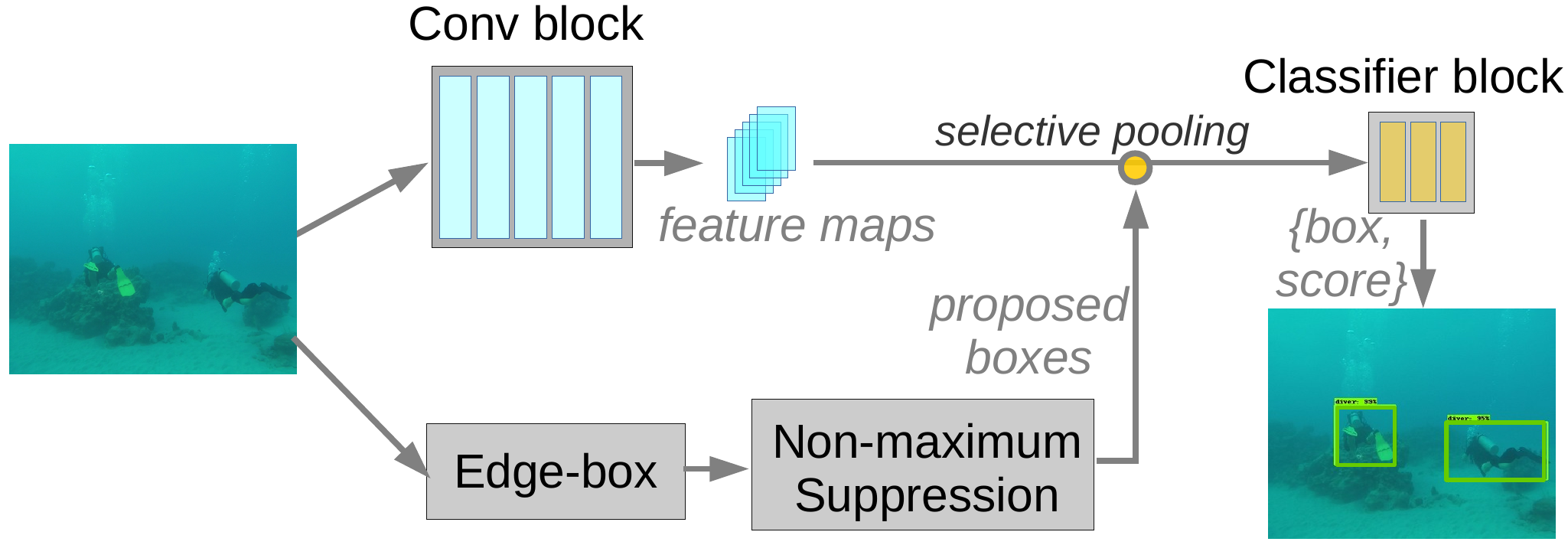}
        \label{dr_detect_mult}
    }
\caption{Schematic diagram of the proposed CNN-based diver detection model~\cite{islam2018towards}.}
\label{dr_detect}
\end{figure}

\subsection{CNN-based Model for Single Diver Detection}
A schematic diagram of the proposed CNN-based model~\cite{islam2018towards} is illustrated in Figure~\ref{dr_detect_single}. It consists of three computational components: a convolutional block, a regressor block, and a classifier block. Five sequential convolutional ({\tt Conv}) layers are used for hierarchical feature extraction. The extracted features are then fed to the classifier and regressor blocks, both consisting of three fully connected layers. 
The regressor learns to generate bounding box (BBox) proposals (one for each object category), while the classifier learns to predict their corresponding \emph{class probability} scores. 
In our implementation, we consider human divers and robots as object categories, and train this model for detecting a single diver/robot in RGB image space. Detailed network parameters and dimensions are specified in Table~\ref{tab:conv}.

%\begin{figure*}[h]
%    \centering
%    \begin{subfigure}[t]{\textwidth}
%       \centering
%        \includegraphics [width=0.65\linewidth]{figs/m1.pdf}
%        \caption{The base model for single diver detection.}
%        \label{dr_detect_single}
%        \end{subfigure}
        
%        \vspace{5mm}
%        \begin{subfigure}[t]{\textwidth}
%        \centering
%        \includegraphics [width=0.65\linewidth]{figs/m2.pdf}
%        \caption{Allowing detection of multiple divers by using a region selector named Edge-box~\cite{zitnick2014edge}.}
%        \label{dr_detect_mult}
%        \end{subfigure}
%\caption{A schematic diagram of the proposed CNN-based model for diver detection in image space.}
%\label{dr_detect}
%\end{figure*}

\begin{table}[h]
\centering
\caption{Specification of the CNN-based model outlined in Figure~\ref{dr_detect} [convolutional layers: {\tt Conv1}-{\tt Conv5}, classifier layers: {\tt FC1}-{\tt FC3}, regression layers: {\tt RC1}-{\tt RC3}; $n$: number of object categories; *an additional pooling layer is used before passing the {\tt Conv5} features-maps to {\tt FC1}].}

\vspace{2mm}
%\scriptsize
\footnotesize
\begin{tabular}{l|llll}
\Xhline{2\arrayrulewidth}
\textbf{Layer} & \textbf{Input feature-map} & \textbf{Kernel size} & \textbf{Strides}  & \textbf{Output feature-map} \\ \Xhline{2\arrayrulewidth}
{\tt Conv1} & $224\times224\times3$ & $11\times11\times3\times64$ & $[1,4,4,1]$   & $56\times56\times64$ \\

{\tt Pool1} & $56\times56\times64$ & $1\times3\times3\times1$ & $[1,2,2,1]$  & $27\times27\times64$  \\ 

{\tt Conv2} & $27\times27\times64$ & $5\times5\times64\times192$ & $[1,1,1,1]$   & $27\times27\times192$ \\

{\tt Pool2} & $27\times27\times192$ & $1\times3\times3\times1$ & $[1,2,2,1]$  &  $13\times13\times192$  \\  

{\tt Conv3} & $13\times13\times192$ & $3\times3\times192\times192$ & $[1,1,1,1]$   & $13\times13\times192$  \\  

{\tt Conv4} & $13\times13\times192$ & $3\times3\times192\times192$ & $[1,1,1,1]$   & $13\times13\times192$  \\  

{\tt Conv5} & $13\times13\times192$ & $3\times3\times192\times128$ & $[1,1,1,1]$   & $13\times13\times128$  \\ \hline 

{\tt FC1} & $4608\times1^*$ & $-$ & $-$   & $1024\times1$  \\

{\tt FC2} & $1024\times1$ & $-$ &  $-$  & $128\times1$  \\

{\tt FC3} & $128\times1$ & $-$ &  $-$  & $n$  \\ \hline

{\tt RC1} & $21632\times1$ & $-$ & $-$   & $4096\times1$  \\

{\tt RC2} & $4096\times1$ & $-$ & $-$   & $192\times1$  \\

{\tt RC3} & $192\times1$ & $-$ & $-$ & $4n$  \\ \Xhline{2\arrayrulewidth} 
\end{tabular}
\label{tab:conv}
\end{table}

\textbf{Design Intuition:} The SOTA deep visual models are designed for generic applications for a large number of object categories. However, for most underwater human-robot collaborative applications including diver-following, only a few object categories (\textit{e.g.}, diver, robot, fish, coral reef, etc.) are relevant. We take advantage of this by designing an architecturally simple model that ensures fast run-time on embedded platforms in addition to providing robust detection performance. As Table~\ref{tab:conv} demonstrates, the proposed model incorporates a set of shallow feature extraction layers and uses a sparse regressor block for object localization rather than using a computationally expensive Region Proposal Network (RPN)~\cite{renNIPS15fasterrcnn,Girshick2014RCNN_CVPR}. The idea is to keep the network computationally light to get high inference rates for single instance object detection as \emph{single robot single diver} interaction scenario is most commonly adopted in practice.

\subsection{Allowing Multiple Detections}
Although following a single diver is the most common diver-following scenario, detecting multiple divers and other objects can be useful in many applications. As shown in Figure~\ref{dr_detect_mult}, we can add multi-object detection capabilities in the proposed model by replacing the regressor with a region selector. In our implementation, we use the SOTA class-agnostic region selector named Edge-box~\cite{zitnick2014edge}. Edge-box utilizes the image-level statistics like edges and contours to measure \emph{objectness scores} in various prospective regions in the image space. The BBox generated by Edge-box are filtered based on their objectness scores and then non-maxima suppression techniques are applied to get the dominant ones in the image space. The corresponding feature maps are then fed to the classifier block to predict their object categories. Although we need additional computation for Edge-box, it runs independently and in parallel with the convolutional block; hence, the overall pipeline is still considerably faster than if we were to use an RPN-based object detection model.

\subsection{SOTA Object Detectors}\label{tf_obj_detect}
Furthermore, we exploit the SOTA deep object detectors to address the inherent difficulties of underwater visual detection. We use the following four models: Faster R-CNN~\cite{renNIPS15fasterrcnn} with Inception V2 \cite{szegedy2016rethinking} as a feature extractor, Single Shot MultiBox Detector (SSD)~\cite{liu2016ssd} with MobileNet V2 \cite{sandler2018inverted, howard2017mobilenets} as a feature extractor, You Only Look Once (YOLO) V2~\cite{redmon2016yolo9000}, and Tiny YOLO \cite{tinyYOLO}. These are the fastest (in terms of processing time of a single frame) among the family of current SOTA models for general object detection; we refer to~\cite{tfzoo,tinyYOLO} for detailed comparisons of their detection performance and run-time. Appendix~\ref{ApenB} briefly discusses their methodologies and the related design choices in terms of major computational components.

%\begin{figure}[ht]
%\vspace{1mm}
%\centering
%\includegraphics[width=0.98\linewidth]{figs/data-comp.pdf} 
%\vspace{-2mm}
%\caption{A few samples from the training dataset are shown; the images are annotated by bounding-box labels for two classes: \textit{i}) human divers and \textit{ii}) robots.}
%\label{fig:data_diver}
%\end{figure}

\subsection{Dataset Preparation}\label{data_col_dive}
We performed numerous diver-following experiments in pools and oceans to prepare training datasets for the deep models. Additionally, we collected data from underwater field trials performed by different research groups over the years in various locations~\cite{islam2018understanding,islam2018towards,islam2019fast}. This variety of experimental setups is crucial to ensure  comprehensiveness of the dataset so that the supervised models can learn the inherent diversity of various application scenarios. In particular, we made sure that the training data capture the following variability - 
\begin{itemize}
\item Natural variability: various waterbody and lighting conditions at varying depths.%
\item Artificial variability: data collected using different robots and cameras.%  
\item Human variability: multiple different humans and appearances, choice and variations of wearable such as suits, flippers, goggles, etc. 
\end{itemize}
We extracted the robots' camera-feed from these experiments and prepared image-based datasets for large-scale supervised training. 
There are over $20$K images in the dataset that are BBox-annotated for two object categories: human divers and robots. 
%(AUVs or ROVs). Several human participants (acknowledged later in the paper) over the period of six months.
%; the annotations include separate bounding-box labels for human divers and robots. A few sample images are provided in Figure~\ref{fig:data_diver}. 

\subsection{Model Training and Evaluation}\label{ddd_train}
We use TensorFlow~\cite{abadi2016tensorflow} and Darknet~\cite{tinyYOLO} libraries to implement the training pipelines for all the models on a Linux machine with four GPU cards (NVIDIA\texttrademark{} GTX 1080). 
For the standard models (\ie, Faster R-CNN, YOLO, and SSD), we use transfer-learning from pre-trained weights by following the recommended configurations provided with their APIs~\cite{tfzoo,tinyYOLO}. In contrast, the proposed models are trained from scratch. RMSProp~\cite{tieleman2012lecture} is used as the global optimizer with an initial learning rate of $0.001$. The standard cross-entropy and $L_2$ loss functions are used by the classifier and regressor blocks, respectively. 
The model approaches convergence with $300$ epochs of training with a batch-size of $16$. Once training is done, the frozen model is transferred to the robot CPU for validation and real-time experiments.

First, we analyze and compare the detection performance of all the models on a test set containing $2.2$K images (exclusive from the training set). We measure the detection accuracy and object localization performance based on mean average precision (mAP) and intersection over union (IoU), respectively; see Appendix~\ref{ApenObj} for their standard definitions. We also evaluate and compare inference rates of all models based on FPS (frames per second) on three different devices: NVIDIA\texttrademark{} GTX 1080 GPU,  Embedded GPU (NVIDIA\texttrademark{} Jetson TX2), and a Robot CPU (Intel\texttrademark{} Core-i3 6100U).

\begin{table}[ht]
\caption{Performance comparison of the deep diver detection models [mAP: mean average precision; IoU: intersection over union; FPS: frames per second].}

\vspace{2mm}
\footnotesize
\centering
  \begin{tabular}{l|c|c|rrr}
  \Xhline{2\arrayrulewidth}
    \multirow{2}{*}{Models} &
      \multirow{2}{*}{mAP (\%)} & \multirow{2}{*}{IoU (\%)} &
      \multicolumn{3}{c}{FPS} \\
      %\cline{4-6}
      &  & & GTX 1080 & Jetson TX2 & Robot CPU \\  
       \Xhline{2\arrayrulewidth}
       Faster R-CNN (Inception V2) & $71.10$ & $78.30$ & $17.30$ & $2.10$ & $0.52$ \\ \hline
       YOLO V2 & $57.84$ & $62.42$ & $73.30$ & $6.20$ & $0.11$ \\ \hline
       Tiny YOLO & $52.33$ & $59.94$ & $220.00$ & $20.00$ & $5.50$ \\ \hline
       SSD (MobileNet V2) & $61.25$ & $69.80$ & $92.00$ & $9.85$ & $3.80$ \\ \hline
       Proposed CNN-based Model & $53.75$ & $67.40$ & $263.50$ & $17.35$ & $6.85$ \\ \Xhline{2\arrayrulewidth}
  \end{tabular}
\label{res_com} 
\vspace{-1mm}
\end{table}

\begin{figure}[t]
\centering
\includegraphics [width=0.98\linewidth]{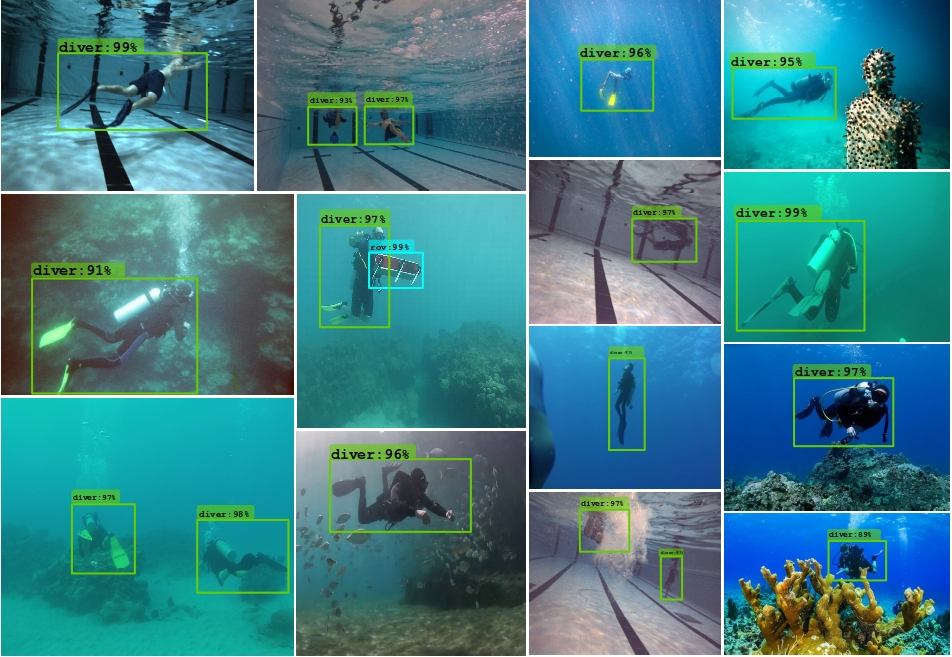}%
\vspace{-2mm} 
\caption{Snapshots of a set of diverse first-person views of robots from different diver-following scenarios. Notice the variation in appearances of the divers and possible noise or disturbances in the scene over different scenarios. The bounding-boxes and text overlaid on the images are the outputs generated by the proposed model at test time.}
\label{fig:ddd}
\end{figure}%

The quantitative performance comparison based on mAP, IoU, and FPS is illustrated in Table~\ref{res_com}. The Faster R-CNN (Inception V2) model achieves much better detection performance compared to other models although it has the slowest inference rates. On the other hand, YOLO V2, SSD (MobileNet V2), and the proposed CNN-based model provide comparable scores for mAP and IoU. Although Tiny YOLO provides fast inference, its detection performance is relatively poor. 
Moreover, the proposed CNN-based model runs at a rate of $6.85$\,FPS on the robot CPU and $17.35$\,FPS on the embedded GPU, which validate its applicability for real-time use. This fast run-time comes at a cost of losing approximately $18\%$ mAP and $11\%$ IoU compared to the Faster R-CNN (Inception V2) model. Nevertheless, this balance between robustness and efficiency is critical for ensuring a reasonable tracking performance in real-time diver-following scenarios.

Figure~\ref{fig:ddd} shows a few qualitative results for the proposed CNN-based model on real-world scenarios. 
Next, we provide the field experimental details and discuss the general applicability of this model from a practical standpoint.

\section{Field Experiments and Feasibility Analysis}\label{servo}
\begin{figure}[!h]
\vspace{2mm}
 \centering
    \includegraphics[width=0.65\linewidth]{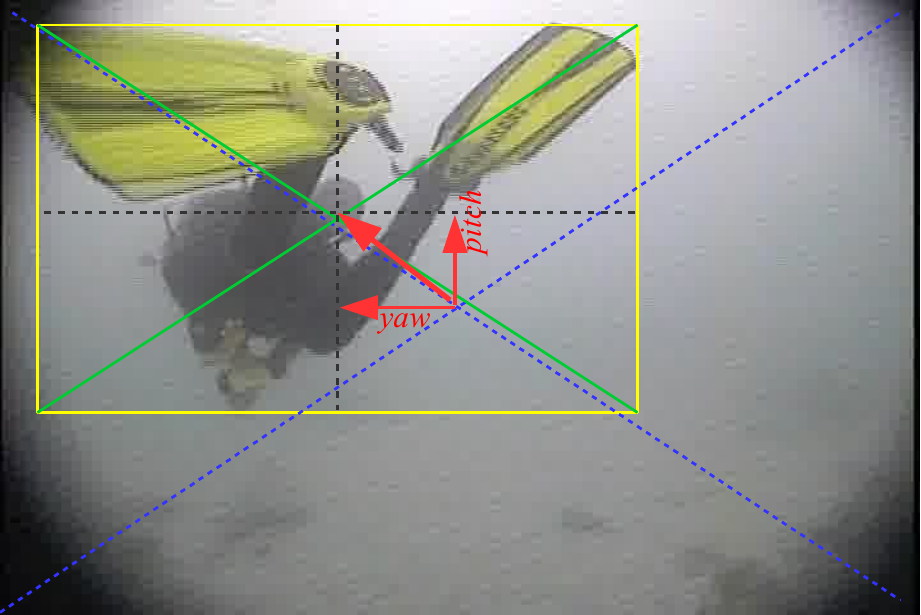}%
\vspace{-2mm}
 \caption{Illustration of how the yaw and pitch commands are generated based on horizontal and vertical displacements of the center of the detected BBox.}
 \label{fig:serv}
\end{figure}

We perform several real-world experiments both in closed-water and in open-water conditions, \ie, in pools and oceans. The Aqua MinneBot AUV~\cite{dudek2007aqua} is used for testing the diver-following modules. During the experiments, a diver swims in front of the robot in arbitrary directions; the task of the robot is to visually detect the diver using its camera feed and follow behind in a smooth motion. The following motion is enabled by a BBox-reactive visual servo controller~\cite{espiau1992new} (see Figure~\ref{fig:serv}). %It uses the BBox generated by the diver detector to regulate robot motion commands in order to follow the diver.
%As mentioned, we use the MinneBot AUV for diver following experiments, which has five DOF of control: three angular (yaw, pitch, and roll) and two linear (surge and heave) controls. 
In our implementation~\cite{islam2018understanding}, we adopt a tracking-by-detection method where the controller tries to bring the observed BBox of the target diver to the center of the robot's camera. The distance of the diver is approximated by the size of the BBox and forward velocity rates are generated accordingly. Additionally, the yaw and pitch commands are normalized based on the horizontal and vertical displacements of the observed BBox center from the image center; these navigation commands are then regulated by separate PID controllers. On the other hand, the roll stabilization and hovering are handled by the robot's autopilot module~\cite{meger20143d}. Such visual servoing is ideal for the Aqua MinneBot as it has five DOF: three angular (yaw, pitch, and roll) and two linear (surge and heave) controls.

\begin{figure}[h]
\centering
\includegraphics [width=0.98\linewidth]{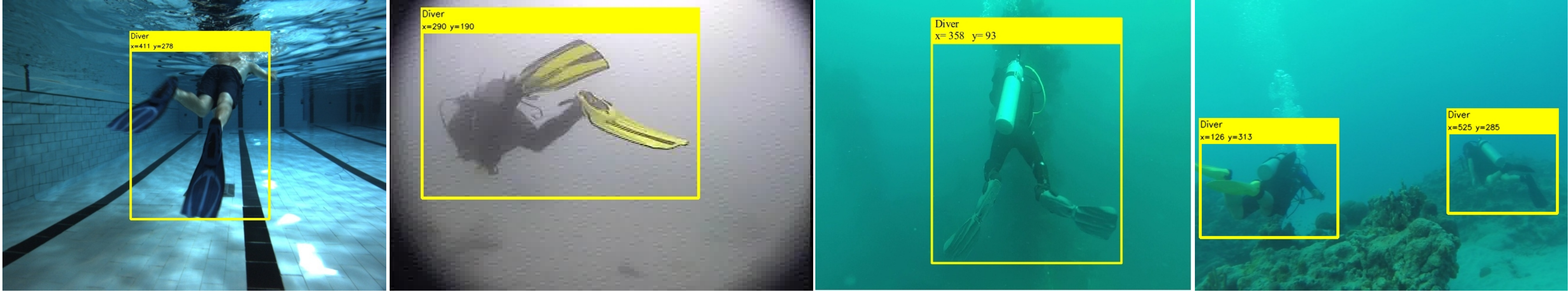}%
\vspace{-2mm}
\caption{Snapshots of a few diver-following scenarios in different field experiments; a video demonstration is available at: \protect\url{https://youtu.be/9xukzT8dqzQ}.}%
\label{fig:ddd2}
\end{figure}%

A few snapshots of the proposed model being used in various diver-following scenarios are illustrated in Figure~\ref{fig:ddd2}. Since we adopt a BBox-reactive servo control, accurate detection of the diver is essential to ensure good tracking performance. During the field experiments, we have found $6$-$7$ positive detection per second on an average, which is sufficient for successfully following a diver in real-time~\cite{islam2018understanding}. 
Moreover, the proposed model is considerably robust to occlusion and noise, in addition to being invariant to divers' appearance and wearable. Lastly, our training data include a large collection of gray-scale and color distorted underwater images; hence the proposed model is considerably robust to noise and color distortions.  

\begin{figure}[h]
\centering
    \subfigure[]{\includegraphics [width=0.475\linewidth]{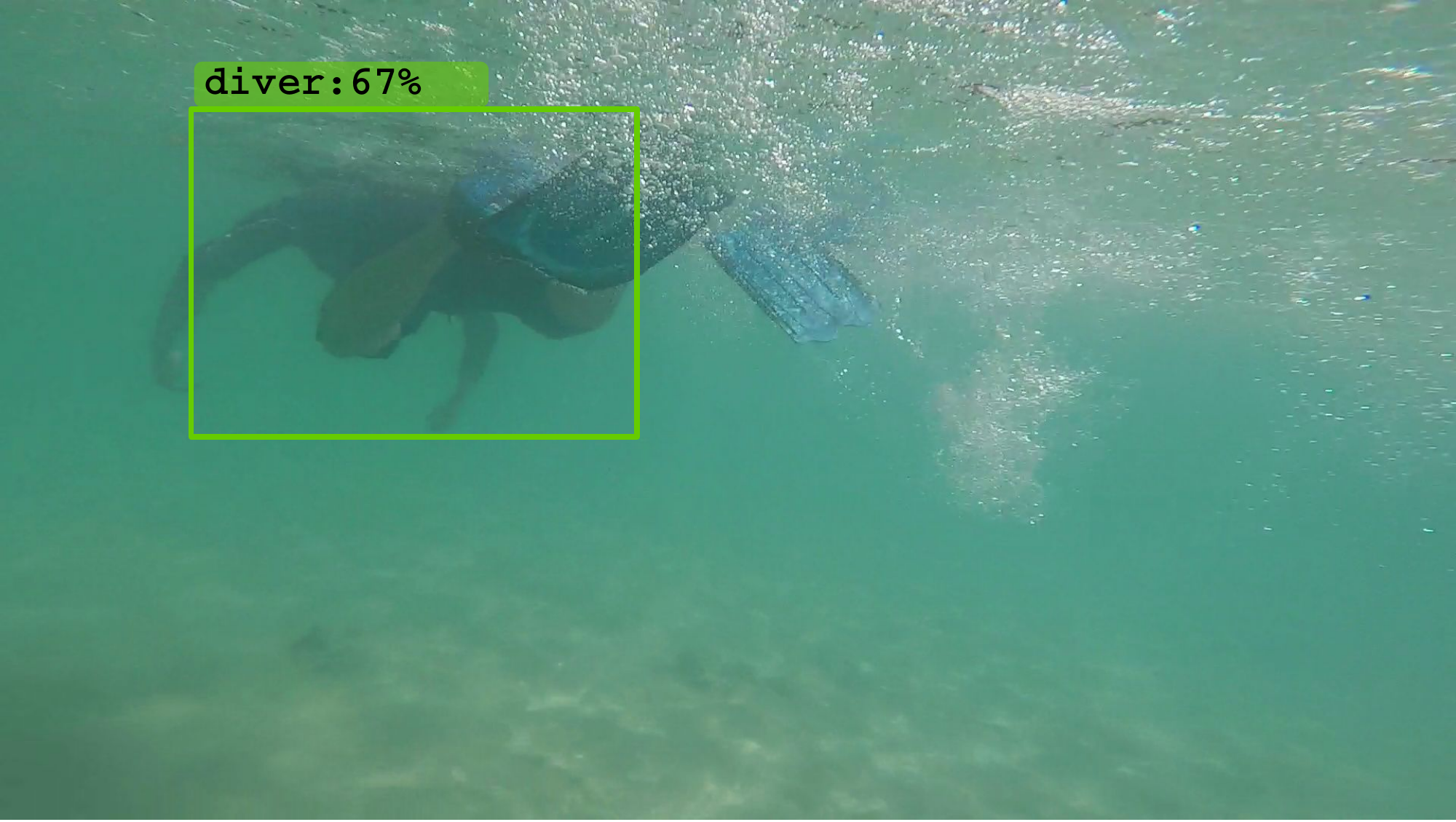}
        \label{fig:bad_a}
    }~
    \subfigure[]{\includegraphics [width=0.36\linewidth]{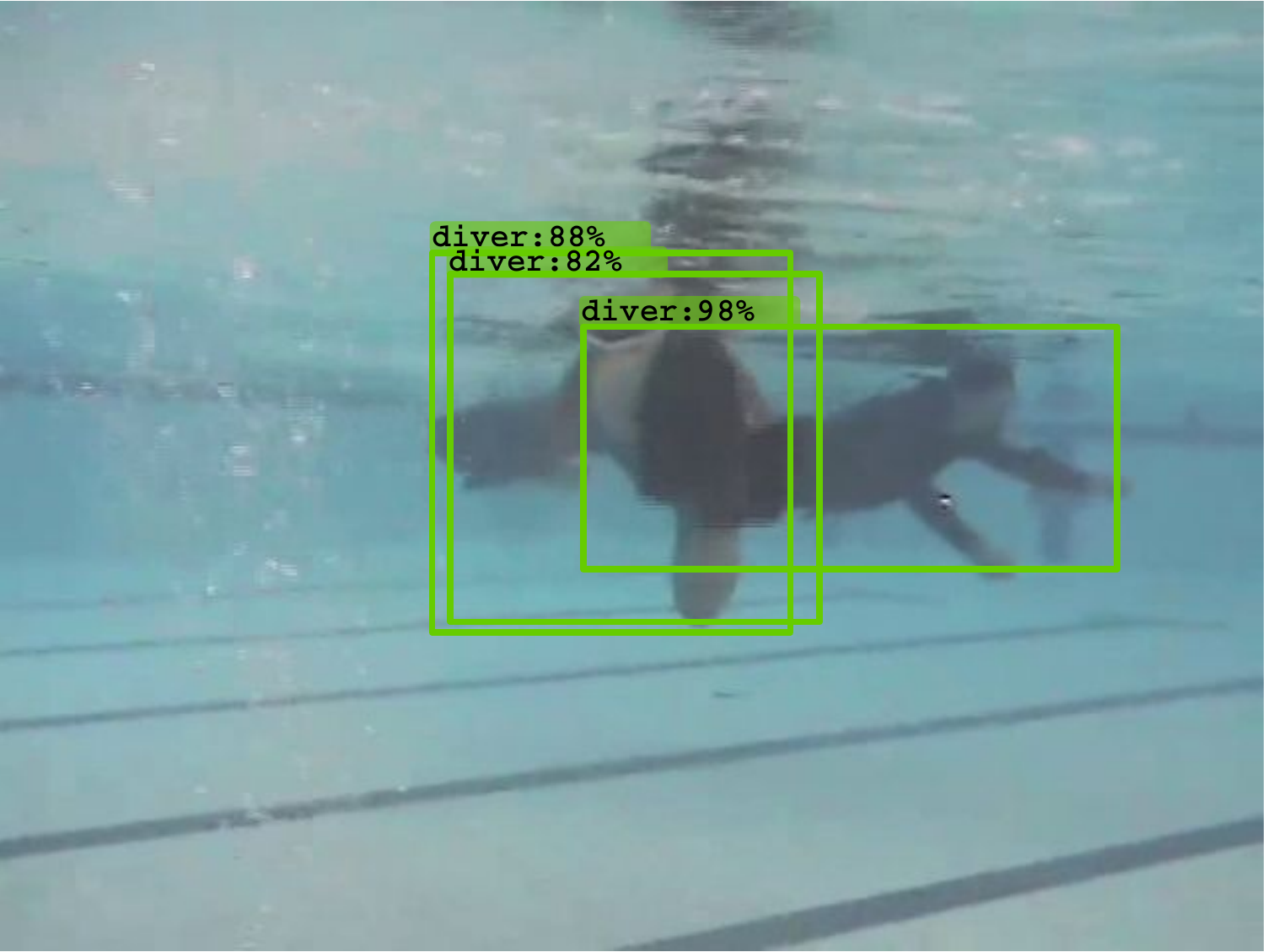}
        \label{fig:bad_b}
    }%
\vspace{-2mm}
\caption{A few cases where the diver-detection performance is considerably affected by noise and occlusion.}
\label{fig:bad}
\end{figure}

Nevertheless, the detection performance can be affected by unfavorable visual conditions; we demonstrate few such cases in Figure~\ref{fig:bad}. In Figure \ref{fig:bad}(a), 
the diver is only partially detected with low confidence ($67\%$) due to a flurry of air-bubbles produced by his flippers' motion while swimming close to the ocean surface, which occluded the robot's view. Suspended particles also cause similar difficulties in diver-following scenarios. 
The visual servo controller can recover from such inaccurate detection as long as the diver is partially visible. However, the continuous tracking may fail if the diver moves away from the robot's field-of-view before it can recover. In this experiment, $27$ consecutive inaccurate detections (\ie, confidence scores less than $50\%$) caused enough drift in the robot's motion for it to lose sight of the diver. Occlusion also affects the detection performance as shown in Figure~\ref{fig:bad}(b); here, the proposed model could not accurately localize the two divers because of occlusion.

\section{Summary and Takeaways}
In this chapter, we presented two diver-following methodologies to address the robustness-efficiency trade-off for autonomous diver-following robots. The first method, named the MDPM tracker, evaluates both spatial and frequency domain features to track human scuba divers swimming in arbitrary trajectories. It incorporates an intuitive representation of human swimming motion into an efficient mixed-domain tracking pipeline based on HMMs. Experimental evaluations point to the utility and effectiveness of this approach, and also provide insight into further performance considerations. Significantly, the tracker is capable of real-time performance and is thus suitable for robotic deployments. The second method uses deep visual features for more reliable detection of divers in noisy conditions. Specifically, we designed a CNN-based model and trained it on a large dataset of hand-annotated images that are collected from various diver-following applications.
The trained model provides near real-time performance in addition to ensuring robustness to noise and invariance to divers' swimming motion, wearables, etc. We also validated its performance margins with respect to several SOTA deep visual object detectors. We further investigated its tracking performance and general applicability through numerous field experiments in pools and oceans.

The MDPM tracker relaxes the directional constraint of existing frequency-domain detectors~\cite{Sattar09RSS} and demonstrates superior performance in tracking arbitrary swimming motions. Moreover, our proposed CNN-based model provides a delicate balance between robustness and efficiency in deep visual diver detection. These diver-following modules are currently used by our Aqua MinneBot AUV~\cite{dudek2007aqua} and LoCO AUV~\cite{LoCOAUV} for underwater human-robot cooperative field experiments.

%In addition, we validate the performance and effectiveness of the proposed model through a number of diver-following experiments in closed-water and open-water environments.

%Our investigations 
%While the discussion of this thesis is limited to autonomous diver-following, our invest
%Relevant publications:~\cite{islam2017mixed},~\cite{islam2018towards},~\cite{islam2018understanding},~\cite{islam2018person}

\chapter{Robot-to-Robot Relative Pose from Human Body-Pose}\label{r2r_pose}
In the previous chapter, we illustrated several scenarios where a team of AUVs follow their companion divers and perform cooperative tasks at various stages of an underwater mission. In such multi-robot cooperative tasks, unless global positioning information is available, the robots need to estimate their positions and orientations relative to each other based on their exteroceptive sensory measurements and odometry~\cite{zhou2008robot}. This process is necessary for registering their measurements to a common frame of reference in order to maintain coordination during task execution. 

\begin{figure}[ht]
\centering
\vspace{1mm}
\includegraphics[width=0.75\linewidth]{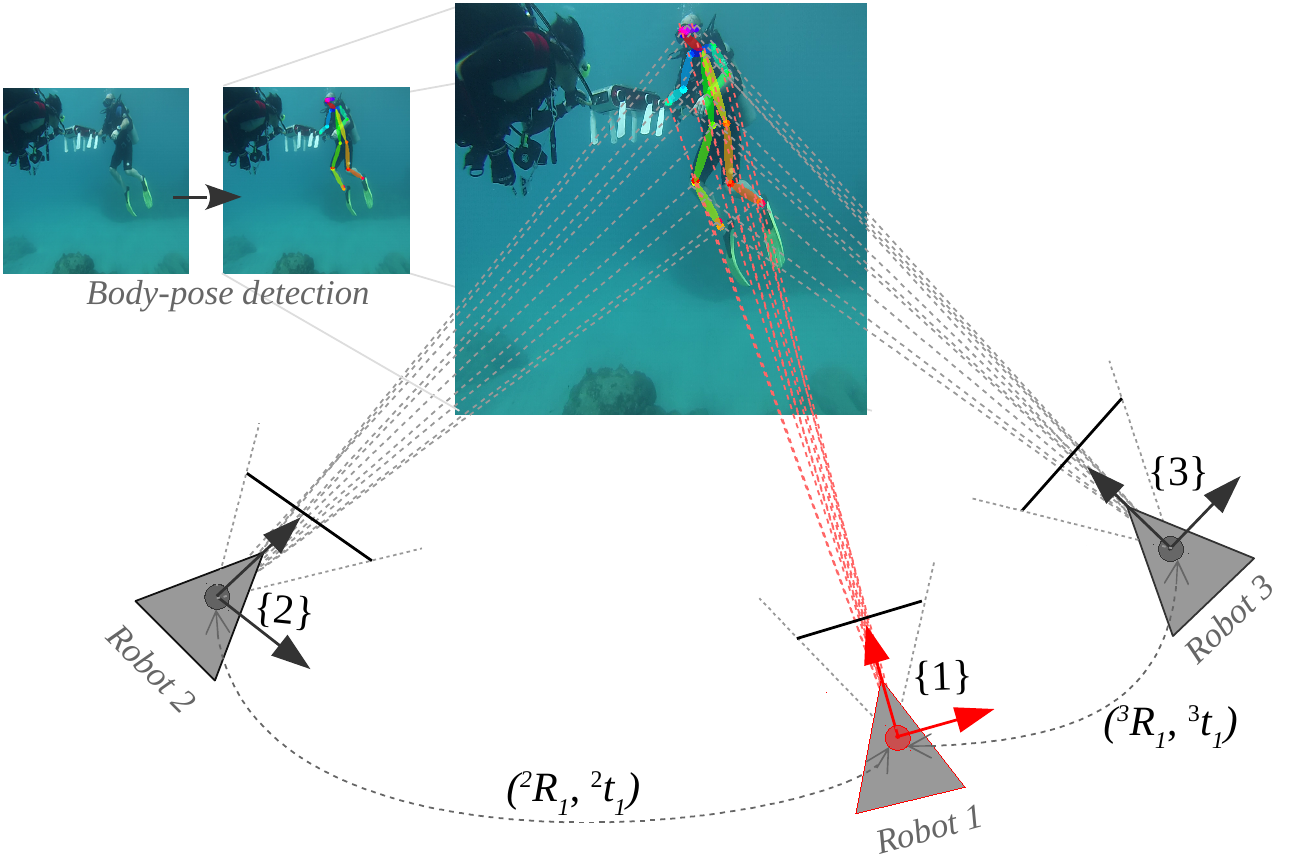}%
\vspace{-1mm} 
\caption{A simplified illustration of 3D relative pose estimation between robot 1 and robot 2 (3). The robots know the transformations between their intrinsically-calibrated cameras and respective global frames, \ie, \{1\}, \{2\}, and \{3\}. Robot 1 is considered as the leader (equipped with a stereo camera) and its pose in global coordinates ($^1R_G$, $^1t_G$) is known. Robot 2 (3) finds its unknown global pose by cooperatively localizing itself relative to robot 1 using the human pose-based key-points as common landmarks.}%
\label{fig1:r2r}
\end{figure}

In a cooperative setting, visually-guided robots solve the relative pose estimation problem by triangulating mutually visible point-based features in the image space. However, a lack of salient features and landmarks in the scene significantly affects the estimation accuracy~\cite{valgren2010sift}, which often arises in poor visibility conditions underwater~\cite{damron2018underwater,sattar2008enabling}. Nevertheless, as mentioned, the presence of human divers in the scene is a fairly common occurrence in underwater human-robot collaborative applications~\cite{islam2018understanding}. Besides, humans are frequently present and visible in many social scenarios~\cite{islam2018person,kummerle2013navigation} where traditional point-based features are not reliably identifiable due to repeated textures, noisy visual conditions, etc. Hence, the problem of having limited natural landmarks can be (potentially) alleviated by using mutually visible humans' body-parts as \emph{markers}, which has not been investigated (see Section~\ref{rel_r2r}).

In this chapter, we present a method for computing six degrees-of-freedom ($6$-DOF) robot-to-robot transformation between pairs of communicating robots by using mutually detected humans' body-poses as correspondences. As illustrated in Figure~\ref{fig1:r2r}, we adopt a \emph{leader-follower} framework where one of the robots (equipped with a stereo camera) is assigned as a leader. First, the leader robot detects and triangulates 3D positions of the pose-based key-points in its own frame of reference. Then the follower robot matches the corresponding 2D projections on its intrinsically calibrated camera and localizes itself by solving the perspective-n-point (PnP) problem~\cite{zheng2013revisiting}. This entire process of \emph{extrinsic calibration} is automatic and does not require prior knowledge about the robots' initial positions. Moreover, it is straightforward to extend the leader-follower framework for multi-robot teams from the pairwise solutions. %Furthermore, if the leader robot has a global positioning system (GPS) or an ultra-short baseline (USBL) receiver, the follower robots can use that information to localize themselves in the global frame as well. 

In addition to the conceptual design, we present an end-to-end system with efficient solutions to the practicalities involved in the proposed robot-to-robot pose estimation method. We use an existing open-source package named OpenPose~\cite{cao2017realtime} for detecting human body-poses in image space. Although it provides state-of-the-art (SOTA) detection performance, the extracted 2D key-points across different views do not necessarily associate as a correspondence. We propose a twofold solution to this:%
\begin{itemize}
    \item First, we design an efficient person re-identification module by evaluating the hierarchical similarities of the key-point regions in image space. It takes advantage of the consistent human pose structures across viewpoints and evaluates their pair-wise similarities for fast body-pose association. We also demonstrate that the SOTA appearance-based person re-identification models fail to provide acceptable performance under single-board real-time constraints.% 
    \item Subsequently, we formulate an iterative optimization algorithm to refine the noisy key-point correspondences by further exploiting their local structural properties in respective images. We demonstrate that the pair-wise key-point refinement is crucial to ensure their validity in a perspective geometric sense.%     
\end{itemize}%
This two-stage process facilitates efficient and robust key-point associations across viewpoints for accurate robot-to-robot relative pose estimation. In this chapter, we primarily focus on these two novel modules because the rest of the computational aspects are generic to all multi-robot cooperative pose estimation systems. Nevertheless, we present a fast implementation of the proposed system and evaluate its end-to-end performance over several terrestrial and underwater field experiments. We present the proposed modules and implementation details in Section~\ref{sec:sys_r2r}; subsequently, we provide an extensive experimental validation in Section~\ref{sec:exp_r2r}. Finally, we discuss the computational aspects and relevant operational considerations in Section~\ref{sec:issues_r2r}.

\section{Related Work}\label{rel_r2r}
The problem of robot-to-robot relative pose estimation has been thoroughly studied for 2D planar robots, particularly for range and bearing sensors. Analytic solutions for determining $3$-DOF robot-to-robot transformation using mutual distance and/or bearing measurements involve solving an over-determined system of nonlinear equations~\cite{zhou2008robot,trawny2010global}. Similar solutions for the 3D case, \ie, for determining 6-DOF transformation using inter-robot distance and/or bearing measurements, has been proposed as well~\cite{zhou2011determining,trawny2010interrobot}. In practice, these analytic solutions are used as an initial estimate for the relative pose, and then iteratively refined by optimization techniques (\eg~nonlinear weighted least-squares) to account for the noise and uncertainty in robot motion.

Robots that rely on visual perception solve the relative pose estimation problem by triangulating mutually visible image-based features~\cite{wang19923d}. Therefore, it reduces to solving the PnP problem by using sets of 2D-3D correspondences between geometric features and their projections on respective image planes~\cite{zheng2013revisiting}. Although high-level geometric features (\eg, lines, conics) have been proposed, point-based features are typically used in practice for relative pose estimation~\cite{janabi2010kalman}. 
Moreover, the PnP problem is solved either using iterative approaches by formulating the over-constrained system ($n$ $>3$) as a nonlinear least-squares problem, or by using sets of three non-collinear points ($n=$ $3$) in combination with Random Sample Consensus (RANSAC) to remove outliers~\cite{fischler1981random}. Besides, vision-based approaches often use temporal-filtering methods, the extended Kalman-filter (EKF) in particular, to reduce the effect of noisy measurements in order to provide near-optimal pose estimates~\cite{wang19923d,janabi2010kalman}. It is also common to simplify the relative pose estimation by attaching specially designed calibration-patterns on each robot~\cite{rekleitis2006simultaneous}. However, this requires that the robots operate at a sufficiently close range, and remain mutually visible.

On the other hand, a large body of existing literature focus on tracking human body-parts relative to a robot for applications such as person/diver following~\cite{islam2018towards,montemerlo2002conditional}, collaborative manipulation~\cite{mainprice2013human}, and behavior imitation~\cite{lei2015whole}. Various forms of \emph{human awareness} are studied for autonomous mobile robots operating in social settings and human-robot collaborative applications as well~\cite{islam2018person,mead2017autonomous}. In underwater domain, a few recent research contributions have focused on the areas of understanding human motion~\cite{islam2018understanding}, gesture~\cite{chavez2018robust,chiarella2015gesture}, instructions~\cite{islam2018dynamic}, etc. However, the feasibility of using human body-poses as geometric feature correspondences for robot-to-robot relative pose estimation has not been explored in the literature.

\section{System Design and Methodology}\label{sec:sys_r2r}
\begin{figure}[h]
\vspace{2mm}
\centering
\includegraphics [width=0.82\linewidth]{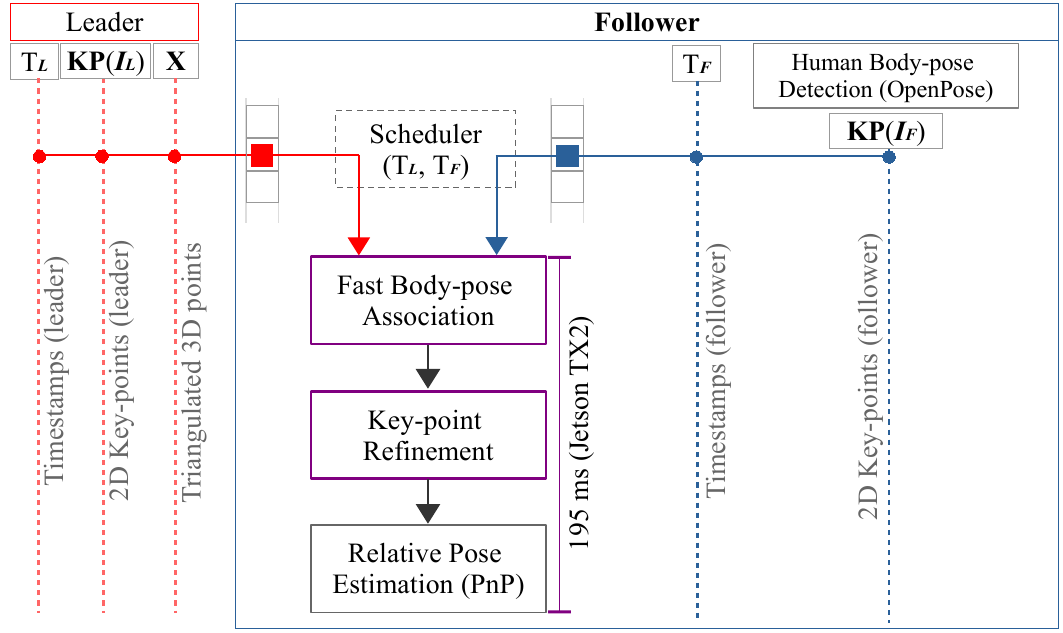}%
\vspace{-2mm} 
\caption{The end-to-end computational pipeline is outlined from the perspective of a follower robot which shares a clock with the communicating leader robot by using a timestamp-based buffer scheduler for synchronized data registration. The mutually visible human body-pose based key-points are then associated and refined for relative pose estimation. We design these two novel components (marked in purple boxes) to establish robust and accurate key-point correspondences at a fast rate ($195$ milliseconds per estimation on a Nvidia Jetson TX2).
}
\label{fig:sys_r2r}
\end{figure}

As shown in Figure~\ref{fig:sys_r2r}, the proposed robot-to-robot relative pose estimation system incorporates several computational components: detection of human body-poses in images captured from different views (by leader and follower robots), pair-wise association of the detected humans across viewpoints, geometric refinement of the key-point correspondences, and 3D pose estimation of the follower robot relative to the leader. We present their methodological details and relevant design choices in the following sections.

\subsection{Human Body-Pose Detection}
OpenPose~\cite{cao2017realtime} is an open-source library for real-time multi-human 2D pose detection in images, originally developed using Caffe and OpenCV libraries~\cite{OpenPoseCMU}. We use a Tensorflow implementation~\cite{OpenPoseTF} based on the \emph{MobileNet model} that provides faster inference compared to the original model (also known as the \emph{CMU model}). Specifically, it processes a $368\times368$ image in $180$ milliseconds on the embedded computing board named NVIDIA\texttrademark{}~Jetson TX2, whereas the original model takes multiple seconds. 

\begin{figure}[t]
\vspace{1mm}
\centering
\includegraphics [width=0.98\linewidth]{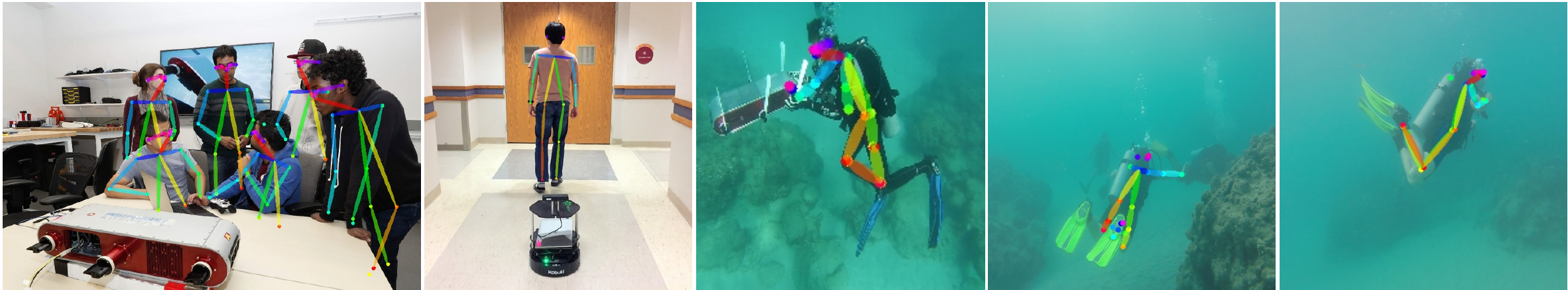}%
\vspace{-2mm} 
\caption{Multi-human 2D body-pose detection using OpenPose in various human-robot collaborative settings.}%
\label{fig:openpose}
\end{figure}%

OpenPose generates $18$ key-points pertaining to the nose, neck, shoulders, elbows, wrists, hips, knees, ankles, eyes, and ears of a human body. As shown in Figure~\ref{fig:openpose}, a subset of these 2D key-points and their pair-wise anatomical relationships are generated for each human. We represent the key-points $\mathbf{KP}(I)$ by a $N_I \times 18$ array where $N_I$ is the number of detected humans in an image $I$. If a particular key-point is occluded or not detected, then the values are left as ($-1$, $-1$). 
We configure $\mathbf{KP}(I)$ in a way that the first row belongs to the left-most person, the second row belongs to the next left-most person, and gradually the last row belongs to the right-most person in the image. This way of sorting the key-points helps to speed up the process of associating the rows of $\mathbf{KP}(I_{leader})$ and $\mathbf{KP}(I_{follower})$. That is, the follower robot needs to make sure that it is pairing the key-points of the \emph{same} individuals. This is important because in practice they might be looking at different individuals, or the same individuals in a different spatial order. Associating multiple persons across different images is a well-studied problem known as \textbf{person re-identification (ReId)}.

\subsection{Person Re-identification using Hierarchical Similarities}
\label{sec:re_id}
\begin{figure*}[t]
\vspace{1mm}
\centering
\includegraphics [width=0.98\linewidth]{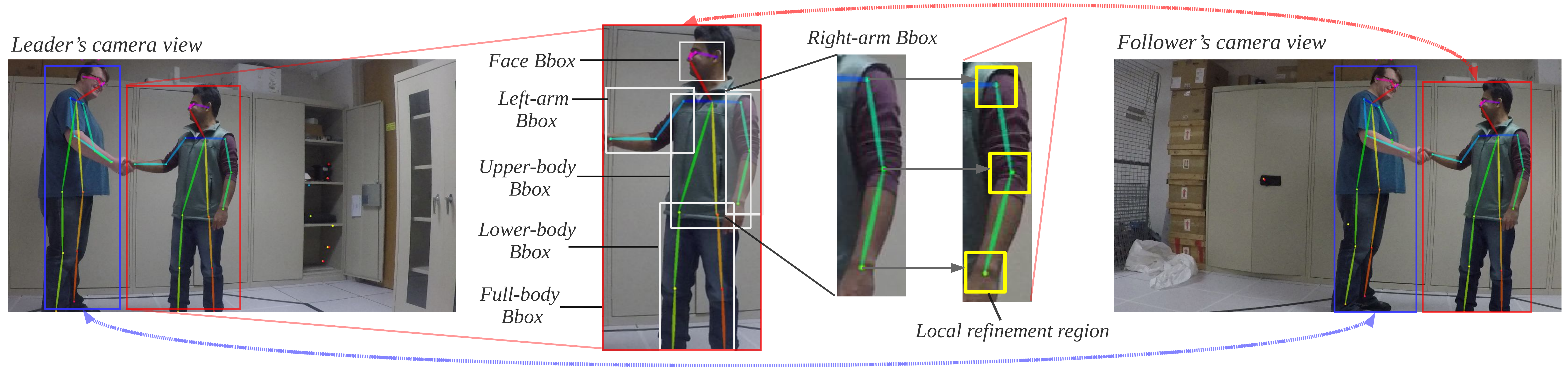}%
\vspace{-2mm}
\caption{An illustration of how the hierarchical body-parts are extracted for person ReId based on their structural similarities; once the persons are associated, the pair-wise key-points are refined and used as correspondences.}%
\label{fig:association}
\end{figure*}

Although several existing deep visual models provide very good solutions for person ReId~\cite{ahmed2015improved,li2014deepreid}, we design a simple and efficient model to meet the real-time single-board computational constraints. The idea is to avoid using a computationally demanding feature extractor by making use of the hierarchical anatomical structures that are already embedded in the key-points. First, we bundle the subsets of key-points into six spatial bounding boxes (BBox) as follows: (i) face BBox: nose, eyes, and ears; (ii) upper-body BBox: neck, shoulders, and hips; (iii) lower-body BBox: hips, knees, and ankles; (iv) left-arm BBox: left shoulder, elbow, and wrist; (v) right-arm BBox: right shoulder, elbow, and wrist; and (vi) full-body BBox: encloses all the key-points. Figure~\ref{fig:association} illustrates the spatial hierarchy of these BBoxes and their corresponding key-points. They are extracted by spanning the key-points' coordinate values in both the $x$ and $y$ dimensions. We use an offset (of additional $10\%$ length) in each dimension to capture more spatial information around the key-points. A BBox is discarded if its area falls below an empirically chosen threshold of $600$ square pixels. We found that BBox areas below this resolution are not always informative and are prone to erroneous results. This happens when the corresponding body-part is either not detected or very far from the camera.  

Once the BBox areas are selected, we exploit their pairwise structural properties as features for person ReId; specifically, we compare the structural similarities~\cite{wang2004image} between image patches pertaining to the face, upper-body, lower-body, left-arm, right-arm, and the full body of a person. Based on their aggregated similarities, we evaluate the pair-wise association between each person as seen by the leader (in $I_{leader}$) and by the follower (in $I_{follower}$). 
The structural similarity~\cite{wang2004image} for a particular pair of single-channel rectangular image-patches ($\mathbf{x}$, $\mathbf{y}$) is evaluated based on three properties: luminance $l(\mathbf{x},\mathbf{y}) = {2 \mathbf{\mu}_\mathbf{x} \mathbf{\mu}_\mathbf{y}}/{(\mathbf{\mu}_\mathbf{x}^2+\mathbf{\mu}_\mathbf{y}^2)}$, contrast $c(\mathbf{x},\mathbf{y}) = {2 \mathbf{\sigma}_\mathbf{x} \mathbf{\sigma}_\mathbf{y}}/{(\mathbf{\sigma}_\mathbf{x}^2+\mathbf{\sigma}_\mathbf{y}^2})$, and structure $s(\mathbf{x},\mathbf{y}) = {\mathbf{\sigma}_{\mathbf{xy}}}/{\mathbf{\sigma}_\mathbf{x}\mathbf{\sigma}_\mathbf{y}}$; here, $\mathbf{\mu}_\mathbf{x}$ ($\mathbf{\mu}_\mathbf{y}$) denotes the mean of image patch $\mathbf{x}$ ($\mathbf{y}$), $\mathbf{\sigma}_\mathbf{x}^2$ ($\mathbf{\sigma}_\mathbf{y}^2$) denotes the variance of $\mathbf{x}$ ($\mathbf{y}$), and $\mathbf{\sigma}_{\mathbf{xy}}$ denotes the cross-correlation between $\mathbf{x}$ and $\mathbf{y}$. The structural similarity metric (SSIM) is then defined as: 
\begin{equation}
\footnotesize
    SSIM(\mathbf{x},\mathbf{y}) = l(\mathbf{x},\mathbf{y}) c(\mathbf{x},\mathbf{y})  s(\mathbf{x},\mathbf{y}) = \frac{2 \mathbf{\mu}_\mathbf{x} \mathbf{\mu}_\mathbf{y} }{\mathbf{\mu}_\mathbf{x}^2+\mathbf{\mu}_\mathbf{y}^2} \times \frac{2 \mathbf{\sigma}_{\mathbf{xy}}}{\mathbf{\sigma}_\mathbf{x}^2+\mathbf{\sigma}_\mathbf{y}^2}.     
\end{equation}
In order to ensure numeric stability, two standard constants $c_1 = (255k_1)^2$ and $c_2 = (255k_2)^2$ are added as: 
\begin{equation}
\footnotesize
    SSIM(\mathbf{x},\mathbf{y}) =  \frac{2 \mathbf{\mu}_\mathbf{x} \mathbf{\mu}_\mathbf{y} + c_1}{\mathbf{\mu}_\mathbf{x}^2+\mathbf{\mu}_\mathbf{y}^2 + c_1} \times \frac{2 \mathbf{\sigma}_{\mathbf{xy}} + c_2}{\mathbf{\sigma}_\mathbf{x}^2+\mathbf{\sigma}_\mathbf{y}^2 + c_2}. 
\label{eq:one}
\end{equation}
We use $k_1=0.01$, $k_2=0.03$, and an $8\times8$ sliding window in our implementation. Additionally, we resize the patches extracted from $I_{leader}$ so that their corresponding pairs in $I_{follower}$ have the same dimensions. Then, we apply Equation~\ref{eq:one} on every channel ($R,G,B$) and use their average value as the similarity metric on a scale of [$0$, $1$]. Specifically, we use this metric for person ReId as follows: 
\begin{itemize}
    \item We only consider the mutually visible body-parts for evaluating the pair-wise SSIM values. This choice is important to enforce meaningful comparisons; otherwise, it is equivalent to using only the full-body BBox, which we found to be highly inaccurate.
    \item Each person in $I_{follower}$ is associated with the most similar person corresponding to the maximum SSIM value in $I_{leader}$. However, the association is discarded if that value is less than a threshold $\delta_{min}=0.4$ which is chosen by an AUC (area under the curve)-based analysis (see Section~\ref{impact}). This reduces the risk of inaccurate associations, particularly when there are mutually exclusive people in the scene.           
\end{itemize}

\subsection{Key-point Refinement}
\label{sec:kp_ref}
Once the specific persons are identified, \textit{i.e.}, the rows of $\mathbf{KP}(I_{leader})$ and $\mathbf{KP}(I_{follower})$ are associated, the mutually visible key-points are paired together to form correspondences. Although the key-points are ordered and OpenPose localizes them reasonably well, they cannot be readily used as geometric correspondences due to perspective distortions and noise.  
We attempt to solve this problem by designing an iterative optimization algorithm that refines the noisy correspondences based on their structural properties in a $32\times32$ neighborhood. By denoting $\mathbf{\phi}_I(\mathbf{p})$ as the $32\times32$ image-patch centered at $\mathbf{p}=[p_x, p_y]^T$ in image $I$, we define a loss function for each correspondence $(\mathbf{p}_l \in I_{leader}, \mathbf{p}_f \in I_{follower})$ as:
\begin{equation}
\footnotesize
L(\mathbf{p}_l, \mathbf{p}_f) = 1 - SSIM(\mathbf{\phi}_{I_{leader}}(\mathbf{p}_{l}), \mathbf{\phi}_{I_{follower}}(\mathbf{p}_{f})).
\end{equation}
Then, we refine each initial key-point correspondence $(\mathbf{p}_l^{0}, \mathbf{p}_f^{0})$ by minimizing the following function:
\begin{equation}
\footnotesize
\mathbf{p}_f^* = \operatorname*{argmin}_{\mathbf{p}} \quad L(\mathbf{p}_{l}^{0}, \mathbf{p}) 
\quad \text{s. t.} \quad ||\mathbf{p}-\mathbf{p}_{f}^{0}||_\infty<32. 
\label{eq:three}
\end{equation}
Here, we fix $\mathbf{p}_l=\mathbf{p}_l^{0}$ and refine $\mathbf{p}_f=\mathbf{p}_f^{0}$ to maximize $SSIM(\mathbf{\phi}_{I_{leader}}(\mathbf{p}_{l}), \mathbf{\phi}_{I_{follower}}(\mathbf{p}_{f}))$. In our implementation, we adopt a gradient-based refinement algorithm that performs the following iterative update:
\begin{equation}
\footnotesize
    \mathbf{p}_f^{t+1} = \mathbf{p}_f^t - \eta \cdot \nabla L(\mathbf{p}_l^0, \mathbf{p}_f^t). 
\end{equation}
We follow the procedures suggested in~\cite{avanaki2009exact,otero2014solving} for computing the gradient of SSIM. For fast processing, we vertically stack all the key-points and their gradients to perform the optimization simultaneously with a fixed learning rate of $\eta=0.003$ for a maximum iteration of $100$. We present empirical validations for the choices of the refinement resolution and other hyper-parameters in Section~\ref{impact}.

\begin{figure}[ht]
\vspace{1mm}
\centering
    \subfigure[A group of people seen from multiple views and their 2D body-poses (detected by OpenPose).]{
        \includegraphics[width=0.95\textwidth]{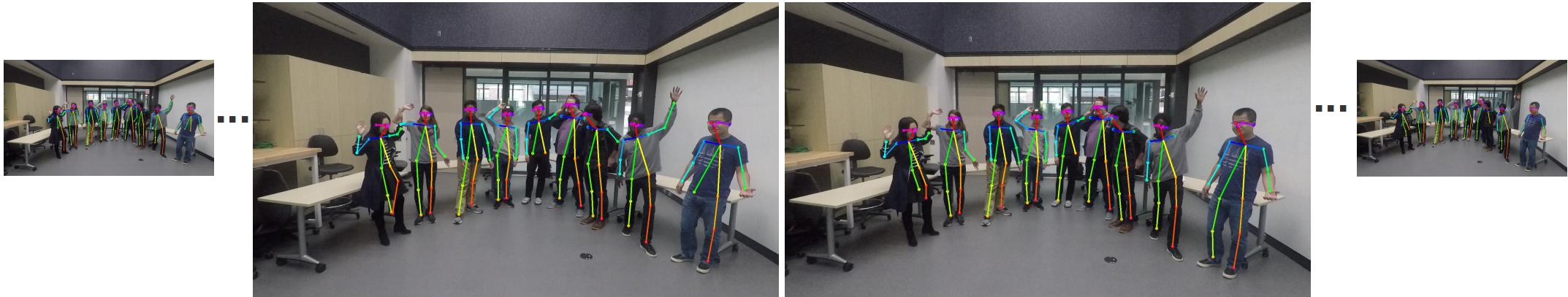}
        \label{fig:exp1a_r2r}
    }
    
    \subfigure[Person association and key-point correspondences for a particular image pair; a unique identifier is assigned to each association, matched key-points are shown for the right-most person.]{
        \includegraphics[width=0.93\textwidth]{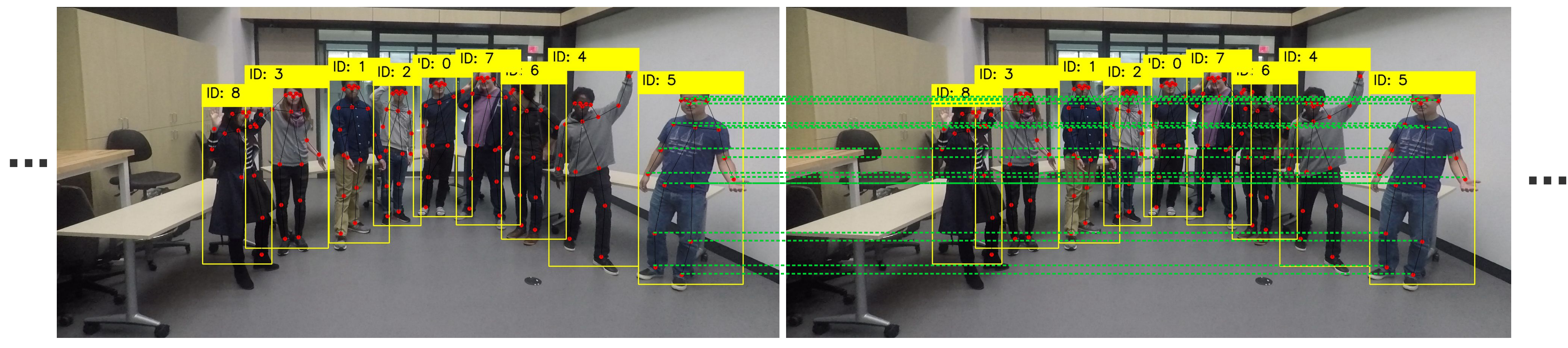}
        \label{fig:exp1b_r2r}
    }
    
    \subfigure[3D reconstruction and the estimated camera poses (up-to scale).]{
        \includegraphics[width=0.8\textwidth]{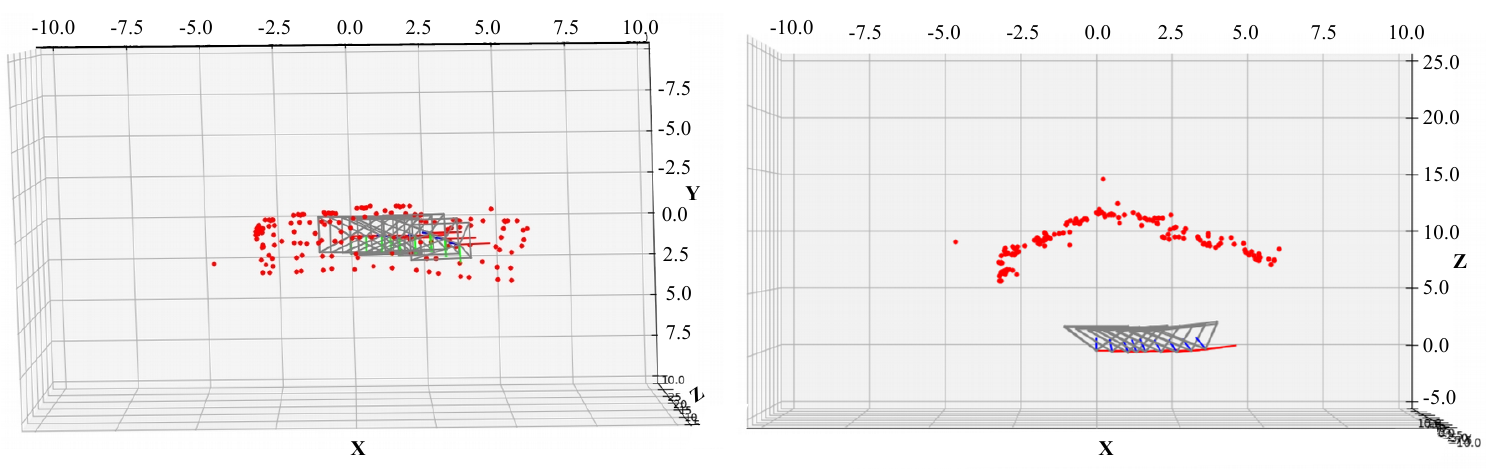}
        \label{fig:exp1c_r2r}
    }
\caption{Results of estimating \emph{structure from motion} using only human pose-based key-points as features.}
\label{fig:exp1_r2r}
\end{figure}

\subsection{Robot-to-robot Pose Estimation}\label{pnp}
Once the mutually visible key-points are associated and refined, the follower robot uses the corresponding 3D positions (provided by the leader) to estimate its relative pose by solving a PnP problem. Thus, we require that the leader robot is equipped with a stereo camera (or an RGBD camera) so that it can triangulate the refined key-points using epipolar constraints (or use the depth sensor) to represent the key-points in 3D. 

Let $\mathbf{x}_l$ denote the 3D locations of the key-points in the leader's coordinate frame, and $\mathbf{p}_f$ denote their corresponding 2D projections on the follower's camera. Then, assuming the cameras are synchronized, the PnP problem is formulated as follows:
\begin{equation}\label{eq:opt_final}
\footnotesize
    \mathbf{T}_f^l = \operatorname*{argmin}_{\mathbf{T}_f^l} ||\mathbf{p}_f - \mathbf{K}_f \mathbf{T}_f^l \mathbf{x}_l||^2.
\end{equation}
Here, $\mathbf{K}_f$ is the intrinsic matrix of the follower's camera and $\mathbf{T}_f^l$ is its 6-DOF transformation relative to the leader. In our implementation, we follow the standard iterative solution for PnP using RANSAC~\cite{zheng2013revisiting}.

\section{Experimental Analysis}\label{sec:exp_r2r}
We conduct several experiments with 2-DOF and 3-DOF robots to evaluate the performance and feasibility of the proposed relative pose estimation method. We present these experimental details, analyze the results, and discuss various operational considerations in the following sections.

\subsection{Proof of Concept: Structure from Motion}
At first, we perform experiments to validate that the human pose-based key-points can be used as geometric correspondences for relative pose estimation. As illustrated in Figure~\ref{fig:exp1a_r2r}, we emulate an experimental set-up for \emph{structure from motion} with humans; we use an intrinsically calibrated monocular camera to capture a group of nine (static) people from multiple views. Here, the goal is to estimate the camera poses and reconstruct the 3D structures of the humans using only their body-poses as features. 

In the evaluation, we first use OpenPose to detect the human pose-based 2D key-points in the images. Then, we utilize the proposed person ReId and key-point refinement modules to obtain the feature correspondences across multiple views, as shown in Figure~\ref{fig:exp1b_r2r}. Subsequently, we follow the standard procedures for structure from motion~\cite{hartley2003multiple}: fundamental matrix computation using 8-point algorithm with RANSAC, essential matrix computation, camera pose estimation by enforcing the Cheirality constraint, and linear triangulation. Finally, the triangulated 3D points and camera poses are refined using bundle adjustment. As demonstrated in Figure~\ref{fig:exp1c_r2r}, the spatial structure of the reconstructed points on the human bodies and the camera poses are consistent with our setup. %Results of another experiment for a \emph{two-view} case are shown in Figure~\ref{fig:appen1_r2r}, which further validate that the estimated camera poses are comparable to the ground truth, \ie, analogous SIFT feature-based estimation. In the next section, we demonstrate the effectiveness of our proposed body-pose association and key-point refinement modules in ensuring this robust pose estimation performance.

\begin{table*}[ht]
	\centering
	\footnotesize
	\caption{A quantitative performance comparison for various person ReId models on standard datasets; two sets of $150$ test images from each dataset are used for the evaluation [R1-Ac: rank-1 accuracy; mAP: mean average precision; FPS: frames per second].}
	\vspace{2mm}
	\begin{tabular}{l||c|c||c|c||c}
		\Xhline{2\arrayrulewidth}
		Person ReId & \multicolumn{2}{c||}{{\tt Market-1501 Dataset}} & \multicolumn{2}{c||}{{\tt CUHK-03 Dataset}} & {FPS} \\ \cline{2-5}
		models &  R1-Ac (\%) & mAP (\%) & R1-Ac (\%) & mAP (\%) & (Jetson TX2) \\
		\Xhline{2\arrayrulewidth}
		Aligned ReId  & $92.90$ & $90.12$  &  $67.15$ & $68.03$ & $0.67$ \\
		%\hline
		Deep person ReId  & $85.86$ & $68.24$      & $62.26$ & $65.15$ & $0.33$ \\
		%\hline
		Tripled-loss ReId  & $85.25$ & $74.88$   & $72.75$ & $60.27$  & $0.62$  \\
		%\hline
		\textbf{Proposed person ReId}  & $75.67$ & $72.26$   & $57.82$ & $54.91$  & \textbf{7.45}  \\
		\Xhline{2\arrayrulewidth} 
	\end{tabular}
	\label{tab:std_r2r}
	\vspace{2mm}
	
	\caption{Effectiveness of the proposed person ReId method on real-world data; each set contains $100$ images of multiple humans in ground and underwater scenarios. }
	\vspace{2mm}
	\begin{tabular}{l||c|c||c|c}
		\Xhline{2\arrayrulewidth}
		Person ReId & \multicolumn{2}{c||}{{\tt Set A (1-2 humans per image)}} & \multicolumn{2}{c}{{\tt Set B (3-5 humans per image)}} \\ \cline{2-5}
		models &  R1-Ac (\%) & FPS (Jetson TX2) & R1-Ac (\%) & FPS (Jetson TX2)  \\
		\Xhline{2\arrayrulewidth}
		Aligned ReId  & $62.75$ & $0.62$  &  $56.65$ & $0.48$  \\
		%\hline
		Deep person ReId   & $55.32$ & $0.29$      & $42.36$ & $0.12$  \\
		%\hline
		Tripled-loss ReId  & $55.15$ & $0.58$   & $44.85$ & $0.44$  \\
		%\hline
		\textbf{Proposed person ReId}  & \textbf{76.55} & \textbf{6.81}   & \textbf{71.56} & \textbf{5.45}    \\
		\Xhline{2\arrayrulewidth}
	\end{tabular}
	\label{tab:our_r2r}
\end{table*}

\begin{figure}[ht]
\vspace{1mm}
\centering
    \subfigure[ROC curve for the person ReID performance ($AUC=0.938$): a total of $20$ thresholds are considered in the evaluation; the point marked in red corresponds to $\delta_{min}=0.4$.]{
        \includegraphics[width=0.48\textwidth]{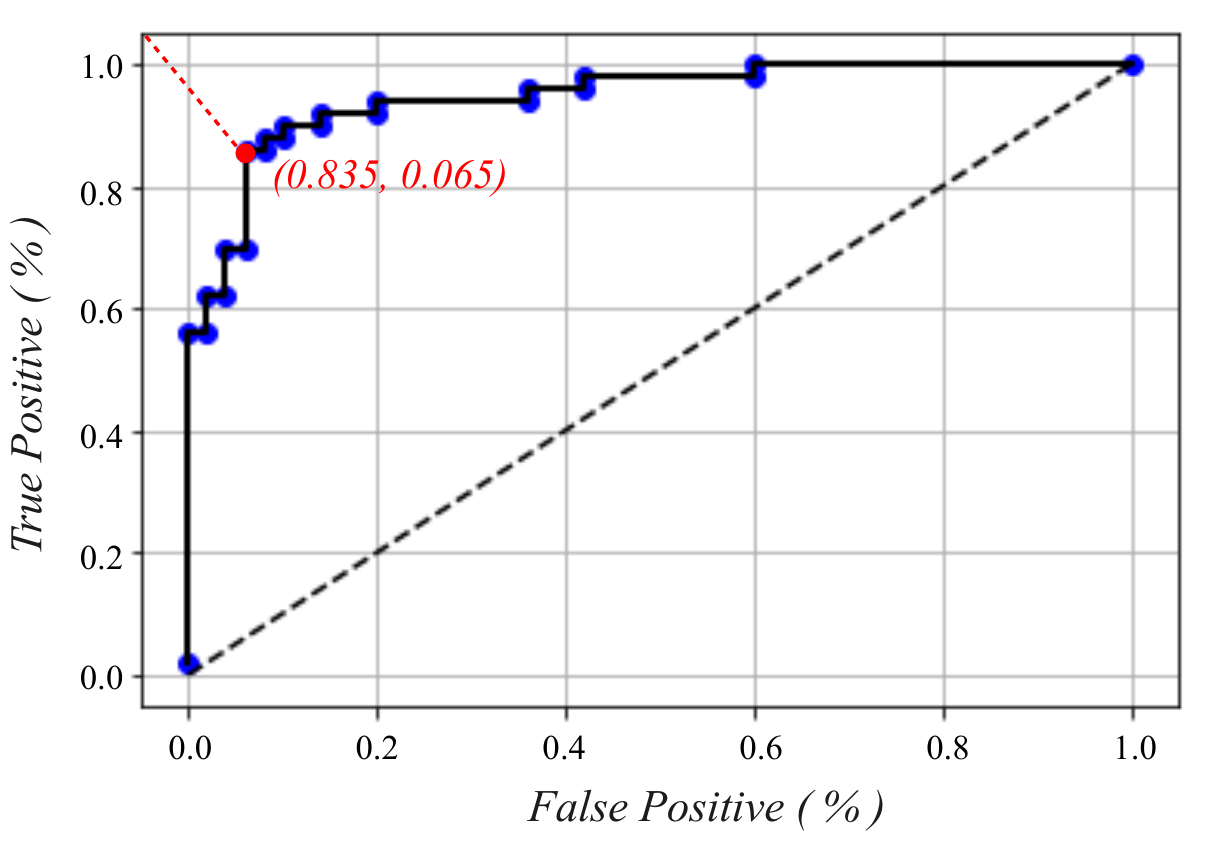}
        \label{fig:param_a_r2r}
    }~ \hspace{1mm}
    \subfigure[Fraction of key-point correspondences that fall within various resolutions of respective refinement regions; the vertical red line corresponds to a refinement region of $32\times32$.]{
        \includegraphics[width=0.48\textwidth]{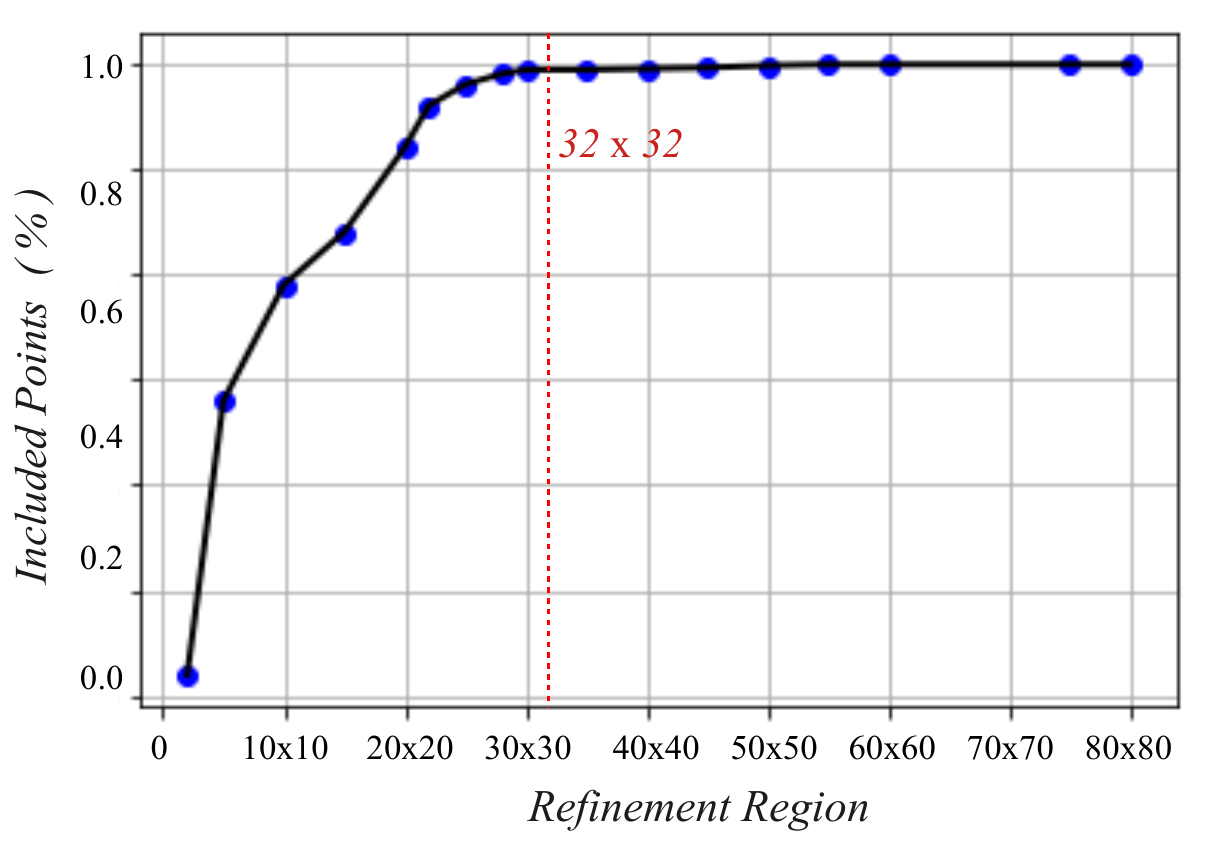}
        \label{fig:param_b_r2r}
    }
    
\caption{Empirical selection of hyper-parameters: (a) SSIM threshold for pose association in the proposed person ReId module and (b) resolution of the key-point refinement region. The evaluation is performed on the combined set of $250$ images containing a total of $687$ person associations with $8256$ key-point correspondences.}
\label{fig:hyper_param_r2r}
\end{figure}

\begin{figure}[ht]
\vspace{1mm}
\centering
    \subfigure[Inaccurate 3D reconstruction using raw key-point correspondences (without refinement).]{
        \includegraphics[width=0.48\textwidth]{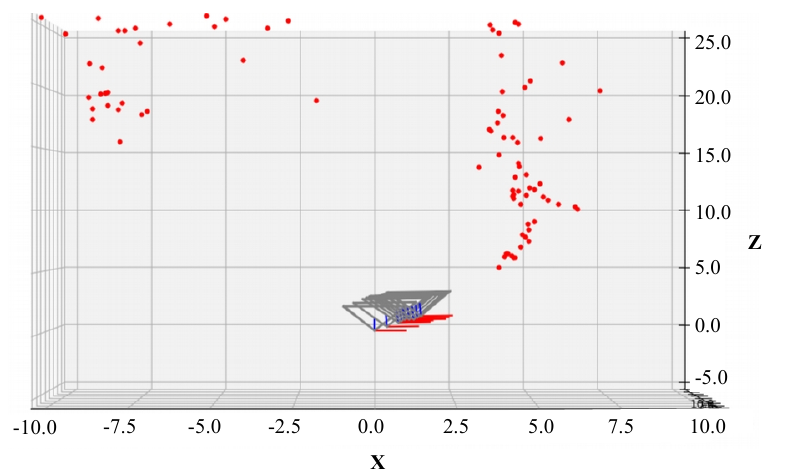}
        \label{fig:issues_b_r2r}
    }~ \hspace{1mm}
    \subfigure[Quantitative performance compared to using SIFT-based features.]{
        \includegraphics[width=0.46\textwidth]{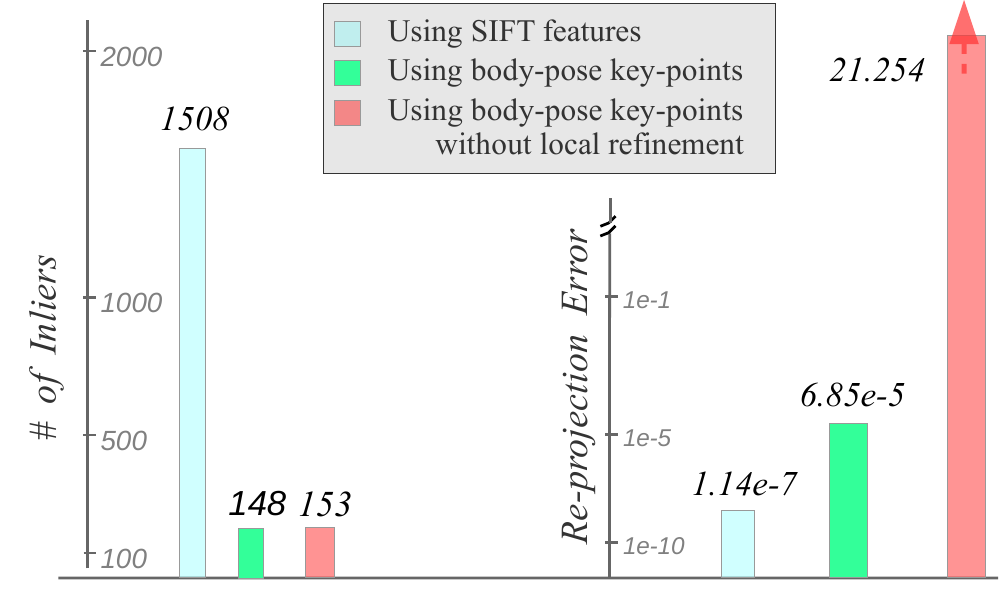}
        \label{fig:issues_a_r2r}
    }
    
\caption{Necessity of the proposed key-point refinement process; results correspond to the experiment illustrated in Figure~\ref{fig:exp1_r2r}.}
\label{fig:issues_r2r}
\end{figure}

\subsection{Utility and Effectiveness of the Proposed Modules}\label{impact}
It is easy to notice that person ReId is essential for associating mutually visible persons across different views. As mentioned in Section~\ref{sec:re_id}, we focus on achieving fast association by making use of the local structural properties around the anatomical key-points in the image space. In contrast, the SOTA person ReId approaches adopt deep visual feature extractors that are computationally demanding. In Table~\ref{tab:std_r2r}, we quantitatively evaluate the SOTA models named Aligned ReId~\cite{zhao2017deeply}, Deep person ReId~\cite{li2014deepreid}, and Tripled-loss ReId~\cite{zheng2011person} based on rank-1 accuracy (R1-Ac) and mean averaged precision (mAP) on two standard datasets. Specifically, a test-set containing $150$ instances from each of the {\tt Market-1501} and {\tt CUHK-03} datasets are used for the evaluation; also, their run-times on a NVIDIA\texttrademark{} Jetson TX2 are shown in the comparison. The results indicate that although these models, once trained on similar data, perform well on standard datasets, they are computationally too expensive for single-board embedded platforms. 

Moreover, as demonstrated in Table~\ref{tab:our_r2r}, these off-the-shelf models do not perform that well on high-resolution real-world images. Although their performance can be improved by training on more comprehensive real-world data, the computational complexity remains a barrier. To this end, the proposed person Reid module provides significantly faster run-time and better portability as it does not require rigorous large-scale training. Its only hyper-parameter is the SSIM threshold $\delta_{min}$ (see Section~\ref{sec:re_id}), which we select by standard AUC-based analysis of ROC (receiver operating characteristic) curve. As shown in Figure~\ref{fig:param_a_r2r}, we choose $\delta_{min}=0.4$, which corresponds to $83.5\%$ true-positive and $6.5\%$ false-positive rates for person ReID on the combined test set of $250$ images containing $687$ person associations. Additionally, we select the key-point refinement resolution through an ablation experiment with $8256$ key-point correspondences. We observe that the optimal key-point location is found within $25\times25$ pixels of the initial estimate by OpenPose for over $96\%$ of the cases. As shown in Figure~\ref{fig:param_b_r2r}, we make a more conservative choice of $32\times32$ refinement region in our implementation.

Finally, we evaluate the utility and effectiveness of the proposed key-point refinement algorithm based on re-projection errors and compare the results with traditional SIFT feature-based reconstruction. 
As Figure~\ref{fig:issues_b_r2r} demonstrates, the 3D reconstruction and camera pose estimation with raw key-points are inaccurate as the unrefined correspondences are invalid in a perspective geometric sense. 
As Figure~\ref{fig:issues_a_r2r} shows, the average re-projection error for the refined key-points reduces to $6.85e^{-5}$ pixels, which is acceptable considering the fact that there are ten times less anatomical key-points than SIFT feature-based key-points. This evaluation corresponds to the experiment presented in Figure~\ref{fig:exp1_r2r}, which shows that the refined key-points constitute accurate scene reconstruction and camera pose estimation. %Another qualitative validation of the iterative key-point refinement algorithm and its convergence behavior can be found in Figure~\ref{fig:appen1_r2r}.  

\begin{figure}[t]
\vspace{1mm}
\centering
    \subfigure[The follower robot's trajectory is marked by red arrows.]{\includegraphics[width=0.58\textwidth]{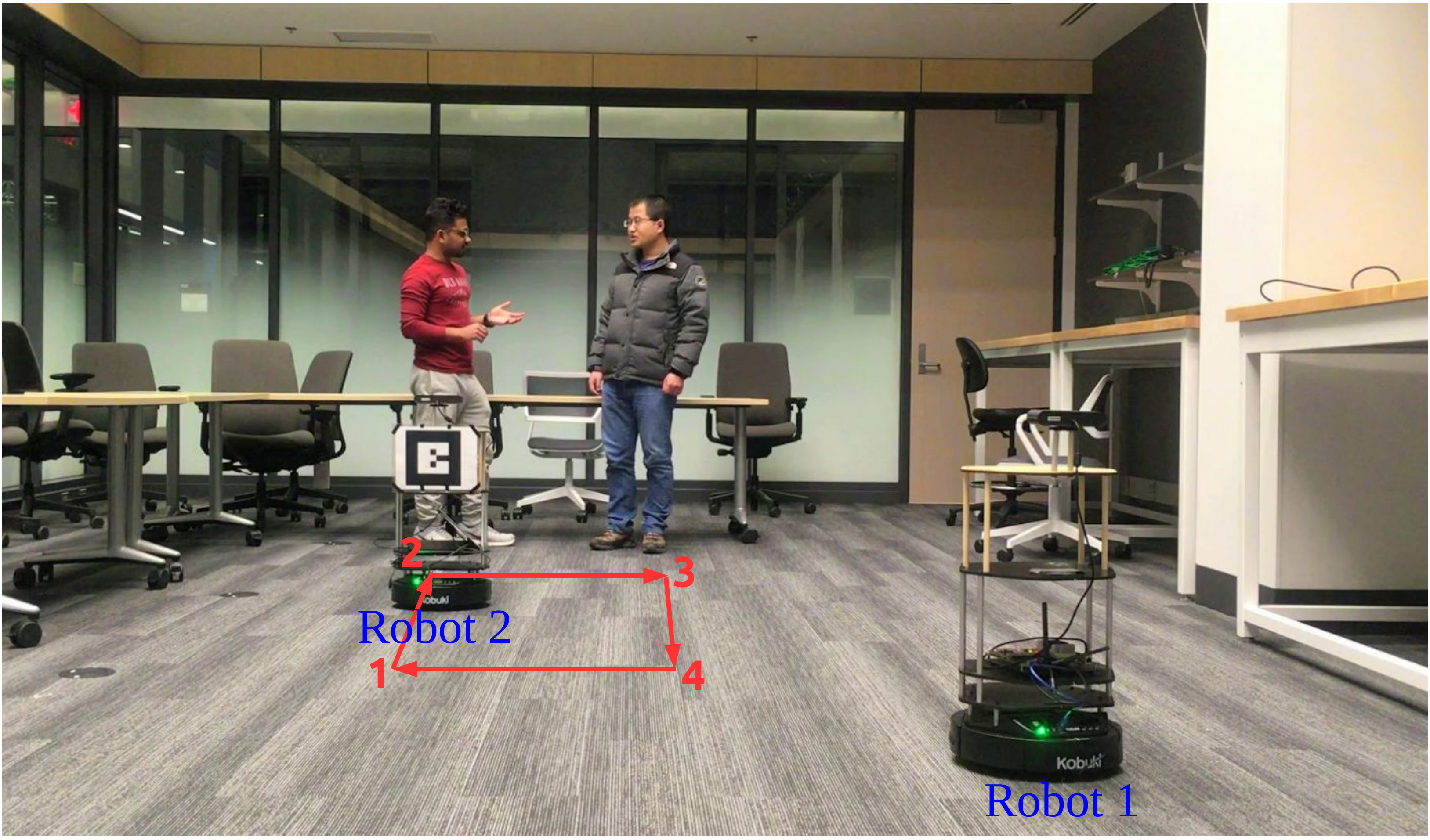}
    \label{fig:setup_a_r2r}
    }

    \subfigure[The leader robot detects the pose-based key-points and shares the 3D locations.]{
        \includegraphics[width=0.42\textwidth]{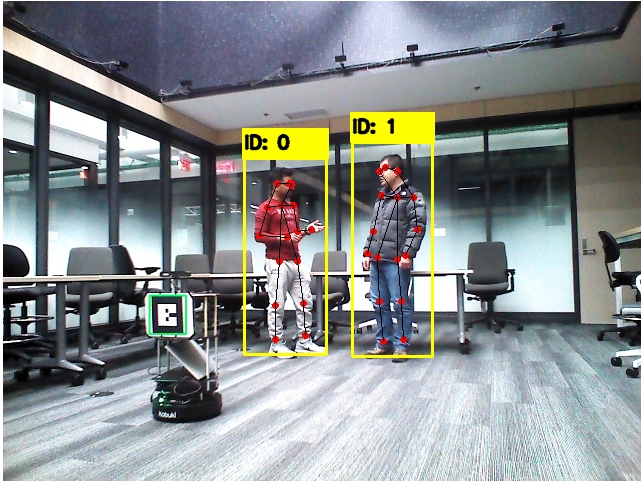}
        \label{fig:ground_a_r2r}
    }~ \hspace{1mm}
    \subfigure[Estimated poses of the follower relative to the leader (the green cones represent the respective ground truth).]{
        \includegraphics[width=0.55\textwidth]{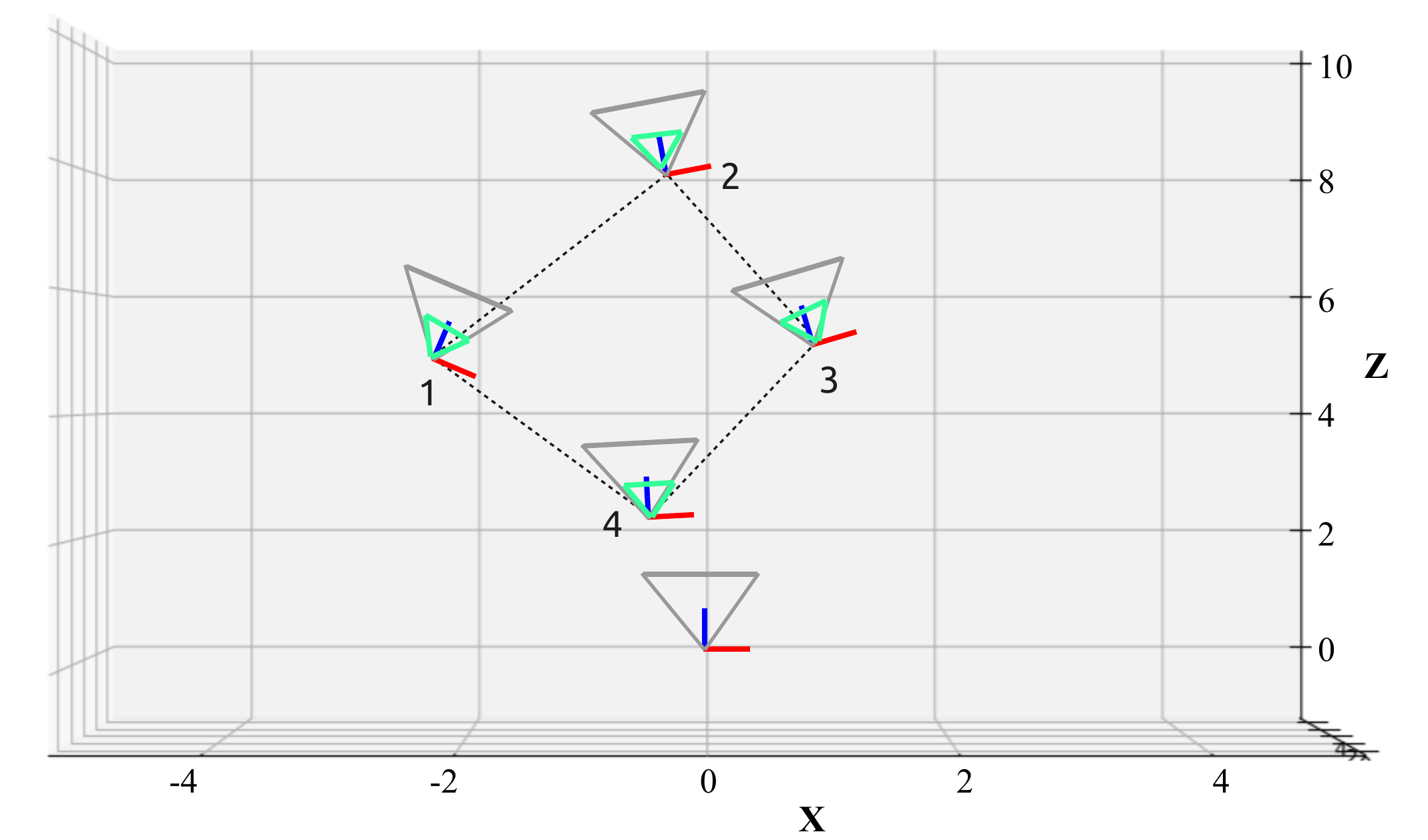}
        \label{fig:ground_b_r2r}
    }
    
\caption{An experiment to evaluate the accuracy of 2D relative pose estimation with two planar robots and two mutually visible humans.}
\label{fig:ground_r2r}
\end{figure}

\subsection{3-DOF Pose Estimation: Ground Robots}
We also perform experiments for 3-DOF robot-to-robot relative pose estimation with 2D robots.
In the particular scenario shown in Figure~\ref{fig:setup_a_r2r}, we use two planar robots (one leader and one follower) and two mutually visible humans in the scene. The robot ($\#1$) with an AR-tag on its back  is used as the follower robot while the other robot ($\#2$) is used as the leader. The AR-tag is used to obtain the follower's ground truth relative pose for comparison. On the other hand, the leader robot is equipped with an RGBD camera; it communicates with the follower and shares the 3D locations of the mutually visible key-points. Specifically, it detects the human pose-based 2D key-points and associates the corresponding depth information to represent them in 3D. Subsequently, the follower robot uses this information to localize itself relative to the leader by following the proposed estimation method.      

As demonstrated in Figure~\ref{fig:setup_a_r2r}, we move the follower robot in a rectangular pattern and evaluate the 3-DOF pose estimates relative to the static leader robot. We present the qualitative results in Figure~\ref{fig:ground_b_r2r}; it shows that the follower robot's pose estimates are very close to their respective ground truth. Overall, we observe an average error of $0.0475\%$ in translation (cm) and a $0.8625^{\circ}$ average error in rotation, which is reasonably accurate. We obtain similar qualitative and quantitative performance with a dynamic leader as well. Next, we present field experimental validations of the relative pose estimation performance in feature-deprived underwater scenarios.

\begin{figure}[h]
\vspace{1mm}
\includegraphics[width=0.98\textwidth]{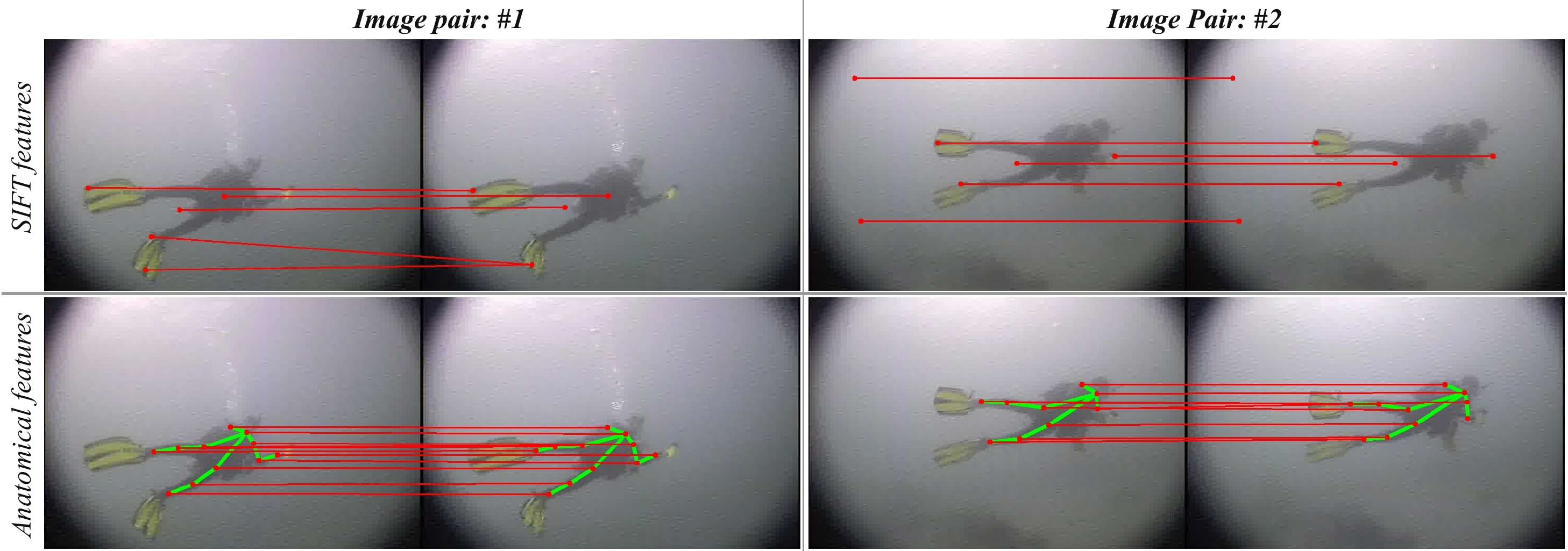}%
\vspace{-3mm}
\caption{Illustrations for two image pairs are shown where a poor underwater visual condition limits the utility of traditional point-based feature detectors~\cite{FLANN}. Nevertheless, the presence of a single human in the scene facilitates considerably more anatomical key-point correspondences than standard SIFT features.}%
\label{fig:exp20_r2r}
\end{figure}

\subsection{6-DOF Pose Estimation: Underwater Robots} 
As seen in Figure~\ref{fig:exp20_r2r}, standard point-based feature detectors fail to generate a large pool of reliable correspondences when there are very few salient features and landmarks in the scene. Consequently, the sampling-based parameter estimation techniques (\eg~RANSAC) often generate inaccurate results in feature-deprived underwater scenarios. However, we demonstrate that human pose-based key-points can still be refined to establish reliable geometric correspondences for robot-to-robot relative pose estimation. Moreover, we get a reasonably large pool of correspondences with only one or two humans in the scene, which is fairly common in cooperative underwater missions.       

\begin{figure}[h]
\vspace{1mm}
\centering
\includegraphics [width=0.75\linewidth]{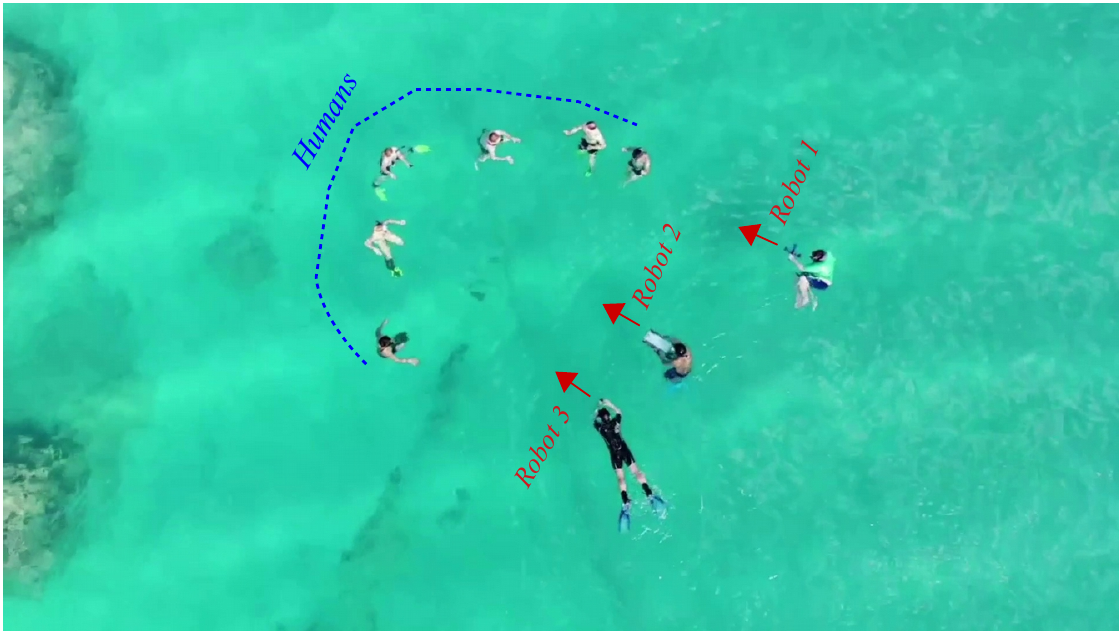}
\vspace{-3mm}
\caption{The setup of a 6-DOF underwater experiment with one leader and two follower robots (aerial view).}%
\label{setup_b_r2r}
\end{figure}%

\begin{figure}[ht]
\vspace{1mm}
\centering
    \subfigure[A group of people seen from multiple perspectives; the detected key-points and their associations are annotated in respective images.]{
        \includegraphics[width=0.98\textwidth]{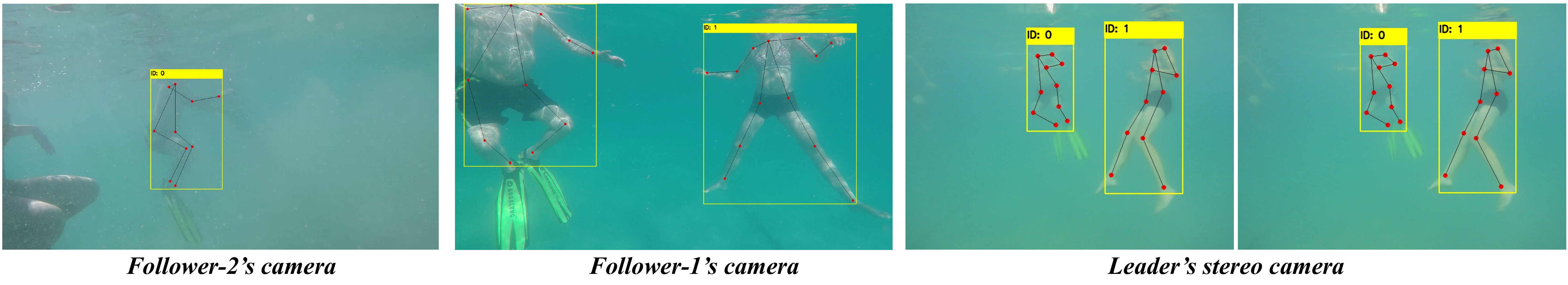}
        \label{fig:exp2a_r2r}
    }
    
    \subfigure[The anatomical key-points are pair-wise associated, refined, and then used to project epipolar lines for the leader's stereo image pair.]{
    \includegraphics[width=0.98\textwidth]{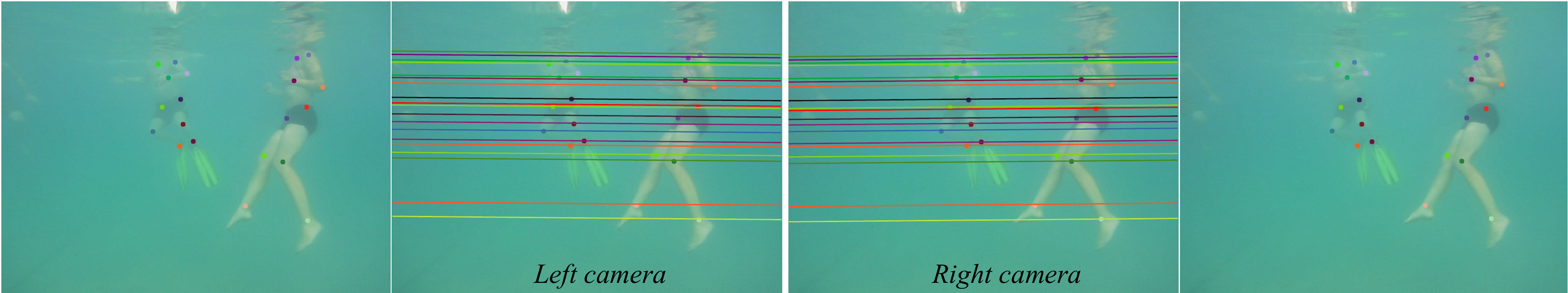}
    \label{fig:exp20e_r2r}
    }
    
    \subfigure[Stereo triangulation by the leader robot.]{
	\includegraphics[width=0.46\linewidth]{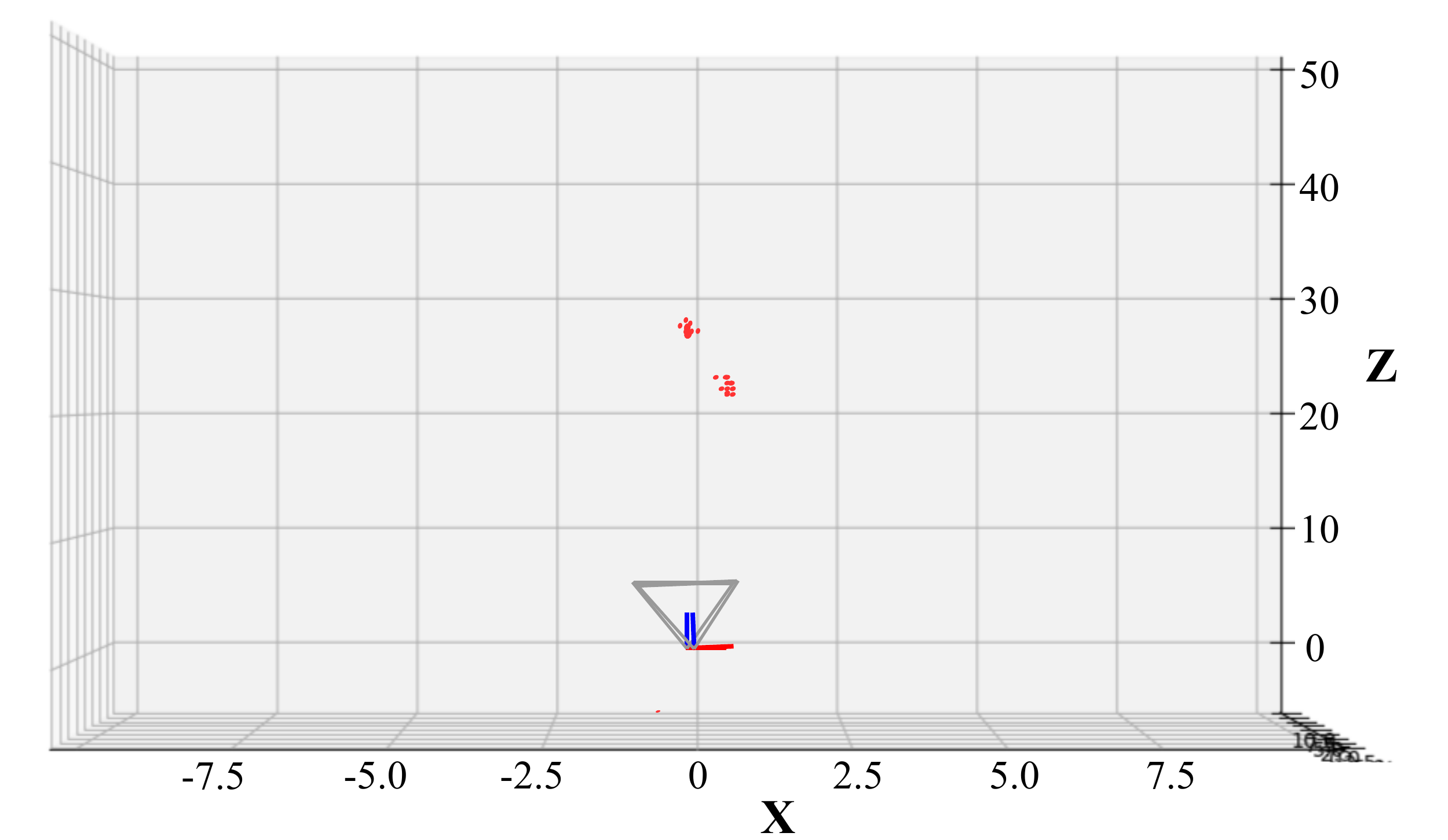}
    \label{fig:exp2b_r2r}
    }~
    \subfigure[Relative poses of the followers robots.]{
    \includegraphics[width=0.46\linewidth]{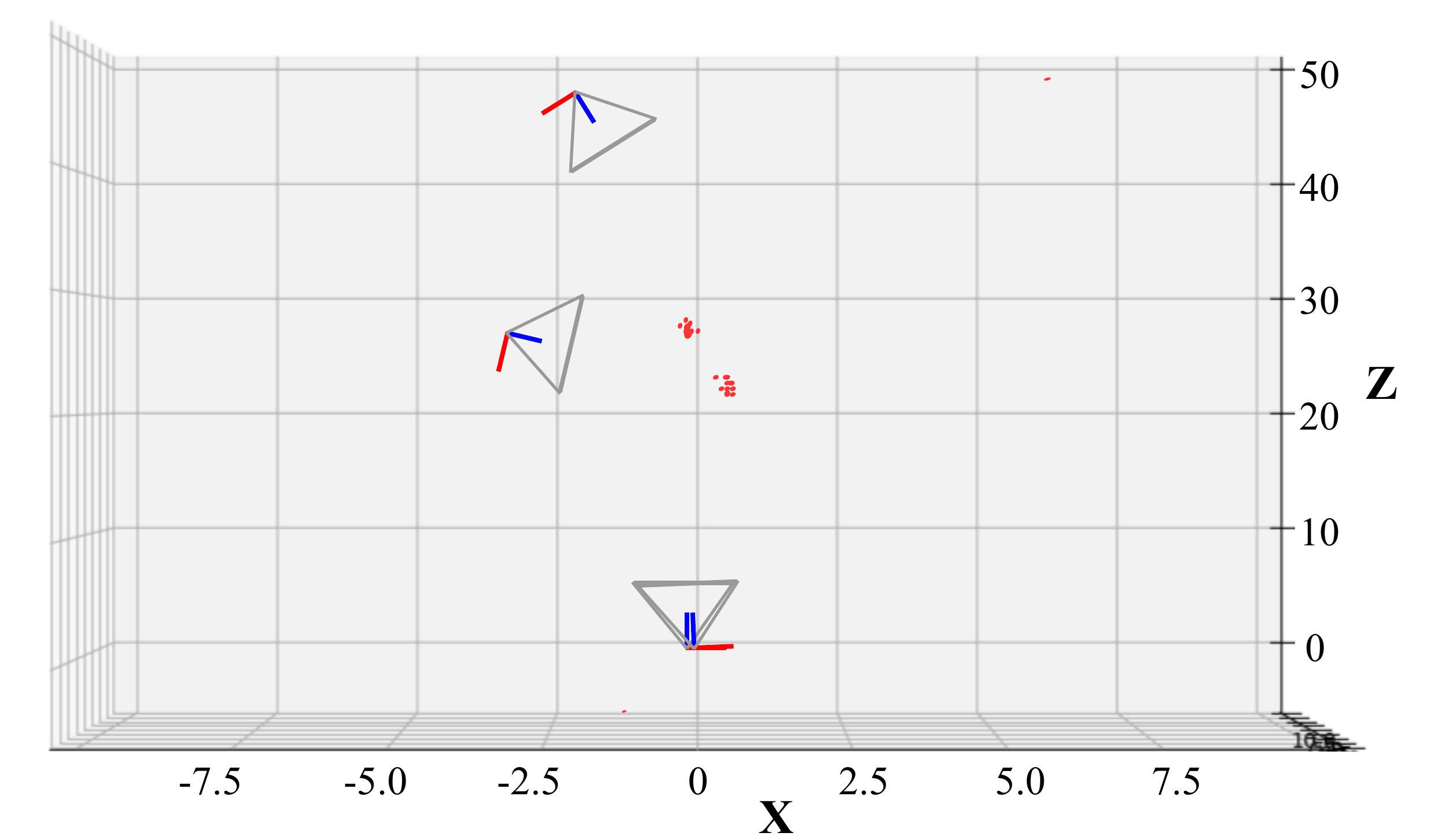}
    \label{fig:exp2c_r2r}
    }
\caption{An underwater experiment for 3D relative pose estimation using one leader and two follower robots.}
\label{fig:exp2_r2r}
\end{figure}

We perform several field experiments in human-robot collaborative setups; Figure~\ref{setup_b_r2r} shows the setup of a particular underwater experiment where we capture human body-poses from different perspectives to estimate the 6-DOF transformations of two follower robots relative to a leader robot. The leader robot is equipped with a stereo camera; hence, the 3D information of the human pose-based key-points is obtained by using stereo triangulation technique. Subsequently, we find the corresponding 2D projections on the follower robots' cameras using the proposed person ReId and key-point refinement processes. Finally, we estimate the follower-to-leader relative poses from their respective PnP solutions.

We present a particular snapshot in Figure~\ref{fig:exp2a_r2r}; it illustrates the leader and follower robots' perspectives and the associated human pose-based key-points. Subsequently, Figure~\ref{fig:exp20e_r2r} demonstrate the geometric validity of those key-point correspondences and the reconstructed 3D points are shown in Figure~\ref{fig:exp2b_r2r}. As seen, the estimated 3D structure is consistent with the mutually visible humans' body-poses. Finally, the estimated 6-DOF poses of the follower robots relative to the leader robot are shown in Figure~\ref{fig:exp2c_r2r}.

Such leader-to-follower pose estimates are useful in cooperative diver following~\cite{islam2018person}, convoying~\cite{shkurti2017underwater}, and other interactive tasks while operating in close proximity. The robust performance and efficient implementation of the proposed modules make it suitable for use by visually-guided underwater robots in human-robot collaborative applications. However, there are a few practicalities involved which can affect the performance; next, we discuss these aspects and their possible solutions based on our experimental findings.

\section{Operational Challenges and Practicalities}\label{sec:issues_r2r}
\textbf{Synchronized cooperation}: A major operational requirement of multi-robot cooperative systems is the ability to register synchronized measurements in a common frame of reference, which can be quite challenging in practice. For problems such as ours, an effective solution is to maintain a buffer of time-stamped measurements and register those as a batch using a temporal \textit{sliding window}. 
We effectively used such timestamp-based buffer schedulers~\cite{RosTime} in our implementation with reasonable robustness. However, the challenge remains in finding instantaneous relative poses, especially when both robots are in motion. Nevertheless, these aspects are independent of the choice of features/key-points for relative pose estimation and more generic requirements to multi-robot cooperative systems.       

\begin{figure*}[t]
\vspace{1mm}
\centering
\includegraphics [width=0.98\linewidth]{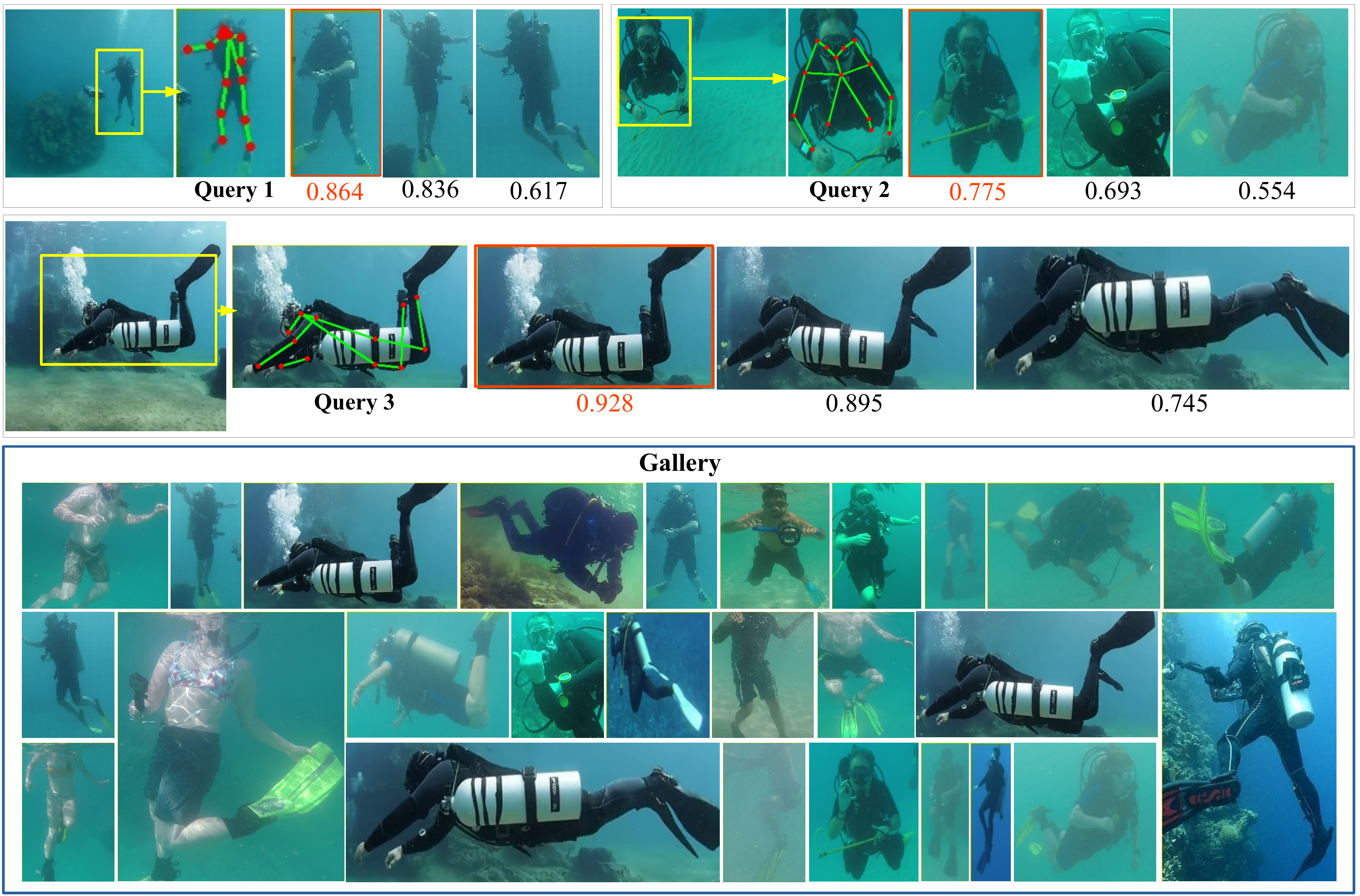}%
\vspace{-2mm}
\caption{Three test cases for the proposed person ReId module are shown: each query is matched with a gallery of candidate images (inside the blue box); the top three matches and respective scores are shown alongside the query image. The scores represent averaged SSIM scores for the mutually visible body-part BBoxes (see Section~\ref{sec:re_id}).     
}
\label{reid_r2r}
\end{figure*}%

\vspace{1mm}
\textbf{Trade-off between robustness and efficiency}: It is quite challenging to ensure a fast yet robust performance for visual feature-based body-pose estimation and person ReId on limited computational resources of embedded platforms. This trade-off between robustness and efficiency led us to design fast body-pose association and refinement modules. These efficient modules enable us to achieve an average end-to-end run-time of $375$-$420$ milliseconds for relative pose estimation on Jetson TX2. Note that the proposed person ReId and key-point refinement account for only $195$-$240$ milliseconds (up to nine humans in the scene). Hence, faster human body-pose detectors than OpenPose will (be able to) significantly boost the end-to-end run-time of the system.

In Section~\ref{impact}, we demonstrated that the proposed person ReId model performs reasonably well in practice despite its simplistic design. One operational benefit in our application is that the humans are seen at once from every perspective; hence, both their appearances and body-poses remain consistent. We provide a demonstration of this benefit in Figure~\ref{reid_r2r}; it shows three \emph{queries} for ReId where humans with similar suit/wearable are seen at various distances and orientations from the camera. Although the \emph{gallery} images contain several humans with similar appearances, we observe that the top three results correspond to best matches both in terms of human appearance and body-pose. We postulate that computing aggregated similarity scores on local pose-based BBoxes contribute to these results. Since the general-purpose person ReId problem is significantly harder and requires more sophisticated computational pipelines~\cite{zhao2017deeply,li2014deepreid}, our proposed module seems to take advantage of the body-pose consistency across viewpoints for a faster run-time.

\vspace{1mm}
\textbf{Number of humans and relative viewing angle}: We observed a couple of other practical issues during the field experiments. First, the presence of multiple humans in the scene helps to ensure reliable pose estimation performance. We found that two or more mutually visible humans are ideal for establishing a large pool of reliable correspondences. Additionally, the pose estimation performance is affected by the relative viewing angle; specifically, it often fails to find correct associations when the $\angle$\textit{leader-human-follower} is larger than (approximately) $135^{\circ}$. This results in a situation where the robots are exclusively looking at opposite sides of the person without enough common key-points. %Moreover, other than temporal lags, we did not observe significant deviations in pose estimation performance with an increasing number of robots within this viewing angle; note that we used up to three follower robots in our experiments.

\section{Concluding Remarks}
In this chapter, we explored the feasibility of using human body-poses as markers to establish reliable multi-view geometric correspondences and to eventually solve the robot-to-robot relative pose estimation problem. 
We adopted a \emph{leader-follower} framework, where at first, the leader robot visually detects and triangulates the human pose-based key-points by using an existing library named OpenPose. Afterward, the follower robots match the corresponding 2D projections on their respective calibrated cameras and find the relative poses by solving their respective PnP problems. In the proposed method, we designed an efficient person ReId technique for associating the mutually visible humans in the scene. Additionally, we presented an iterative optimization algorithm to refine the associated key-points based on their local structural properties in the image space. We demonstrated that these novel modules are essential to establish accurate geometric correspondences and to ensure robust end-to-end performance at a fast rate.

We performed several field experiments in terrestrial and underwater environments to validate the proposed robot-to-robot relative pose estimation system. This is particularly useful for underwater robots operating in challenging visual conditions when a global positioning sensor (\eg~USBL) is not available. We demonstrated its utility as an \textit{implicit} way to maintain spatial coordination by cooperative robots and discussed the relevant operational considerations. In the next chapter, we present a system that allows divers to \emph{explicitly} interact with their companion robots to communicate live instructions during a mission.   
\chapter{Underwater Human-to-Robot Communication System}\label{hrc}
The ability to interact with companion divers and understand their instructions is essential for underwater robots operating in a cooperative mission. A simple yet robust human-to-robot communication system~\cite{dudek2007visual, xu2008natural, chiarella2015gesture} is particularly important in collaborative exploration and data collection processes. Moreover, in complex missions such as surveillance and rescue operations, robots need to dynamically adjust their mission parameters by regularly interacting with the diver~\cite{islam2018person,dudek2007visual}. The typical instructions communicated to a robot can be to invoke following, tracking a certain target, recording snapshots, taking samples, changing motion parameters, etc. 

Such direct human supervision eliminates the necessity of teleoperation (and teleoperators). Otherwise, since Wi-Fi or electromagnetic communication is severely degraded underwater~\cite{dudek2008sensor}, the current task needs to be interrupted, and the robot needs to be brought to the surface for reprogramming, which is inconvenient and often expensive in terms of time and physical resources. Therefore, dynamic reconfiguration of mission parameters based on human input while the robot is underwater is a practical, simpler, and more efficient approach. Modulating robot motion based on human input in the form of speech, free-form gestures, or keyboard interfaces has been explored extensively for terrestrial environments~\cite{coronado2017gesture,chen2015hand,wolf2013gesture}. However, these communication modules are not readily applicable in the underwater domain due to environmental and operational constraints~\cite{islam2018person,dudek2008sensor}. As more feasible alternatives, several vision-based interaction systems have been proposed~\cite{dudek2007visual,xu2008natural,chiarella2015gesture}, which we elaborately discuss in Section~\ref{rel_robochatgest}.

In this chapter, we address the limitations of existing vision-based diver-to-robot communication systems by presenting a real-time programming and parameter reconfiguration method for Autonomous Underwater Vehicles (AUVs). By using a set of intuitive and meaningful hand-gestures to instruction mapping rules, we develop a syntactically simple \emph{visual language} that is computationally more efficient than a complex, grammar-based approach. In the proposed system, a Convolutional Neural Network (CNN)-based model is designed for fast hand gesture recognition; we also provide the option to use state-of-the-art (SOTA) deep visual detectors (\eg, SSD~\cite{liu2016ssd}, Faster R-CNN~\cite{renNIPS15fasterrcnn}) for more reliable hand gesture recognition. Subsequently, a Finite-State Machine (FSM)-based deterministic model performs efficient gesture-to-instruction mapping and further improves the robustness of the interaction scheme. One key aspect of this system is that it can be easily adopted by divers for communicating simple instructions to AUVs without carrying and manipulating artificial tags or markers to program complex language rules during underwater missions. The detailed language mapping rules and implementation details are provided in Section~\ref{human_robot}.

Subsequently, we demonstrate the robustness, efficiency, and portability of our proposed human-to-robot communication system through simulation and by real-world field experiments. We also validate its operational and usability benefits compared to other existing visual languages by a user interaction study. The experimental validations and analyses of the results are presented in Section~\ref{perf_gest}.

\section{Existing Systems: Utility and Limitations}\label{rel_robochatgest}
In traditional systems, a set of \emph{fiducial markers} is used by human divers to communicate simple instructions to their companion robots. These systems typically use a predefined mapping between various markers and atomic instructions as \emph{language rules}. 
The most commonly used fiducial markers are those with square, black-and-white patterns that are easily detectable as visual targets, such as AR-Tags~\cite{fiala2005artag} and April-Tags~\cite{olson2011apriltag}, among others. Circular markers with similar patterns such as the photomodeler coded targets and Fourier tags~\cite{sattar2007fourier} have also been used in practice.  
%These markers are popular because their visual detection is easy and robust even in noisy conditions. 
%Besides, the simple one-to-one instruction mapping is simple and easily programmable to the robot. Nevertheless,   
%Nevertheless, the use of tags can be avoided for tethered operations, where a diver can   
%A similar system based free-form gestures is also proposed   
%For , A gesture-based framework for underwater visual servo control was introduced in~\cite{dudek2005visually}, where a human operator on the surface was required to interpret human gestures and modulate the corresponding robot movements. 

\begin{figure}[ht]
\centering
\vspace{3mm}
\includegraphics [width=0.85\linewidth]{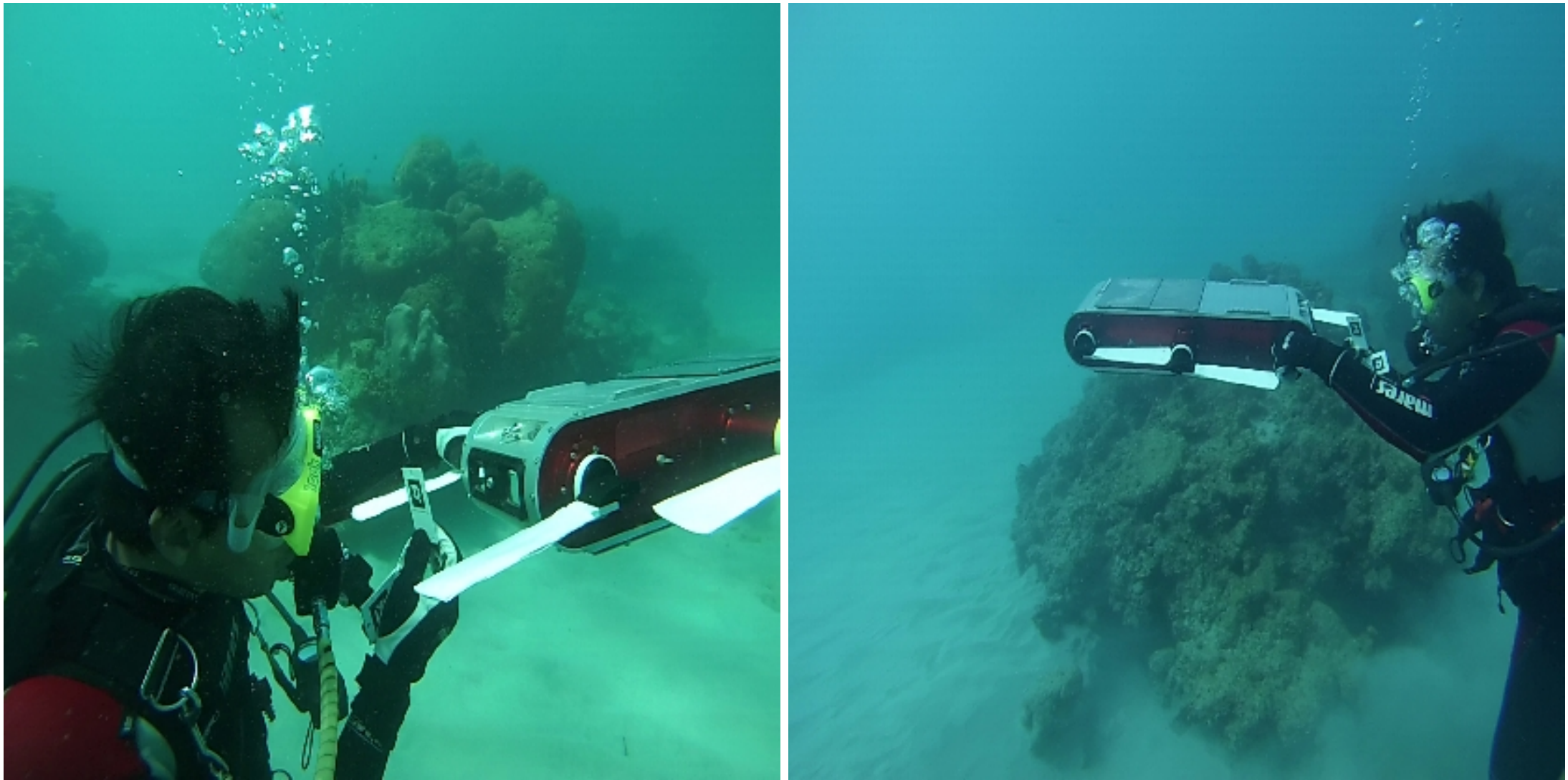}%
\vspace{-1mm}
\caption{A diver is using AR-Tags to communicate instructions to an underwater robot via RoboChat.}%
\label{fig:tag}
\end{figure}%

Grammar-based diver-robot communication systems use more expressive language rules. For instance, RoboChat~\cite{dudek2007visual} allows programming both atomic action commands as well as procedural statements by using sequences of visual symbols encoded into AR-Tag markers. Each marker represents one symbolic {\tt token}, and specific sequences of such {\tt tokens} constitute various programs to modulate robot behavior. Since the AR-Tag markers can be robustly and efficiently detected in noisy visual conditions, RoboChat eliminates the need for a human operator on the surface for interpreting instructions to the robot, as done in many tethered operations~\cite{dudek2005visually}. As shown in~Figure~\ref{fig:tag}, divers can reliably use it for dynamic reconfiguration of motion parameters on-the-fly.

Despite its utility, RoboChat has two major limitations. First, because a separate marker is required for each token, a large number of marker cards need to be securely carried during the mission, and specific cards are required to be searched and ordered to formulate a syntactically correct instruction; this whole process imposes a rather high cognitive load on divers. Secondly, the symbol-to-instruction mapping is not intuitive, which makes it inconvenient for rapidly programming a robot. The first limitation is addressed in~\cite{xu2008natural}, in which a set of discrete motions of a pair of fiducial markers is interpreted as language rules. Different features such as shape, orientation, and size of these motions are extracted and mapped to various instructions. Since more information is embedded in each token, a large number of instructions can be supported by using only two fiducial markers. However, this method introduces additional computational overhead to track the markers' shape, orientation, and motion trajectory. These are extremely challenging in real-world scenarios as both the divers and robots are suspended in a six-degrees-of-freedom (6-DOF) environment. Besides, the symbol-to-instruction mapping remains counter-intuitive.

We attempt to address these limitations by 
designing \textbf{RoboChatGest}~\cite{islam2018dynamic}, a hand gesture-based diver-to-robot communication system that is syntactically simpler and more intuitive than existing grammar-based systems. Moreover, since the conventional method of communication between scuba divers is with hand gestures, the proposed system can be easily adopted by divers for communicating instructions to underwater robots without using fiducial markers or requiring memorization of complex grammars. 
Although several hand gesture-based HRI frameworks exist for terrestrial and aerial robots~\cite{coronado2017gesture,chen2015hand,wolf2013gesture,naseer2013followme}, the design and feasibility of such systems for underwater human-robot cooperation have not yet been explored in-depth~\cite{islam2018person}. 

%we to the simplicity of the language rules and associated cognitive load (on the users), and the computational efficiency are      

%The robust visual detection of hand gestures is a challenge, which we    

%There exist several hand gesture-based HRI frameworks \cite{coronado2017gesture,chen2015hand,wolf2013gesture} for terrestrial robots. Additionally, recent visual hand gesture recognition techniques \cite{molchanov2015hand,neverova2014multi} based on CNNs are highly accurate and robust to noise and visual distortions. A number of such visual recognition and tracking techniques have been successfully used for underwater tracking \cite{shkurti2017underwater} and have proven to be more robust than other purely feature-based methods \cite{islam2018person}. However, the feasibility of these models for hand gesture-based human-robot communication has not yet been explored in-depth, which we attempt to address in~\cite{islam2018dynamic}. Moreover, we demonstrate that off-the-shelf deep visual detection models~\cite{tfzoo} can be utilized in our framework to ensure robust performance.

\section{Proposed System: RoboChatGest}
\label{human_robot}
RoboChatGest~\cite{islam2018dynamic} explores the challenges involved in the design and development of a hand gesture-based human-robot communication system for underwater robots. In particular, a simple interaction framework is developed where a diver can use a set of intuitive and meaningful hand gestures to program the accompanying robot or to reconfigure program parameters. 
%A CNN-based robust hand gesture recognizer is used with a simple set of gesture-to-instruction mapping. A finite-state machine-based interpreter ensures predictable robot behavior by eliminating spurious inputs and incorrect instruction compositions.
It is built on several components: the hand gesture to instruction mapping rules, multiple hand gesture recognition modules, and a FSM-based instruction decoder. Their design and computational aspects are elaborately discussed in the following sections.

\begin{figure*}[ht]
\centering
\vspace{3mm}
\includegraphics [width=0.98\linewidth]{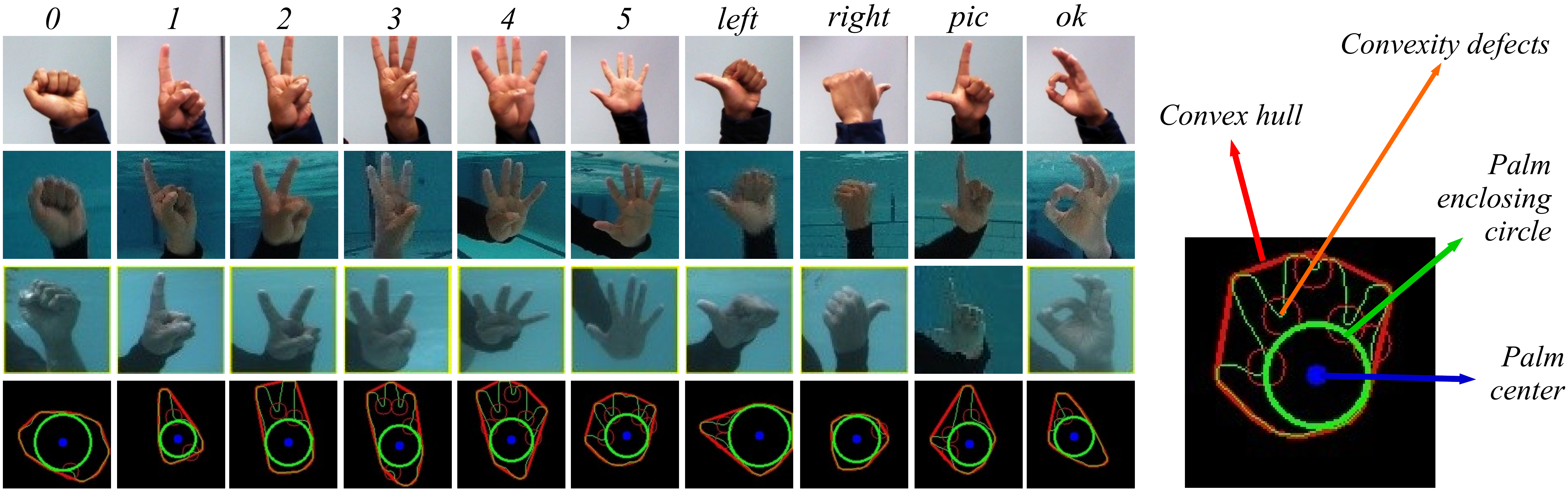}%
\vspace{-1mm}
\caption{The first three rows on the left show a few sample training images for the ten classes of hand gestures used in RoboChatGest; the bottom row shows the expected hand contours with different curvature markers for each class of gestures. The annotated curvature markers for a particular example are shown on the right.}
\label{data_gest}
\end{figure*}

\begin{figure}[t]
\centering
\includegraphics [width=0.7\linewidth]{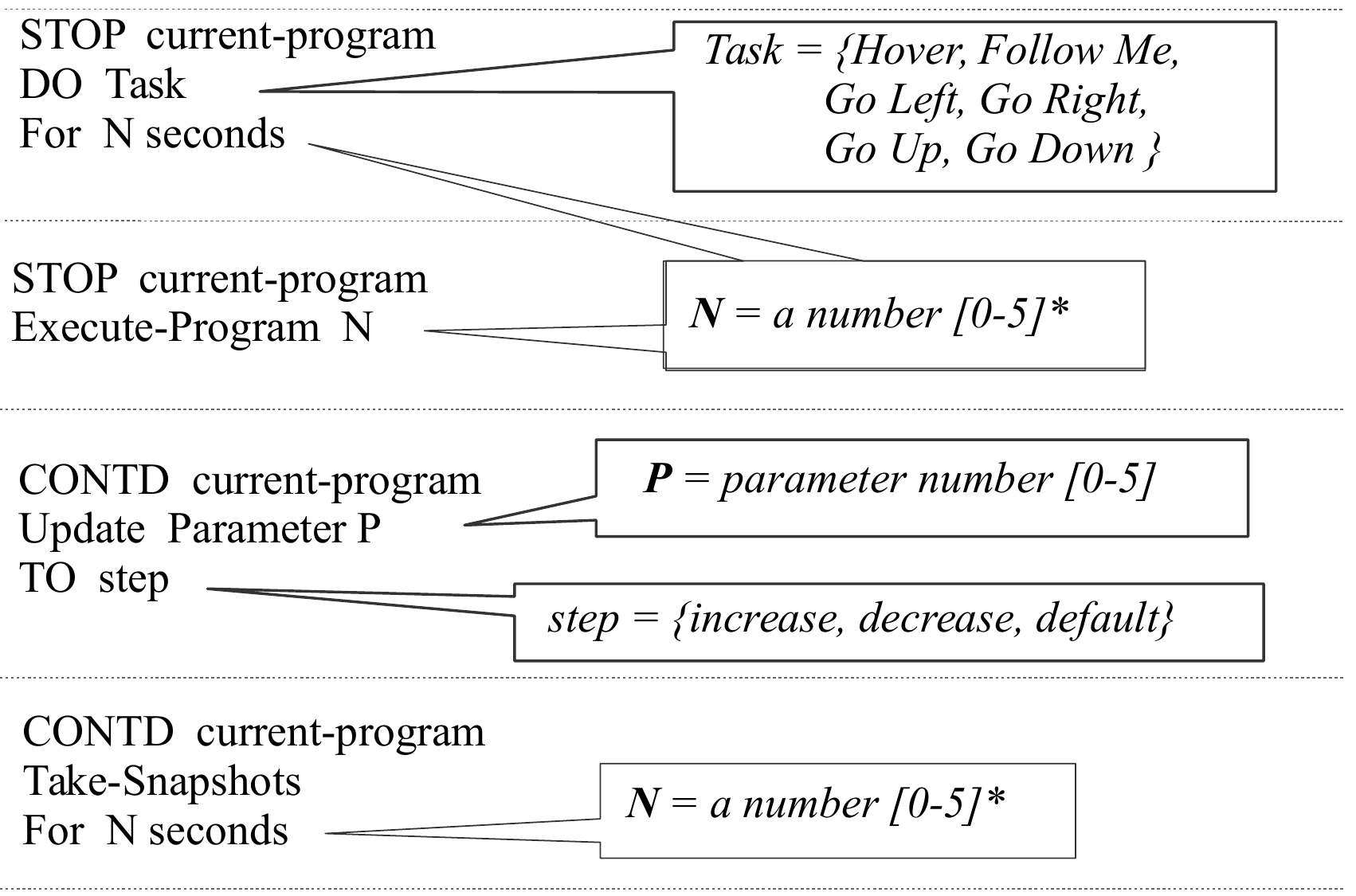}%
\vspace{-1mm}
\caption{Formation of various `task switching' and `parameter reconfiguration' instructions that are currently supported by RoboChatGest.}
\label{fig:ins_rule}
\end{figure}

\subsection{Hand Gestures to Instructions Mapping Rules}
\label{sec:gesture-to-token}
Intending to design simple yet expressive language rules, we choose a small collection of visually distinctive and intuitive hand gestures as symbolic {\tt tokens}. 
We use the ten {\tt gesture-tokens} shown in Figure~\ref{data_gest}, various combinations of which, formed with both hands, are mapped to specific {\tt instruction-tokens}. We also choose the admissible instruction sets in a way that can be easily interpreted and adopted by divers. As illustrated in Figure~\ref{fig:ins_rule}, we concentrate on the following two sets of instructions: 

\begin{itemize}
\item \textbf{Task switching} instructions are communicated to a robot to stop its current program execution and start a new task specified by the diver such as hovering, following, or moving left/right/up/down, etc. These commands are atomic behaviors that the robot is capable of executing. An optional argument can be provided to specify the duration of the new task (in seconds). An operational requirement is that the desired programs need to be numbered and known to the robot beforehand.

\item \textbf{Parameter reconfiguration} refers to instructing a robot to continue its current program with updated parameter values. This enables underwater missions to continue unimpeded, without interrupting the current task or requiring the robot to be brought to the surface. Here, the requirement is that the \textit{tunable} parameters and their choice of values need to be specified to the robot beforehand. %The robot can also be instructed to take pictures (for some time) while executing its current program. 
\end{itemize}

\begin{figure}[t]
\centering
\includegraphics [width=0.95\linewidth]{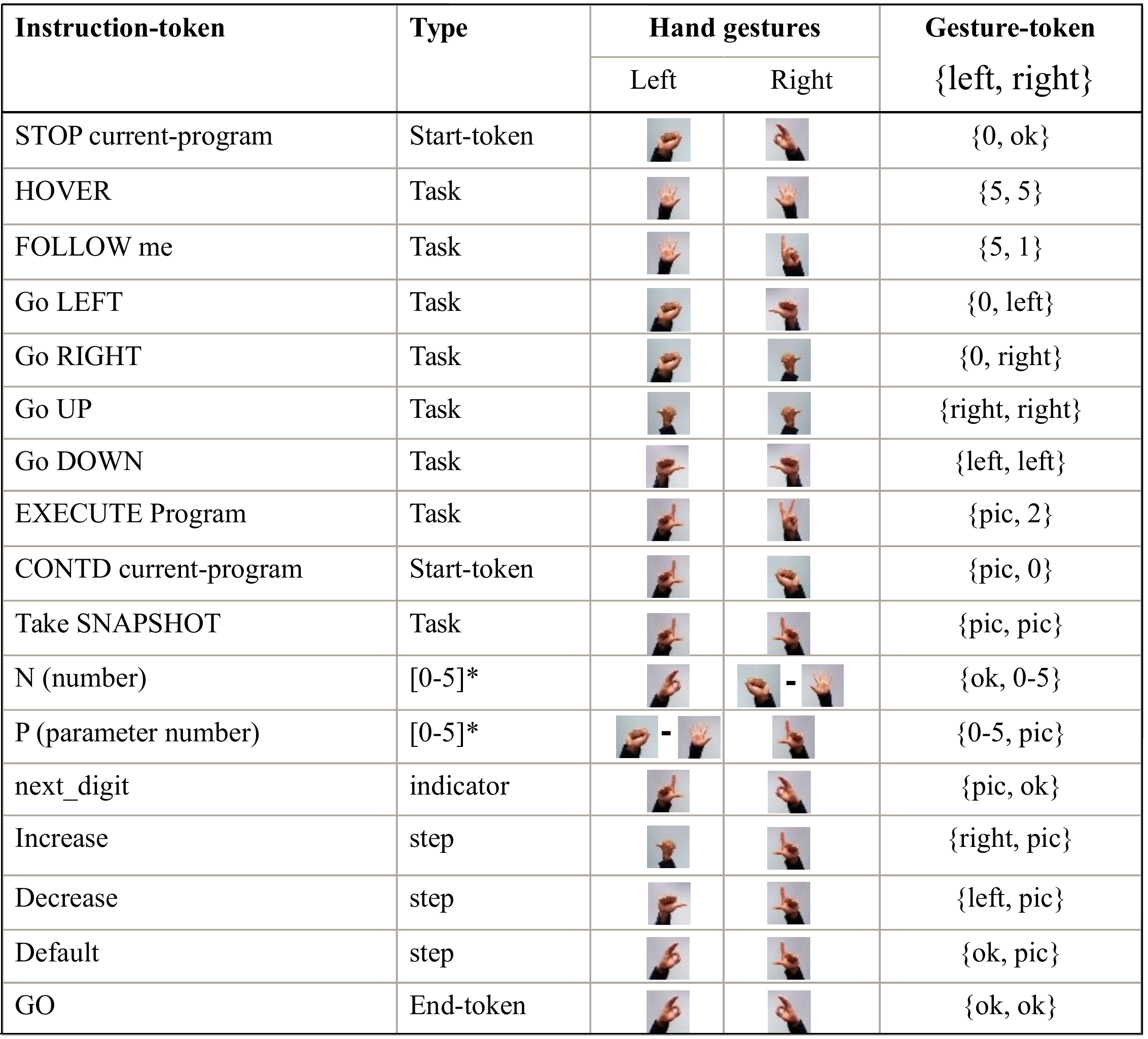}%
\vspace{-3mm}
\caption{The one-to-one mapping of {\tt gesture-tokens} to {\tt instruction-tokens} used in RoboChatGest.}
\label{fig:ins_map}
\end{figure}

\begin{figure}[t]
\centering
\includegraphics [width=0.8\linewidth]{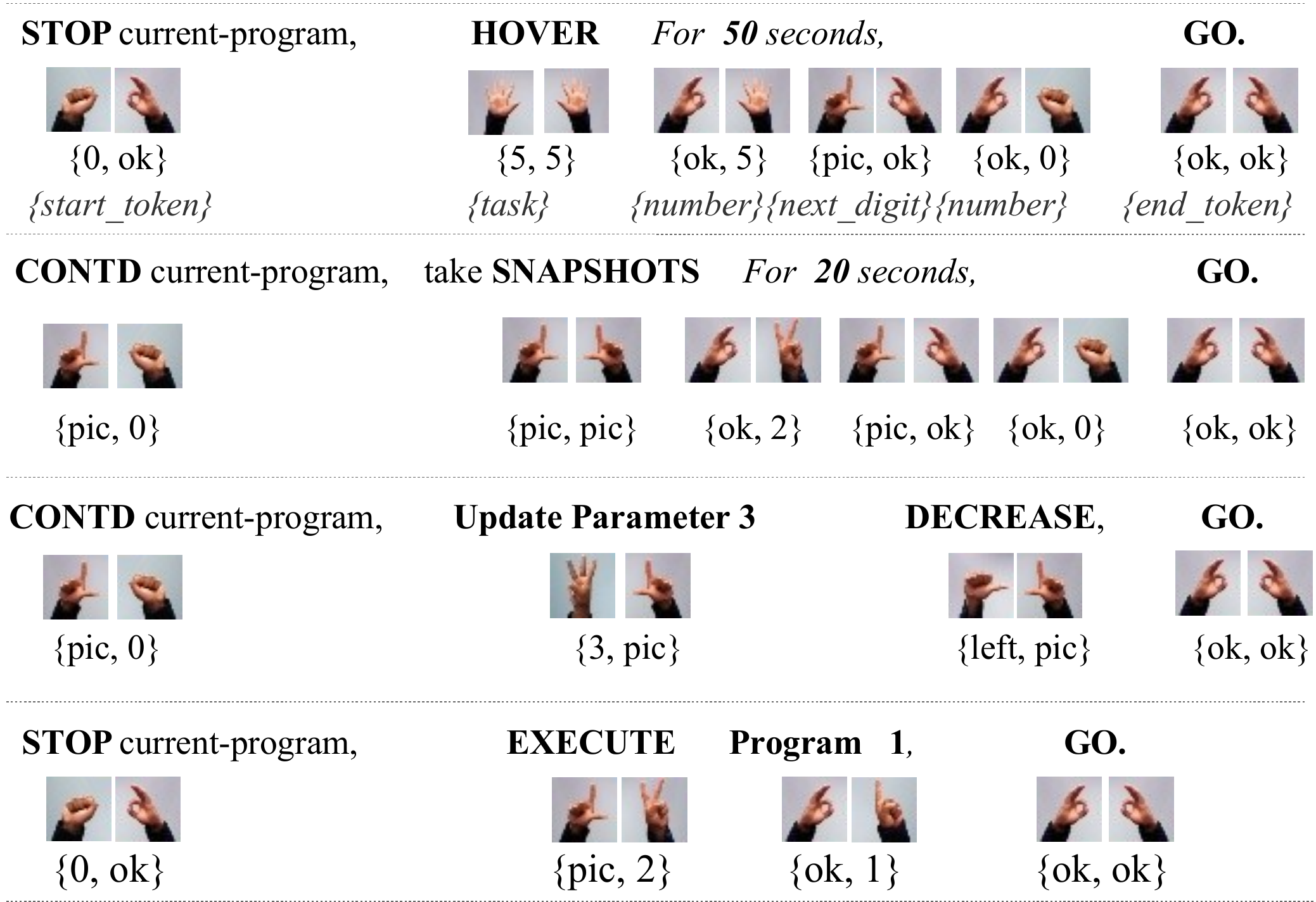}%
\vspace{-1mm}
\caption{A few examples of valid instruction generation by sequences of {\tt gesture-tokens} of the form: ({\tt start-token}, {\tt instruction-tokens}, {\tt end-token}).}
%\vspace{1mm}
\label{fig:ins_exm}
\end{figure}

RoboChatGest supports a variety of instructions for task switching and parameter reconfiguration, which can be extended by simply modifying a user-editable configuration file for application-specific use. 
Also, the instruction formats are carefully designed so that the robot triggers executable programs only when intended by the diver. This is done by attributing specific hand gestures as sentinels, \ie, {\tt start-tokens} or {\tt end-tokens}. Figure~\ref{fig:ins_map} provides the full list of {\tt gesture-token} to {\tt instruction-token} mapping rules that are adopted in our implementation. 
Moreover, Figure~\ref{fig:ins_exm} shows several examples that demonstrate how sequences of {\tt gesture-tokens} of the form ({\tt start-token}, {\tt instruction-tokens}, {\tt end-token}) can form complete instructions.  

\subsection{Hand Gesture Recognition Modules}
\label{HandGest}
Fast and accurate hand gesture recognition is the primary computational requirement of using RoboChatGest is real-world applications. 
Aiming a simplistic design, we adopt a two-step recognition process where, first, the hand gestures are localized in the image space by a `hand contour'-based region selector; then, the segmented gestures are identified by a shallow CNN-based classifier. This minimal design facilitates fast hand gesture recognition on resource-constrained single-board platforms. Nonetheless, RoboChatGest provides the flexibility to use more powerful hand gesture recognition models for high-end platforms. %The following sections discuss both of these options and their methodological details 

\begin{figure}[ht]
\centering
    \subfigure[A simple RoI selection mechanism for hand gestures in image space.]{
        \includegraphics[width=0.72\textwidth]{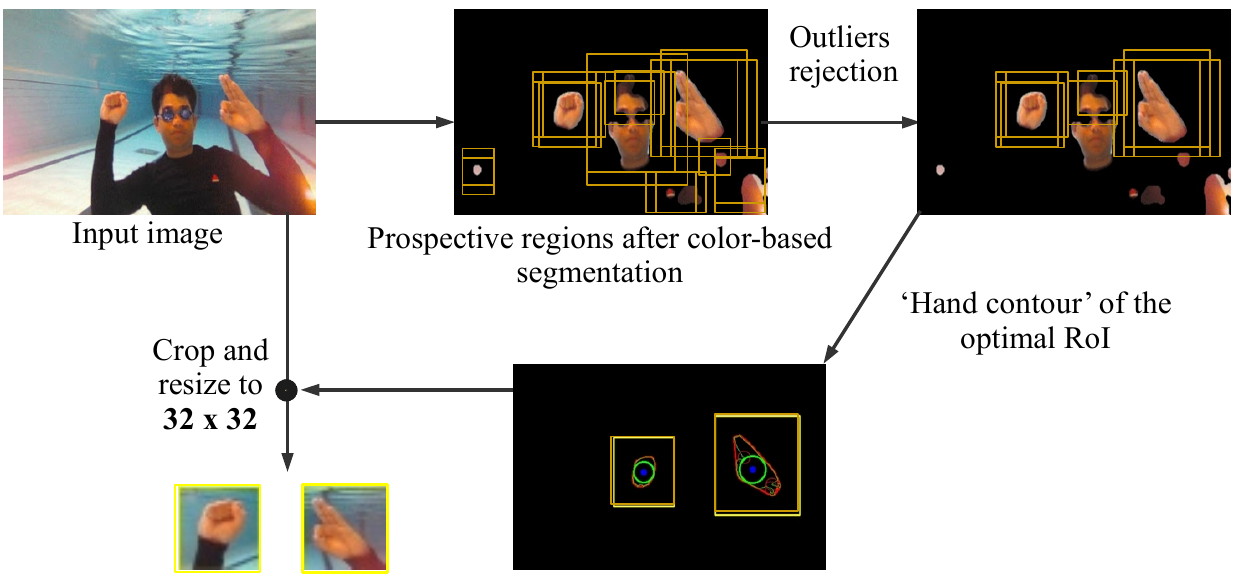}
        \label{fig:reg}
    }
    
    \vspace{2mm}
    \subfigure[An efficient CNN-based model for hand gesture classification.]{
        \includegraphics[width=0.98\textwidth]{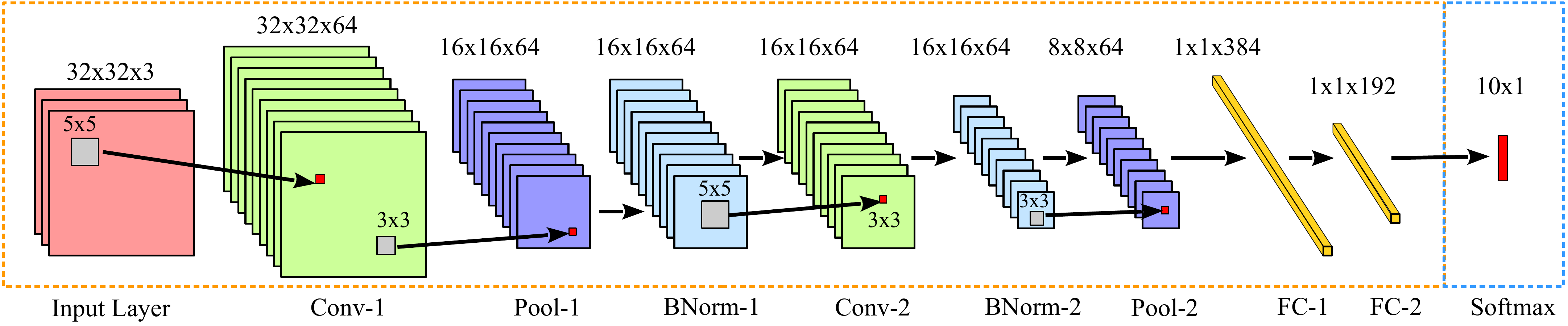}
        \label{fig:cnn}
    }%
\vspace{-1mm}
\caption{A fast hand gesture recognition module:~\subref{fig:reg} first, region selection is performed by color-based filtering followed by hand contour matching;~\subref{fig:cnn} subsequently, the selected RoIs are classified by a CNN-based model. This minimal design of a simple $32\times32$ RoI selector and a shallow 10-class classifier facilitates fast operation on single-board machines. Note that, any hand gesture recognition module can be independently integrated into the RoboChatGest system.}
\label{fig:hand_gest_recog}
\end{figure}

\vspace{1mm}
\textbf{Hand contour matching for region selection}: 
As illustrated in Figure~\ref{fig:reg}, we exploit classical image processing techniques to select prospective regions of interest (RoIs) in RGB image space. 
First, skin color-based filtering~\cite{oliveira2009skin} is performed on the Gaussian blurred input image (note that, if the diver is to wear gloves, the color thresholding range needs to be adjusted accordingly). Then, contour boundaries of these segmented image regions are identified and respective convex hull, convexity defects, and other curvature points are extracted; we refer to~\cite{yeo2015hand} for details about the properties and significance of these curvature markers. 
Subsequently, the outliers are rejected using cached information about the scale and location of hand gestures detected in the previous frame (if available). The hand contours of potential regions are then matched with a bank of ground truths (see Figure~\ref{data_gest}). Finally, the image regions for both hands are selected by sorting the proximity values of the closest contour match.

\vspace{1mm}
\textbf{CNN-based model for hand gesture classification}: Following region selection, the cropped and resized $32\times 32$ image patches are fed to a CNN-based model for classification. Figure~\ref{fig:cnn} shows the architecture and detailed specification of this model. It contains only two composite layers (\ie, convolutional layer followed by spatial down-sampling and batch normalization) for hierarchical feature extraction. The extracted features are exploited by two shallow fully connected layers to learn decision hyper-planes for the $10$-class hand gesture classification. This shallow network has only $1755$K parameters, and provides real-time inference on single-board devices; we provide details on its performance evaluation and feasibility analysis in Section~\ref{perf_gest}. 
%within the distribution of training data. Finally, a soft-max layer provides the output probabilities for each class, given the input data. Note that similar CNN-based models are known to perform well for small-scale (\textit{i.e.}, $10$-class classification) problems that are analogous to ours. %The dimensions of each layer and associated hyper-parameters are specified in Figure~\ref{Arch}; details about the training process will be provided in Section~\ref{training}.     

\textbf{Applicability of SOTA object detectors}: Any standard hand gesture recognition module can be integrated into the RoboChatGest pipeline for platform-specific use. In particular, we demonstrate that standard deep visual models such as Faster RCNN (Inception v2) and SSD (MobileNet v2) can be effectively used in RoboChatGest on high-end robot CPUs and embedded GPU platforms. 
Unlike the two-step approach shown in Figure~\ref{fig:hand_gest_recog}, these models learn hand gesture localization and classification simultaneously on RGB image space. We already discussed their methodological details for diver detection in Section~\ref{tf_obj_detect} and Appendix~\ref{ApenB}. We follow a similar procedure to train these models end-to-end by considering the ten hand gestures as separate object categories. Further details of their training and evaluation processes are presented in Section~\ref{setup:gest} and Section~\ref{setup:eval}, respectively.

%Although these models offer at a few FPS rates on ,  inference           
%An operational convenience of hand gesture-based programming is that the robot stays in `hover' mode, and the overall operation is not as time-critical as in the diver-following scenario. Hence, deeper and denser models can be applied for more robust hand gesture recognition performance. In particular, we demonstrate that SOTA deep object detection models, or any standard model that provided only a few FPS inference, can be ported into the RoboChatGest pipeline for effective use. In our implementation, hand gesture recognizers based on Faster RCNN (Inception v2) and SSD (MobileNet v2) are provided as optional .        
%Specifically, we explore the applicability of the SOTA deep visual models for hand gesture recognition and try to balance the trade-offs between accuracy and run-time. We use two object detection models that are discussed in Section~\ref{tf_obj_detect}: Faster RCNN with Inception v2 and Single Shot MultiBox Detector (SSD) with MobileNet v2; their methodological details are provided in Appendix~\ref{ApenB}. 

\begin{figure*}[ht]
\centering
\includegraphics [width=0.98\linewidth]{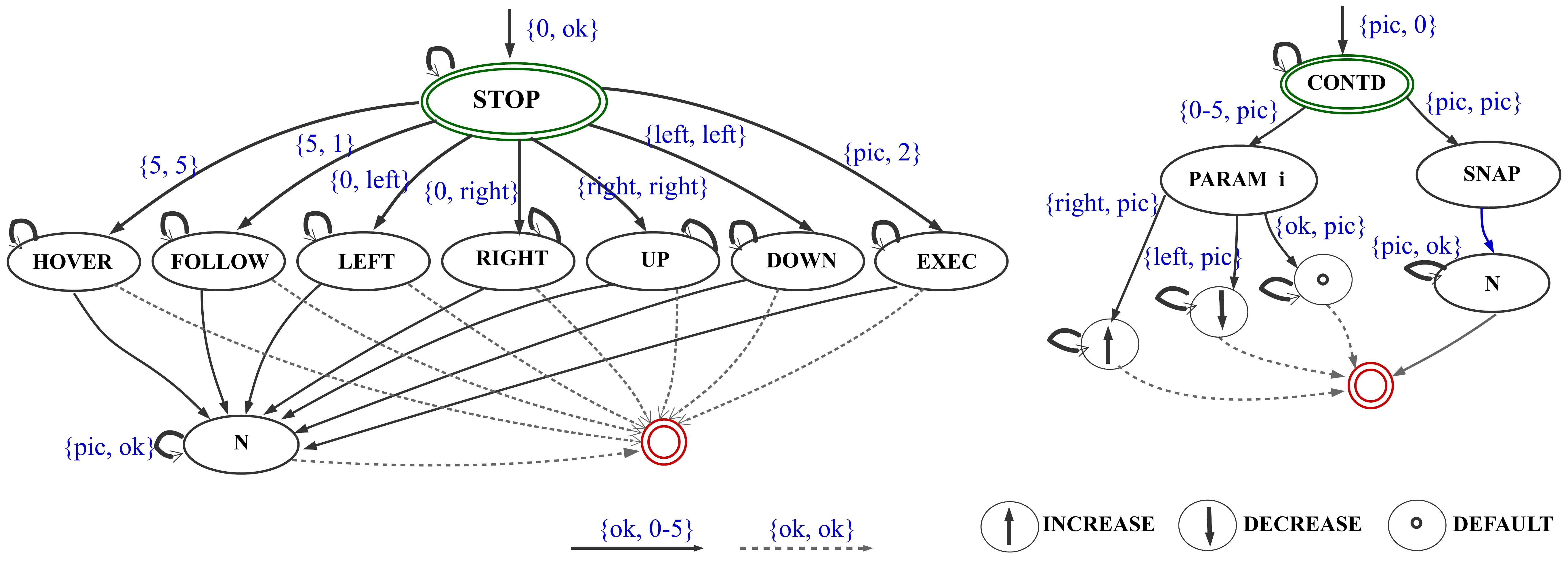}
\caption{FSM-based deterministic decoding of both kinds of instruction sets.}
\label{Fsm}
\end{figure*}

\subsection{FSM-based Instruction Decoder}\label{sec:fsm}
We design an FSM-based deterministic model to decode sequences of hand {\tt gesture-tokens} and generate complete instructions.  
As illustrated in Figure~\ref{Fsm}, the transitions between the states (\ie, {\tt instruction-tokens}) are defined as functions of {\tt gesture-tokens} based on the rules defined in Figure~\ref{fig:ins_map}. We ensure robustness of the instruction decoding by adopting the following transition rules: 
\begin{itemize}
\item State transitions are activated only if the corresponding {\tt gesture-tokens} are detected for $10$ consecutive frames. Therefore, an incorrect recognition has to happen $10$ consecutive times to generate an incorrect {\tt instruction-token}, which is highly unlikely. Additionally, it helps to discard noisy tokens which may be detected when a diver changes from one hand gesture to the next.

\item There are no transition rules for incorrect {\tt gesture-tokens}. Since the mapping is one-to-one, a wrong instruction cannot be generated even if a diver mistakenly performs some inaccurate gestures because there are no transition rules other than the correct ones at each state.  
\end{itemize}

\begin{figure}[ht]
\centering
\includegraphics[width=0.95\linewidth]{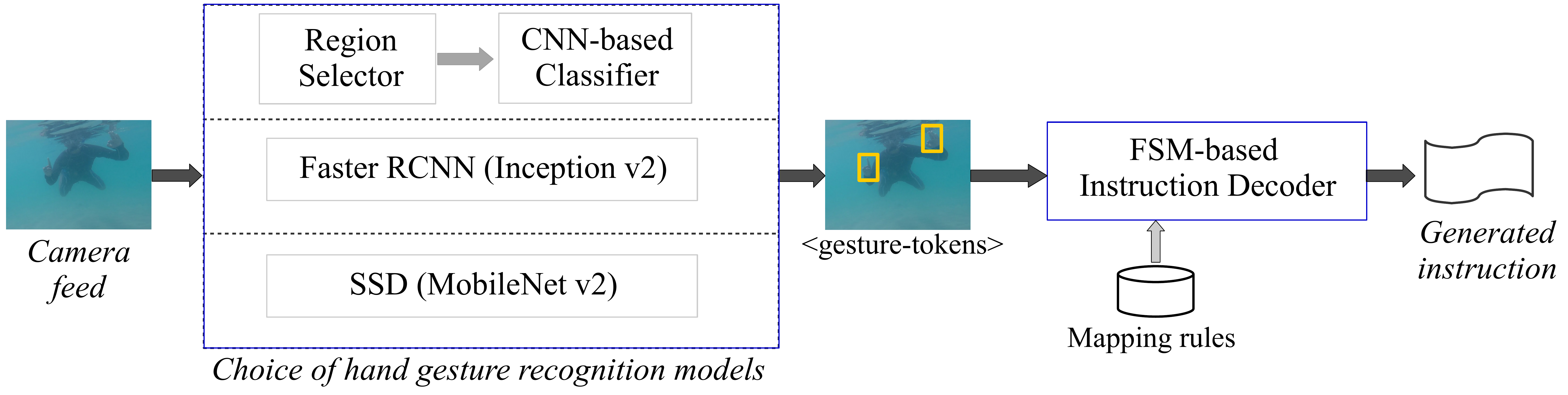}%
\vspace{-1mm}
\caption{Outline of the proposed RoboChatGest system~\cite{islam2018dynamic}.}
\label{Proc_gest}
\end{figure}

\section{Experimental Validation}\label{perf_gest}
The end-to-end pipeline of RoboChatGest is outlined in Figure~\ref{Proc_gest}. As discussed, sequences of hand gestures performed by a diver are recognized as {\tt gesture-tokens}, and then mapped to corresponding {\tt instruction-tokens}; valid sequences of such tokens are then decoded into complete instructions. Since the mapping rules are one-to-one, the FSM-based instruction decoder is deterministic and operates at a constant rate. Hence, the performance variability of RoboChatGest mostly depends on the hand gesture recognition module. In the following sections, we present the training and evaluation processes of the three choices of hand gesture recognition modules in RoboChatGest and analyze their performance trade-offs. Subsequently, we demonstrate the 
effectiveness of RoboChatGest through several field experiments and validate its usability benefits by a user interaction study.

\begin{figure}[ht]
\centering
    \subfigure[Convergence behavior based on training loss and classification accuracy.]{
        \includegraphics[width=0.46\textwidth]{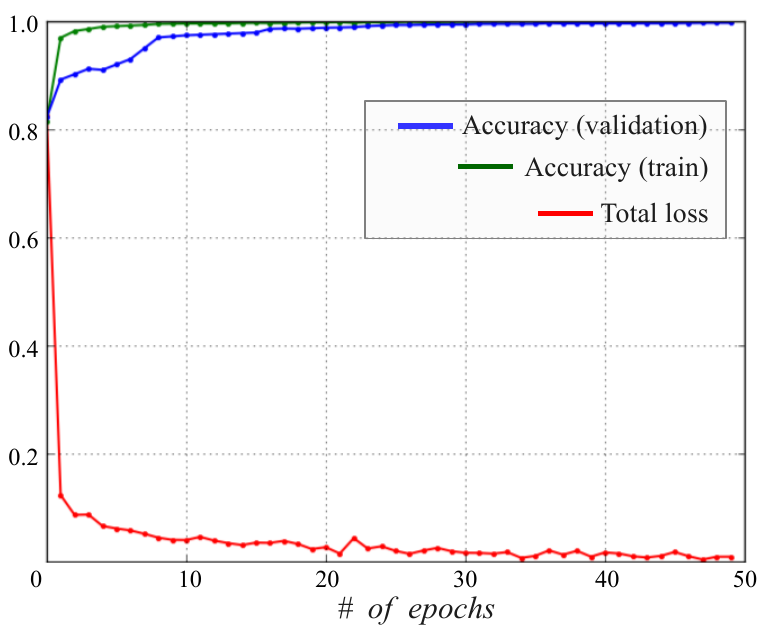}
        \label{fig:cnn_gest_conv}
    }~
    \subfigure[Confusion matrix for test-set accuracy across all hand gesture categories.]{
        \includegraphics[width=0.46\textwidth]{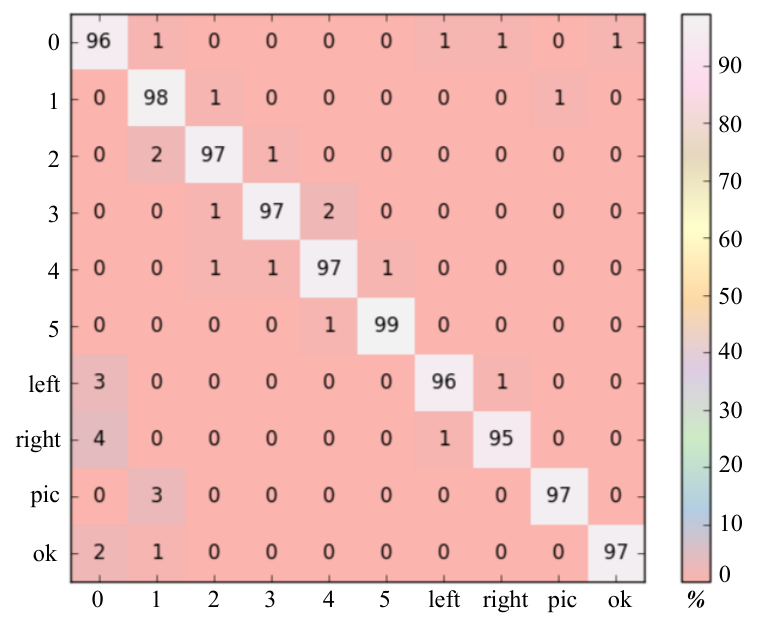}
        \label{fig:cnn_gest_conf}
    }
\vspace{-1mm}
\caption{Training behavior and test-set evaluation of the CNN-based model (shown in Figure~\ref{fig:hand_gest_recog}) for hand-gesture classification in RoboChatGest.}
\label{fig:cnn_gest_train}
\end{figure}

\subsection{Data Preparation and Model Training}\label{setup:gest}
We conducted several experiments in pools and oceans to mimic human-robot cooperative scenarios for data collection~\cite{islam2018understanding}.   
We provided details on our field experimental setups and discussed some of the key aspects of 
preparing image-based datasets for large-scale supervised training in Section~\ref{data_col_dive}. For RoboChatGest, in particular, snapshots of divers performing various hand gestures to underwater robots over a diverse set of scenarios are accumulated. A dataset of $35$K RGB images are thus prepared for training; two additional sets with $4$K and $1$K images are prepared for validation and testing, respectively. The images are then BBox-annotated for ten hand gesture categories considered in RoboChatGest. 
%,  by following a transfer-learning pipeline presented in Section~\ref{ddd_train}.   

The raw RGB images and BBox labels are used to train the Faster RCNN (Inception v2) and SSD (MobileNet v2) models with their standard TensorFlow APIs~\cite{tfzoo} by following the notion of transfer learning. In contrast, the BBox image regions are cropped and resized to $32\times32$ for training the CNN-based classifier model from scratch. We use TensorFlow~\cite{abadi2016tensorflow} libraries to implement the training pipeline on a Linux machine with one NVIDIA\texttrademark{} GTX 1080 GPU. 
Figure~\ref{fig:cnn_gest_conv} illustrates its learning behavior in terms of training loss and classification accuracy. It reaches a maximum validation accuracy of $0.986$ after $50$ epochs of training with a batch-size of $128$. Moreover, Figure~\ref{fig:cnn_gest_conv} demonstrates its test-set accuracy across all hand gesture categories.

\begin{figure}[ht]
\centering
    \subfigure[Instructing a robot to stop executing its current program and hover.]{ \includegraphics[width=0.8\linewidth]{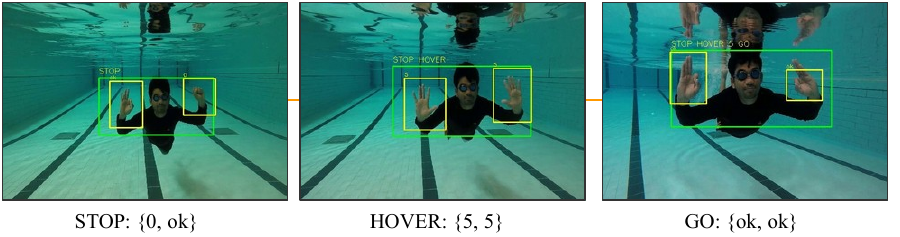}
    }
    \subfigure[Instructing a robot to continue its program but increase the value of parameter \#2 (by one step).]{
        \includegraphics[width=0.98\linewidth]{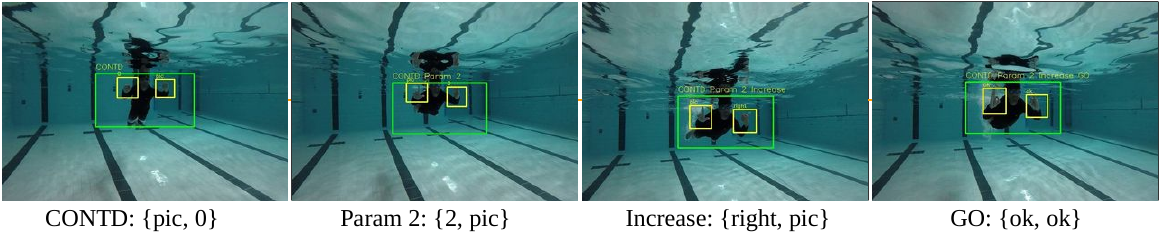}
    }%
 \vspace{-2mm}
 \caption{Snapshots of hand gesture-based robot programming by RoboChatGest; the yellow BBoxes represent {\tt gesture-tokens} detected by the CNN-based model.}
 \label{fig:robogest}
\end{figure}

\subsection{Interaction Scenario and Model Evaluation}\label{setup:eval}
Figure~\ref{fig:robogest} demonstrates how a diver can perform sequences of hand gestures of the form ({\tt start-token}, {\tt instruction-tokens}, {\tt end-token}) to communicate complete instructions to a robot. 
We conduct a series of such trials for instruction generation and subsequently evaluate the RoboChatGest performance for different choice of hand gesture recognition modules. 
Table~\ref{comp_gest} summarizes the results for a total of $30$ different instructions involving $162$ {\tt gesture-tokens}. 
%on a diverse set of video footage containing 

As Table~\ref{comp_gest} demonstrates, the CNN-based module performs significantly faster, but achieves a lower detection accuracy compared to the standard models. It correctly detected $24$ out of $30$ instructions with a hand gesture recognition accuracy of $78\%$. We observed two major issues while inspecting the failed cases: (i) Divers' hands often appeared in front of their faces or only partially appeared to the camera. These caused the region selector to fail, which subsequently triggered wrong hand gesture classification, \eg, classifying `{\tt ok}' as `{\tt 0}', `{\tt pic}' as `{\tt 1}', etc. 
(ii) Surface reflection and air bubbles also caused some inaccurate classifications despite the correct image regions were selected. Suspended particles can cause similar problems in deep water scenarios as well.

\begin{table}[t]
\caption{Performance of RoboChatGest for different choices of hand gesture recognition modules [$N_{I} \odot N_{G}$: \# of tested instructions $\odot$ \# of {\tt gesture-tokens} involved; FPS: frames per second (on Intel\texttrademark{} Core-i3 CPU)].}
\footnotesize
\centering

\vspace{2mm}
\begin{tabular}{c|l||c|c|l} \Xhline{2\arrayrulewidth}
$N_{I} \odot N_{G}$ & Model & Correct detection & Accuracy (\%) & FPS  \\ \Xhline{2\arrayrulewidth}
& CNN-based Model  & $24 \odot 128$ & $80.0 \odot 78.0$ & $17.2$  \\ 
$30 \odot 162$ & Faster RCNN (Inception V2)  & $29 \odot 152$ & $96.6 \odot 93.8$ & $0.52$ \\ 
& SSD (MobileNet V2) & $27 \odot 144$ & $90.0 \odot 88.8$ & $3.80$ \\ \Xhline{2\arrayrulewidth}
\end{tabular}
\label{comp_gest}
\end{table}

\begin{figure}[t]
\centering
    \subfigure[Detections using Faster RCNN (Inception v2).]{ \includegraphics[width=0.98\linewidth]{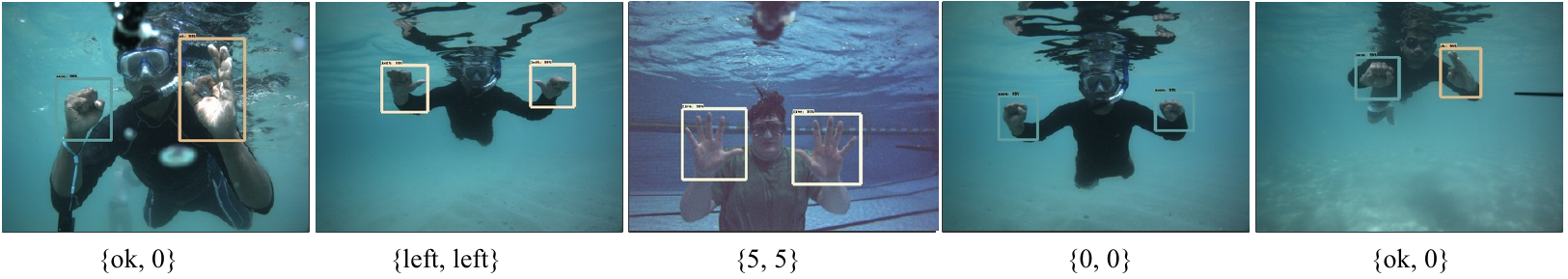}
    }
    \subfigure[Detections using SSD (MobileNet v2).]{  \includegraphics[width=0.98\linewidth]{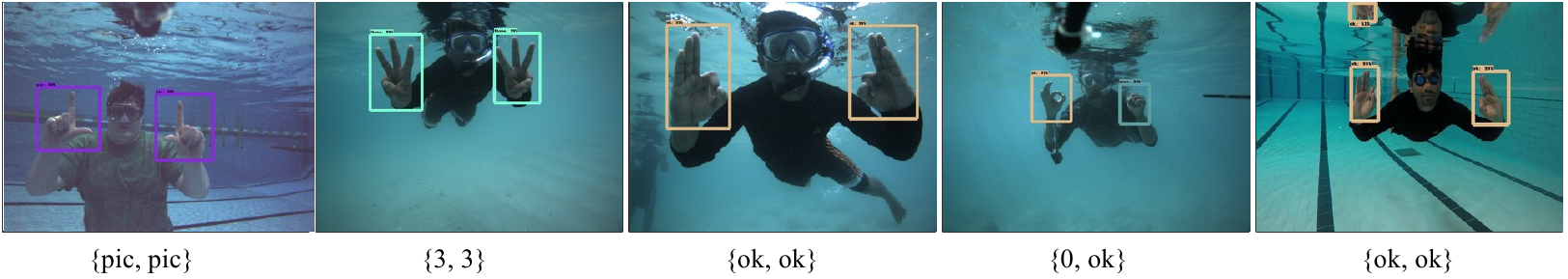}
    }
    
 \vspace{-1mm}
 \caption{A few snapshots of robust hand gesture recognition by the SOTA object detectors used in RoboChatGest.}
 \label{fig:robogestDeep}
\end{figure}

Comparatively, the deep visual detectors perform much better in such challenging conditions. As demonstrated in Table~\ref{comp_gest}, Faster RCNN (Inception v2) correctly detected $29$ out of $30$ instructions with a hand gesture recognition accuracy of $93.8\%$. On the other hand, SSD (MobileNet v2) correctly detected $27$ instructions with $88.8\%$ hand gesture recognition accuracy. They are significantly more robust than the CNN-based module, and hence, are ideal choices for high-end robot CPUs and embedded GPU platforms. As Figure~\ref{fig:robogestDeep} indicates, we perform further experiments in real-world settings to explore their utility in the RoboChatGest system. We noticed that even with hand gesture recognition rates as low as $2$--$3$\,FPS, the RoboChatGest system can be effectively used in practice. This is due to the operational convenience that the robot stays in ‘hover’ mode during the programming process, and the overall operation is not as time-critical as in tracking or diver-following scenario. To date, we use SSD (MobileNet V2) as the default hand gesture recognizer on Aqua MinneBot AUV~\cite{dudek2007aqua} (has an Intel\texttrademark{} Core-i3 6100U CPU) and LoCO AUV~\cite{LoCOAUV} (has an NVIDIA\texttrademark{}Jetson TX2).      
%to balance the trade-offs between performance and run-time. 

\begin{figure}[ht]
\centering
    \includegraphics [width=0.98\linewidth]{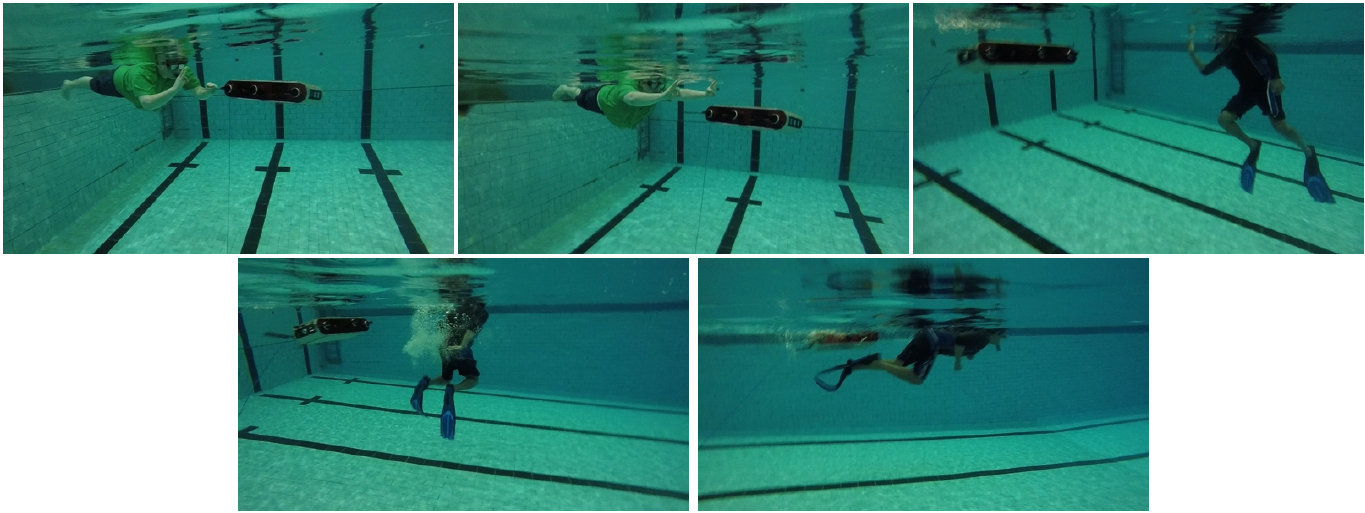}
    \vspace{-3mm}
    \caption{Image sequence demonstrates RoboChatGest in use (left-to-right, top-to-bottom): a diver instructs the robot to switch to diver-following mode from its current (hovering) mode; the full video is available at: \protect\url{https://youtu.be/An4IdMV_VtU}. }
\label{combo_gest}
\end{figure}

\subsection{Operational Feasibility Analysis}
The qualitative and quantitative results illustrated so far are based on numerous field experiments in pools and oceans~\cite{islam2018understanding}. Besides, we have been using RoboChatGest as a de facto diver-to-robot communication system
on the Aqua MinneBot AUV and LoCO AUV in general-purpose human-robot cooperative tasks. The most common usage lies in communicating various atomic instructions such as calibrating the flippers, directional swimming, hovering, etc. 
We heavily use RoboChatGest to program task-switching instructions as well. A sample demonstration is shown in Figure~\ref{combo_gest}, in which, the following instruction is communicated to Aqua MinneBot: 
\begin{quote}
{\tt STOP current-program and EXECUTE the DIVER-FOLLOWING module.}   
\end{quote}
Since Aqua MinneBot has a menu selection screen on the rear end, its back camera is used for human-robot interaction. As seen in Figure~\ref{combo_gest}, it detects the hand gestures performed by a diver through its back camera, decodes the instruction, and eventually invokes the diver-following module. 
Then, it detects the diver in front (using its front camera) and starts following him. It is important to note that the overall operation is independent of this setup; the same diver can perform hand gestures and start leading the robot using front cameras only. However, we have found the current setup operationally convenient for our experiments.

\begin{figure}[ht]
\centering
\includegraphics [width=0.95\linewidth]{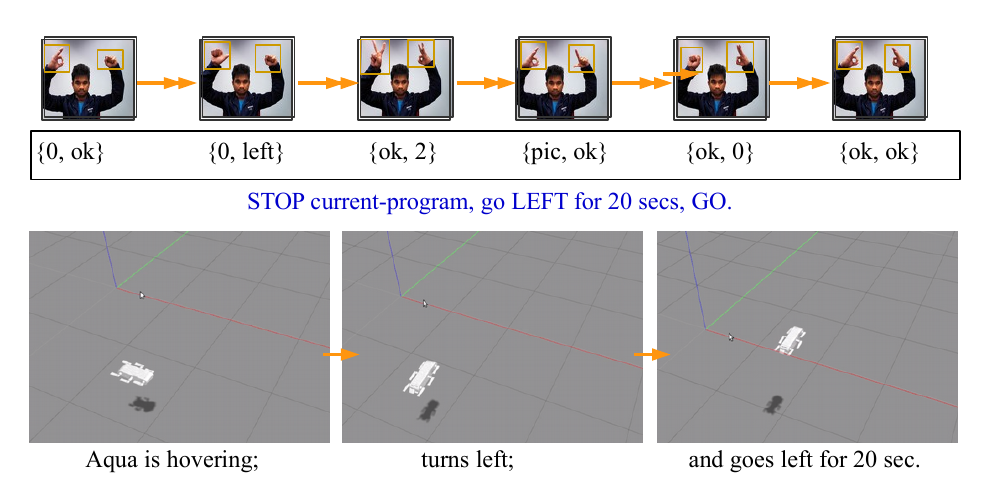}
\vspace{-3mm}
\caption{Controlling an Aqua robot via RoboChatGest in a ROS Gazebo simulation environment.}
\label{simu}
\end{figure}

\subsection{Gazebo Simulation and User Study}
We also perform simulation experiments for controlling an Aqua robot via RoboChatGest on the Gazebo (ROS Kinetic) platform~\cite{GazROS}. In our setup, hand gesture sequences performed by a human are captured (through a webcam), decoded into respective instructions via RoboChatGest, and then sent to the simulator for robot control; a particular instance is shown in Figure~\ref{simu}. Although a noise-free simulation environment does not pose many real-world challenges, it helps to set benchmarks for expected performance bounds and is useful in human interaction studies.

We conduct a particular study where the participants are introduced to three diver-to-robot communication systems: RoboChatGest, RoboChat~\cite{dudek2007visual}, and RoboChat-Motion~\cite{xu2008natural}. AprilTag markers~\cite{olson2011apriltag} are used in the latter two systems' trials to deliver commands for various symbolic {\tt tokens}. A total of ten individuals participated in the study, who were grouped into three categories: 
\begin{itemize}
\item Beginner user ($2$): unfamiliar with dynamic robot programming.
\item Medium user ($7$): familiar with gestures/markers-based robot programming.
\item Expert user ($1$): familiar and practicing these systems for some time.
\end{itemize}

Our user interaction study is similar to the one conducted by Xu~\etal~\cite{xu2008natural}. In the first set of trials, participants are asked to program the following instructions: 
\begin{enumerate}[$\hspace{5mm}$N1.]
\setlength{\itemsep}{2pt}
\setlength{\parskip}{0pt}
\setlength{\parsep}{0pt}
\item {\tt STOP current-program, HOVER for 50 seconds, GO}.
\item {\tt CONTD current-program, take SNAPSHOTS for 20 seconds, GO}.
\item {\tt CONTD current-program, Update Parameter \#3:\,DECREASE one step, GO}.
\item {\tt STOP current-program, EXECUTE Program \#1, GO}.
\end{enumerate}
In the second set of trials, the participants are asked to program the robot for the following two relatively more complex scenarios:
\begin{enumerate}[\hspace{5mm}Sa.]
\setlength{\itemsep}{2pt}
\setlength{\parskip}{0pt}
\setlength{\parsep}{0pt}
\item {\tt The robot has to stop its current task and execute program \#2 while taking snapshots}.
\item {\tt The robot has to take snapshots for 50 seconds and then start following its companion diver}.
\end{enumerate}

\begin{figure}[ht]
\vspace{1mm}
\centering
\includegraphics [width=0.8\linewidth]{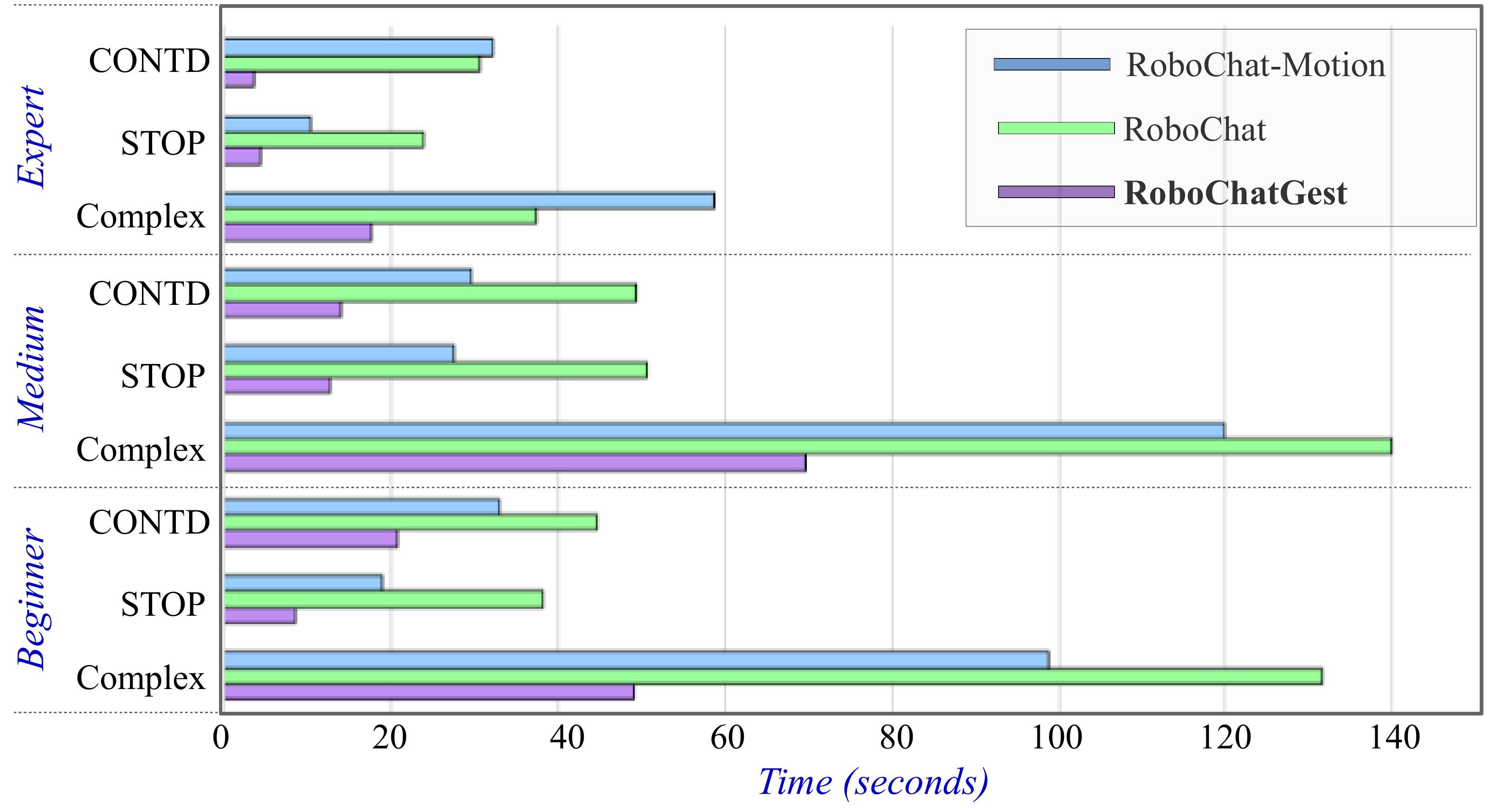}
\vspace{-2mm}
\caption{Comparison of average time taken by the participants to successfully generate various instructions [{\tt STOP}: instructions {\tt 1} and {\tt 4}; {\tt CONTD}: instructions {\tt 2} and {\tt 3}; {\tt Complex}: scenarios {\tt a} and {\tt b}].}
\label{hri}
\end{figure} 

\begin{table}[t]
\caption{Average number of mistakes made by participants before correctly generating each instruction for the (\textbf{RoboChatGest}, RoboChat, RoboChat-Motion) systems [{\tt STOP}: instructions {\tt N1} \& {\tt N4}; {\tt CONTD}: instructions {\tt N2} \& {\tt N3}; {\tt Complex}: scenarios {\tt Sa} \& {\tt Sb}].}
\footnotesize
\centering

\vspace{2mm}
\begin{tabular}{l||c|c|c} \Xhline{2\arrayrulewidth}
Instruction Type & Beginner User & Medium  User & Expert User \\  \Xhline{2\arrayrulewidth}
{\tt STOP}  & $(\mathbf 2,1,3)$ & $(\mathbf 1,0,1)$ & $(\mathbf 0,0,0)$ \\ 
{\tt CONTD} & $(\mathbf 0,0,1)$ & $(\mathbf 0,0,0)$ & $(\mathbf 0,0,0)$   \\ 
{\tt Complex} & $(\mathbf 2,3,7)$ & $(\mathbf 2,2,3)$ & $(\mathbf 0,0,0)$  \\ \Xhline{2\arrayrulewidth}
\end{tabular}
\label{hri_dudes}
\end{table}

The participants used hand gestures, AprilTags, and AprilTags with motion to program these instructions by following the language rules of RoboChatGest, RoboChat, and RoboChat-Motion, respectively. As Figure~\ref{hri} suggests, participants quickly adopted the RoboChatGest syntax and took significantly less time to finish programming compared to the other two alternatives. Specifically, participants found it inconvenient to search through all the tags while using RoboChat. On the other hand, although performing a set of discrete motions with only two AprilTags saves time, RoboChat-Motion rules were less intuitive to the participants. 
As a result, it still took a long time to formulate the complex instructions, as evident from the results. 
One interesting result is that the `beginner' users took less time to complete the instructions compared to `medium' users. This is probably because unlike the beginner users, medium users were trying to interpret and learn the syntax while performing the gestures. As illustrated by Table~\ref{hri_dudes}, beginner users made more mistakes on an average before completing the instructions successfully. Since there are no significant differences in the number of mistakes for any type of user, we conclude that simplicity, efficiency, and intuitiveness are the major advantages of RoboChatGest over other methods.

\section{Summary and Discussion}
In this chapter, we presented a hand gesture-based diver-to-robot communication system for cooperative task execution in underwater missions. The proposed system, named RoboChatGest, includes the design and development of a syntactically simple and intuitive language, various hand gesture recognition modules, and a robust instruction decoder module. We elaborately discussed the entire computational pipeline and validated its performance and effectiveness through extensive field experiments in real-world settings. Additionally, we conducted a user interaction study which reveals that RoboChatGest can be easily adopted by divers without using fiducial markers or requiring memorization of complex language rules.

\begin{figure}[ht]
\vspace{2mm}
\centering
\includegraphics [width=0.86\linewidth]{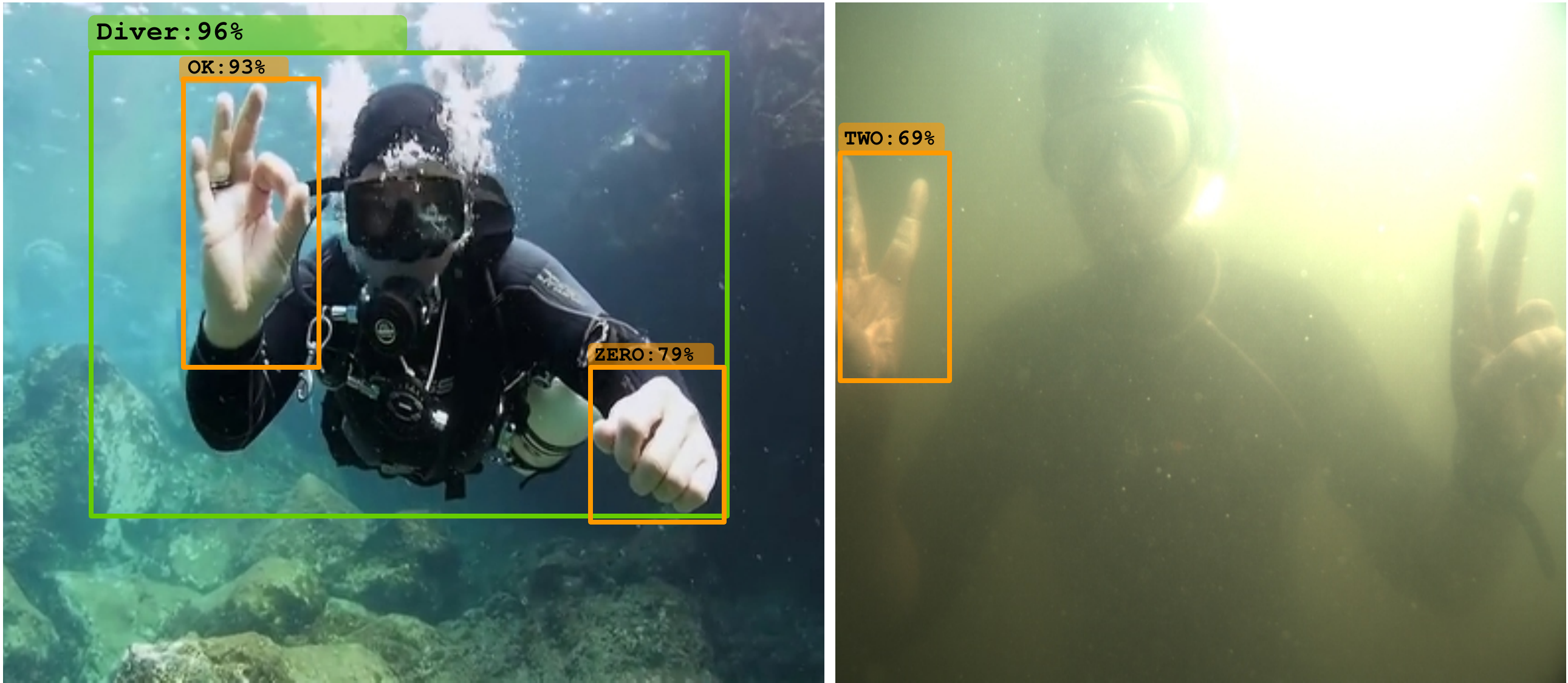}
\vspace{-2mm}
\caption{Affects of poor visibility, lighting condition, and image distortion on detection performance of the proposed models; notice the drastic difference in image quality for a near-ideal scenario versus a rather poor one (on the left and right, respectively).}
\label{bad_gest}
\end{figure}

The hand gesture-based communication system, along with the diver-following and relative pose estimation modules presented in the previous two chapters, enables an underwater robot to visually detect, follow, and interact with a diver in cooperative tasks. It is evident from our field experimental results that the proposed methodologies are considerably robust and suitable for real-time robotic deployments~\cite{islam2018understanding}. Nevertheless, their effectiveness still gets impaired by optical noise and image distortions in unfavorable visual conditions, as shown in Figure~\ref{bad_gest}. Essentially, the visual perception capabilities are constrained by the captured image quality, which depends on the visibility and lighting conditions as well as the optical waterbody properties. We have explored several aspects of underwater image formation in adverse visual conditions, particularly the lack of contrast and color degradation, to model the underlying distortions. These eventually led us to investigate whether automatic image quality enhancement techniques can alleviate these effects to improve underwater visual perception performance. We explore the relevant research question and present our findings in the next chapter.

%Such adversity can be caused by poor  depending on  

%The underwater image distortions are hard to model as they depend on the       

%Essentially, the  of any underwater visual perception module are  

% can g
%  chromatic distortion       
%performance of the visual perception modules
%low-contrast, often blurred, and color-degraded images

%These are essential capabilities for a visually-guided companion robot.  

%In a broader sense, these modules understanding human motion and gestures, visually-guided companion robots.  Similar to our field tests for ,  
%We also investigated the extent of their performances over various unfavorable visual conditions. 
%visibility and lighting condition

%We trained the proposed models for diver-detection and hand gesture recognition on comprehensive datasets in order to deal with the challenges involved in underwater visual perception. our training datasets include a large collection of gray-scale and color distorted underwater images, the proposed models are considerably robust to noise and color distortions ().  

\chapter{Fast Underwater Image Enhancement: FUnIE-GAN}\label{en_sr}

% why do we need this: challenges and scope
One major operational challenge for visually-guided underwater robots is that despite using high-end cameras, visual sensing is often greatly affected by poor visibility, light refraction, absorption, and scattering~\cite{lu2013underwater,zhang2017underwater,islam2018understanding}. These optical artifacts trigger non-linear distortions in the captured images, which severely affect the performance of vision-based tasks such as detection and tracking, segmentation, and visual servoing. Fast and accurate image enhancement techniques can alleviate these problems by restoring the perceptual and statistical qualities~\cite{zhang2017underwater,fabbri2018enhancing} of the distorted images in real-time.   

\begin{figure}[ht]
\centering
\includegraphics[width=0.85\textwidth]{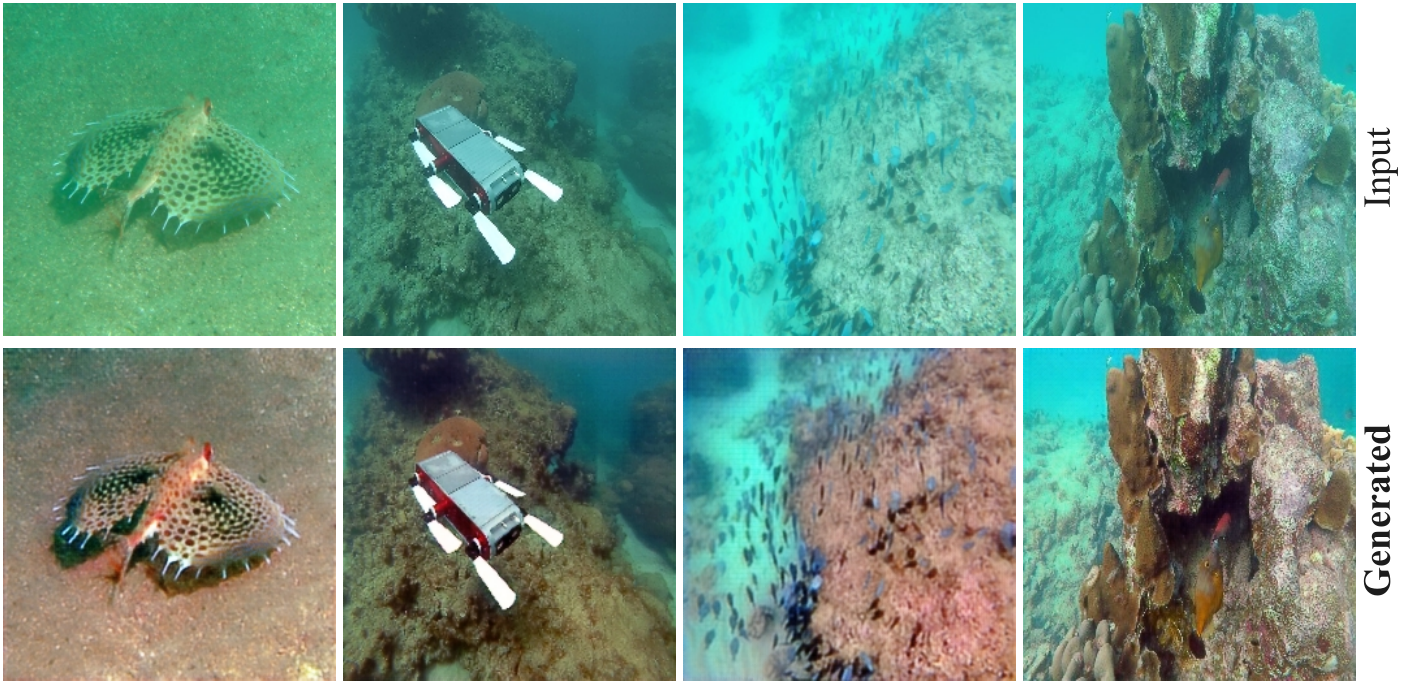}
\vspace{-2mm}
\caption{Perceptual image enhancement by our proposed model: FUnIE-GAN~\cite{islam2019fast}.}
\label{fig:4.1}
\end{figure}

% physics-based to learning-based
As light propagation differs underwater (than in the atmosphere), a unique set of non-linear image distortions occur which are propelled by a variety of factors. For instance, underwater images tend to have a dominating green or blue hue~\cite{fabbri2018enhancing} because red wavelengths get absorbed in deep water as light travels further. Such wavelength dependant attenuation~\cite{akkaynak2018revised}, scattering, and other optical properties of the waterbodies cause irregular non-linear distortions~\cite{guo2019underwater,zhang2017underwater} which result in low contrast, often blurred, and color degraded images (see Figure~\ref{fig:4.1}). Some of these aspects can be modeled and well estimated by physics-based solutions, particularly for dehazing and color correction~\cite{bryson2016true,berman2018underwater}. However, information such as the scene depth and optical water-quality measures are not always available in many robotic applications. Besides, these models are often computationally too demanding for real-time deployments.   

% learning-based to our contributions
A practical alternative is to approximate the underlying solution by learning-based methods, which demonstrated remarkable success in recent years. Several models based on deep Convolutional Neural Networks (CNNs) and Generative Adversarial Networks (GANs) provide state-of-the-art (SOTA) performance~\cite{ignatov2017dslr,chen2018deep,isola2017image,zhu2017unpaired} in learning to enhance perceptual image quality from large-scale datasets. For underwater imagery, in particular, a number of GAN-based models~\cite{fabbri2018enhancing,yu2018underwater} and CNN-based residual models~\cite{liu2019underwater} report inspiring progress for automatic color enhancement, dehazing, and contrast adjustment. However, there is significant room for improvement as learning perceptual enhancement for underwater imagery is a more challenging ill-posed problem (than terrestrial imagery).    
Additionally, due to the high costs and difficulties in acquiring large-scale underwater data, most learning-based models use small-scale and only synthetic images that fail to capture a wide range of natural variability. Moreover, designing robust yet efficient image enhancement models and investigating their applicability for improving real-time underwater visual perception have not been explored in depth.

%In this chapter, we delineate our 
We attempt to address these challenges by designing a fast underwater image enhancement model and analyzing its feasibility for real-time applications. The proposed model, named \textbf{FUnIE-GAN}~\cite{islam2019fast}, has a fully-convolutional conditional GAN-based architecture that is end-to-end trainable by both paired and unpaired data. To supervise the adversarial training, we formulate an objective function that evaluates perceptual image quality based on global content, color, local texture, and style information. We provide the model specifications and associated optimization pipelines in Section\ref{funie_gan}. We also present \textbf{EUVP}, a large-scale dataset of a paired and an unpaired collection of underwater images (of `poor' and `good' quality) that are captured using seven different cameras over various visibility conditions during oceanic explorations and human-robot collaborative experiments. The proposed model and dataset are released for academic research purposes at: \url{http://irvlab.cs.umn.edu/resources}.

%Detailed information about the data collection and preparation processes for is proided in Section~\ref{}.      

In addition to presenting the conceptual model of FUnIE-GAN, we analyze important design choices and relevant practicalities for its efficient implementation. We also perform thorough experimental evaluations to validate its effectiveness for improving the real-time perception performance of visually-guided underwater robots. The experimental processes and results are presented in Section~\ref{exp_funie}.

\section{Related Work}
Underwater image enhancement is an active research problem that deals with correcting optical image distortions to recover true pixel intensities~\cite{akkaynak2019sea,bryson2016true}. Classical approaches use hand-crafted filters to improve local contrast and enforce color constancy. These approaches are inspired by the \textit{Retinex theory} of human visual perception~\cite{jobson1997multiscale,zhang2017underwater,fu2014retinex}, and mainly focus on restoring background illumination and lightness rendition. Another class of physics-based approaches uses an atmospheric dehazing model to estimate true \textit{transmission} and \textit{ambient} light in a scene~\cite{cho2018model,he2010single}.  Additional prior knowledge or statistical assumptions (\eg, haze-lines, dark channel prior~\cite{berman2018underwater}, etc.) are utilized for global enhancements as well. Recent work by Akkaynak~\etal~\cite{akkaynak2018revised,akkaynak2019sea} introduces a revised image formation model that accounts for the unique characteristics of underwater light propagation; this contributes to a more accurate estimation of range-dependent attenuation and backscatter~\cite{roznere2019real}. 
Nevertheless, these methods require scene depth (or multiple images) and optical waterbody measurements as prior.  

%However, these methods require scene depth and optical water-body measures as prior. %, which are not always available in robotic applications. 
%Besides, these approaches tend to be computationally too demanding for real-time deployments. 
%Nevertheless, 

While accurate underwater image recovery remains a challenge, the learning-based approaches for \textit{perceptual enhancement} have made remarkable progress in recent years. Driven by large-scale supervised training~\cite{islam2019fast,yu2018underwater}, these approaches learn sequences of non-linear filters to approximate the underlying pixel-to-pixel (Pix2Pix) mapping~\cite{isola2017image} between the \textit{distorted} and \textit{enhanced} image domains.  The contemporary deep CNN-based generative models provide SOTA performance in learning such image-to-image translation for both terrestrial~\cite{cheng2015deep,cai2016dehazenet} and underwater domains~\cite{islam2019fast,liu2019underwater}. Moreover, the GAN-based models attempt to improve generalization performance by employing a two-player min-max game~\cite{goodfellow2014generative}, where an adversarial \textit{discriminator} evaluates the \textit{generator}-enhanced images compared to ground truth samples. This forces the generator to learn realistic enhancement while evolving with the discriminator toward equilibrium. Several GAN-based underwater image enhancement models have reported impressive results from both paired~\cite{fabbri2018enhancing,li2018watergan} and unpaired training~\cite{islam2019fast}. 

However, the GAN-based models are prone to training instability, and hence require careful hyper-parameter choices, and intuitive loss function adaptation~\cite{arjovsky2017wasserstein,mao2017least} to ensure convergence. 
For instance, Wasserstein GAN~\cite{arjovsky2017wasserstein} improves the training stability by using the earth-mover distance to measure the distance between the data distribution and the model distribution. Energy-based GANs~\cite{zhao2016energy} also improve training stability by modeling the discriminator as an energy function, whereas the Least-Squared GAN~\cite{mao2017least} addresses the vanishing gradients problem by adopting a least-square loss function for the discriminator. On the other hand, conditional GANs~\cite{mirza2014conditional} allow constraining the generator to produce samples that follow a pattern or belong to a specific class, which is particularly useful for image-to-image translation problems.  
%to learn a pixel-to-pixel (Pix2Pix) mapping~\cite{isola2017image} between an arbitrary input domain (\eg, distorted images) and the desired output domain (\eg, enhanced images).
%%%%%%%%%%%%%%%%%%%%%%%%%%%%%%%%%%%%%%%

Another major limitation of the existing learning-based approaches is that they typically use only synthetically distorted images for paired training, which often limits their generalization performance. 
Although various models based on two-way GANs (\eg, CycleGAN~\cite{zhu2017unpaired}, DualGAN~\cite{yi2017dualgan}, etc.) have been proposed for learning perceptual enhancement on unpaired terrestrial data~\cite{chen2018deep}, the extent of large-scale unpaired training on naturally distorted underwater images have not been explored in the literature. Moreover, most existing models fail to ensure fast inference on single-board platforms, hence are not feasible for real-time robotic deployments. We attempt to address these aspects in the design and implementation of FUnIE-GAN.

% %Furthermore, Ignatov \etal~\cite{ignatov2017dslr} showed that additional loss-terms for preserving the high-level feature-based content improve the quality of image enhancement using GANs. 

\section{FUnIE-GAN Model}\label{funie_gan}
\begin{figure}[h]
\vspace{2mm}
\centering
    \subfigure[Generator: five encoder-decoder pairs with mirrored skip-connections (inspired by the success of U-Net~\cite{ronneberger2015u}; however, it is a much simpler model).]{
        \includegraphics[width=0.95\linewidth]{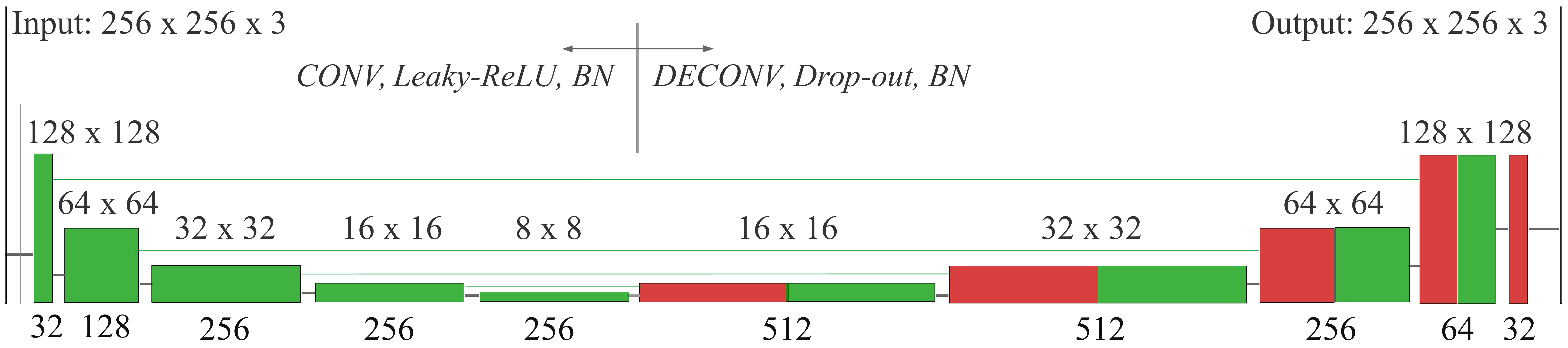}
        \label{fig:model_a}
    }
    
    \vspace{1mm}
    \subfigure[Discriminator: a Markovian PatchGAN~\cite{isola2017image} with s and a patch-size of $16$$\times$$16$.]{
        \includegraphics[width=0.6\linewidth]{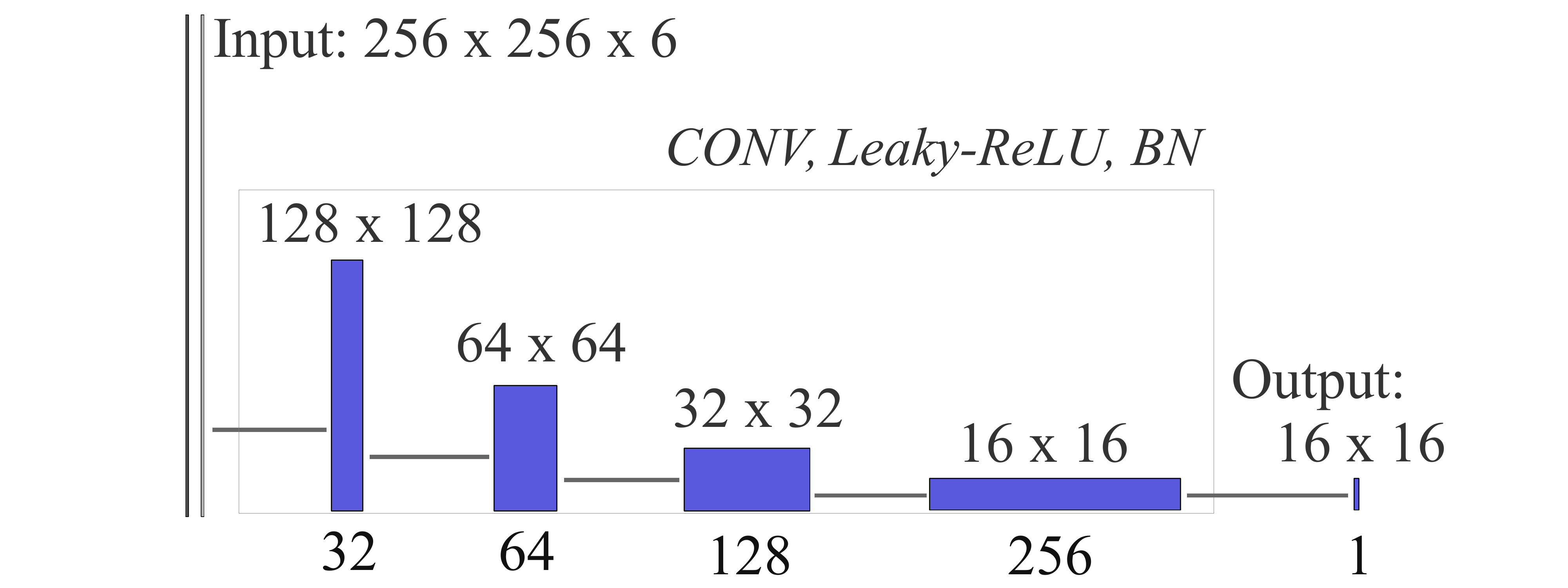}
        \label{fig:model_b}
    }
\caption{Schematic diagram of the FUnIE-GAN architecture.}
\label{fig:model_funie}
\end{figure}

Given a source domain $X$ (of distorted images) and desired domain $Y$ (of enhanced images), our goal is to learn a mapping ${\small G: X \rightarrow Y}$ to perform automatic image enhancement. We adopt a conditional GAN-based model where the generator tries to learn this mapping by evolving with an adversarial discriminator through an iterative min-max game. A simplified sketch of the network architecture of FUnIE-GAN is presented in Figure~\ref{fig:model_funie}.

\subsection{Network Architecture}
As illustrated in Figure~\ref{fig:model_a}, we design a generator network by following the principles of U-Net~\cite{ronneberger2015u}. 
It is an encoder-decoder network ($e_1$-$e_5$,$d_1$-$d_5$) with connections between the mirrored layers, \ie, between ($e_1$, $d_5$), ($e_2$, $d_4$), ($e_3$, $d_2$), and ($e_4$, $d_4$). Specifically, the outputs of each encoders are concatenated to the respective mirrored decoders. This idea of \textit{skip-connections} in the generator network is shown to be very effective~\cite{isola2017image,chen2018deep,fabbri2018enhancing} 
for image-to-image translation and image quality enhancement problems. In FUnIE-GAN, however, we employ a much simpler model with fewer parameters in order to achieve fast inference. The input to the network is set to $256\times256\times3$ and the encoder ($e_1$-$e_5$) learns only $256$ feature-maps of size $8\times8$. The decoder ($d_1$-$d_5$) utilizes these feature-maps and inputs from the skip-connections to learn to generate a $256\times256\times3$ enhanced image as output. The network is fully-convolutional as no fully-connected layers are used. Additionally, 2D convolutions with $4\times4$ filters are applied at each layer, which is then followed by a Leaky-ReLU non-linearity~\cite{maas2013rectifier} and Batch Normalization (BN)~\cite{ioffe2015batch}. The feature-map sizes in each layer and other model parameters are annotated in Figure~\ref{fig:model_a}.

For the discriminator, we employ a Markovian PatchGAN~\cite{isola2017image} architecture that assumes the independence of pixels beyond the patch-size, \ie, only discriminates based on the patch-level information. This assumption is important to effectively capture high-frequency features such as local texture and style~\cite{yi2017dualgan}. In addition, this configuration is computationally more efficient as it requires fewer parameters compared to discriminating globally at the image level. As shown in Figure~\ref{fig:model_b}, four convolutional layers are used to transform a $256\times256\times6$ input (real and generated image) to a $16\times16\times1$ output that represents the averaged \textit{validity} responses of the discriminator. At each layer, $3\times3$ convolutional filters are used with a stride of $2$; then the non-linearity and BN are applied the same way as the generator.  %Traditionally, PatchGANs use $70\times70$ patches for $256\times256$ images (\eg, in Pix2Pix~\cite{isola2017image}, DualGAN~\cite{yi2017dualgan}, etc.). However, we use a patch-size of only $16$$\times$$16$ in FUnIE-GAN.                         

\subsection{Objective Function Formulation}\label{obj_fun}
A standard conditional GAN-based model learns a mapping ${\small G:\{X, Z\} \rightarrow Y}$, where $X$ ($Y$) represents the source (desired) domain, and $Z$ denotes random noise. The conditional adversarial loss function~\cite{mirza2014conditional} is expressed as:   
\begin{equation} %\label{eq:cgan}
\small
\centering
  \begin{aligned}
	\mathcal{L}_{cGAN}(G,D) = \mathbb{E}_{X,Y} \big[\log D(Y)\big] + \mathbb{E}_{X,Y} \big[\log (1-D(X, G(X, Z)))\big].
  \end{aligned}
\end{equation}
Here, the generator $G$ tries to minimize $\mathcal{L}_{cGAN}$ while the discriminator $D$ tries to maximize it. In FUnIE-GAN, we associate three additional aspects, \ie, global similarity, image content, and local texture and style information in the objective to quantify perceptual image quality.    

\vspace{1mm}
\textbf{I.} \textbf{Global similarity}: existing methods have shown that adding an $\mathcal{L}_1$ ($\mathcal{L}_2$) loss to the objective function enables $G$ to learn to sample from a globally similar space in an $\mathcal{L}_1$ ($\mathcal{L}_2$) sense~\cite{isola2017image,yu2018underwater}. Since the $L_1$ loss is less prone to introduce blurring, we add the following loss term in the objective:     
    \begin{equation} \label{eq:l1}
    \small
    \centering
    	\mathcal{L}_{1}(G) = \mathbb{E}_{X,Y,Z} \big[\big|\big|Y-G(X, Z)\big|\big|_1\big].
    \end{equation}

\vspace{1mm}
\textbf{II.} \textbf{Image content}: we add a \textit{content loss} term in the objective to encourage $G$ to generate enhanced image that has similar content (\ie, feature representation) as the target (\ie, real) image. Being inspired by~\cite{johnson2016perceptual,ignatov2017dslr}, we define the image content function $\Phi_{VGG}(\cdot)$ as the high-level features extracted by the {\tt block5\_conv2} layer of a pre-trained VGG-19 network. Then, we formulate the content loss as:    
    \begin{equation} \label{eq:lcontent}
    \small
   \centering
	\mathcal{L}_{con}(G) = \mathbb{E}_{X,Y,Z} \big[\big|\big|\Phi_{VGG} (Y)-\Phi_{VGG} (G(X, Z)) \big|\big|_2\big].
   \end{equation}

\vspace{1mm}
\textbf{III.} \textbf{Local texture and style}: as mentioned, Markovian PatchGANs are effective in capturing high-frequency information of local texture and style~\cite{isola2017image}. We rely on $D$ to enforce the local texture and style consistency in adversarial fashion.

\subsubsection{Paired Training}
For paired training, we formulate an objective function that guides $G$ to learn perceptual image quality enhancement so that the generated images are close to their respective ground truth in terms of global appearance and high-level feature representation. On the other hand, $D$ will discard a generated image that has locally inconsistent texture and style. Specifically, we use the following objective function for paired training: 
\begin{equation*} %\label{eq:pix2pix_final}
\small
\centering
	G^* = \argmin\limits_{G}\max\limits_{D} \mathcal{L}_{cGAN}(G,D)+\lambda_1 \, \mathcal{L}_{1}(G) + \lambda_c \, \mathcal{L}_{con}(G). 
\end{equation*}
Here, $\lambda_1=0.7$ and $\lambda_c=0.3$ are scaling factors that we empirically tuned as hyper-parameters.

\subsubsection{Unpaired Training}
For unpaired training, we do not enforce the global similarity and content loss constraints as pairwise ground truth is not available. Instead, the objective is to learn both forward mapping ${\small G_{F}:\{X, Z\} \rightarrow Y}$ and the reconstruction ${\small G_{R}:\{Y, Z\} \rightarrow X}$ simultaneously by maintaining cycle-consistency. As suggested by Zhu \etal~\cite{zhu2017unpaired}, we formulate the cycle-consistency loss as follows:   
\begin{equation} \label{eq:cycle}
\small
\centering
  \begin{aligned}
	\mathcal{L}_{cyc}(G_{F}, G_{R}) = \mathbb{E}_{X,Y,Z} \big[\big|\big|X-G_{R}(G_{F}(X, Z))\big|\big|_1\big] + \text{ } \mathbb{E}_{X,Y,Z} \big[\big|\big|Y-G_{F}(G_{R}(Y, Z))\big|\big|_1\big].
  \end{aligned}
\end{equation}
Therefore, our objective for unpaired training is: 
\begin{equation*} %\label{eq:pix2pix_final}
\small
\centering
   \begin{aligned}
	G_{F}^*, G_{R}^* = \argmin\limits_{G_{F}, G_{R}}\max\limits_{D_Y,D_X} \mathcal{L}_{cGAN}(G_{F}, D_Y) \text{ } + \text{ } \mathcal{L}_{cGAN}(G_{R}, D_X) \text{ } + \lambda_{cyc} \, \mathcal{L}_{cyc}(G_{F}, G_{R}).
   \end{aligned}	  
\end{equation*}
Here, $D_Y$ ($D_X$) is the discriminator associated with generator $G_{F}$ ($G_{R}$), and the scaling factor $\lambda_{cyc}=0.1$ is an empirically tuned hyper-parameter. We do not enforce additional global similarity loss-term because the $\mathcal{L}_{cyc}$ computes analogous reconstruction loss for each domain in $L_1$ space.

\begin{figure}[h]
\centering
    \subfigure[Paired instances: ground truth images and their respective distorted pairs are shown on the top and bottom row, respectively.]{
        \includegraphics[width=0.95\linewidth]{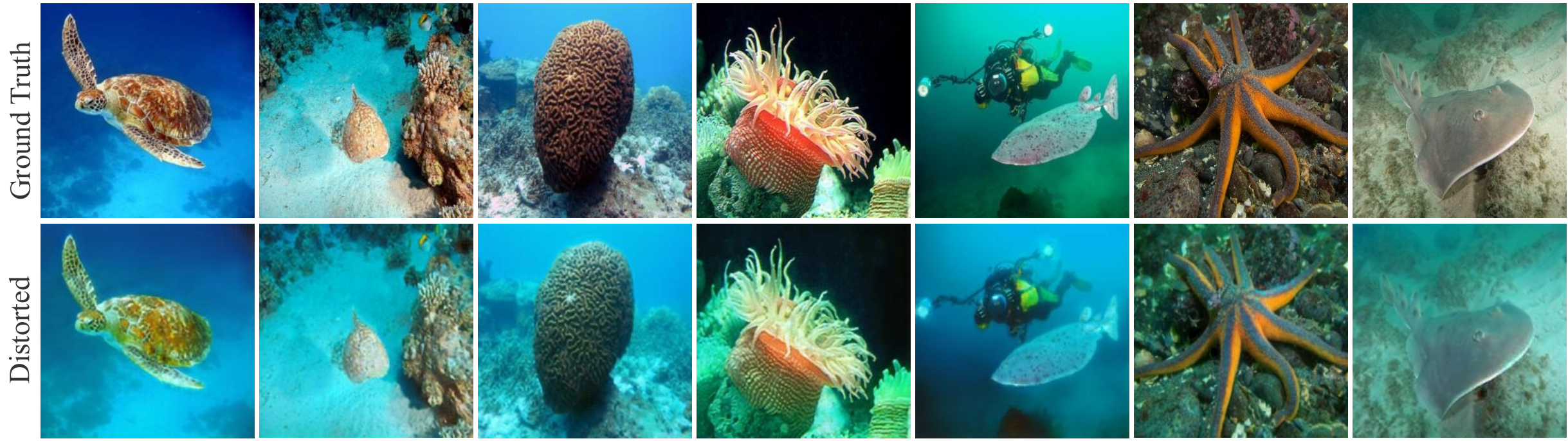}
    }
    
    \subfigure[Unpaired instances: good and poor quality images are shown on the top and bottom row (in no particular order), respectively.]{
        \includegraphics[width=0.95\linewidth]{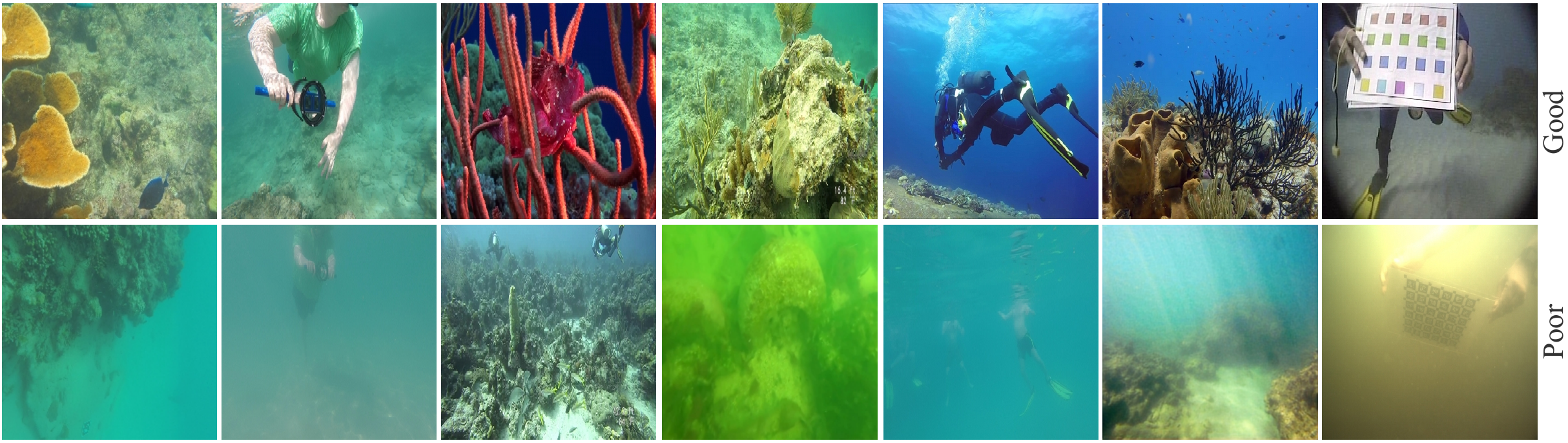}
    }
    \caption{A few sample images from the EUVP dataset are shown.}
    \label{fig:data_funie}
\end{figure}

\section{EUVP Dataset}\label{EUVP}
The EUVP dataset contains a large collection of paired and unpaired underwater images of poor and good perceptual quality. We used seven different cameras including multiple GoPros~\cite{gopro}, Aqua AUV's uEye cameras~\cite{dudek2007aqua}, low-light USB cameras~\cite{lowlight}, and Trident ROV's HD camera~\cite{trident}, to capture images for this dataset. The data was collected during oceanic explorations and human-robot cooperative experiments in different locations under various visibility conditions. Additionally, images extracted from a few publicly available YouTube\texttrademark{} videos were included in the dataset. 
The images were carefully selected to accommodate a wide range of natural variability in the data, \ie, diverse scenes, waterbodies, lighting conditions, etc.

The unpaired data was prepared (\ie, good, and poor quality images were separated) based on visual inspection by six human participants. They inspected several image properties (\eg, color, contrast, and sharpness) and considered whether a scene was visually interpretable and the foreground objects were identifiable. Hence, the unpaired training endorses the modeling of human perceptual preferences of underwater image quality. On the other hand, the paired data was prepared by following a procedure suggested in~\cite{fabbri2018enhancing}. Specifically, a CycleGAN~\cite{zhu2017unpaired}-based model was trained on the unpaired data to learn domain transformation between the good and poor quality images. Subsequently, the good quality images were distorted by the learned model to generate respective pairs; we also augmented a set of underwater images from the ImageNet dataset~\cite{deng2009imagenet} and from Flickr\texttrademark.

There are over $12$K paired and $8$K unpaired instances in the EUVP dataset; a few samples are provided in Figure~\ref{fig:data_funie}. It is to be noted that our focus is to facilitate `perceptual image enhancement' for boosting robotic scene understanding, not to model the underwater optical degradation process for `image restoration', which requires scene depth and waterbody properties.  
%The images are of various resolutions, \eg, $800 \times 600$, $640 \times 480$, $256 \times 256$, and $224 \times 224$. 

\begin{figure}[ht]
\centering
    \subfigure[True color and sharpness is restored in the enhanced image.]{
        \includegraphics[width=0.7\linewidth]{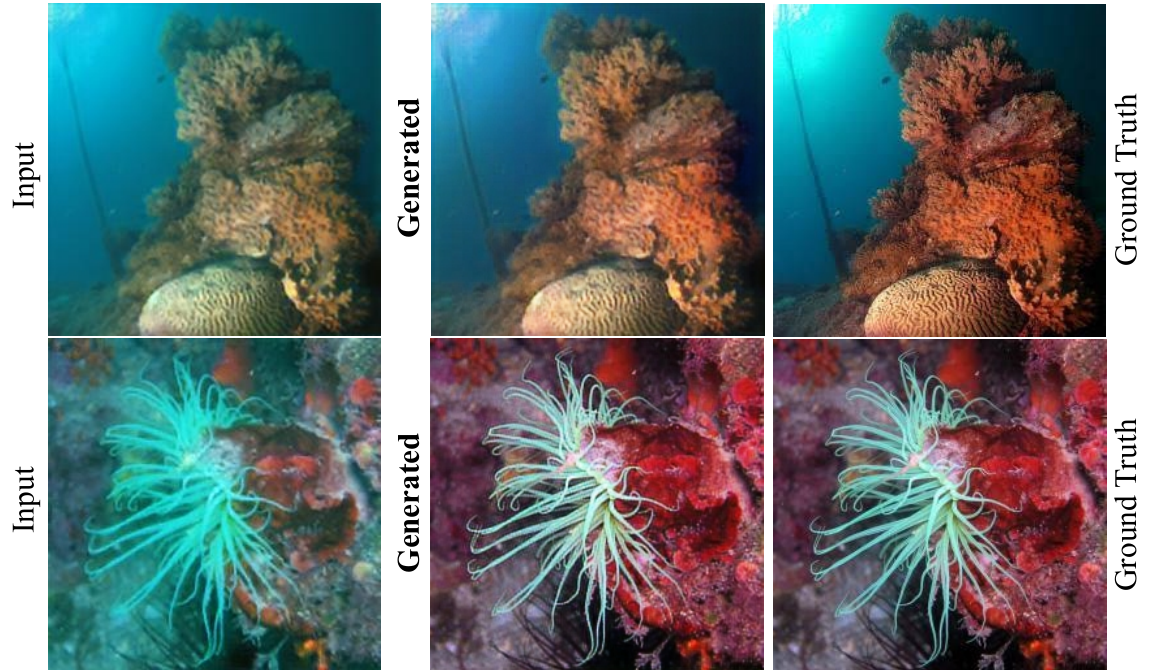}
        \label{fig:res2a}
    }
    
    \subfigure[The greenish hue is rectified and global contrast is enhanced.]{
        \includegraphics[width=0.7\linewidth]{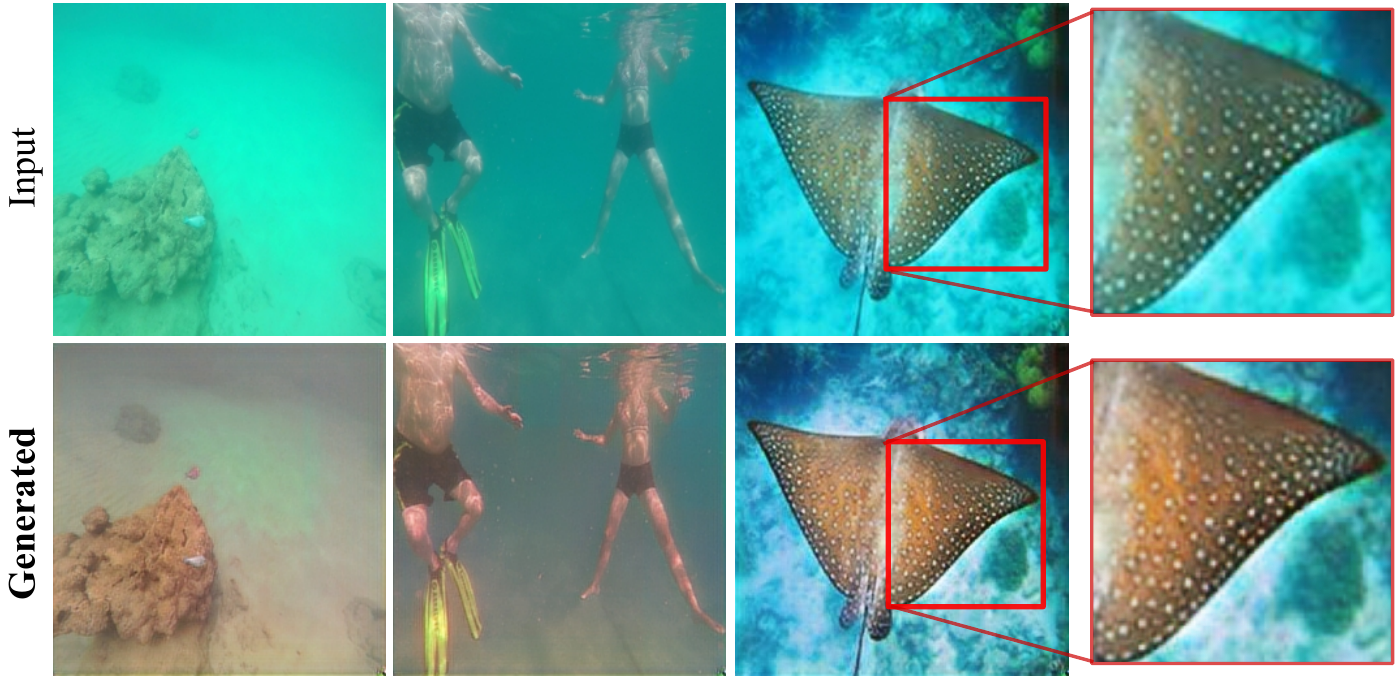}
        \label{fig:res2b}
    }
    \caption{Demonstration of improved image attributes by FUnIE-GAN in terms of color, sharpness, and contrast:~\subref{fig:res2a} paired training;~\subref{fig:res2b} unpaired training.}
    \vspace{-1mm}
    \label{fig:res1}
\end{figure}

\section{Experimental Validation}\label{exp_funie}
  
\subsection{Implementation Details}
We use TensorFlow libraries~\cite{abadi2016tensorflow} to implement the optimization pipeline of FUnIE-GAN on a Linux machine with Four NVIDIA\texttrademark{} GTX 1080 graphics cards. The paired and unpaired models are trained separately on $11$\,K paired and $7.5$\,K unpaired instances that are randomly chosen from the EUVP dataset; the rest are used for respective validation and testing. Adam optimizer~\cite{kingma2014adam} is used for the global iterative learning with a rate of $0.0003$ and a momentum of $0.5$; both models are trained for $60$K--$70$K iterations with a batch-size of $8$. Next, we present the experimental evaluations based on a qualitative analysis, standard quantitative metrics, ablation studies, and a human preference study.

\subsection{Qualitative Analysis}
\begin{figure}[t]
    \centering
    \includegraphics[width=0.98\linewidth]{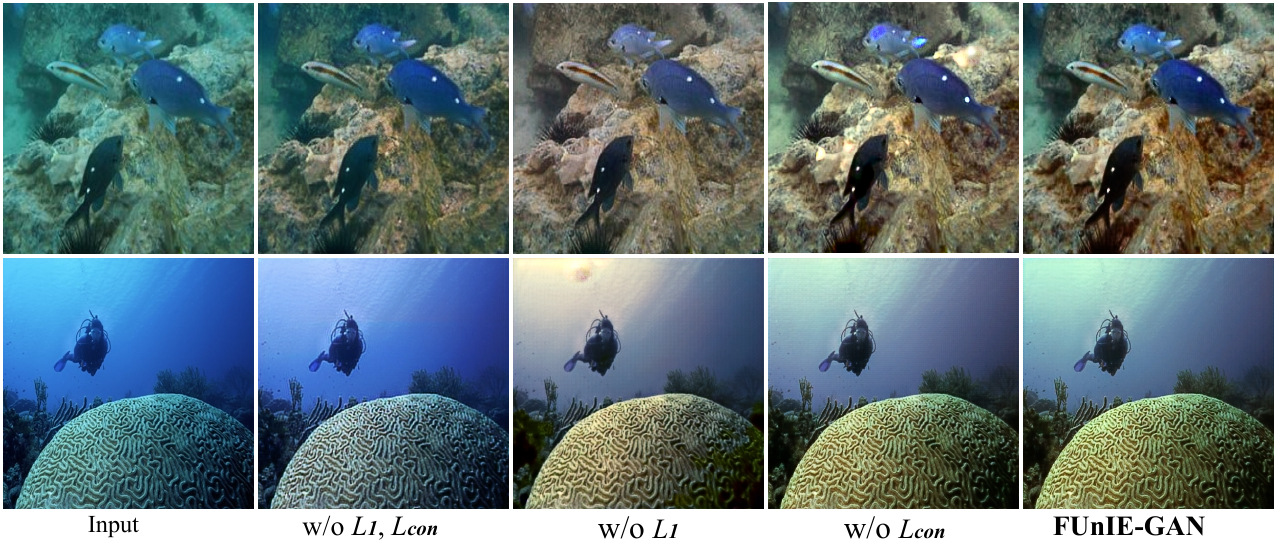}
    \vspace{-3mm}
    \caption{Ablation experiment: learning enhancement without (w/o) $\mathcal{L}_{1}$ and $\mathcal{L}_{con}$, w/o $\mathcal{L}_{1}$, and w/o $\mathcal{L}_{con}$ loss-terms in FUnIE-GAN.}
        \label{fig:res2c}
\end{figure}

We first qualitatively analyze the enhanced color and sharpness of FUnIE-GAN-generated images compared to their respective ground truths. As Figure~\ref{fig:res2a} shows, the true color, and sharpness is mostly recovered in the enhanced images. Additionally, as shown in Figure~\ref{fig:res2b}, the greenish hue in underwater images are rectified and the global contrast is enhanced. These are the primary characteristics of an effective underwater image enhancer. 
We further demonstrate the contributions of each loss-terms of FUnIE-GAN: global similarity loss ($\mathcal{L}_{1}$), and image content loss ($\mathcal{L}_{con}$), for learning the enhancement. We observe that the $\mathcal{L}_{1}$ term helps to generate sharper images, while the $\mathcal{L}_{con}$ term contributes to furnishing finer texture details (see Figure~\ref{fig:res2c}). Moreover, we found slightly better numeric stability for $\mathcal{L}_{con}$ with the {\tt block5\_conv2} layer of VGG-19 compared to its last feature extraction layer {\tt block5\_conv4}.

\begin{figure*}[t]
    \centering
    \includegraphics[width=0.99\linewidth]{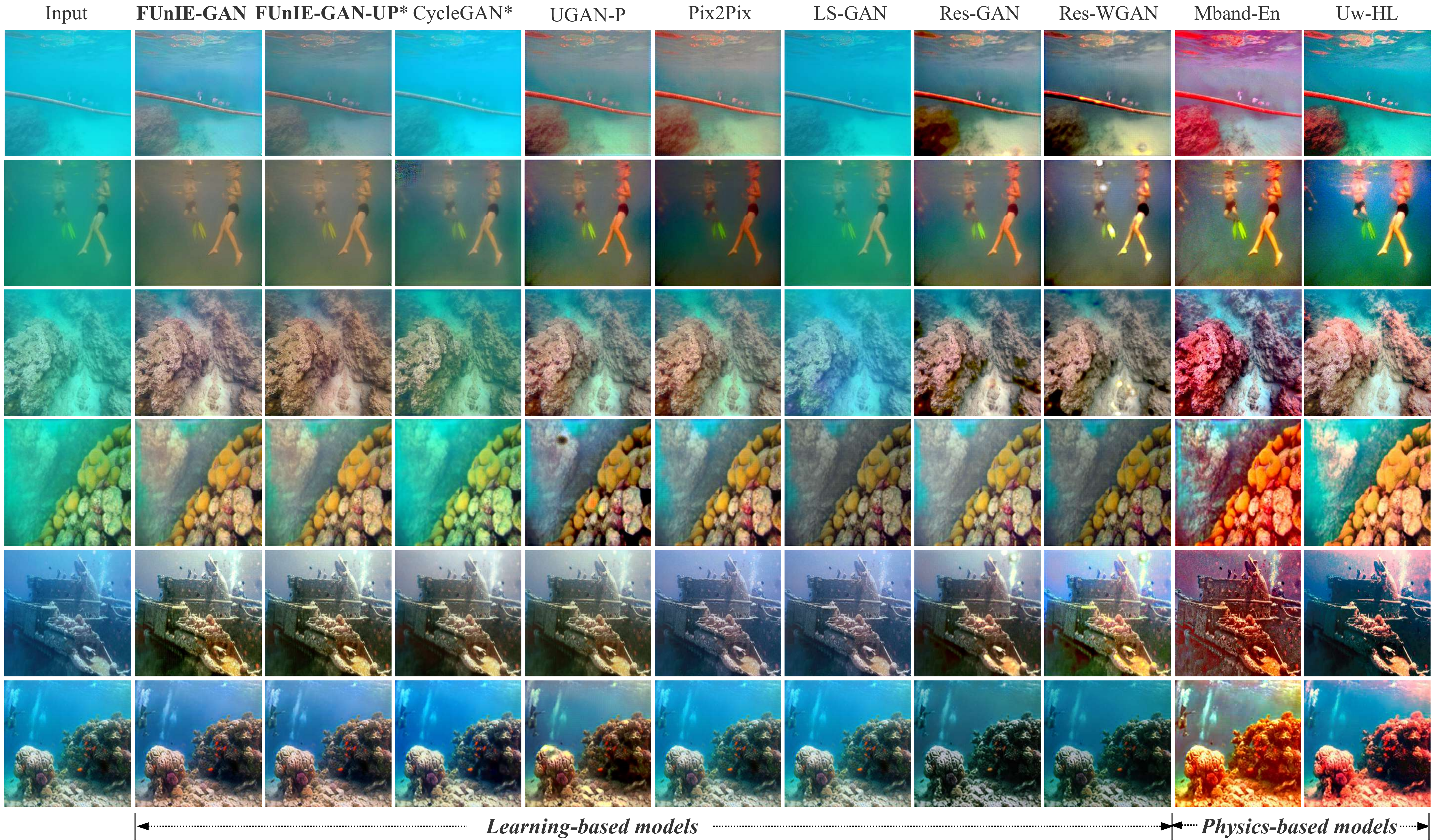} 
    \vspace{-3mm}
    \caption{Qualitative performance comparison of FUnIE-GAN and FUnIE-GAN-UP with learning-based methods: CycleGAN~\cite{zhu2017unpaired}, UGAN-P~\cite{fabbri2018enhancing}, Pix2Pix~\cite{isola2017image}, LS-GAN~\cite{mao2017least}, Res-GAN~\cite{li2017perceptual}, and Res-WGAN~\cite{arjovsky2017wasserstein}; the super-scripted asterisk ($\mathbf{\ast}$) denotes unpaired training. Two physics-based models: Mband-EN~\cite{cho2018model} and Uw-HL~\cite{berman2018underwater}, are also included in the comparison. (Best viewed at $400\%$ zoom)}
    \label{fig:comp}
\end{figure*}

Next, we present a qualitative comparison of perceptual image enhancement by FUnIE-GAN with several SOTA models. We consider five learning-based models: (i) underwater GAN with gradient penalty (UGAN-P~\cite{fabbri2018enhancing}), (ii) Pix2Pix~\cite{isola2017image}, (iii) least-squared GAN (LS-GAN~\cite{mao2017least}), (iv) GAN with residual blocks~\cite{li2017perceptual} in the generator (Res-GAN), and (v) Wasserstein GAN~\cite{arjovsky2017wasserstein} with residual blocks in the generator (Res-WGAN). These models are implemented with $8$ encoder-decoder pairs (or $16$ residual blocks) in the generator network and $5$ convolutional layers in the discriminator. They are trained on the paired EUVP dataset using the same setup as the FUnIE-GAN. Additionally, we consider CycleGAN~\cite{zhu2017unpaired} as a baseline for comparing the performance of unpaired training, \ie, FUnIE-GAN-UP. We also include two physics-based models in the comparison: Multi-band fusion-based enhancement (Mbad-EN~\cite{cho2018model}), and haze-line-aware color restoration (Uw-HL~\cite{berman2018underwater}).   
A common test set with $1$K images are used for the qualitative evaluation; it also includes $72$ images with known waterbody types~\cite{berman2018underwater}. A few sample comparisons are illustrated in Figure~\ref{fig:comp}.

As demonstrated in Figure~\ref{fig:comp}, Res-GAN, Res-WGAN, and Mbad-EN often suffer from over-saturation, while LS-GAN generally fails to rectify the greenish hue in images. UGAN-P, Pix2Pix, and Uw-HL perform reasonably well and their enhanced images are comparable to that of FUnIE-GAN; however, UGAN-P over-saturates bright objects in the scene while Pix2Pix fails to enhance global brightness in some cases. 
On the other hand, we observe that achieving color consistency and hue rectification are relatively more challenging through unpaired learning. This is mostly because of the lack of reference color or texture information in the loss function. Nevertheless, FUnIE-GAN-UP still outperforms CycleGAN in general.     
Overall, FUnIE-GAN performs as well and often better without using scene depth or prior waterbody information as the physics-based models, and despite having a much simpler network architecture compared to the existing learning-based models.

\begin{table}[H]
\centering
\caption{Quantitative comparison for average PSNR and SSIM values on $1$K paired test images of EUVP dataset.}
\footnotesize
\vspace{4mm}
\begin{tabular}{l||c|c}
  \Xhline{2\arrayrulewidth}
  \textbf{Model} & $PSNR\big(G(\mathbf{x}),\mathbf{y}\big)$ & $SSIM\big(G(\mathbf{x}),\mathbf{y}\big)$ \\  
   & Input: $17.27 \pm 2.88$ & Input: $0.62 \pm 0.075$  \\ \Xhline{2\arrayrulewidth}
  Uw-HL & $18.85 \pm 1.76$  &  $0.7722 \pm 0.066$  \\ \hline 
  Mband-EN & $12.11 \pm 2.55$  &  $0.4565 \pm 0.097$ \\ \hline
  Res-WGAN & $16.46 \pm 1.80$ & $0.5762 \pm 0.014$ \\ \hline
  Res-GAN & $14.75 \pm 2.22$ & $0.4685 \pm 0.122$ \\ \hline
  LS-GAN & $17.83 \pm 2.88$ & $0.6725 \pm 0.062$ \\ \hline
  Pix2Pix & $20.27 \pm 2.66$ & $0.7081 \pm 0.069$ \\ \hline
  UGAN-P & $19.59 \pm 2.54$ & $0.6685 \pm 0.075$ \\ \hline
  CycleGAN & $17.14 \pm 2.65$ & $0.6400 \pm 0.080$  \\ \hline
  \textbf{FUnIE-GAN-UP} & $21.36 \pm 2.17$ & $0.8164 \pm 0.046$ \\ \hline
  \textbf{FUnIE-GAN} & $21.92 \pm 1.07$ & $0.8876 \pm 0.068$ \\ \Xhline{2\arrayrulewidth}
\end{tabular}
\label{tab:psnr_ssim}
\vspace{2mm}
\end{table}

\subsection{Quantitative Evaluation}\label{funiegan-quant}
We consider two standard metrics~\cite{hore2010image,islam2019fast} named Peak Signal-to-Noise Ratio (PSNR) and Structural Similarity (SSIM) in order to quantitatively compare FUnIE-GAN-enhanced images with their respective ground truth. We conduct a similar analysis for Underwater Image Quality Measure (UIQM)~\cite{panetta2016human,liu2019real}, which quantifies underwater image colorfulness, sharpness, and contrast. Their definitions and relevant details are provided in Appendix~\ref{ApenD}.

\begin{table}[H]
\centering
\caption{Quantitative comparison for average UIQM values on $1$K paired and $2$K unpaired  test images of EUVP dataset.}
\footnotesize
\vspace{4mm}
\begin{tabular}{l||c|c}
  \Xhline{2\arrayrulewidth}
   & \textbf{Paired data} & \textbf{Unpaired data} \\ 
  \textbf{Model} & Input: $2.20 \pm 0.69$ & Input: $2.29 \pm 0.62$  \\
  & G. Truth: $2.91 \pm 0.65$ & G. Truth: N/A  \\ \Xhline{2\arrayrulewidth} 
  Uw-HL & $2.62 \pm 0.35$ & $2.75 \pm 0.32$ \\ \hline 
  Mband-EN & $2.28 \pm 0.87$  & $2.34 \pm 0.45$ \\ \hline
  Res-WGAN & $2.55 \pm 0.64$ & $2.46 \pm 0.67$ \\ \hline
  Res-GAN & $2.62 \pm 0.89$ & $2.28 \pm 0.34$ \\ \hline
  LS-GAN & $2.37 \pm 0.78$ & $2.59 \pm 0.52$ \\ \hline
  Pix2Pix & $2.65 \pm 0.55$ & $2.76 \pm 0.39$ \\ \hline
  UGAN-P & $2.72 \pm 0.75$ & $2.77 \pm 0.34$ \\ \hline
  CycleGAN & $2.44 \pm 0.71$ & $2.62 \pm 0.67$ \\ \hline
  \textbf{FUnIE-GAN-UP} & $2.56 \pm 0.63$ & $2.81 \pm 0.65$ \\ \hline
  \textbf{FUnIE-GAN} &  $2.78 \pm 0.43$ & $2.98 \pm 0.51$ \\ \Xhline{2\arrayrulewidth}
\end{tabular}
\label{tab:uiqm}
%\vspace{2mm}
\end{table}

%We use a set of $1$K paired test images $(\mathbf{x} \in X, \mathbf{y} \in Y)$ in our evaluation. At first, we use FUnIE-GAN to generate enhanced images $G(\mathbf{x})$ for each $\mathbf{x}$ and then compute $PSNR\big(G(\mathbf{x}),\mathbf{y}\big)$ and $SSIM\big(G(\mathbf{x}),\mathbf{y}\big)$ using Eq.~\ref{eq:psnr} and~\ref{eq:ssm}, respectively. 
In Table~\ref{tab:psnr_ssim}, we provide the averaged PSNR and SSIM values over $1$K test images for FUnIE-GAN and compare the same models used in our qualitative evaluation. The results indicate that FUnIE-GAN performs best on both PSNR and SSIM metrics. 
We present a similar analysis for UIQM in Table~\ref{tab:uiqm}, which indicates that although FUnIE-GAN-UP performs better than CycleGAN, its UIQM values on the paired dataset are relatively poor. Interestingly, the models trained on paired data, particularly FUnIE-GAN, UGAN-P, and Pix2Pix, produce better results on both paired and unpaired test images. We postulate that the global similarity loss in FUnIE-GAN and Pix2Pix, or the gradient-penalty term in UGAN-P contribute to this enhancement, as they all add $\mathcal{L}_{1}$ terms in their objective. Our ablation experiments of FUnIE-GAN (see Figure~\ref{fig:res2c}) reveal that $\mathcal{L}_{1}$ contributes to $4.58\%$ improvements in UIQM values, while $\mathcal{L}_{con}$ contributes $1.07\%$, on an average. Moreover, without both $\mathcal{L}_{1}$ and $\mathcal{L}_{con}$, the average UIQM values drop by $17.6\%$; we observe similar statistics for PSNR and SSIM as well.

\begin{figure}[ht]
    \centering
        \includegraphics[width=0.95\linewidth]{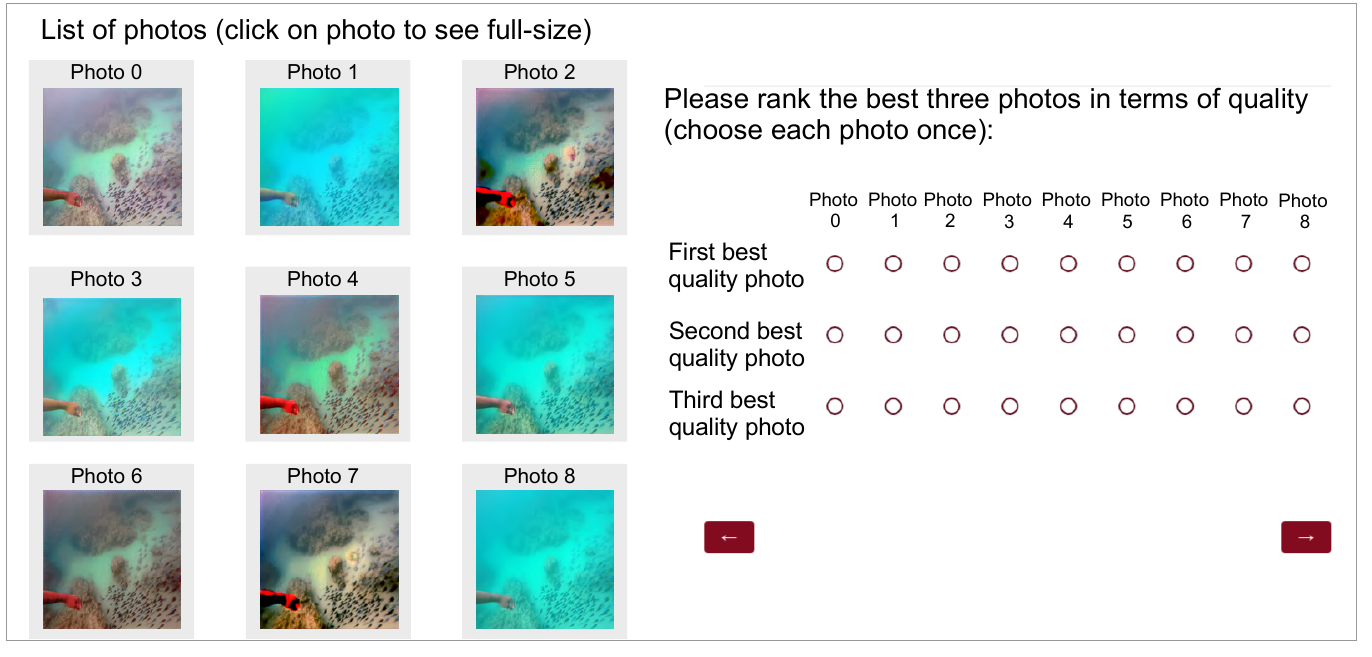}
   \vspace{-4mm}
    \caption{A snapshot of the user interface used in our study.}
    \label{fig:study}
\end{figure}

\subsection{Human Preference Study}
We also conduct a user study to add human preferences to our performance analysis. As Figure~\ref{fig:study} illustrates, the participants are shown different sets of $9$ images (one for each learning-based models) and asked to rank top $3$ best quality images. A total of $78$ individuals participated in the study and a total of $312$ responses are recorded. Table~\ref{tab:study} compares the average rank-1, rank-2, and rank-3 accuracy of the top $4$ categories. The average rank-3 accuracy of the original images is recorded to be $6.67$, which suggests that the users clearly preferred enhanced images over the original ones. Moreover, the results indicate that the users prefer the images enhanced by FUnIE-GAN, UGAN-P, and Pix2Pix compared to the other models; these statistics are consistent with our qualitative and quantitative analysis.   

\begin{table}[H]
\centering
\caption{Rank-$n$ accuracy ($n=1,2,3$) for the top four models based on $312$ responses provided by $78$ individuals.}
\footnotesize
\vspace{2mm}
\begin{tabular}{l||c|c|c}
  \Xhline{2\arrayrulewidth}
  \textbf{Model} & Rank-1 ($\%$) & Rank-2 ($\%$) & Rank-3 ($\%$) \\ \Xhline{2\arrayrulewidth}
  \textbf{FUnIE-GAN} & $24.50$  & $68.50$   & $88.60$  \\ \hline
  \textbf{FUnIE-GAN-UP} & $18.67$  & $48.25$   & $76.18$  \\ \hline
   UGAN-P & $21.25$ & $65.75$  &  $80.50$ \\ \hline
   Pix2Pix & $11.88$ & $45.15$  &  $72.45$ \\ \Xhline{2\arrayrulewidth}
\end{tabular}
\label{tab:study}
%\vspace{-2mm}
\end{table}%%%%%%%%%%%%%%%%%%%%%%%%%%%%%%%%

\begin{figure}[h]
    \centering
        \includegraphics[width=0.98\linewidth]{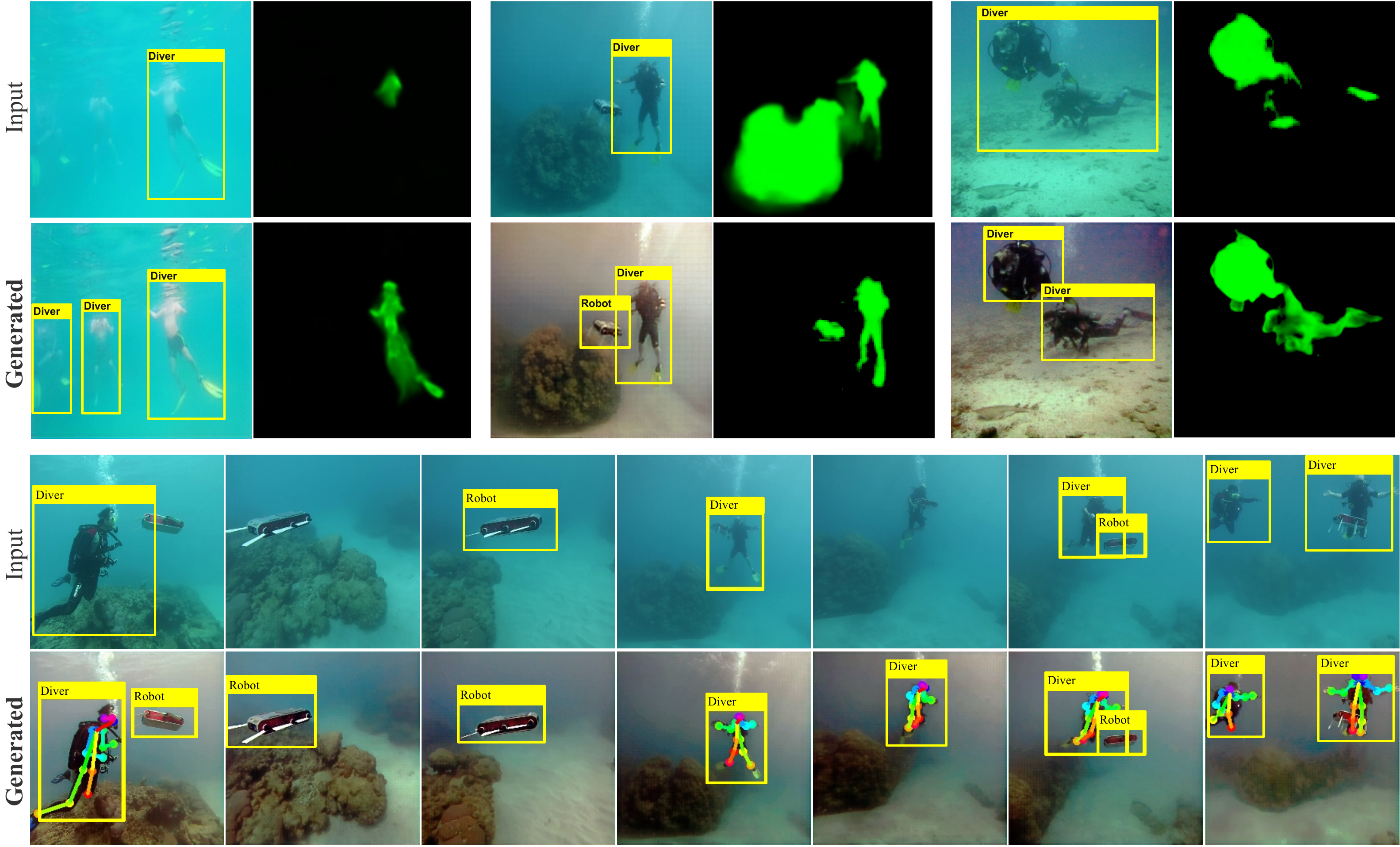}
   \vspace{-4mm}
    \caption{A few snapshots showing qualitative improvement for object detection, saliency prediction, and human body-pose estimation on FUnIE-GAN-generated images; a detailed demonstration can be found at: \url{https://youtu.be/1ewcXQ-jgB4}.}
    \label{fig:per_a}
\end{figure}

\begin{figure}[h]
    \centering
        \includegraphics[width=0.98\linewidth]{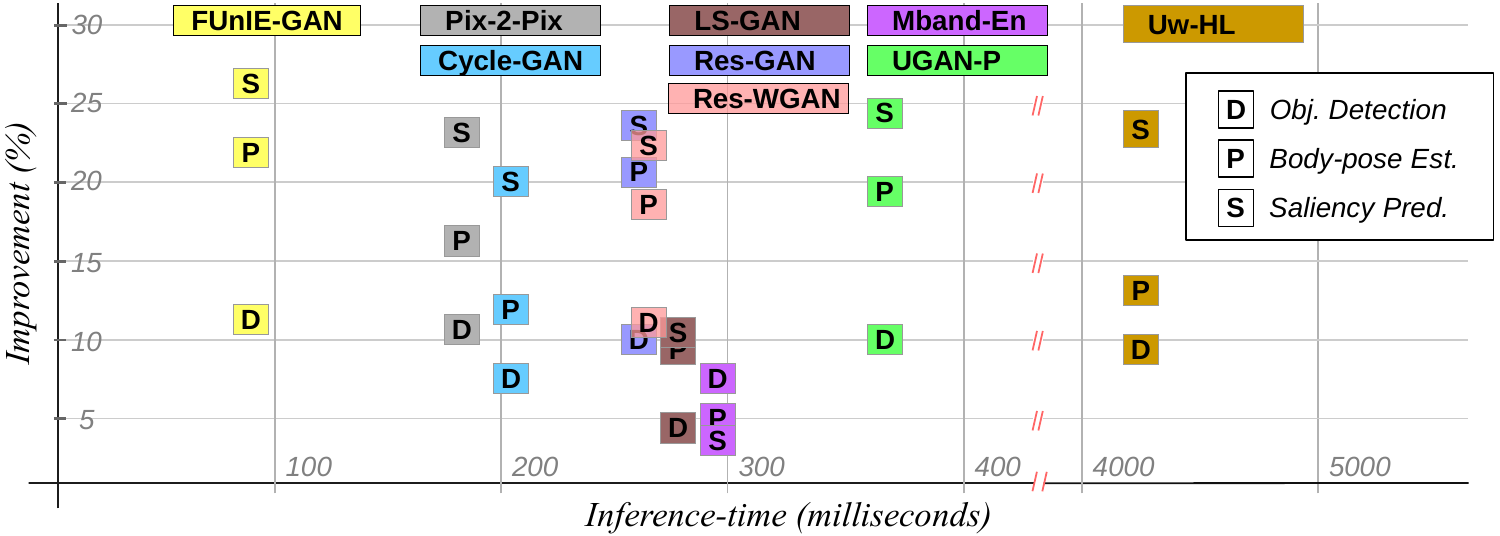}
   \vspace{-4mm}
    \caption{Improvement versus inference-time comparison with the SOTA models; FUnIE-GAN offers over 10 FPS speed (on common platform: Intel\texttrademark{} Core-i5 3.6GHz CPU); the run-times are evaluated on $256\times256$ image patches for all the models.}
    \label{fig:per_b}
\end{figure}

%\begin{figure}[h]
%\centering
%    \subfigure[A few snapshots showing qualitative improvement on FUnIE-GAN-generated images; a detailed demonstration can be found at: \url{https://youtu.be/1ewcXQ-jgB4}.]{
%        \includegraphics[width=0.8\linewidth]{figs/per.pdf}
%        \label{fig:per_a}
%    }
    
%    \vspace{2mm}
%    \subfigure[Improvement versus inference-time comparison with the SOTA models; FUnIE-GAN offers over 10 FPS speed (on common platform: Intel\texttrademark{} Core-i5 3.6GHz CPU); note that the run-times are evaluated on $256\times256$ image patches for all the models.]{
%        \includegraphics[width=0.8\linewidth]{figs/exp4.pdf}
%        \label{fig:per_b}
%    }
%    \caption{Improved performance for object detection, saliency prediction, and human body-pose estimation on enhanced images.}
    %
%    \label{fig:per}
%\end{figure}

\subsection{Real-time Feasibility Analysis}
As demonstrated in Figure~\ref{fig:per_a}, we conduct further experiments to quantitatively interpret the effectiveness of FUnIE-GAN-enhanced images for underwater visual perception over a variety of test cases. We analyze the performance of standard deep visual models for underwater object detection~\cite{islam2018towards}, 2D human body-pose estimation~\cite{cao2017realtime}, and visual attention-based saliency prediction~\cite{wang2018salient}; although results vary depending on the image qualities of a particular test set, on an average, we observe $11$-$14\%$, $22$-$28\%$, and $26$-$28\%$ improvements, respectively. We also evaluate other SOTA models on the same test sets; as Figure~\ref{fig:per_b} suggests, images enhanced by UGAN-P, Res-GAN, Res-WGAN, Uw-HL, and Pix2Pix also achieve considerable performance improvements. However, these models offer significantly slower inference rates than FUnIE-GAN, most of which are not suitable for real-time deployments. Specifically, 
%FUnIE-GAN offers $2$--$4$ times faster run-times than other models and more than $40$ times faster than Uw-HL. 
FUnIE-GAN's memory requirement is $17$\,MB and it operates at a rate of $25.4$\,FPS (frames per second) on a single-board computer (NVIDIA\texttrademark{} Jetson TX2), $148.5$\,FPS on a graphics card (NVIDIA\texttrademark{} GTX 1080), and $7.9$\,FPS on a robot CPU (Intel\texttrademark{} Core-i3 6100U). These computational aspects are ideal for it to be used as an image processing pipeline by visually-guided underwater robots in real-time applications.

\begin{figure}[t]
    \centering
        \includegraphics[width=0.95\linewidth]{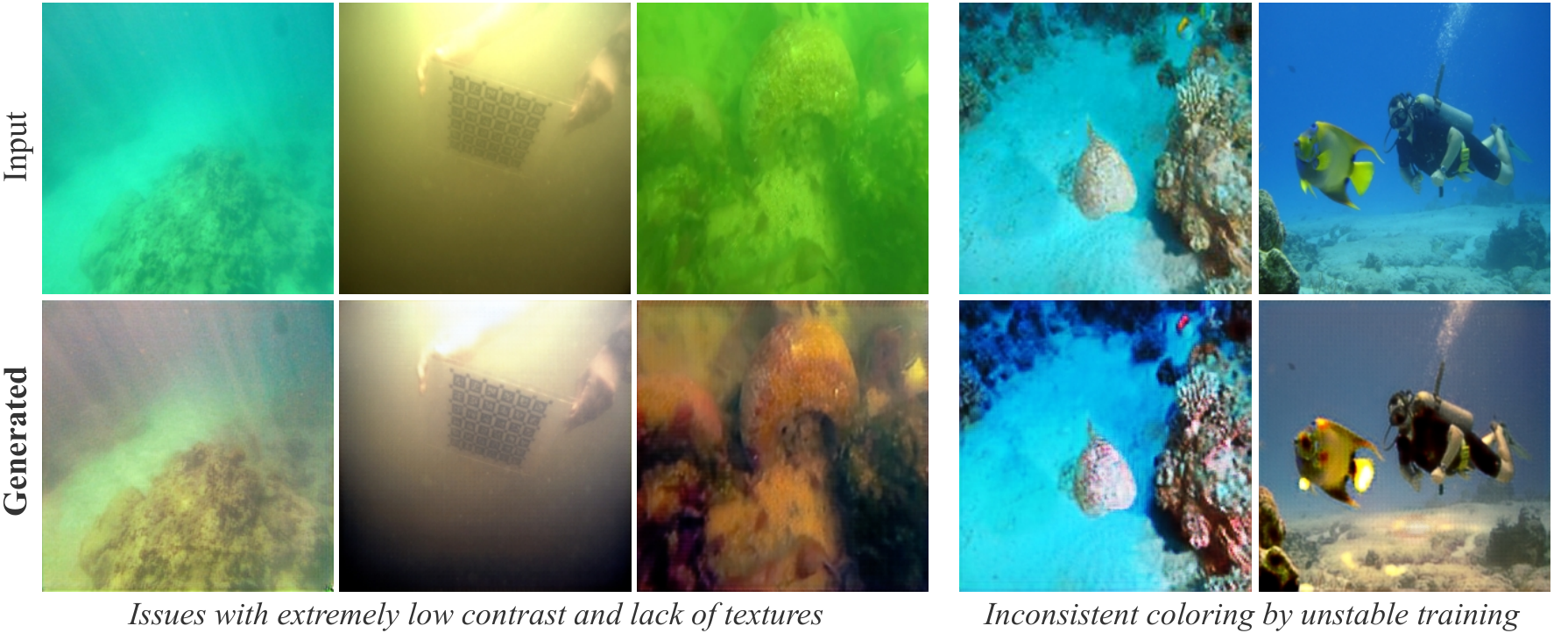}
    \vspace{-3mm}
    \caption{Extremely low-contrast and texture-less images are generally challenging for FUnIE-GAN, whereas FUnIE-GAN-UP often suffers from inconsistent coloring due to training instability.}
    \label{fig:bad_funie}
\end{figure}

Nevertheless, we observe a couple of challenging cases for FUnIE-GAN. First, FUnIE-GAN is not very effective for enhancing severely degraded texture-less images. The generated images in such cases are often over-saturated by noise amplification. Although the hue rectification is generally correct, color and texture recovery remains poor. Secondly, FUnIE-GAN-UP is prone to training instability. Our investigations suggest that the discriminator often converges too early, causing a \textit{diminishing gradient} effect that halts the generator's learning. As shown in Figure~\ref{fig:bad_funie}, the generated images in such cases lack color consistency and accurate texture details. This is a common issue in unpaired GAN training~\cite{chen2018deep,ignatov2017dslr}, and requires meticulous hyper-parameter tuning.

\section{Concluding Remarks}
In this chapter, we presented a fully-convolutional conditional GAN-based model for fast underwater image enhancement, which we refer to as FUnIE-GAN. We formulated a multi-modal objective function to train this model by evaluating the perceptual quality of an image based on its global content, color, local texture, and style information. 
Additionally, we presented a large-scale dataset named EUVP, which has paired and unpaired collections of underwater images (of poor and good quality) to facilitate \textit{one-way} or \textit{two-way} adversarial training. Through a series of qualitative and quantitative experiments, we demonstrated that FUnIE-GAN can learn to enhance perceptual image quality from both paired and unpaired training. Moreover, the FUnIE-GAN-enhanced images significantly boost the performance of standard visual perception tasks such as object detection, human pose estimation, and saliency prediction. Lastly, we analyzed the computational aspects of FUnIE-GAN to validate that it can be used by underwater robots to improve real-time perception in adverse visual conditions.

FUnIE-GAN balances a trade-off between robustness and efficiency which limits its performance to a certain degree. More powerful deep models (\ie, denser architectures with more parameters) can be adopted for non-real-time applications; moreover, the input/output layers can be modified with additional bottleneck layers for learning enhancement at higher resolution than $256\times256$. On the other hand, FUnIE-GAN does not guarantee the recovery of true pixel intensities as it is designed for perceptual image quality enhancement. If scene depth and optical waterbody properties are available, underwater light propagation and image formation characteristics~\cite{akkaynak2018revised,berman2018underwater,bryson2016true} can be incorporated into the optimization for more accurate image restoration. 

\chapter{Simultaneous Enhancement and Super-Resolution: SESR}\label{sesr}
In Chapter~\ref{intro}, we briefly discussed how the single image super-resolution (SISR) capability can facilitate detailed visual perception on salient image regions for better scene understanding and attention modeling. However, if the low resolution (LR) image regions suffer from noise and optical distortions, those get amplified by SISR, resulting in uninformative high resolution (HR) outputs. Hence, restoring perceptual and statistical image qualities is essential for robust perception in noisy underwater visual conditions, as we demonstrated in the previous chapter. Nevertheless, separately processing visual data for enhancement and super-resolution, even with the fastest available solutions, is not computationally feasible on single-board robotic platforms.

To this end, we introduce simultaneous enhancement and super-resolution (SESR) for underwater robot vision and provide an efficient solution for near real-time applications. We present \textbf{Deep SESR}~\cite{islam2020sesr}, a residual-in-residual network-based generative model that can learn to restore perceptual image qualities at $2\times$, $3\times$, or $4\times$ higher spatial resolution. We supervise its training by formulating a multi-modal objective function that addresses the chrominance-specific underwater color degradation, lack of image sharpness, and loss in high-level feature representation. It is also supervised to learn salient foreground regions in the image, which in turn guides the network to learn global contrast enhancement. We design an end-to-end training pipeline to jointly learn the saliency prediction and SESR on a shared hierarchical feature space for fast inference. Moreover, we present \textbf{UFO-120}, the first dataset to facilitate large-scale SESR learning; it contains over $1500$ training samples and a benchmark test set of $120$ samples.

\begin{figure}[t]
    \centering
\includegraphics[width=0.98\linewidth]{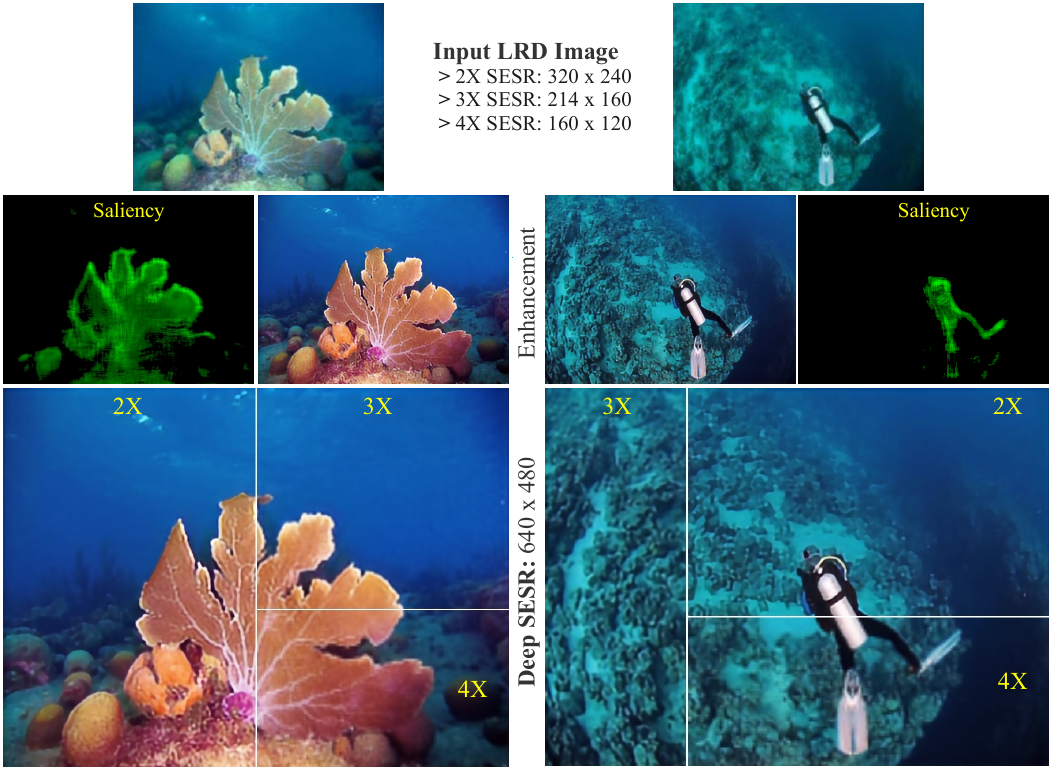}%
\vspace{-2mm}
 \caption{The proposed `Deep SESR' model~\cite{islam2020sesr} offers perceptually enhanced HR image generation and saliency prediction by a single efficient inference. The enhanced images restore color, contrast, and sharpness at higher spatial scales for detailed visual perception, whereas the saliency map can be further exploited for attention modeling.}%
 \label{fig:sesr_intro}
 \vspace{2mm}
\end{figure}

By thorough experimental evaluation on UFO-120 and several other standard datasets, we demonstrate that Deep SESR outperforms the existing solutions for underwater image enhancement and super-resolution. 
As shown in Fig.~\ref{fig:sesr_intro}, it learns to restore perceptual image qualities at higher spatial scales (up to $4\times$); as a byproduct, it learns to identify salient foreground regions in the image. We further validate its generalization performance on several test cases that include underwater images with diverse spectral and spatial degradation levels, and also terrestrial images with unseen natural objects. Lastly, we analyze its computational feasibility for single-board deployments and demonstrate its operational benefits for visually-guided underwater robots. The proposed model and dataset are released for
academic research purposes at: \url{http://irvlab.cs.umn.edu/resources}.

In the following section, we formulate the SESR problem and specify the data preparation processes of UFO-120. Subsequently, in Section~\ref{sesr_net}, we present the Deep SESR architecture, design its objective function, and discuss the associated loss components. Then, we provide the implementation details and experimental results in Section~\ref{sesr_implement}. Finally, we demonstrate its generalization performance in Section~\ref{sesr_general} and discuss various design choices for practical use cases in Section~\ref{sesr_feasibility}.

\section{Problem Formulation}
\subsection{Learning SESR}
SESR refers to the task of generating perceptually enhanced HR images from their LR and possibly distorted (LRD) input measurements. We formulate the problem as learning a pixel-to-pixel mapping from a source domain $X$ (of LRD images) to its target domain $Y$ (of enhanced HR images); we represent this mapping as a generative function $G: X \rightarrow Y$. We adopt an extended formulation by considering the task of learning SESR and saliency prediction on a shared feature space. Specifically, Deep SESR learns the generative function $G: X \rightarrow S, E, Y$; here, the additional outputs $S$ and $E$ denote the predicted saliency map, and enhanced image (in the same resolution as the input $X$), respectively. Additionally, it offers up to $4\times$ SESR for the final output $Y$.

\begin{figure}[t]
\centering
    \subfigure[A few sample ground truth images and corresponding saliency maps are shown on the top, and bottom row, respectively.]{
        \includegraphics[width=0.98\textwidth]{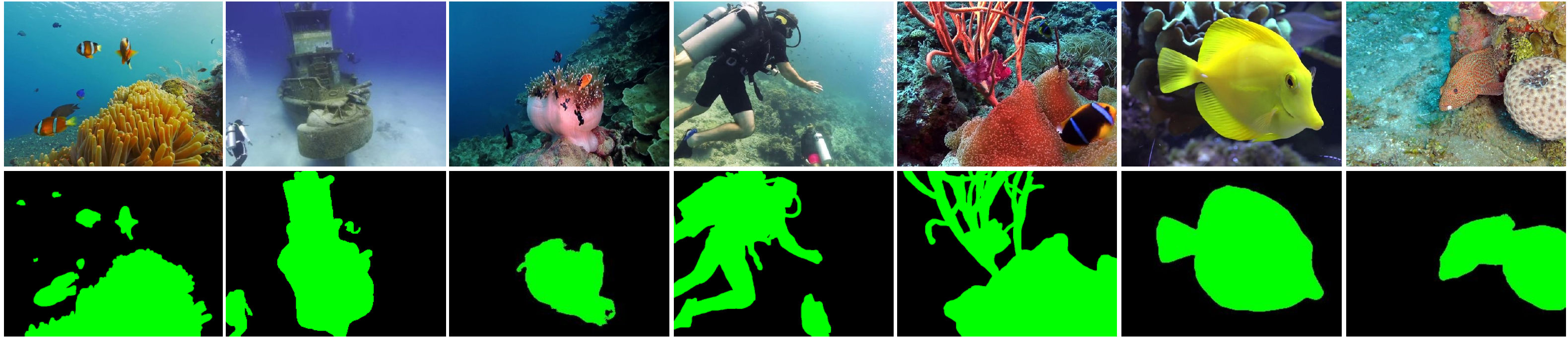}
        \label{fig:data_sesr_a}
    }
    \vspace{1mm}
    
    \subfigure[Two particular instances are shown: the HR ground truth images are of size $640\times480$; their corresponding LR distorted (LRD) images are of size $320\times240$, $214\times160$, and $160\times120$.]{
    \includegraphics[width=0.96\textwidth]{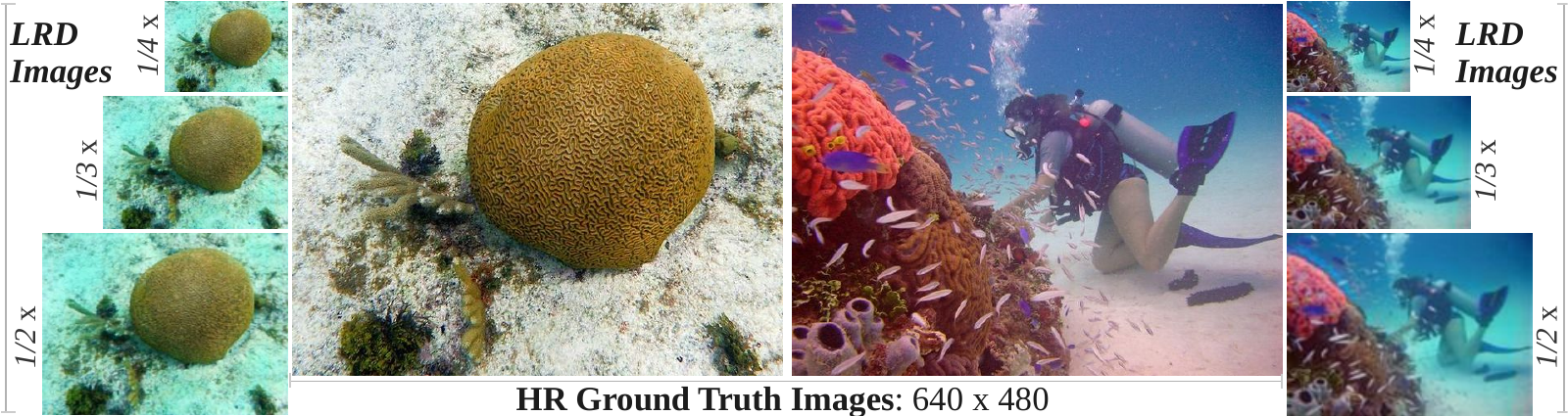}
    \label{fig:data_sesr_b}
    }
    
    \caption{The UFO-120 dataset facilitates paired training of $2\times$, $3\times$, and $4\times$ SESR models; it also contains salient pixel annotations for all training samples. The combined data is used for the supervised training of Deep SESR model.}
    \label{fig:data_sesr}
\end{figure}

\subsection{Data Preparation: The UFO-120 Dataset}\label{sec:data_prep}
We utilize several existing underwater image enhancement and super-resolution datasets to supervise the SESR learning with paired data of the form ($\{X\}$, $\{S, E, Y\}$). We prepare over $1500$ samples for training and another $120$ for testing in the UFO-120 dataset. It contains HR natural underwater images collected from oceanic explorations in multiple locations having different waterbodies, as seen in Figure~\ref{fig:data_sesr_a}. 
The saliency maps are annotated by human participants, whereas standard procedures for optical/spatial image degradation~\cite{islam2019fast,li2019feedback,fabbri2018enhancing} are followed to create the respective LRD samples.

Specifically, we adopt a widely used domain-transfer technique~\cite{islam2019fast,fabbri2018enhancing} that deploys a CycleGAN~\cite{zhu2017unpaired}-based model trained on unpaired natural data to generate distorted images by mimicking underwater optical distortion characteristics. Subsequently, we prepare the LRD samples by Gaussian blurring ({\tt GB}) and bicubic down-sampling ({\tt BD}). Based on their relative order, we group the data into three sets:
\begin{itemize}
\setlength{\itemsep}{2pt}
\setlength{\parskip}{0pt}
\setlength{\parsep}{0pt}
    \item \textbf{Set-U:} {\tt GB} is followed by {\tt BD},
    \item \textbf{Set-F:} the order is interchanged with a $0.5$ probability, and
    \item \textbf{Set-O:} {\tt BD} is followed by {\tt GB}. 
\end{itemize}
We use a $7\times7$ kernel and a noise level of $20\%$ for the {\tt GB}. As Figure~\ref{fig:data_sesr_b} illustrates, we use $2\times$, $3\times$, and $4\times$ {\tt BD} to generate the LRD samples from the synthetically distorted HR pairs. Hence, there are nine available training combinations for SESR. Note that the UFO-120 dataset can also be used for training underwater SISR ($E$$\rightarrow$$Y$), image enhancement ($X$$\rightarrow$$E$), or saliency prediction ($E$$\rightarrow$$S$) models.

\section{Deep SESR Model}\label{sesr_net}
%\subsection{Network Architecture}
The major components of our Deep SESR model are: residual dense blocks (RDBs), a feature extraction network (FENet), and an auxiliary attention network (AAN). These components are tied to an end-to-end architecture for the combined SESR learning.

\begin{figure}[ht]
\centering
\vspace{1mm}
\includegraphics[width=0.75\linewidth]{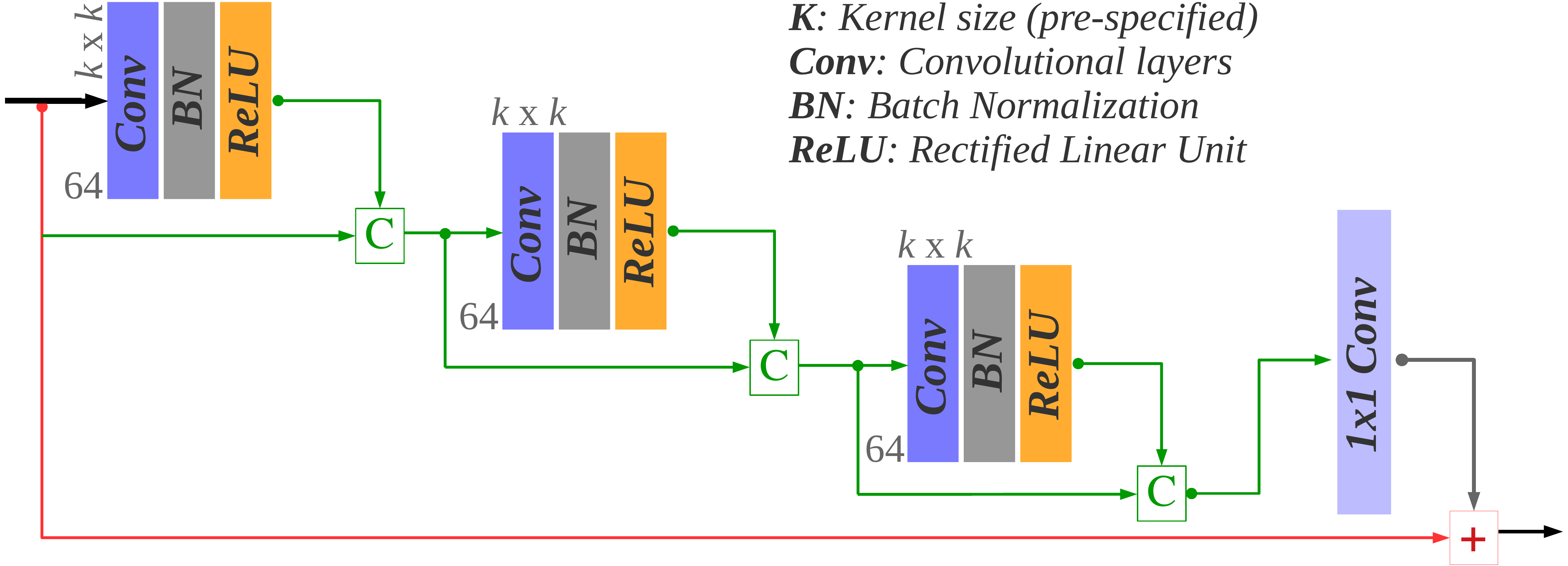}%
\vspace{-2mm}
 \caption{A residual dense block (RDB)~\cite{zhang2018residual}.}%
 %\vspace{1mm}
 \label{fig:model_rdb}
\end{figure}

\vspace{1mm}
\textbf{Residual Dense Blocks (RDBs)} consist of three sets of convolutional ({\tt Conv}) layers, each followed by Batch Normalization ({\tt BN})~\cite{ioffe2015batch} and {\tt ReLU} non-linearity~\cite{nair2010rectified}. As Figure~\ref{fig:model_rdb} illustrates, the input and output of each layer is concatenated to subsequent layers. This architecture is inspired by Zhang \etal~\cite{zhang2018residual} who demonstrated that such dense skip connections facilitate an improved hierarchical feature learning. Each {\tt Conv} layer learns $64$ filters of a given kernel size; their outputs are then fused by a $1\times1$ {\tt Conv} layer for local residual learning.

\begin{figure}[ht]
\centering
    \subfigure[The feature extraction network (FENet).]{
    \includegraphics[width=0.86\textwidth]{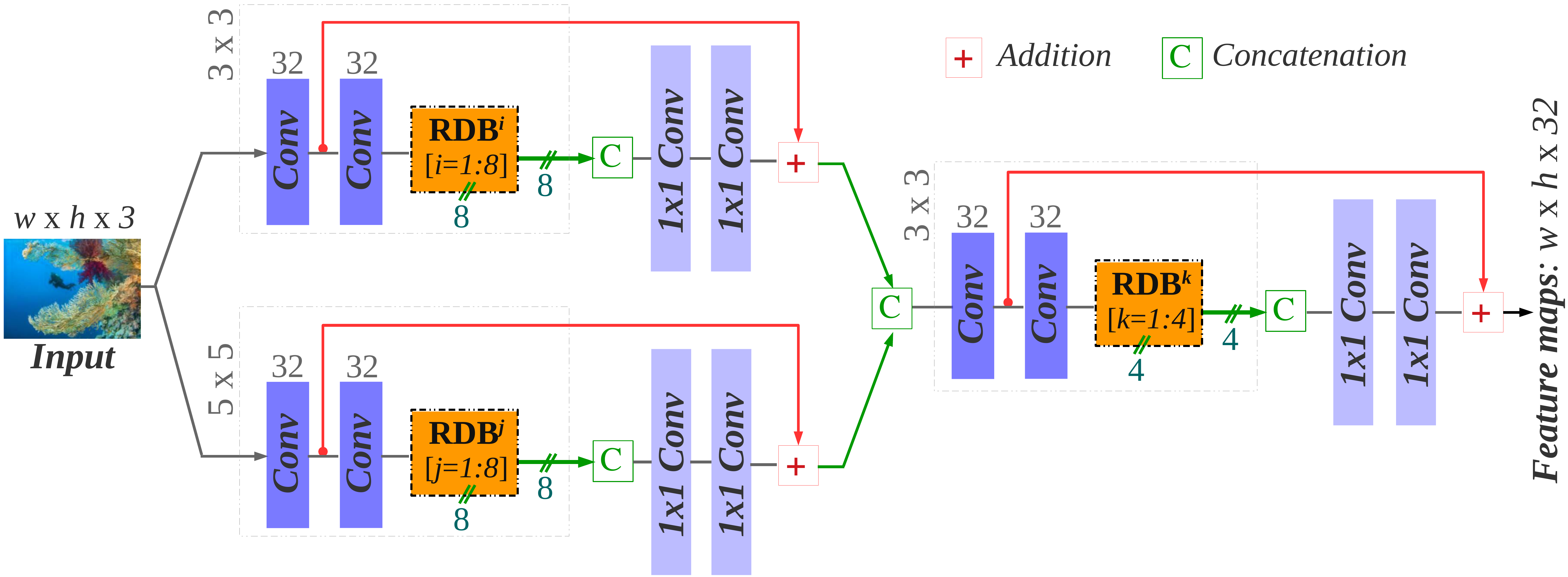}
    \label{fig:model_fnet}
    }
    \vspace{2mm}
    
    \subfigure[The end-to-end architecture is shown. FENet-extracted feature maps are propagated along two branches: \textit{i)} to AAN for learning saliency, and \textit{ii)} to an intermediate convolutional layer for learning enhancement. Another convolutional layer and subsequent upsampling layers learn SESR along the main branch.]{
    \includegraphics[width=0.823\textwidth]{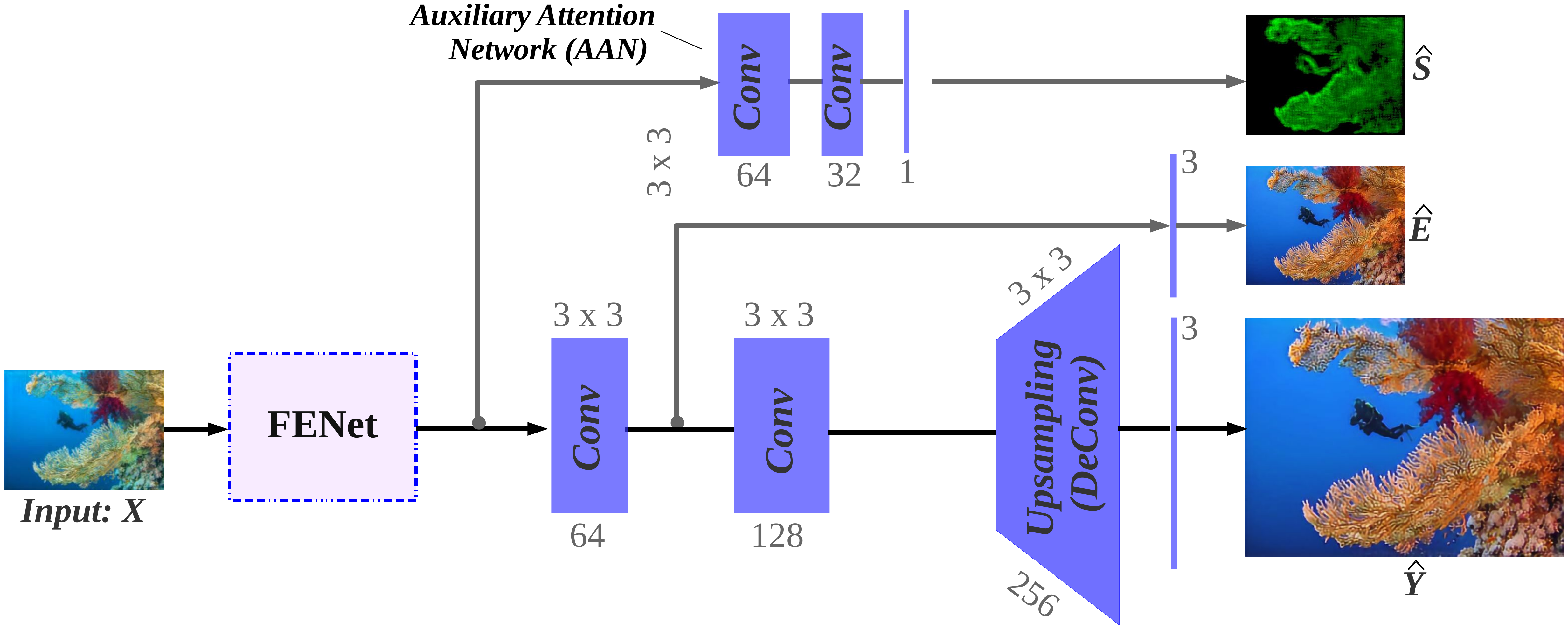}
    \label{fig:deep_sesr}
    }
    
\caption{Network architecture and detailed parameter specification of the proposed Deep SESR model.}
\label{fig:model_sesr}
\end{figure}

\textbf{Feature Extraction Network (FENet)} uses RDBs as building blocks to incorporate two-stage residual-in-residual learning. As shown in Figure~\ref{fig:model_fnet}, on the first stage, two parallel branches use eight RDB blocks each to separately learn $3\times3$ and $5\times5$ filters in input image space; these filters are then concatenated and passed to a common branch for the second stage of learning. Four RDB blocks with $3\times3$ filters are used in the later stage which eventually generates $32$ feature maps. % as the final output. 
Our motive for such design is to have the capacity to learn locally dense informative features while still maintaining a globally shallow architecture to ensure fast feature extraction.
%during inference.               

\vspace{1mm}
\textbf{Auxiliary Attention Network (AAN)} learns to model visual attention in the FENet-extracted feature space. As shown in Figure~\ref{fig:deep_sesr}, two sequential {\tt Conv} layers learn to generate a single channel output that represents saliency (probabilities) for each pixel. We show the predicted saliency map as \textit{green} intensity values; the black pixels represent background regions.

 \vspace{1mm}
\textbf{The Deep SESR learning} is guided along the primary branch by a series of {\tt Conv} and {\tt DeConv} (de-convolutional) layers. As Figure~\ref{fig:deep_sesr} demonstrates, the enhanced image (LR), and the SESR image (HR) are generated by separate output layers at different stages in the network. The enhanced image is generated from the {\tt Conv} layer that immediately follows FENet; it is supervised to learn enhancement by dedicated loss functions applied at the shallow output layer. The \textit{enhanced} features are also propagated to another {\tt Conv} layer, followed by {\tt DeConv} layers for upsampling. The final SESR output is generated from upsampled features based on the given scale: $2\times$, $3\times$, or $4\times$. %The loss functions applied at the final layer and
Other model parameters, \eg, the number of filters, kernel sizes, etc., are annotated in Figure~\ref{fig:model_sesr}.

\subsection{Loss Function Formulation}\label{loss_fun}
The end-to-end training of Deep SESR is supervised by seven loss components that address various aspects of learning the function $G: X\rightarrow S, E, Y$. By denoting $\hat{S},\hat{E},\hat{Y} = G(X)$ as the generated output, we formulate the loss terms as follows:      

\vspace{1mm}
\textit{i)} \textbf{Information Loss} for saliency prediction is measured by a standard cross-entropy function~\cite{lu2016hierarchical,wang2017deep}. 
It quantifies the dissimilarity in pixel intensity distributions between the generated saliency map ($\hat{S}$) and its ground truth ($S$). For a total of $N_p$ pixels in $\hat{S}$, it is calculated as   
\begin{equation}\small
    \mathcal{L}_{Saliency}^{AAN} = \frac{1}{N_p}  \sum_{p=1}^{N_p}\big[ -S_p \log \hat{S}_p - (1-S_p) \log (1-\hat{S}_p) \big].
\end{equation}

\vspace{1mm}
\textit{ii)} \textbf{Contrast loss (LR)} evaluates the hue and luminance recovery in the enhanced images. The dominating green/blue hue in distorted underwater images often causes low-contrast and globally dim foreground pixels. We quantify this loss of relative strength (\ie, intensity) in foreground pixels in RGB space by utilizing a differentiable function: Contrast Measurement Index (CMI)~\cite{shaus2017potential,trivedi2013novel}. The CMI measures the average intensity of foreground pixels ($F_I$) relative to the background ($B_I$) for an image $I$, as 
$CMI(I) = \frac{(F_I - B_I)}{(F_I + B_I)} \text{ } \propto (F_I - B_I)$. We exploit the saliency map $S$ (or $\hat{S})$ to find the foreground pixels in $E$ (or $\hat{E})$, as $F_E = E\odot S$ and $F_{\hat{E}} = \hat{E}\odot \hat{S}$; here, $\odot$ denotes element-wise multiplication. Subsequently, we compute the contrast loss as 
\begin{equation} \small
    \mathcal{L}_{Contrast}^{LR} = \big|\big| CMI(E)-CMI(\hat{E}) \big|\big|_2.
\end{equation}

\begin{figure}[ht]
\centering
    \subfigure[Contrast measure: $CMI(I)$.]{
    \includegraphics[width=0.36\textwidth]{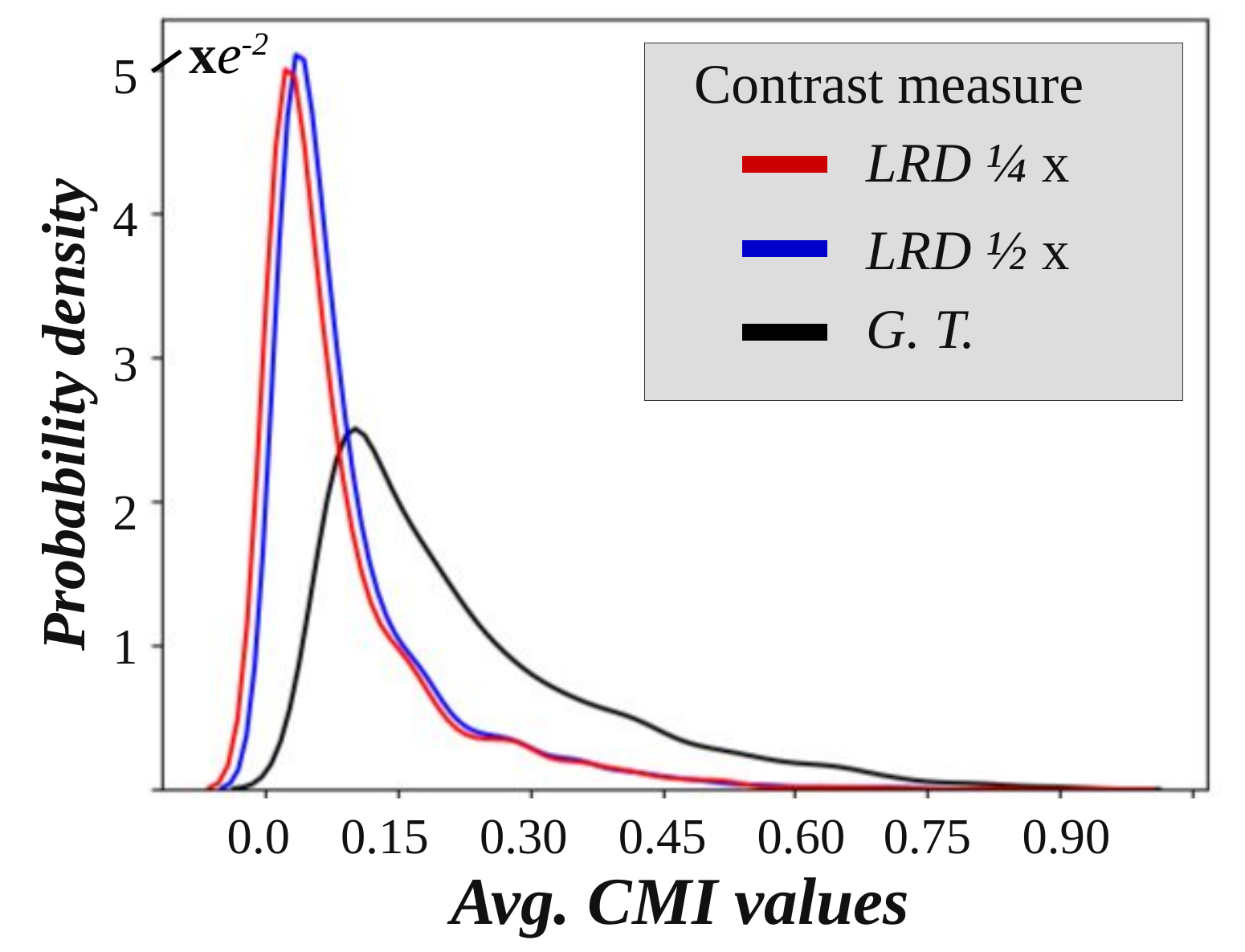}
    \label{fig:con_a}
    }~\subfigure[Sharpness measure: $|\nabla I|$.]{
    \includegraphics[width=0.36\textwidth]{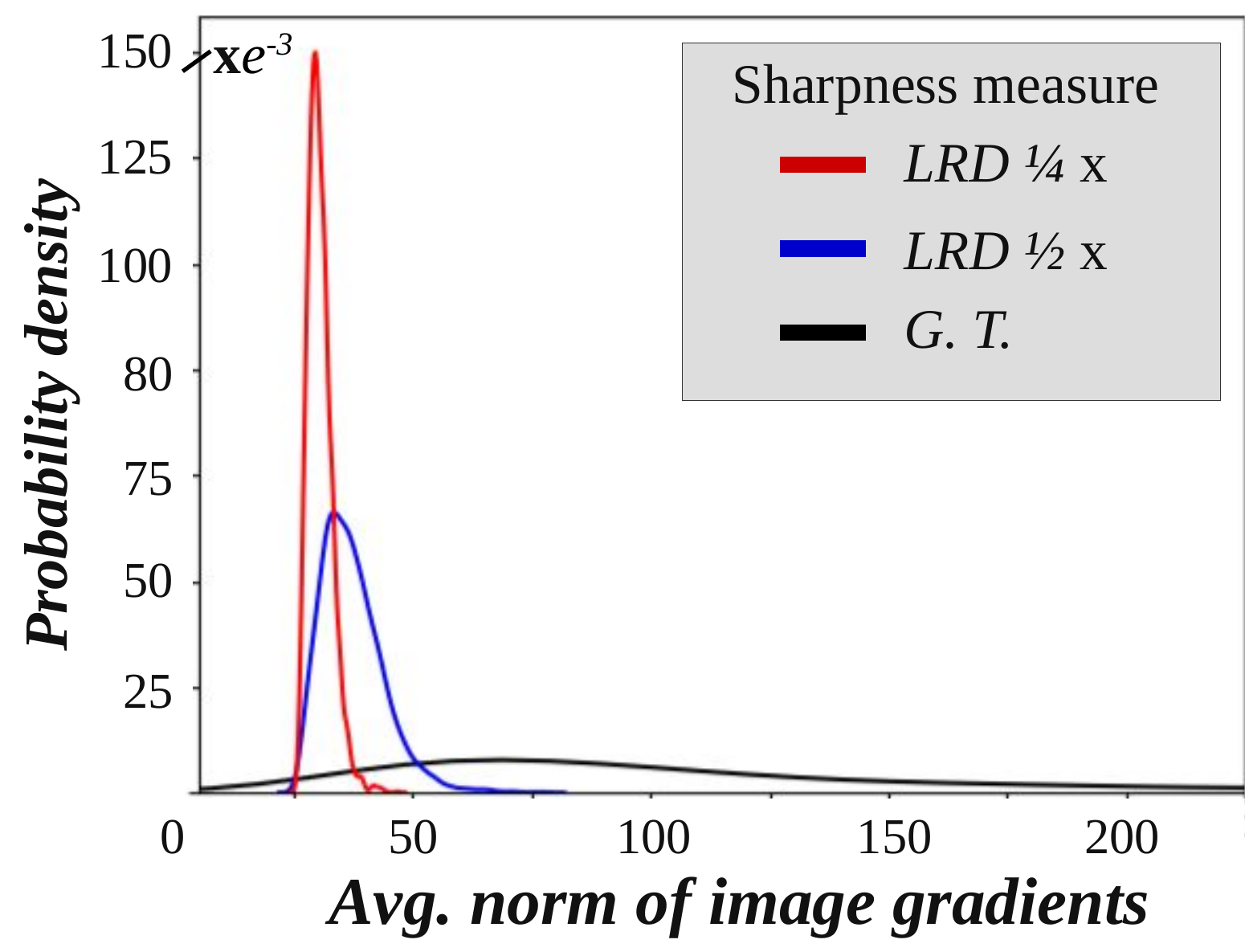}
    \label{fig:sharp_b}
    }
    \vspace{1mm}
    
    \subfigure[Image contrast and sharpness properties of a particular sample compared to its ground truth measurement.]{
    \includegraphics[width=0.75\textwidth]{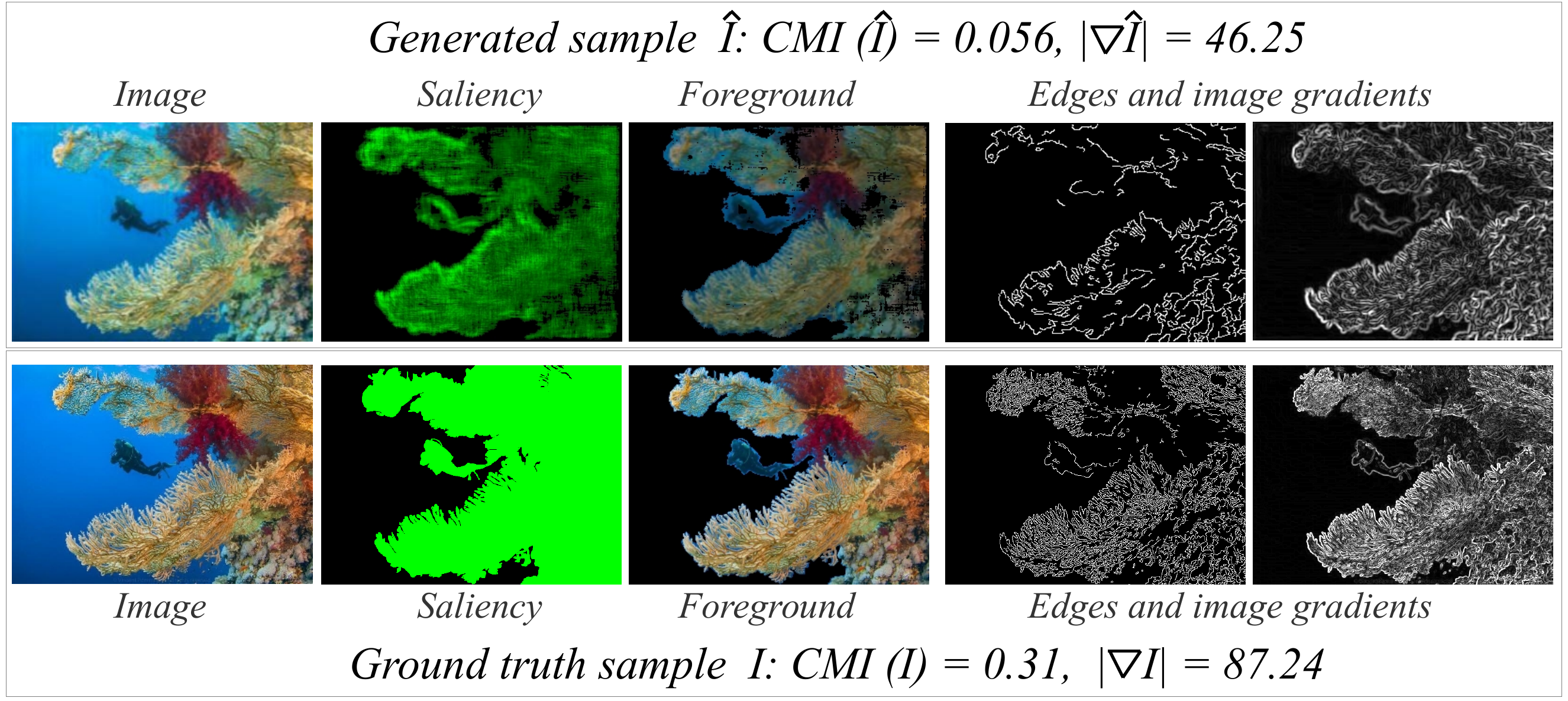}
    \label{fig:cons_sharp_exp}
    }
    
\caption{The lack of contrast and sharpness in LRD samples of UFO-120 dataset (compared to their ground truth) are shown in (a) and (b); as seen, distributions for LRD samples are densely skewed to lower values, whereas the ground truth distributions span considerably higher values. A qualitative interpretation of this numeric disparity is illustrated in (c).}
\label{fig:loss}
\end{figure}

An immediate consequence of using $\mathcal{L}_{Contrast}^{LR}$ is that AAN can directly influence learning enhancement despite being on a separate branch. Such coupling also provides better training stability (otherwise AAN tends to converge too early and starts over-fitting). Moreover, in Figure~\ref{fig:con_a}, we show the distributions of CMI for training samples of the UFO-120 dataset, which suggests that the distorted samples' CMI scores are skewed to much lower values compared to the ground truth. Hence, $\mathcal{L}_{Contrast}^{LR}$ forces the CMI distribution to shift toward higher values for learning contrast enhancement.

\vspace{1mm}
\textit{iii)} \textbf{Color Loss (LR/HR)} evaluates global similarity of the enhanced ($\hat{E}$) and SESR ($\hat{Y}$) output with respective ground truth measurements in RGB space. The standard $\mathcal{L}_{2}$ loss terms are: $	\mathcal{L}_{2}^{LR} = \big|\big|E-\hat{E}\big|\big|_2, \text{ and  } \mathcal{L}_{2}^{HR} = \big|\big|Y-\hat{Y}\big|\big|_2$. Additionally, we formulate two \textit{perceptual loss} functions that are particularly designed for learning underwater image enhancement and super-resolution. First, we utilize two wavelength-dependent chrominance terms: $C_{rg}=(r- g)$, and $C_{yb}=\frac{1}{2}(r+g)-b$, which are core elements of the Underwater Image Colorfulness Measure (UICM)~\cite{panetta2016human,liu2019real}. By denoting $\Delta r$, $\Delta g$, and $\Delta b$, as the per-channel numeric differences between $\hat{E}$ and $E$, we formulate the loss as:  
\begin{equation}\small
	\mathcal{L}_{P}^{LR} = \big|\big| 4{ (\Delta r - \Delta g})^2 + ({\Delta r + \Delta  g - 2\Delta b})^2  \big|\big|_2.
\end{equation}
On the other hand, being inspired by~\cite{compuphase,islam2019underwater}, we evaluate the perceptual similarity at HR as   
\begin{equation}\small
	\mathcal{L}_{P}^{HR} = \big|\big| \frac{(512+\bar{R})}{256} \Delta R^2 + 4 \Delta G^2+ \frac{(767-\bar{R})}{256}\Delta B^2  \big|\big|_2.
\end{equation}
Here, $\bar{R} = (R_{Y} + R_{\hat{Y}})/2$, whereas $\Delta R$, $\Delta G$, and $\Delta B$ are the per-channel disparities between $\hat{Y}$ and $Y$. Finally, we adopt the color loss terms for enhancement and SESR as   
\begin{align}\small
    \mathcal{L}_{Color}^{LR} &= 0.25 \text{ } \mathcal{L}_{P}^{LR} + 0.75 \text{ } \mathcal{L}_{2}^{LR}, \text{ and } \\
    \mathcal{L}_{Color}^{HR} &= 0.25 \text{ } \mathcal{L}_{P}^{HR} + 0.75 \text{ } \mathcal{L}_{2}^{HR},\text{ respectively.}
\end{align}

\vspace{1mm}
\textit{iv)} \textbf{Content loss (LR/HR)} forces the generator to restore a similar \textit{feature content} as the ground truth in terms of high-level representation. Such feature preservation has been found to be very effective for image enhancement, style transfer, and SISR problems~\cite{ignatov2017dslr,islam2019underwater}; as suggested in~\cite{johnson2016perceptual}, we define the image content function $\Phi_{VGG}(\cdot)$ as high-level features extracted by the last {\tt Conv} layer of a pre-trained VGG-19 network. Then, we formulate the content loss for enhancement and SESR as
\begin{align}\small
    \mathcal{L}_{Content}^{LR} &= \big|\big|\Phi_{VGG} (E) - \Phi_{VGG} (\hat{E}) \big|\big|_2, \text{ and } \\
    \mathcal{L}_{Content}^{HR} &= \big|\big|\Phi_{VGG} (Y) - \Phi_{VGG} (\hat{Y}) \big|\big|_2,\text{ respectively.}
\end{align}

\vspace{1mm}
\textit{v)} \textbf{Sharpness loss (HR)} measures the blurriness recovery in SESR output by exploiting local image gradients.  The literature offers several solutions for evaluating image sharpness based on norm/histogram of gradients or frequency-domain analysis. In particular, the notions of Just Noticable Blur (JNB)~\cite{ferzli2009no} and Perceptual Sharpness Index (PSI)~\cite{feichtenhofer2013perceptual} are widely used; they apply non-linear transformation and thresholding on local contrast or gradient-based features to quantify perceived blurriness based on the characteristics of human visual system.   
However, we found better results and numeric stability by using the norm of image gradients directly; specifically, we use the standard $3\times3$ Sobel operator~\cite{gao2010improved} for computing spatial gradient $\nabla I = \sqrt{I_x^2 + I_y^2}$ for an image $I$. Subsequently, we formulate the sharpness loss for SESR as     
\begin{equation} \small
    \mathcal{L}_{Sharpness}^{HR} = \big|\big| \text{ } |\nabla Y|^2 - |\nabla \hat{Y}|^2 \text{ } \big|\big|_1.
\end{equation}

In Figure~\ref{fig:sharp_b}, we present a statistical validity of $\mathcal{L}_{sharpness}^{LR}$ as a loss component; also, edge gradient features for a particular sample are provided in Figure~\ref{fig:cons_sharp_exp}. As shown, numeric disparities for the norm of gradients between distorted images and their HR ground truth are significant, which we quantify by $\mathcal{L}_{sharpness}^{LR}$ to encourage sharper image generation.

\subsection{End-to-end Training Objective}
We use a linear combination of the above-mentioned loss components to formulate the unified objective function as   
\begin{align} \label{eq:final}
\centering
\small
	G^* = \argmin\limits_{G} \big\{ 
	 \lambda_s^{AAN} \mathcal{L}_{Saliency}^{AAN} + \mathcal{L}_{SESR}^{LR} + \mathcal{L}_{SESR}^{HR}  \big\}\text{;}  
\end{align}
where $\mathcal{L}_{SESR}^{LR}$ and $\mathcal{L}_{SESR}^{HR}$ are expressed by   
\begin{align*} %\label{eq:pix2pix_final}
\centering
\small
	\mathcal{L}_{SESR}^{LR} &= 
	 \lambda_c^{LR} \mathcal{L}_{Color}^{LR}+\lambda_f^{LR}  \mathcal{L}_{Content}^{LR}+\lambda_t^{LR} \mathcal{L}_{Contrast}^{LR},\text{ and} \\
	\mathcal{L}_{SESR}^{HR} &= 
	 \lambda_c^{HR}  \mathcal{L}_{Color}^{HR}+\lambda_f^{HR} \mathcal{L}_{Content}^{HR}+\lambda_g^{HR} \mathcal{L}_{Sharpness}^{HR}.
\end{align*}
Here, $\lambda_{\odot}^{\Box}$ symbols are scaling factors that represent the contributions of respective loss components; their values are empirically tuned as hyper-parameters.

\section{Experimental Results}
\subsection{Implementation Details}\label{sesr_implement}
As mentioned in Section~\ref{sec:data_prep}, Deep SESR training is supervised by paired data of the form $(\{X\}, \{S, E, Y\})$. We use TensorFlow libraries~\cite{abadi2016tensorflow} to implement the optimization pipeline (of Eq.~\ref{eq:final}); a Linux host with two Nvidia\texttrademark{ }GTX 1080 graphics cards are used for training. Adam optimizer~\cite{kingma2014adam} is used for the global iterative learning with a rate of $10^{-4}$ and a momentum of $0.5$; the network converges within $23$-$26$ epochs of training in this setup with a batch-size of $2$. In the following sections, we present the experimental results based on qualitative analysis, quantitative evaluations, and ablation studies. Since there are no existing SESR methods, we compare the Deep SESR performance separately with SOTA image enhancement and super-resolution models. Note that, all models in comparison are trained on the same train-validation splits (of respective datasets) by following their recommended parameter settings. Also, for datasets other than UFO-120, the AAN (and $\mathcal{L}_{Contrast}^{LR}$) is not used by Deep SESR as their ground truth saliency maps are not available.

\begin{figure}[ht]
\vspace{1mm}
\centering
        \includegraphics[width=0.98\linewidth]{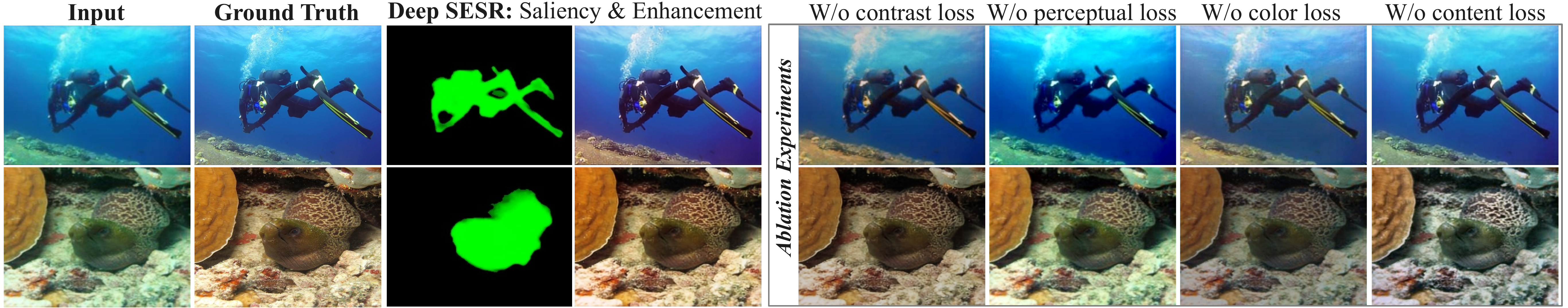}
        \vspace{-3mm}
        \caption{Each row demonstrates perceptual enhancement and saliency prediction by Deep SESR on respective LRD input images; the corresponding results of an ablation experiment shows contributions of various loss-terms in the learning.}%
\label{fig:qual_en}
\end{figure}

\subsection{Evaluation: Enhancement}
We first qualitatively analyze the Deep SESR-generated images in terms of color, contrast, and sharpness. As Figure~\ref{fig:qual_en} illustrates, the enhanced images are perceptually similar to the respective ground truth. Specifically, the greenish underwater hue is rectified, true pixel colors are mostly restored, and the global image sharpness is recovered. 
Moreover, the generated saliency map suggests that it focused on the right foreground regions for contrast improvement. We further demonstrate the contributions of each loss-term: $\mathcal{L}_{Contrast}^{LR}$, $\mathcal{L}_{P}^{LR}$, $\mathcal{L}_{Color}^{LR}$, and $\mathcal{L}_{Content}^{LR}$ for learning the enhancement. 
We observe that the color rendition gets impaired without $\mathcal{L}_{P}^{LR}$ and $\mathcal{L}_{Color}^{LR}$, whereas, $\mathcal{L}_{Content}^{LR}$ contributes to learning finer texture details. We also notice a considerably low-contrast image generation without $\mathcal{L}_{Contrast}^{LR}$, which validates the utility of saliency-driven contrast evaluation via CMI (see Section~\ref{loss_fun}).

%\begin{landscape}
\begin{table}[H]
\centering
\caption{Quantitative performance comparison for enhancement baesd on PSNR, SSIM, and UIQM; scores are shown as $\textit{mean} \pm \sqrt{\textit{variance}}$. %; the first and second best scores (in each row) are colored red, and blue, respectively.
}
%\footnotesize
\scriptsize
\vspace{2mm}

\begin{tabular}{c|l||c|c|c|c|c}
  \multicolumn{7}{c}{(a) Comparison with physics-based models.} \\
  \Xhline{2\arrayrulewidth}
  & \textbf{Dataset} & RGHS & UCM & MS-Fusion & MS-Retinex & \textbf{Deep SESR}\\ \Xhline{2\arrayrulewidth}
  \parbox[t]{1mm}{\multirow{3}{*}{\rot{\textit{PSNR}}}} & UFO-120 & $20.05 \pm 3.1$ & $20.99 \pm 2.2$ & $21.32 \pm 3.3$ & $21.69 \pm 3.6$ & {$27.15 \pm 3.2$} \\ %\hline
  & EUVP & $20.12 \pm 2.9$ & $20.55 \pm 1.8$  & $19.85 \pm 2.4$ & $21.27 \pm 3.1$ & {$25.25 \pm 2.1$} \\ %\hline
  & UImNet & $19.98 \pm 1.8$  & $20.48 \pm 2.2$ & $19.59 \pm 3.2$ & $22.63 \pm 2.5$ & {$25.52 \pm 2.7$} \\ 
  \hline
  \parbox[t]{1mm}{\multirow{3}{*}{\rot{\textit{SSIM}}}} & UFO-120 & $0.75 \pm 0.06$ & $0.78 \pm 0.07$ & $0.79 \pm 0.09$ & $0.75 \pm 0.10$ & {$0.84 \pm 0.03$} \\ %\hline
  & EUVP &  $0.69 \pm 0.11$ & $0.73 \pm 0.14$  & $0.70 \pm 0.05$ & $0.69 \pm 0.15$ & {$0.75 \pm 0.07$} \\ %\hline
  & UImNet & $0.61 \pm 0.08$  & $0.67 \pm 0.06$ & $0.64 \pm 0.11$ & $0.74 \pm 0.04$ & {$0.81 \pm 0.05$} \\ 
  \hline
  \parbox[t]{1mm}{\multirow{3}{*}{\rot{\textit{UIQM}}}} & UFO-120 & $2.36 \pm 0.33$ & $2.41 \pm 0.53$ & $2.76 \pm 0.45$ & $2.69 \pm 0.59$ & {$3.13 \pm 0.45$} \\ 
  & EUVP &  $2.45 \pm 0.46$ &  $2.48 \pm 0.77$ & $2.51 \pm 0.36$ & $2.48 \pm 0.09$ & {$2.98 \pm 0.28$} \\ 
  & UImNet & $2.32 \pm 0.48$ & $2.38 \pm 0.42$ & $2.79 \pm 0.55$ & $2.84 \pm 0.37$ & {$3.26 \pm 0.36$} \\ 
 \Xhline{2\arrayrulewidth}
\end{tabular}

\vspace{3mm}
\begin{tabular}{c|l||c|c|c|c|c}
  \multicolumn{7}{c}{(b) Comparison with learning-based models.} \\
  \Xhline{2\arrayrulewidth}
  & \textbf{Dataset} & Water-Net & UGAN & Fusion-GAN & FUnIE-GAN & \textbf{Deep SESR}\\ \Xhline{2\arrayrulewidth}
  \parbox[t]{1mm}{\multirow{3}{*}{\rot{\textit{PSNR}}}} & UFO-120 & $22.46 \pm 1.9$ & $23.45 \pm 3.1$ & $24.07 \pm 2.1$ & {$25.15 \pm 2.3$} & {$27.15 \pm 3.2$} \\ %\hline
  & EUVP & $20.14 \pm 2.3$ & $23.67 \pm 1.5$ & $23.77 \pm 2.4$ & {$26.78 \pm 1.1$} & {$25.25 \pm 2.1$} \\ %\hline
  & UImNet & $21.02 \pm 1.6$ & $23.88 \pm 2.1$ & $23.12 \pm 1.9$ & {$24.68 \pm 2.4$} & {$25.52 \pm 2.7$} \\ 
  \hline
  \parbox[t]{1mm}{\multirow{3}{*}{\rot{\textit{SSIM}}}} & UFO-120 & $0.79 \pm 0.05$ & $0.80 \pm 0.08$ & {$0.82 \pm 0.07$} & {$0.82 \pm 0.08$} & {$0.84 \pm 0.03$} \\ %\hline
  & EUVP & $0.68 \pm 0.18$ & $0.67 \pm 0.11$ & $0.68 \pm 0.05$ & {$0.86 \pm 0.05$} & {$0.75 \pm 0.07$} \\ %\hline
  & UImNet & $0.71 \pm 0.07$ & {$0.79 \pm 0.08$} & $0.75 \pm 0.07$ & $0.77 \pm 0.06$ & {$0.81 \pm 0.05$} \\ 
  \hline
  \parbox[t]{1mm}{\multirow{3}{*}{\rot{\textit{UIQM}}}} & UFO-120 & $2.83 \pm 0.48$ & $3.04 \pm 0.28$ & $2.98 \pm 0.28$ & {$3.09 \pm 0.51$} & {$3.13 \pm 0.45$} \\ 
  & EUVP & $2.55 \pm 0.06$ & $2.70 \pm 0.31$ & $2.58 \pm 0.07$ & {$2.95 \pm 0.38$} & {$2.98 \pm 0.28$} \\ 
  & UImNet & $2.92 \pm 0.35$ & {$3.32 \pm 0.55$} & $3.19 \pm 0.27$ & $3.23 \pm 0.32$ & {$3.26 \pm 0.36$} \\ 
 \Xhline{2\arrayrulewidth}
\end{tabular}

\label{tab:quan_se}
\end{table}

%\end{landscape}

\begin{figure}[t]
    \centering
        \includegraphics[width=0.98\linewidth]{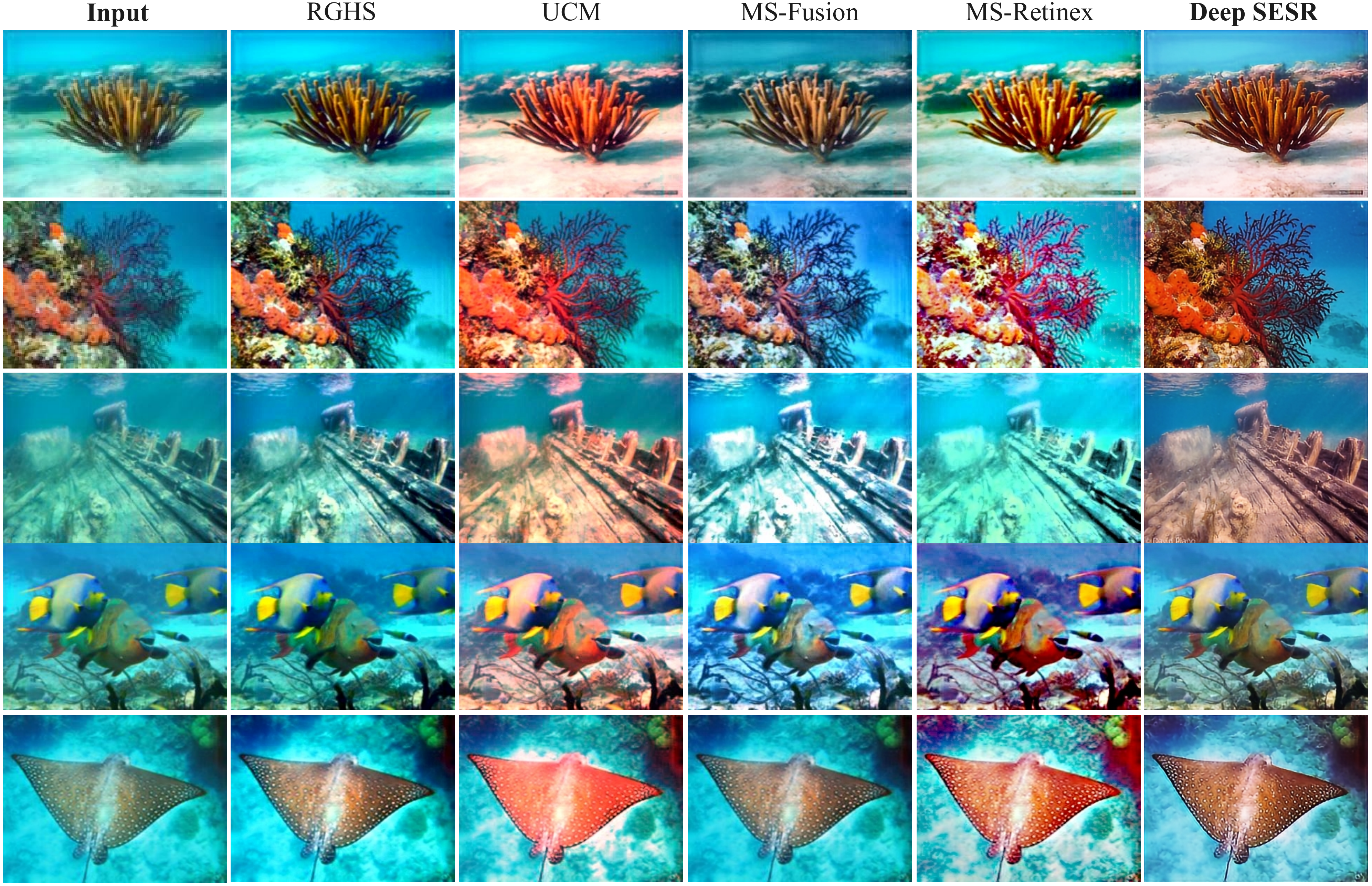}%
        \vspace{-3mm}
        \caption{Qualitative comparison of Deep SESR-enhanced images with physics-based models: RGHS~\cite{huang2018shallow}, UCM~\cite{iqbal2010enhancing}, MS-Fusion~\cite{ancuti2012enhancing}, and MS-Retinex~\cite{zhang2017underwater}. (Best viewed at 300\% zoom)}%
        \label{fig:quan_en_comp1}
\end{figure}

\begin{figure}[t]
    \centering
        \includegraphics[width=0.98\linewidth]{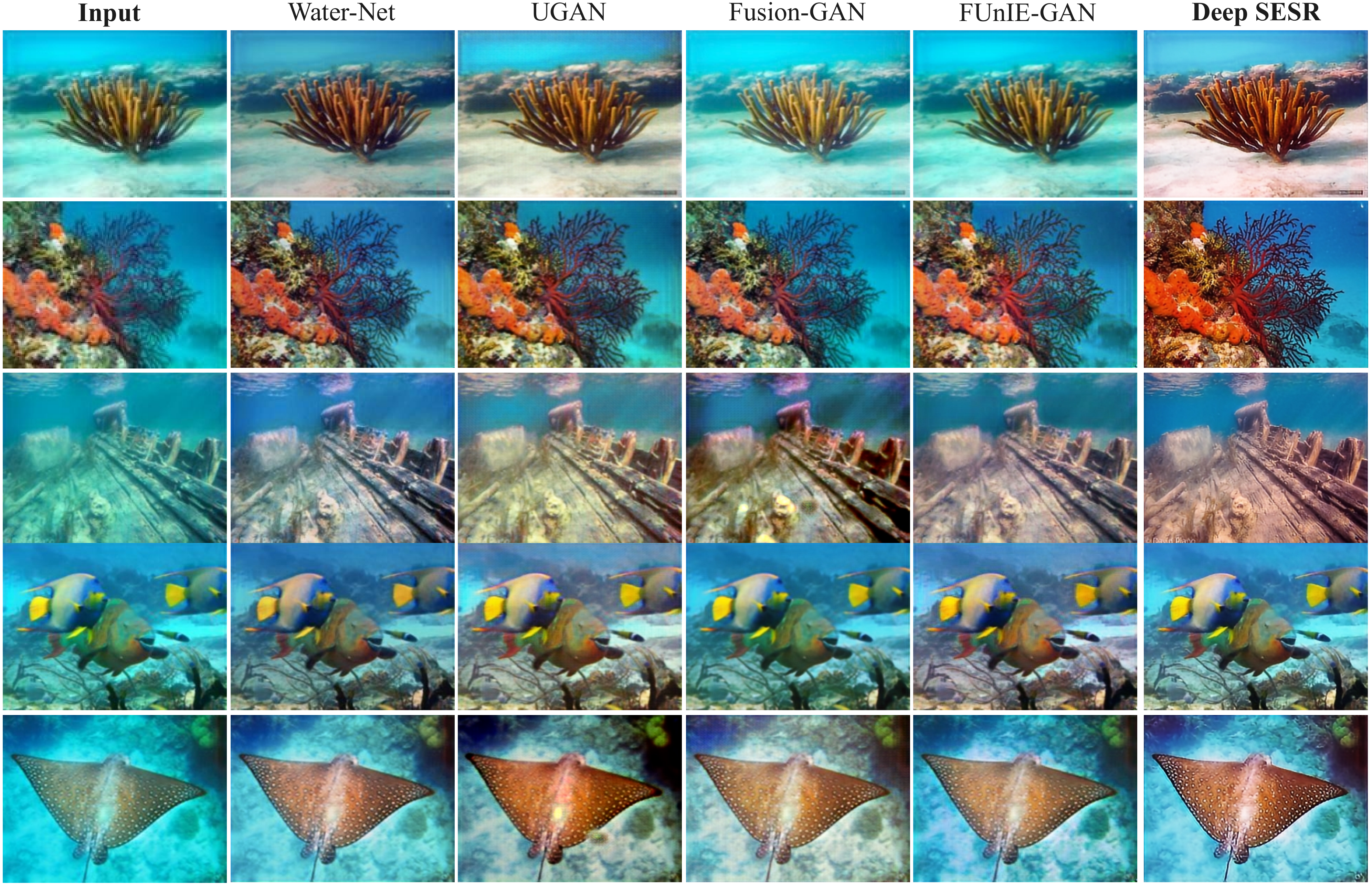}%
        \vspace{-3mm}
        \caption{Qualitative comparison of Deep SESR-enhanced images with learning-based models: Water-Net~\cite{li2019underwater}, UGAN~\cite{fabbri2018enhancing}, Fusion-GAN~\cite{li2019fusion}, and FUnIE-GAN~\cite{islam2019fast}. (Best viewed at 300\% zoom)}%
        \label{fig:quan_en_comp2}
\end{figure}

Next, we compare the perceptual image enhancement performance of Deep SESR with the following models: (i) relative global histogram stretching (RGHS)~\cite{huang2018shallow}, (ii) unsupervised color correction (UCM)~\cite{iqbal2010enhancing}, (iii) multi-scale fusion (MS-Fusion)~\cite{ancuti2012enhancing}, (iv) multi-scale Retinex (MS-Retinex)~\cite{zhang2017underwater}, (v) Water-Net~\cite{li2019underwater}, (vi) UGAN~\cite{fabbri2018enhancing}, (vii) Fusion-GAN~\cite{li2019fusion}, and (viii) FUnIE-GAN~\cite{islam2019fast}. The first four are physics-based models and the rest are learning-based models; they provide SOTA performance for underwater image enhancement in RGB space (without requiring scene depth or optical waterbody measures). Their performance is quantitatively evaluated on common test sets of each dataset based on standard metrics~\cite{islam2019fast,liu2019real}: peak signal-to-noise ratio (PSNR)~\cite{hore2010image}, structural similarity measure (SSIM)~\cite{wang2004image}, and underwater image quality measure (UIQM)~\cite{panetta2016human}. The PSNR and SSIM quantify reconstruction quality and structural similarity of generated images with respect to ground truth, whereas the UIQM evaluates image qualities based on colorfulness, sharpness, and contrast. We already introduced these standard metrics in Chapter~\ref{en_sr} (Section~\ref{funiegan-quant}); their definitions and relevant details are provided in Appendix~\ref{ApenD}. The evaluation is summarized in Table~\ref{tab:quan_se}; moreover, a few qualitative comparisons are shown in Figure~\ref{fig:quan_en_comp1} and Figure~\ref{fig:quan_en_comp2}.    

%\end{landscape}

%\end{landscape}

As Figure~\ref{fig:quan_en_comp1}--\ref{fig:quan_en_comp2} demonstrates, UCM and MS-Retinex often suffer from over-saturation, whereas RGBH, MS-Fusion, and Water-Net fall short in hue rectification. In comparison, the color restoration and contrast enhancement of UGAN, Fusion-GAN, and FUnIE-GAN are generally better. In addition to achieving comparable color recovery and hue rectification, the Deep SESR-generated images are considerably sharper. Since the boost in performance is rather significant for UFO-120 dataset (suggested by the results of Table~\ref{tab:quan_se}), it is likely that the additional knowledge about foreground pixels through $\mathcal{L}_{Contrast}^{LR}$ helps in this regard. Deep SESR achieves competitive and often better performance in terms of PSNR and SSIM as well. In particular, it generally attains better UIQM scores; we postulate that $\mathcal{L}_{P}^{LR}$ contributes to this enhancement, as it is designed to improve the UICM (see Section~\ref{loss_fun}). Further ablation investigations reveal a $9.47\%$ drop in UIQM values without using $\mathcal{L}_{P}^{LR}$ in the learning objective.

\begin{table*}[t]
\centering
\caption{Quantitative performance comparison for super-resolution for PSNR, SSIM, and UIQM; scores are shown as $\textit{mean} \pm \sqrt{\textit{variance}}$. (Results for $3\times$ SISR are omitted due to space constraints; those results and other supplementary materials can be found in this pre-print: \url{https://arxiv.org/pdf/2002.01155.pdf})}
%\footnotesize
\scriptsize
\vspace{2mm}
\begin{tabular}{c|l||c|c||c|c||c|c}
  \Xhline{2\arrayrulewidth}
   & & \multicolumn{2}{c||}{\textit{PSNR}} & \multicolumn{2}{c||}{\textit{SSIM}} & \multicolumn{2}{c}{\textit{UIQM}} \\  
  \cline{3-8} 
  & \textbf{Model} & $2\times$ & $4\times$ & $2\times$ & $4\times$ & $2\times$  & $4\times$ \\ \Xhline{2\arrayrulewidth}
  \parbox[t]{1mm}{\multirow{6}{*}{\rot{{UFO-120}}}} & SRCNN & $24.75 \pm 3.7$ & $19.05 \pm 2.3$  & $0.72 \pm .07$  & $0.56 \pm .12$  & $2.39 \pm 0.35$  & $2.02 \pm 0.47$    \\ 
  & SRResNet & $25.23 \pm 4.1$  & $19.13 \pm 2.4$  & $0.74 \pm .08$  &  $0.56 \pm .05$  & $2.42 \pm 0.37$ &  $2.09 \pm 0.30$   \\ 
  & SRGAN & $26.11 \pm 3.9$  & $21.08 \pm 2.3$  & $0.75 \pm .06$  & $0.58 \pm .09$  & $2.44 \pm 0.28$  &  $2.26 \pm 0.17$   \\ 
  & RSRGAN & $25.25 \pm 4.3$ & $20.25 \pm 2.4$  & { $0.79 \pm .08$} & $0.58 \pm .04$  & $2.41 \pm 0.29$  &  $2.27 \pm 0.22$   \\ 
  & SRDRM & $26.23 \pm 4.4$ & { $22.26 \pm 2.5$}  & { $0.79 \pm .09$} & { $0.59 \pm .05$}  & { $2.45 \pm 0.43$}  & { $2.28 \pm 0.35$}  \\ 
  & SRDRM-GAN & { $26.26 \pm 4.3$} & $22.21 \pm 2.4$  & $0.78 \pm .08$ & $0.58 \pm .13$  & $2.42 \pm 0.30$   &  $2.27 \pm 0.44$   \\ 
  & \textbf{Deep SESR} & { $28.57 \pm 3.5$}  & { $24.75 \pm 2.8$}  & { $0.85 \pm .09$}  & { $0.66 \pm .05$}  & { $3.09 \pm 0.41$}  & { $2.55 \pm 0.35$}  \\ \hline
\parbox[t]{1mm}{\multirow{6}{*}{\rot{{USR-248}}}} & SRCNN & $24.88 \pm 4.4$  & $23.75 \pm 3.2$  &  $0.73 \pm .08$ & $0.69 \pm .12$  & $2.38 \pm 0.38$  & $2.21 \pm 0.68$   \\ 
  & SRResNet &  $24.96 \pm 3.7$ & $22.21 \pm 3.6$  &  $0.74 \pm .07$ & $0.70 \pm .08$  &  $2.42 \pm 0.48$ & $2.27 \pm 0.70$    \\ 
  & SRGAN & $25.76 \pm 3.5$  & $24.36 \pm 4.3$  &  $0.77 \pm .06$ & $0.69 \pm .13$  & $2.53 \pm 0.42$  &  $2.75 \pm 0.66$\\ 
  & RSRGAN & $25.11 \pm 2.9$  &  $24.15 \pm 2.9$ & $0.75 \pm .06$  & { $0.71 \pm .09$} & $2.42 \pm 0.35$  & $2.55 \pm 0.47$    \\ 
  & SRDRM & $26.16 \pm 3.5$  & { $24.96 \pm 3.3$}  & $0.77 \pm .10$ &  { $0.72 \pm .11$} & $2.47 \pm 0.69$ & $2.35 \pm 0.51$     \\ 
  & SRDRM-GAN & { $26.77 \pm 4.1$}  & { $24.77 \pm 3.4$}  & { $0.82 \pm .07$} & $0.70 \pm .12$  & { $2.87 \pm 0.55$} & { $2.81 \pm 0.56$}     \\ 
  & \textbf{Deep SESR} & { $27.03 \pm 2.9$}  &  $24.59 \pm 3.8$ & { $0.88 \pm .05$} & { $0.71 \pm .08$}  & { $3.15 \pm 0.44$} & { $2.96 \pm 0.28$}  \\ \Xhline{2\arrayrulewidth}
\end{tabular}
\label{tab:quan_sr}
\end{table*}

\begin{figure}[t]
\vspace{1mm}
    \centering
    \includegraphics[width=0.98\linewidth]{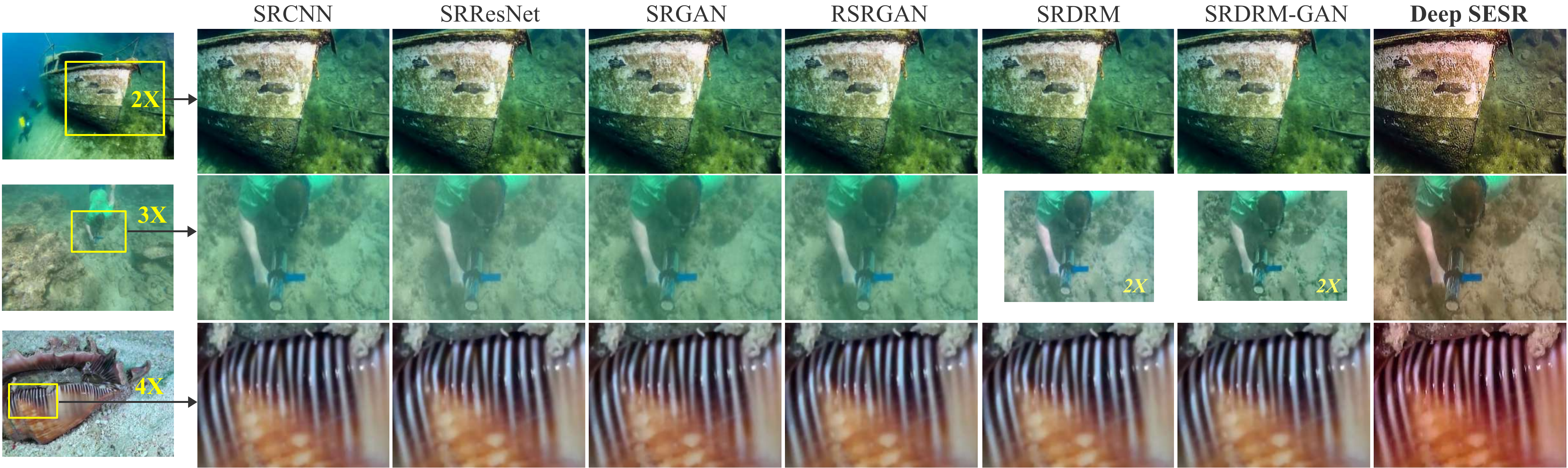}%
      \vspace{-3mm}
        \caption{Qualitative comparison for SISR performance of Deep SESR with existing solutions and SOTA models: SRCNN~\cite{dong2015image}, SRResNet~\cite{ledig2017photo}, SRGAN~\cite{ledig2017photo}, RSRGAN~\cite{chen2019recovering}, SRDRM~\cite{islam2019underwater}, and SRDRM-GAN~\cite{islam2019underwater}.}%
    \label{fig:comp_sesr}
\end{figure}

\subsection{Evaluation: Super-Resolution}
We follow similar experimental procedures for evaluating the super-resolution performance of Deep SESR. We consider the existing underwater SISR models named RSRGAN~\cite{chen2019recovering}, SRDRM~\cite{islam2019underwater}, and SRDRM-GAN~\cite{islam2019underwater} for performance comparison. 
We also include the standard (terrestrial) SISR models named SRCNN~\cite{dong2015image}, SRResNet~\cite{ledig2017photo}, and SRGAN~\cite{ledig2017photo} in the evaluation as benchmarks. We compare their $2\times$, $3\times$, and  $4\times$ SISR performance on two large-scale datasets: UFO-120, and USR-248. The quantitative results SISR are presented in Table~\ref{tab:quan_sr} and a few samples are shown in Figure~\ref{fig:comp_sesr}. Note that, the test images of USR-248 dataset are left undistorted for a fair comparison.

\begin{figure}[ht]
\vspace{2mm}
    \centering
    \includegraphics[width=0.98\linewidth]{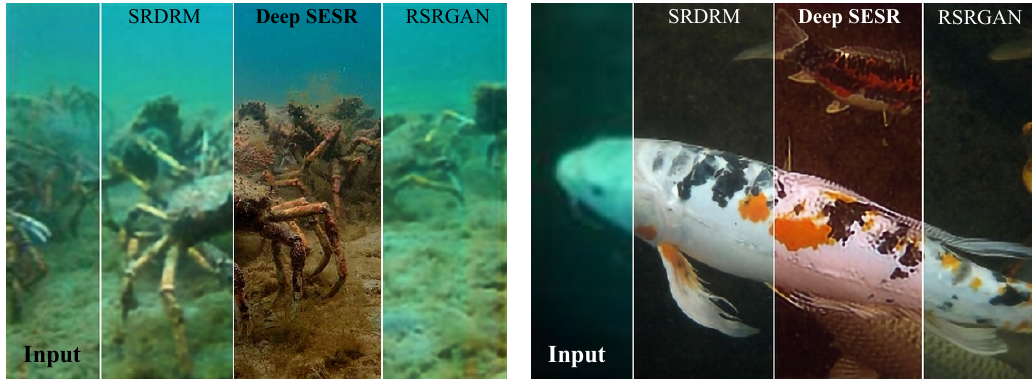}%
    \vspace{-3mm}
        \caption{Color and texture recovery of Deep SESR: comparison shown with two best-performing SISR models (as of Table~\ref{tab:quan_sr}).}%
    \label{fig:qual_sr}
\end{figure}

As Table~\ref{tab:quan_sr} demonstrates, Deep SESR outperforms other models in comparison by considerable margins on UIQM. This is due to the fact that it enhances perceptual image qualities in addition to spatial resolution. As shown in Figure~\ref{fig:comp_sesr}, Deep SESR generates much sharper and better quality HR images from both distorted and undistorted LR input patches, which contributes to its competitive PSNR and SSIM scores on the USR-248 dataset. 
Figure~\ref{fig:qual_sr} further demonstrates that it does not introduce noise by unnecessary over-correction, which is a prevalent limitation of existing solutions.
Lastly, we observe similar performance trends for all three types of spatial down-sampling, \ie, for Set-U, Set-F, and Set-O (see Section~\ref{sec:data_prep}); we present the relative quantitative scores in Table~\ref{tab:ufo_quant}.

\begin{table}[ht]
\footnotesize
%\scriptsize
    \centering
    \caption{Deep SESR performance on the UFO-120 test dataset; set-wise mean scores are shown for $2\times$ / $3\times$ / $4\times$ SESR.}
\vspace{2mm}

    \begin{tabular}{l||c|c|c}
    \Xhline{2\arrayrulewidth}
      & \textit{PSNR} & \textit{SSIM} & \textit{UIQM} \\ \Xhline{2\arrayrulewidth} 
        \textbf{Set-U} & $28.55$ / $26.77$ / $24.25$ & $0.86$ / $0.75$ / $0.66$ & $3.07$ / $2.89$ / $2.54$ \\
        \textbf{Set-F} & $27.93$ / $26.33$ / $24.87$ & $0.85$ / $0.73$ / $0.63$ & $3.10$ / $2.84$ / $2.52$ \\
        \textbf{Set-O} & $28.95$ / $27.15$ / $25.45$ & $0.84$ / $0.79$ / $0.68$ & $3.09$ / $2.86$ / $2.58$ \\ \Xhline{2\arrayrulewidth}
    \end{tabular}
    \label{tab:ufo_quant}
\end{table}

\begin{figure}[ht]
    \centering
    \includegraphics[width=0.98\linewidth]{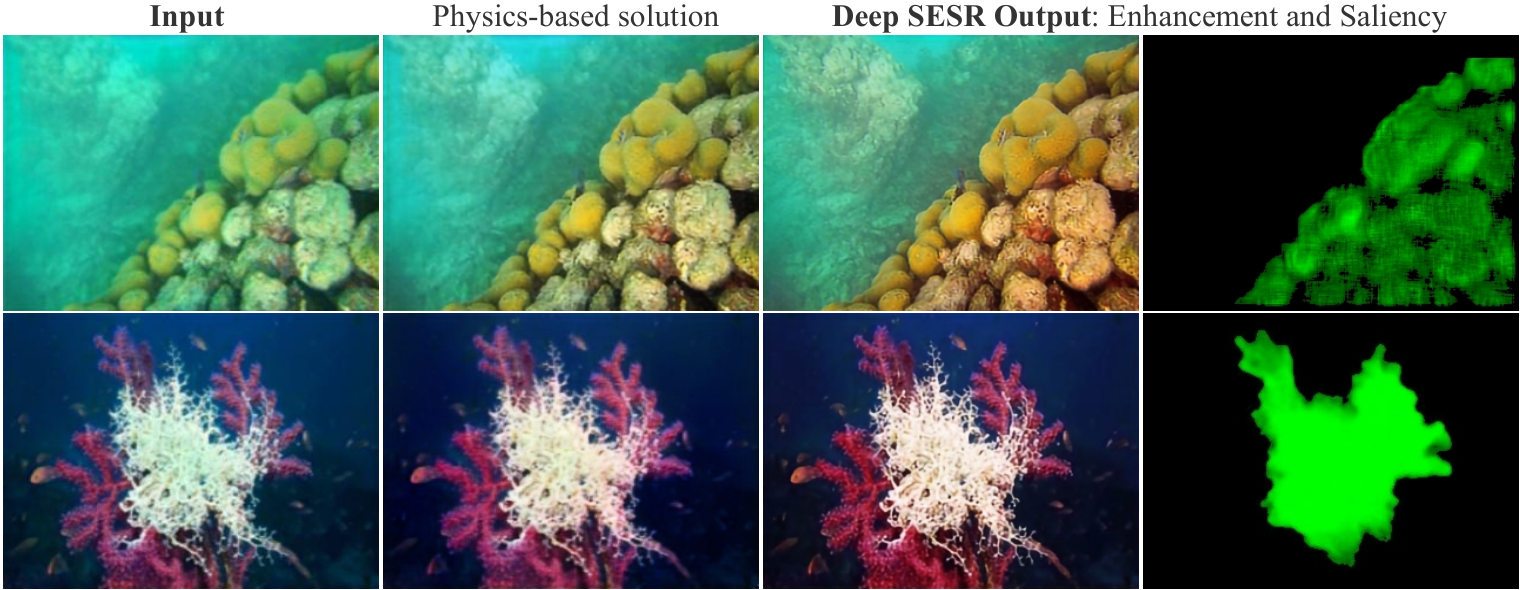}%
    \vspace{-3mm}
        \caption{Comparison with a physics-based color restoration method~\cite{berman2018underwater} that uses spectral waterbody measures and haze-lines prior.}%
    \label{fig:qual_ph}
\end{figure}

\begin{figure}[ht]
    \centering
    \includegraphics[width=0.95\linewidth]{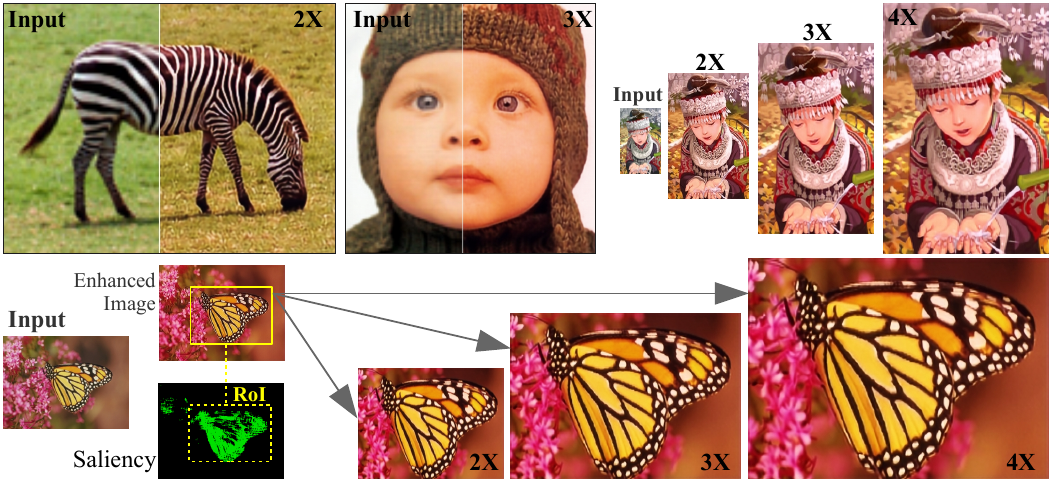}%
    \vspace{-2mm}
        \caption{Performance for $2\times$, $3\times$, and $4\times$ SESR on terrestrial images.}%
    \label{fig:qual_terr}
\end{figure}

\section{Generalization Performance}\label{sesr_general}
Due to the ill-posed nature of modeling underwater image distortions without scene-depth and optical waterbody measurements, learning-based solutions often fail to generalize beyond supervised data. In addition to the already-presented results, we demonstrate the color and texture recovery of Deep SESR on unseen natural images in Figure~\ref{fig:qual_ph}. As shown, Deep SESR-enhanced pixel intensities are perceptually similar to a comprehensive physics-based approximation~\cite{berman2018underwater}. Additionally, it generates the respective HR images and saliency maps, and still offers more than $10$ times faster run-time.

Deep SESR also provides reasonable performance on terrestrial images. As demonstrated in Figure~\ref{fig:qual_terr}, the color and texture enhancement of unseen objects (\eg, grass, face, clothing, etc.) are perceptually coherent. Moreover, as Table~\ref{tab:terr_quant} indicates, its performance in terms of sharpness and contrast recovery for $2\times$, $3\times$, and $4\times$ SISR are competitive with SOTA benchmark results~\cite{li2019feedback,zhang2018residual}. Note that, much-improved performance can be achieved by further tuning and training on terrestrial datasets. Nevertheless, these results validate that the proposed architecture has the capacity to learn a generalizable solution of the underlying SESR problem. 
%The PSNR scores are slightly lower, which is likely due to the additional color enhancement performed by Deep SESR (as PSNR is calculated based on mean-squared error in pixel intensities).             

\begin{table}[ht]
\footnotesize
%\scriptsize
    \centering
    \caption{Deep SESR performance on terrestrial test data; boldfaced scores represent $3\%$ margins with SOTA benchmark results for $2\times$ / $3\times$ / $4\times$ SISR~\cite{li2019feedback,zhang2018residual}.}%
    \vspace{2mm}
    \begin{tabular}{l||c|c}
    \Xhline{2\arrayrulewidth}
      & \textit{PSNR} & \textit{SSIM} \\ \Xhline{2\arrayrulewidth} %\hline  
        \textbf{Set5}~\cite{bevilacqua2012low} & $29.87$ / {$\mathbf{28.77}$} / $\mathbf{26.14}$ & $\mathbf{0.925}$ / ${ \mathbf{0.908}}$ / $\mathbf{0.855}$  \\
        \textbf{Set14}~\cite{zeyde2010single} & {$\mathbf{28.78}$} / $27.34$ / $\mathbf{26.89}$ & ${\mathbf{0.914}}$ / $\mathbf{0.801}$ / $\mathbf{0.756}$  \\
        \textbf{Sun80}~\cite{sun2012super} & $25.73$ / $23.18$ / $21.05$ & $\mathbf{\mathbf{0.802}}$ / $\mathbf{0.755}$ / $0.704$  \\ \Xhline{2\arrayrulewidth}
    \end{tabular}
    \label{tab:terr_quant}
\end{table}

\section{Operational Feasibility \& Design Choices}\label{sesr_feasibility}
\begin{figure}[ht]
\vspace{1mm}
    \centering
        \includegraphics[width=0.98\linewidth]{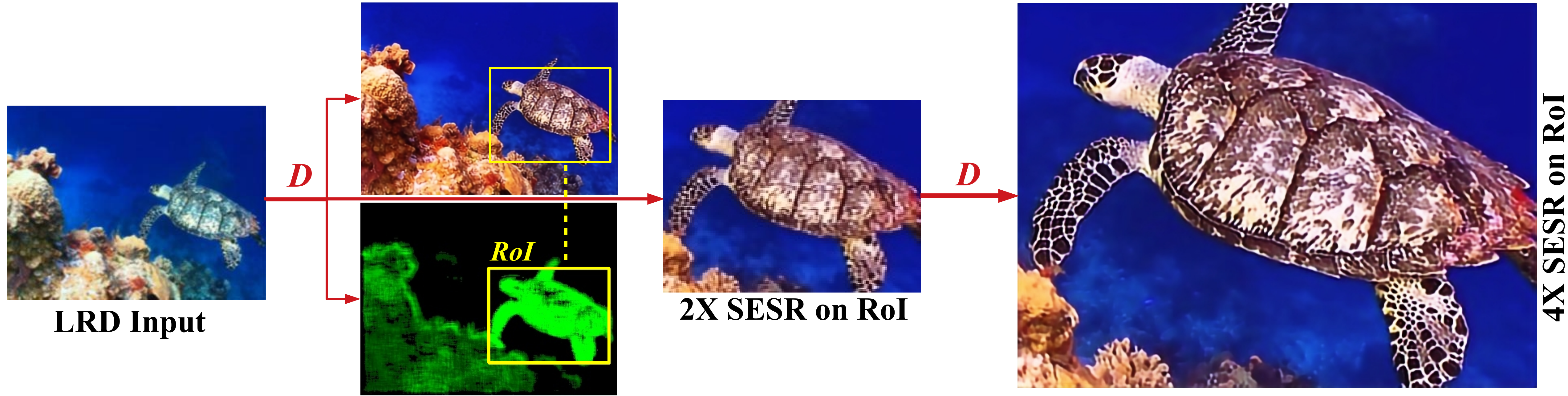} 
        \vspace{-3mm}
        \caption{Automatic RoI selection based on local intensity values in the saliency map; Deep SESR (\textit{\textbf{D}}) can be applied again on the enhanced RoI for a detailed perception.}
        \vspace{1mm}
    \label{fig:qual_sal}
\end{figure}

Deep SESR's on-board memory requirement is only $10$ MB, and it offers a run-time of $129$ milliseconds (ms) per-frame, \ie, $7.75$ frames-per-second (FPS) on a single-board computer named Nvidia\texttrademark AGX Xavier. 
As shown in Table~\ref{tab:time_sesr}, it provides much faster speeds for the following design choices: 

\vspace{1mm}
\textit{i)} \textbf{Learning $\hat{E}$ and $\hat{S}$ on separate branches} facilitates a faster run-time when HR perception is not required. Specifically, we can decouple the $X$ $\rightarrow$ $S$, $E$ branches from the frozen model, which operates at $10.02$ FPS ($22\%$ faster) to perform enhancement and saliency prediction. As shown in Figure~\ref{fig:qual_sal}, the predicted saliency map can be exploited for automatic RoI selection by using density gradient estimation techniques such as mean-shift~\cite{comaniciu1999mean}. The SESR output corresponding to the RoI can be generated with an additional $25$ ms of processing time.

\begin{table}[t]
\centering
\caption{Deep SESR run-time on Nvidia\texttrademark AGX Xavier.}
\vspace{2mm}
\footnotesize
\begin{tabular}{l|c|c}
  \Xhline{2\arrayrulewidth}
   & $X$ $\rightarrow$ $S$, $E$  & $X$ $\rightarrow$ $S$, $E$ , $Y$   \\ \Xhline{2\arrayrulewidth}
  \textbf{With FENet-1d} &  $87.3$ ms ($11.45$ FPS)  &  $113$ ms ($8.85$ FPS)   \\
    \textbf{With FENet-2d} &  $99.8$ ms ($10.02$ FPS)  &  $129$ ms ($7.75$ FPS)   \\ \Xhline{2\arrayrulewidth}
\end{tabular}
\label{tab:time_sesr}
\vspace{2mm}
\end{table}

\vspace{1mm}
\textit{ii)} \textbf{FENet-1d and FENet-2d} are two design choices for the FENet (see Figure~\ref{fig:model_fnet}); FENet-2d is the default architecture that learns $3\times3$ and $5\times5$ filters in two parallel branches, whereas, FENet-1d refers to using a single branch of $3\times3$ filters. As shown in Table~\ref{tab:time_sesr}, faster feature extraction by FENet-1d facilitates a $12.5\%$ speed-up for Deep SESR. 
However, we observe a slight drop in performance, \eg, $1.8\%$/$1.5\%$/$1.8\%$ lower scores for PSNR/SSIM/UIQM on UFO-120 dataset. Nevertheless, the generated images are qualitatively indistinguishable and the trade-off is admissible in practical applications. Overall, Deep SESR offers use-case-specific design choices and ensures computational efficiency with robust SESR performance. These features make it suitable for near real-time robotic deployments; more detailed demonstrations can be found at: \url{https://youtu.be/wEkTu2CPW-g}.

\section{Summary}
In this chapter, we introduced the SESR problem and presented an efficient learning-based solution for underwater imagery. The proposed generative model, named Deep SESR, can learn $2\times$$-$$4\times$ SESR and saliency prediction on a shared feature space. We also presented its detailed network architecture, associated loss functions, and end-to-end training pipeline. Additionally, we contributed over $1500$ annotated samples to facilitate large-scale SESR training on the UFO-120 dataset. We performed a series of qualitative and quantitative experiments, which suggest that Deep SESR: \textit{i)} provides SOTA performance on underwater image enhancement and super-resolution, \textit{ii)} exhibits significantly better generalization performance on natural images than existing solutions, \textit{iii)} provides competitive results on terrestrial images, and \textit{iv)} achieves fast inference on single-board platforms. 
The inspiring performance, computational efficiency, and availability of application-specific design choices make Deep SESR suitable for near real-time use by visually-guided underwater robots. 
\chapter{Saliency-guided Visual Attention Modeling: SVAM}\label{svam}
Salient object detection (SOD) aims at modeling human visual attention behavior to highlight the most important and distinct objects in a scene. It is a well-studied problem in the domains of robotics and computer vision~\cite{borji2019salient,kim2013real,liu2018picanet,liu2019simple} for its usefulness in identifying regions of interest (RoI) in an image. The SOD capability is essential for visually-guided robots because they need to make critical navigational and operational decisions based on the relative \emph{importance} of various objects in their field-of-view. The Autonomous Underwater Vehicles (AUVs), in particular, rely heavily on saliency-guided visual attentioin modeling for tasks such as exploration and surveying~\cite{girdhar2014autonomous,koreitem2020one,kaeli2014visual,shkurti2012multi,johnson2010saliency}, ship-hull inspection~\cite{kim2013real}, event detection~\cite{edgington2003automated}, place recognition~\cite{maldonado2019learning}, target localization/tracking~\cite{zhu2020saliency,islam2020suim}, and more.

In the pioneering work on SOD, Itti~\etal~\cite{itti1998model} used local feature contrast in image regions to infer visual saliency. Numerous methods have been subsequently proposed~\cite{edgington2003automated,klein2011center,cheng2014global} that utilize local point-based features and also global contextual information as reference for saliency estimation. In recent years, the state-of-the-art (SOTA) approaches have used powerful deep visual models~\cite{wang2019salient,wang2019iterative} to imitate human visual information processing through top-down or bottom-up computational pipelines. 
The bottom-up models learn to gradually infer high-level semantically rich features~\cite{wang2019iterative}; hence the shallow layers' structural knowledge drives their multi-scale saliency learning. 
Conversely, the the top-down approaches~\cite{liu2018picanet,feng2019attentive} progressively integrate high-level semantic knowledge with low-level features for learning \emph{coarse-to-fine} saliency estimation. 
Moreover, the contemporary models have introduced various techniques to learn boundary refinement~\cite{qin2019basnet,feng2019attentive,wang2018detect}, pyramid feature attention~\cite{wang2019salient}, and contextual awareness~\cite{liu2018picanet}, which significantly boost the SOD performance on benchmark datasets.

\begin{figure}[t]
    \centering
    \includegraphics[width=0.8\linewidth]{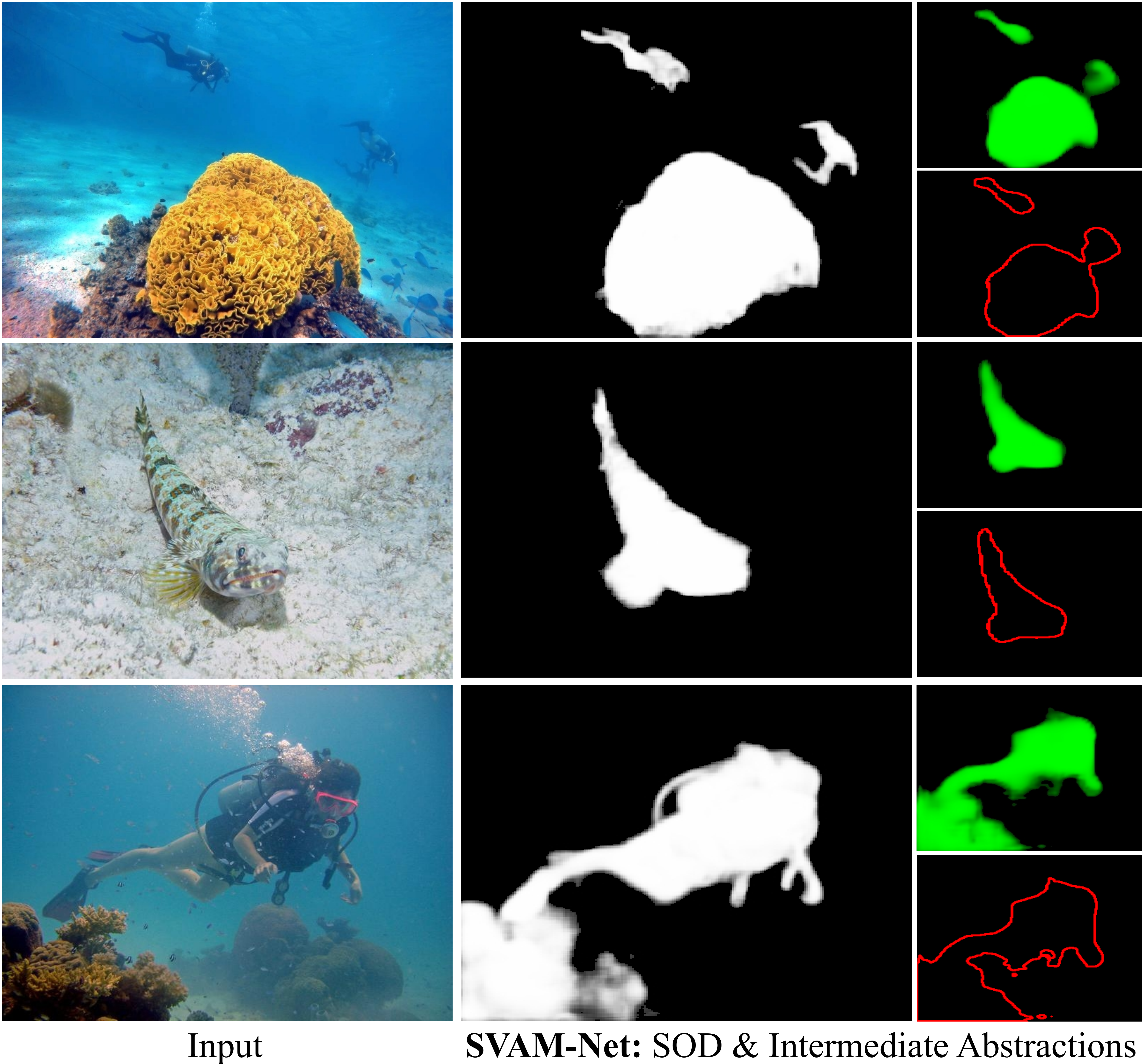}%
    \vspace{-2mm}
    \caption{The proposed SVAM-Net model identifies salient objects and interesting image regions to facilitate effective visual attention modeling by autonomous underwater robots. 
    It also generates abstract saliency maps (shown in green intensity channel and red object contours) from an early bottom-up SAM which can be used for fast processing on single-board devices.
    }
    \label{svam_fig_intro}
\end{figure}

However, the applicability of such powerful learning-based SOD models in real-time underwater robotic vision is rather limited. The underlying challenges and practicalities are twofold. First, the visual content of underwater imagery is uniquely diverse due to domain-specific object categories, background waterbody patterns, and a host of optical distortion artifacts~\cite{akkaynak2018revised,islam2019fast}; hence, the SOTA models trained on terrestrial data are not transferable off-the-shelf. A lack of large-scale annotated underwater datasets aggravates the problem; the existing datasets and relevant methodologies are tied to specific applications such as coral reef classification and coverage estimation~\cite{beijbom2012automated,alonso2019coralseg,VAIME}, object detection~\cite{ravanbakhsh2015automated,islam2018towards,chuang2011automatic}, and foreground segmentation~\cite{li2016mapreduce,padmavathi2010non,zhu2017underwater}. Consequently, these do not provide a comprehensive data representation for effective learning of underwater SOD. Secondly, learning a generalizable SOD function demands the extrapolation of multi-scale hierarchical features by high-capacity deep network models. This results in a heavy computational load and makes real-time inference impossible, particularly on single-board robotic platforms.        

%The existing literature...
To this end, traditional approaches based on various feature contrast evaluation techniques~\cite{girdhar2014autonomous,zhu2020saliency,maldonado2016robotic} are often practical choices for saliency estimation by visually-guided underwater robots. These techniques encode low-level image-based features (\eg, color, texture, object shapes or contours) into super-pixel descriptors~\cite{kumar2019saliency,jian2018integrating,maldonado2016robotic,wang2013saliency,johnson2010saliency} to subsequently infer saliency by quantifying their relative \textit{distinctness} on a global scale. 
Such bottom-up approaches are computationally light and are useful as pre-processing steps for faster visual search~\cite{kumar2019saliency,kim2013real} and exploration tasks~\cite{girdhar2014autonomous,maldonado2019learning}. 
However, they do not provide a standalone generalizable solution for SOD in underwater imagery. 
A few recently proposed approaches attempt to address this issue by learning more generalizable SOD solutions from large collection of annotated underwater data~\cite{islam2020suim,jian2019extended,islam2020sesr,jian2017ouc,rizzini2015investigation}. These approaches and other SOTA deep models have reported inspiring results for underwater SOD and relevant problems~\cite{jian2018integrating,islam2020sesr,islam2020suim}. 
Nevertheless, their utility and performance margins for underwater robotic applications have not been explored in the literature; more in-depth analyses are presented in Section~\ref{related_work}. 

In this chapter, we formulate a robust and efficient solution for saliency-guided visual attention modeling (SVAM) by harnessing the power of both bottom-up and top-down learning in a novel encoder-decoder model named \textbf{SVAM-Net}. 
We design two spatial attention modules (SAMs) named \textbf{SAM\textsuperscript{bu}} and \textbf{SAM\textsuperscript{td}} to effectively exploit the coarse-level and fine-level semantic features along the bottom-up and top-down learning pathways, respectively. 
SAM\textsuperscript{bu} utilizes the semantically rich low-dimensional features extracted by the encoder to perform an abstract yet reasonably accurate saliency estimation. 
Concurrently, SAM\textsuperscript{td} combines the multi-scale hierarchical features of the encoder to progressively decode the information for robust SOD. A residual refinement module (\textbf{RRM}) further sharpens the initial SAM\textsuperscript{td} predictions to provide fine-grained localization of the salient objects. To balance the high degree of \textit{refined} gradient flows from the later SVAM-Net layers, we deploy an auxiliary SAM named  \textbf{SAM\textsuperscript{aux}} that guides the spatial activations of early encoding layers and ensures smooth end-to-end learning. The model and detailed training pipeline is presented in Section~\ref{svam_model}.

In addition to sketching the conceptual design, we present a holistic training pipeline of SVAM-Net and its variants. The end-to-end learning is supervised by six loss functions which are selectively applied at the final stages of SAM\textsuperscript{aux}, SAM\textsuperscript{bu}, SAM\textsuperscript{td}, and RRM. 
These functions evaluate information loss and boundary localization errors in the respective SVAM-Net predictions and collectively ensure effective SOD learning. 
In our evaluation, we analyze SVAM-Net's performance in standard quantitative and qualitative terms on three benchmark datasets named UFO-120~\cite{islam2020sesr}, MUED~\cite{jian2019extended}, and SUIM~\cite{islam2020suim}. 
We also conduct performance evaluation on \textbf{USOD}, which we prepare as a new challenging test set for underwater SOD. Without data-specific tuning or task-specific model adaptation, SVAM-Net outperforms other existing solutions on these benchmark datasets (see Section~\ref{svam_beval}); more importantly, it exhibits considerably better generalization performance on random unseen test cases of natural underwater scenes.

\section{Background and Related Work}\label{related_work}
\subsection{Salient Object Detection (SOD)}
SOD is a successor to the human fixation prediction (FP) problem~\cite{itti1998model} that aims to identify \textit{fixation points} that human viewers would focus on at first glance. While FP originates from research in cognition and psychology~\cite{kruthiventi2017deepfix,le2006coherent,wang2018salient}, SOD is more of a visual perception problem explored by the computer vision and robotics community~\cite{borji2019salient,kim2013real,liu2018picanet,liu2019simple}. The history of SOD dates back to the work of Liu~\etal~\cite{liu2010learning} and Achanta~\etal~\cite{achanta2009frequency}, which make use of multi-scale contrast, center-surround histogram, and frequency-domain cues to (learn to) infer saliency in image space. 
Other traditional SOD models rely on various low-level saliency cues such as point-based features~\cite{edgington2003automated}, local and global contrast~\cite{cheng2014global,klein2011center}, background prior~\cite{yang2013saliency}, etc. Please refer to~\cite{borji2015salient} for a more comprehensive overview of non-deep learning-based SOD models.

Recently, deep convolutional neural network (CNN)-based models have set
new SOTA for SOD~\cite{borji2019salient,wang2019salientsurvey}. Li~\etal~\cite{li2015visual,li2016deep} and Zhao~\etal~\cite{zhao2015saliency} use sequential CNNs to extract multi-scale hierarchical features to infer saliency on global
and local contexts. 
Recurrent fully convolutional networks (FCNs)~\cite{wang2016saliency,bazzani2016recurrent} are also used to progressively refine saliency estimates. In particular, Wang~\etal~\cite{wang2018salient} use multi-stage convolutional LSTMs for saliency estimation guided by fixation maps. Later in~\cite{wang2019iterative}, they explore the benefits of integrating bottom-up and top-down recurrent modules for co-operative SOD learning. Since the feed-forward computational pipelines lack a feedback strategy~\cite{li2018contour,le2006coherent}, recurrent modules offer more learning capacity via self-correction. However, they are prone to the vanishing gradient problem and also require meticulous design choices in their feedback loops~\cite{zhang2018progressive,wang2019iterative}.    
To this end, top-down models with UNet-like architectures~\cite{liu2018picanet,feng2019attentive,luo2017non,zhang2017amulet,liu2016dhsnet} provide more consistent learning behavior. These models typically use a powerful backbone network (\eg, VGG~\cite{simonyan2014very}, ResNet~\cite{he2016deep}) to extract a hierarchical pyramid of features, then perform a coarse-to-fine feature distillation via mirrored skip-connections. Subsequent research introduces the notions of short connections~\cite{hou2017deeply} and guided super-pixel filtering~\cite{hu2017deep} 
to learn to infer compact and uniform saliency maps. 

Moreover, various \textit{attention mechanisms} are incorporated by contemporary models to intelligently guide the SOD learning, particularly for tasks such as image captioning and visual question answering~\cite{xu2015show,yu2017multi,lu2016hierarchical,li2016attentive}. 
Additionally, techniques like pyramid feature attention learning~\cite{wang2019salient,zhao2019pyramid}, boundary refinement modules~\cite{qin2019basnet,feng2019attentive,wang2018detect}, contextual awareness~\cite{liu2018picanet}, and cascaded partial decoding~\cite{wu2019cascaded} have significantly boosted the SOTA SOD performance margins. However, this domain knowledge has not been applied or explored in-depth for saliency-guided visual attention modeling (SVAM) by autonomous underwater robots, which we attempt to address in the design and implementation of SVAM-Net.

\begin{figure}[t]
    \centering
\includegraphics[width=0.75\linewidth]{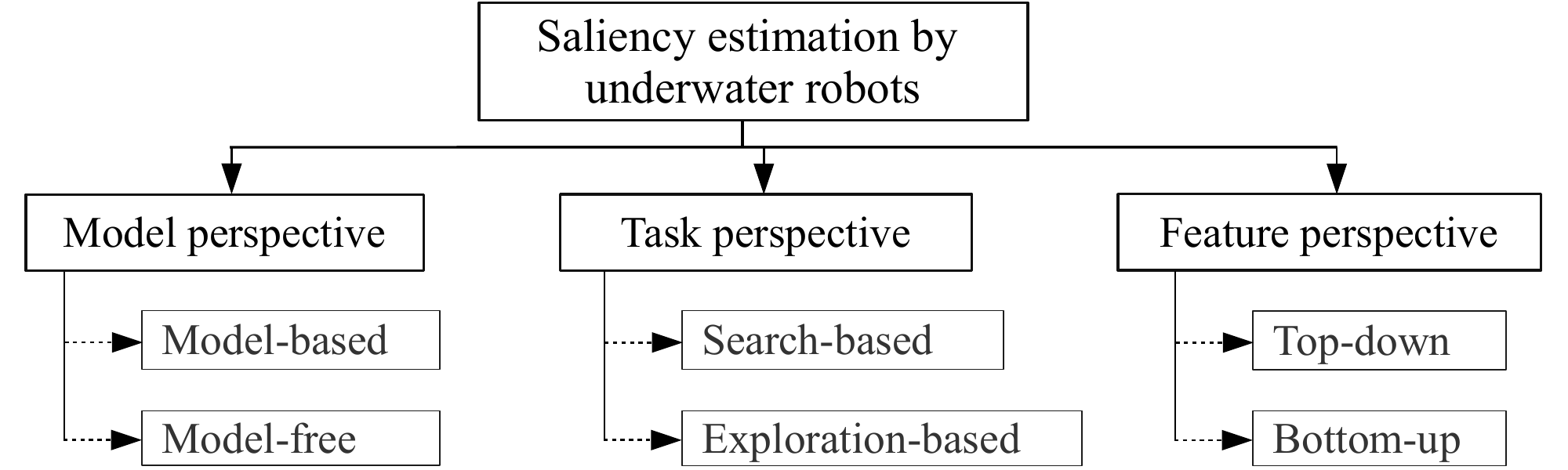}%
\vspace{-2mm}
 \caption{A categorization of underwater saliency estimation techniques based on model adaptation, high-level tasks, and feature evaluation.}%
 \label{fig:svam_fig_category}
 \vspace{2mm}
\end{figure}

\subsection{SOD and SVAM by Underwater Robots}\label{uw_sod_svam}
The most essential capability of visually-guided AUVs is to
identify interesting and relevant image regions to make effective operational decisions. As shown in Figure~\ref{fig:svam_fig_category}, the existing systems and solutions for visual saliency estimation can be categorically discussed from the perspectives of model adaptation~\cite{koreitem2020one,islam2020suim}, high-level robot tasks~\cite{girdhar2014autonomous,rehman2019salient}, and feature evaluation pipeline~\cite{jian2018integrating,maldonado2019learning}. Since we already discussed the particulars of bottom-up and top-down computational pipelines, our following discussion is schematized based on the \textit{model} and \textit{task} perspectives.

Visual saliency estimation approaches can be termed as either \textit{model-based} or \textit{model-free}, depending on whether the robot models any prior knowledge of the target salient objects and features. 
The model-based techniques are particularly beneficial for fast visual search~\cite{koreitem2020one,johnson2010saliency}, enhanced object detection~\cite{zhu2020saliency,rizzini2015investigation}, and monitoring applications~\cite{modasshir2020enhancing,manderson2018vision}. 
For instance,
Maldonado-Ram{\'\i}rez~\etal~\cite{maldonado2019learning} use ad hoc visual descriptors learned by a convolutional autoencoder to identify salient landmarks for fast place recognition. Moreover, Koreitem~\etal~\cite{koreitem2020one} use a bank of pre-specified image patches (containing interesting objects or relevant scenes) to learn a similarity operator that guides the robot's visual search in an unconstrained setting. Such similarity operators are essentially spatial saliency predictors which assign a degree of \textit{relevance} to the visual scene based on the prior model-driven knowledge of what may constitute as salient, \eg, coral reefs~\cite{alonso2019coralseg,ModasshirFSR2019}, companion divers~\cite{islam2018towards,zhu2020saliency}, wrecks or ruins~\cite{islam2020suim}, specific
species of fish~\cite{ravanbakhsh2015automated}, etc.

On the other hand, model-free approaches are more feasible for autonomous exploratory applications~\cite{girdhar2016modeling,rekleitis2001multi}. 
The early approaches date back to the work of Edgington~\etal~\cite{edgington2003automated} that uses binary morphology filters to extract salient features for automated event detection. Subsequent approaches adopt various feature contrast evaluation techniques that encode low-level image-based features (\eg, color, luminance, texture, object shapes) into super-pixel descriptors~\cite{maldonado2016robotic,kumar2019saliency,wang2013saliency}. These low-dimensional representations are then exploited by heuristics or learning-based models to infer global saliency. For instance, Girdhar~\etal~\cite{girdhar2014autonomous} formulate an online topic-modeling scheme that encodes visible features into a low-dimensional semantic descriptor, then adopt a probabilistic approach to compute a \textit{surprise score} for the current observation based on the presence of high-level patterns in the scene. Moreover, Kim~\etal~\cite{kim2013real} introduce an online bag-of-words scheme to measure intra- and inter-image saliency estimation for robust key-frame selection in SLAM-based navigation. Wang~\etal~\cite{wang2013saliency} encode multi-scale image features into a topographical descriptor, then apply Bhattacharyya~\cite{Bhattacharyya43} measure to extract salient RoIs by segmenting out the background. These bottom-up approaches are effective in pre-processing raw visual data to identify point-based or region-based salient features; however, they do not provide a generalizable object-level solution for underwater SOD.

Nevertheless, several contemporary research~\cite{chen2019underwater,jian2018integrating,zhu2017underwater,li2016saliency} report inspiring results for object-level saliency estimation and foreground segmentation in underwater imagery. Chen~\etal~\cite{chen2019underwater} use a level set-based formulation that exploits various low-level features for underwater SOD. Moreover, Jian~\etal~\cite{jian2018integrating} perform principal components analysis (PCA) in quaternionic space to compute pattern distinctness and local contrast to infer directional saliency. 
These methods are also model-free and adopt a bottom-up feature evaluation pipeline. 
In contrast, our earlier work~\cite{islam2020sesr} incorporates multi-scale hierarchical features extracted by a top-down deep residual model to identify salient foreground pixels for global contrast enhancement. In this chapter, we formulate a generalized solution for underwater SOD and demonstrate its utility for SVAM by visually-guided underwater robots. 
It combines the benefits of bottom-up and top-town feature evaluation in a compact end-to-end pipeline, provides SOTA performance, and ensures computational efficiency for robotic deployments in both search-based and exploration-based applications.

\begin{figure*}[t]
    \centering
    \includegraphics[width=0.98\linewidth]{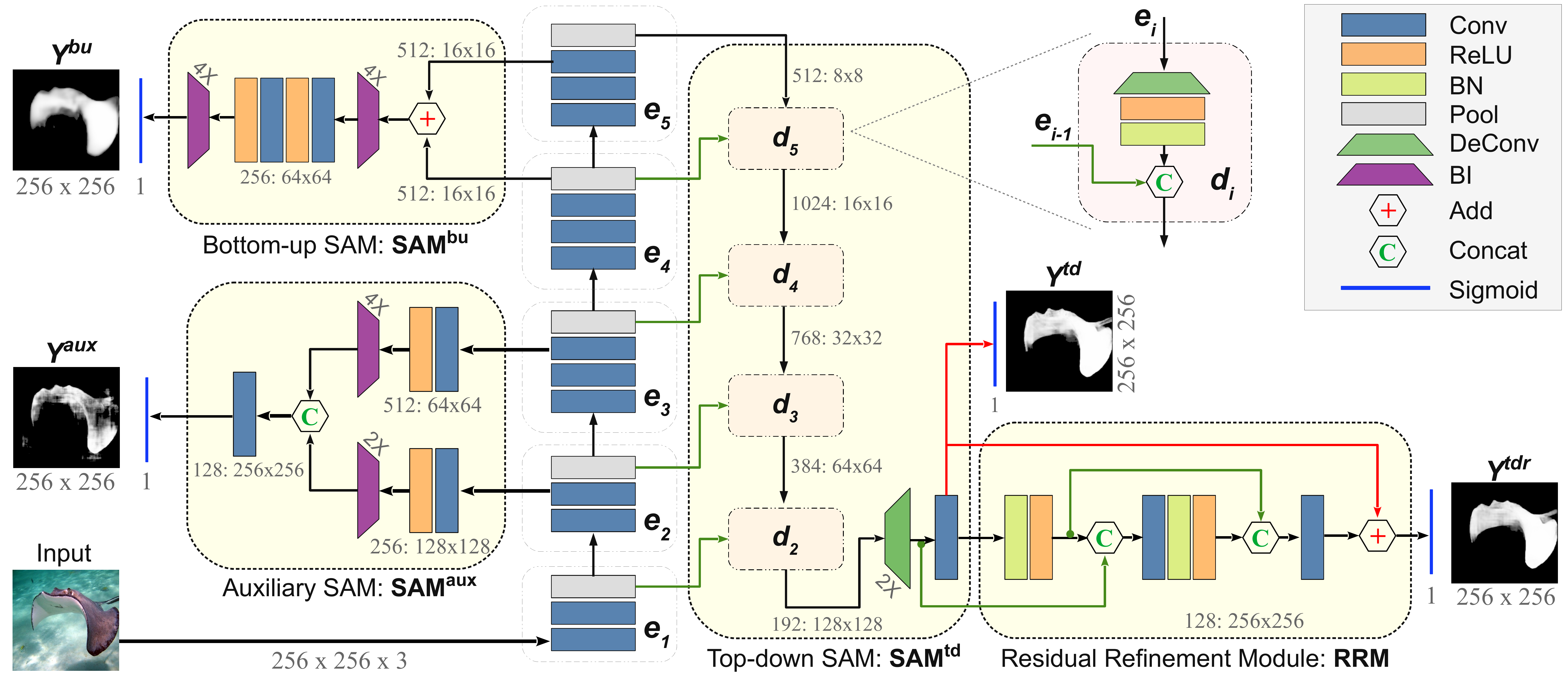}%
    \vspace{-2mm}
    \caption{The detailed architecture of SVAM-Net is shown. The input image is passed over to the sequential encoding blocks $\{\mathbf{e_1} \rightarrow \mathbf{e_5}\}$ for multi-scale convolutional feature extraction. Then, SAM\textsuperscript{td} gradually up-samples these hierarchical features and fuses them with mirrored skip-connections along the top-down pathway $\{\mathbf{d_5} \rightarrow \mathbf{d_2}\}$ to subsequently generate an intermediate output $Y^{td}$; the RRM refines this intermediate representation and produces the final SOD output $Y^{tdr}$. Moreover, SAM\textsuperscript{bu} exploits the features of $\mathbf{e_4}$ and $ \mathbf{e_5}$ to generate an abstract SOD prediction $Y^{bu}$ along the bottom-up pathway; additionally, SAM\textsuperscript{aux} performs an auxiliary refinement on the $\mathbf{e_2}$ and $ \mathbf{e_3}$ features that facilitates a smooth end-to-end SOD learning. 
    }
    \label{svam_fig_model}
    \vspace{2mm}
\end{figure*}

\section{Model and Training Pipeline}\label{svam_model}
\subsection{SVAM-Net Architecture}
As illustrated in Figure~\ref{svam_fig_model}, the major components of our SVAM-Net model are: the backbone encoder network, the top-down SAM (SAM\textsuperscript{td}), the residual refinement module (RRM), the bottom-up SAM (SAM\textsuperscript{bu}), and the auxiliary SAM (SAM\textsuperscript{aux}). These components are tied to an end-to-end architecture for a supervised SOD learning.

\vspace{1mm}
\subsubsection{Backbone Encoder Network}
We use the first five sequential blocks of a standard VGG-16 network~\cite{simonyan2014very} as the backbone encoder in our model. Each of these blocks consist of two or three convolutional ({\tt Conv}) layers for feature extraction, which are then followed by a pooling ({\tt Pool}) layer for spatial down-sampling. For an input dimension of $256\times256\times3$, the composite encoder blocks $\mathbf{e_1} \rightarrow \mathbf{e_5}$ learn $128\times128\times64$, $64\times64\times128$, $32\times32\times256$, $16\times16\times512$, and $8\times8\times512$ feature-maps, respectively. These multi-scale deep visual features are jointly exploited by the attention modules of SVAM-Net for effective learning.          

\vspace{1mm}
\subsubsection{Top-Down SAM (SAM\textsuperscript{td})}
Unlike the existing U-Net-based architectures~\cite{liu2018picanet,feng2019attentive,luo2017non}, we adopt a partial top-down decoder $\mathbf{d_5} \rightarrow \mathbf{d_2}$ that allows skip-connections from mirrored encoding layers. We consider the mirrored conjugate pairs as $\mathbf{e_4}  \sim \mathbf{d_5}$, $\mathbf{e_3}  \sim \mathbf{d_4}$, $\mathbf{e_2}  \sim \mathbf{d_3}$, and $\mathbf{e_1}  \sim \mathbf{d_2}$. Such asymmetric pairing facilitates the use of a standalone de-convolutional ({\tt DeConv}) layer~\cite{zeiler2010deconvolutional} following $\mathbf{d_2}$ rather than using another composite decoder block, which we have found to be redundant (during ablation experiments). The composite blocks $\mathbf{d_5} \rightarrow \mathbf{d_2}$ decode $16\times16\times1024$, $32\times32\times768$, $64\times64\times384$, and $128\times128\times192$ feature-maps, respectively. Following $\mathbf{d_2}$ and the standalone {\tt DeConv} layer, an additional {\tt Conv} layer learns $256\times256\times128$ feature-maps to be the final output of SAM\textsuperscript{td} as 
\begin{equation} \small
\mathbf{S}^{td}_{coarse} = \mathbf{SAM}^{\text{td}}(\mathbf{e_1}:\mathbf{e_5}).
\end{equation}
These feature-maps are passed along two branches (see Figure~\ref{svam_fig_model}); on the shallow branch, a {\tt Sigmoid} layer is applied to generate an intermediate SOD prediction $Y^{td}$, while the other deeper branch incorporates residual layers for subsequent refinement.

\vspace{1mm}
\subsubsection{Residual Refinement Module (RRM)}
We further design a residual module to effectively refine the top-down coarse saliency predictions by learning the desired residuals as
\begin{equation} \small
\mathbf{S}^{tdr}_{refined} = \mathbf{S}^{td}_{coarse} + \mathbf{S}^{rrm}_{residual}.
\end{equation} 
Such refinement modules~\cite{deng2018r3net,qin2019basnet,wang2018detect} are designed to address the loss of regional probabilities and boundary localization in intermediate SOD predictions. While the existing methodologies use iterative recurrent modules~\cite{deng2018r3net} or additional residual encoder-decoder networks~\cite{qin2019basnet}, we deploy only two sequential residual blocks and a {\tt Conv} layer for the refinement. Each residual block consists of a {\tt Conv} layer followed by batch normalization ({\tt BN})~\cite{ioffe2015batch} and a rectified linear unit ({\tt ReLU}) activation~\cite{nair2010rectified}. The entire RRM operates on a feature dimension of $256\times256\times128$; following refinement, a {\tt Sigmoid} layer squashes the feature-maps to generate a single-channel output $Y^{tdr}$, which is the final SOD prediction of SVAM-Net.

\vspace{1mm}
\subsubsection{Bottom-Up SAM (SAM\textsuperscript{bu})}
A high degree of supervision at the final layers of RRM and SAM\textsuperscript{td} forces the backbone encoding layers to learn effective multi-scale features. In SAM\textsuperscript{bu}, we exploit these low-resolution yet semantically rich features for efficient bottom-up SOD learning. Specifically, we combine the feature-maps of dimension $16\times16\times512$ from $\mathbf{e_4}$ ({\tt Pool4}) and $\mathbf{e_5}$ ({\tt Conv53}), and subsequently learn the bottom-up spatial attention as 
\begin{equation} \small
\mathbf{S}^{bu} = \mathbf{SAM}^{\text{bu}}(\mathbf{e_4}.{\tt Pool4}, \text{ } \mathbf{e_5}.{\tt Conv53}).
\end{equation}
On the combined input feature-maps, SAM\textsuperscript{bu} incorporates $4\times$ bilinear interpolation ({\tt BI}) followed by two {\tt Conv} layers with {\tt ReLU} activation to learn $64\times64\times256$ feature-maps. Subsequently, another {\tt BI} layer performs $4\times$ spatial up-sampling to generate $\mathbf{S}^{bu}$; lastly, a {\tt Sigmoid} layer is applied to generate the single-channel output $Y^{bu}$.         

\vspace{1mm}
\subsubsection{Auxiliary SAM (SAM\textsuperscript{aux})}
We excluded the features of early encoding layers for bottom-up SOD learning in SAM\textsuperscript{bu} for two reasons: $i$) they lack important semantic details despite their higher resolutions~\cite{zhao2019pyramid,wu2019cascaded}, and $ii$) it is counter-intuitive to our goal of achieving fast bottom-up inference. Nevertheless, we adopt a separate attention module that refines the features of $\mathbf{e_2}$ ({\tt Conv22}) and $\mathbf{e_3}$ ({\tt Conv33}) as  
\begin{equation}\small
\mathbf{S}^{aux} = \mathbf{SAM}^{\text{aux}}(\mathbf{e_2}.{\tt Conv22}, \text{ } \mathbf{e_3}.{\tt Conv33}).
\end{equation}
Here, a {\tt Conv} layer with {\tt ReLU} activation is applied separately on these inputs, followed by a $2\times$ or $4\times$ {\tt BI} layer (see Figure~\ref{svam_fig_model}). Their combined output features are passed to a {\tt Conv} layer to subsequently generate $\mathbf{S}^{aux}$ of dimension $256\times256\times128$. The sole purpose of this module is to backpropagate additional \textit{refined} gradients via supervised loss applied to the {\tt Sigmoid} output $Y^{aux}$. This auxiliary refinement facilitates smooth feature learning while adding no computational overhead in the bottom-up inference through SAM\textsuperscript{bu} (as we discard SAM\textsuperscript{aux} after training).

\subsection{Learning Objectives and Training}\label{svam_training}
SOD is a pixel-wise binary classification problem that refers to the task of identifying all \textit{salient} pixels in a given image. We formulate the problem as learning a function $f: X \rightarrow Y$, where $X$ is the input image domain and $Y$ is the target saliency map, \ie, saliency probability for each pixel. As illustrated in Figure~\ref{svam_fig_model}, SVAM-Net generates saliency maps from four output layers, namely $Y^{aux} = \sigma(\mathbf S^{aux})$, $Y^{bu} = \sigma(\mathbf S^{bu})$, $Y^{td} = \sigma(\mathbf S^{td}_{coarse})$, and $Y^{tdr} = \sigma(\mathbf S^{tdr}_{refined})$ where $\sigma$ is the {\tt Sigmoid} function. Hence, the learning pipeline of SVAM-Net is expressed as $f : X \rightarrow Y^{aux}, \text{ } Y^{bu}, \text{ } Y^{td}, \text{ } Y^{tdr}$. %It is to note that we use $Y^{bu}$ and $Y^{tdr}$ as the final bottom-up and top-down output of SVAM-Net, respectively. 

We adopt six loss components to collectively evaluate the information loss and boundary localization error for the supervised training of SVAM-Net. To quantify the information loss, we use the standard binary cross-entropy (BCE) function~\cite{de2005tutorial} that measures the disparity between predicted saliency map $\hat{Y}$ and ground truth $Y$ as   
\begin{equation}\small
    \mathcal{L}_{BCE} (\hat{Y}, {Y}) = \mathbb{E} \big[ - Y_p \log \hat{Y}_p - (1-Y_p) \log (1-\hat{Y}_p) \big].
\label{svam_net_eq_bce}
\end{equation}
We also use the analogous weighted cross-entropy loss function $\mathcal{L}_{WCE} (\hat{Y}, {Y})$, which is widely adopted in SOD literature to handle the imbalance problem of the number of salient pixels~\cite{borji2019salient,wang2018salient,zhao2019pyramid}. While $\mathcal{L}_{WCE}$ provides general guidance for accurate saliency estimation, we use the 2D Laplace operator~\cite{gilbarg2015elliptic} to further ensure robust boundary localization of salient objects. Specifically, we utilize the 2D Laplacian kernel ${K}_{Laplace}$ to evaluate the divergence of image gradients~\cite{zhao2019pyramid} in the predicted saliency map and respective ground truth as  
\begin{align}\small
    \Delta {\hat{Y}} &= \big| \text{{\tt tanh}} \big( \text{{\tt conv}} ( {\hat{Y}}, \text{ } {K}_{Laplace}) \big) \big|, \text{ and} \\
    \Delta {{Y}} &= \big| \text{{\tt tanh}} \big( \text{{\tt conv}} ( {{Y}}, \text{ } {K}_{Laplace}) \big) \big|. \qquad 
\label{svam_net_eq_lap1}
\end{align}
Then, we measure the boundary localization error as
\begin{equation}\small
    \mathcal{L}_{BLE} (\hat{Y}, {Y}) = \mathcal{L}_{WCE}(\Delta {\hat{Y}}, \Delta {{Y}}).
\label{svam_net_eq_ble}
\end{equation}

\begin{figure}[t]
    \centering
    \includegraphics[width=0.9\linewidth]{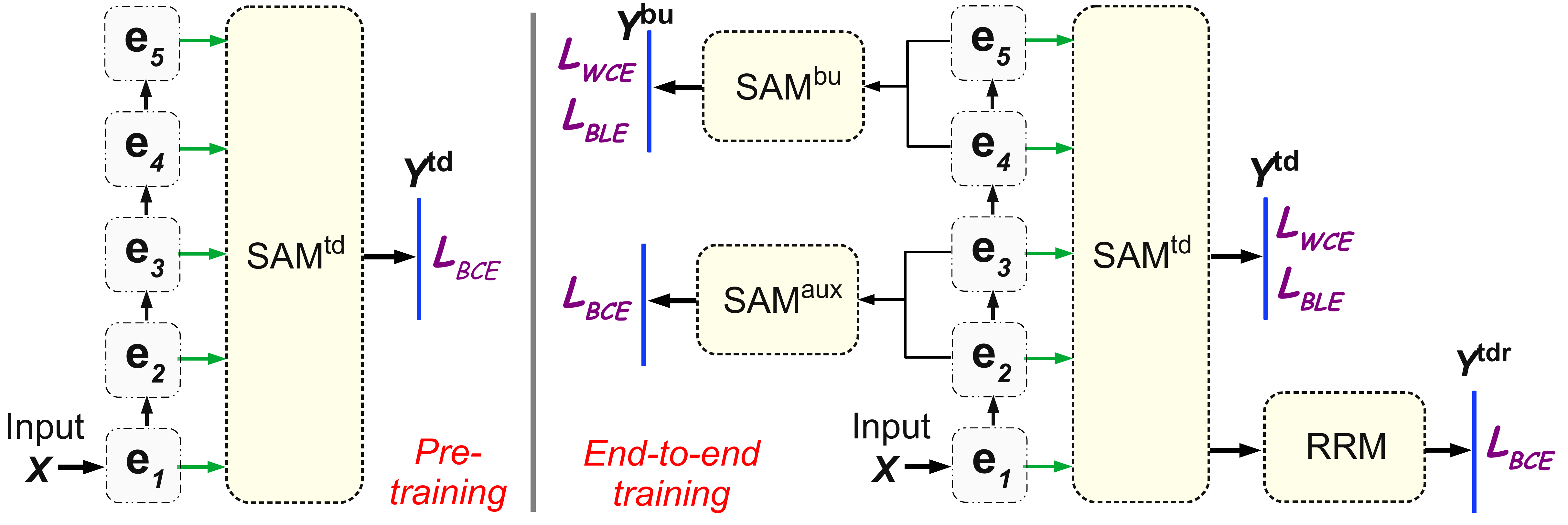}%
\vspace{-2mm}
\caption{Training configurations of SVAM-Net are shown. At first, the backbone and top-down modules are pre-trained holistically on combined terrestrial and underwater data; subsequently, the end-to-end model is fine-tuned by further training on underwater imagery (see Table~\ref{svam_tab_train}). Information loss and/or boundary localization error terms applied at various output layers are annotated by purple letters.  %$\mathcal{L}_{BCE}^{td}$ is used for the supervised pre-training, while the end-to-end training is supervised by $\mathcal{L}_{E2E}$ (see Equation~\ref{svam_net_eq_bce}-\ref{svam_net_eq_bce}).
}
\vspace{2mm}
\label{svam_fig_train_eval}  
\end{figure}

As demonstrated in Figure~\ref{svam_fig_train_eval}, we deploy a two-step training process for SVAM-Net to ensure robust and effective SOD learning. 
First, the backbone encoder and SAM\textsuperscript{td} are pre-trained holistically with combined terrestrial (DUTS~\cite{wang2017learning}) and underwater data (SUIM~\cite{islam2020suim},  UFO-120~\cite{islam2020sesr}). The DUTS training set (DUTS-TR) has $10553$ terrestrial images, whereas the SUIM and UFO-120 datasets contain a total of $3025$ underwater images for training and validation. This large collection of diverse training instances facilitates a comprehensive learning of a generic SOD function (more details in~\ref{svam_gen}). We supervise the training by applying $\mathcal{L}_{PT} 	\equiv \mathcal{L}_{BCE} ({Y}^{td}, {Y})$ loss at the sole output layer of $SAM\textsuperscript{td}$. The SGD optimizer~\cite{kingma2014adam} is used for the iterative learning with an initial rate of $1e^{-2}$ and $0.9$ momentum, which is decayed exponentially by a drop rate of $0.5$ after every $8$ epochs; other hyper-parameters are listed in Table~\ref{svam_tab_train}.

\begin{table}[H]
\centering
\caption{The two-step training process of SVAM-Net and corresponding learning parameters [$b$: batch size; $e$: number of epochs; $N_{train}$: size of the training data; $f_{opt}$: global optimizer; $\eta_{o}$: initial learning rate; $m$: momentum;  $\tau$: decay drop rate].}
\footnotesize
\vspace{2mm}
\begin{tabular}{l||l|l}
  \Xhline{2\arrayrulewidth}
   & Backbone Pre-training & End-to-end Training  \\ 
   \Xhline{2\arrayrulewidth}
   Pipeline & $\{\mathbf{e_{1:5}}\rightarrow \text{SAM\textsuperscript{td}}\}$  & Entire SVAM-Net  \\
   Objective & $\mathcal{L}_{PT} 	\equiv \mathcal{L}_{BCE} ({Y}^{td}, {Y})$  & $\mathcal{L}_{E2E}$ (see Equation~\ref{svam_net_eq_e2e}) \\
   Data & DUTS + SUIM + UFO-120 & SUIM + UFO-120  \\ 
   $b \odot e \text{ } / \text{ } N_{train}$ & $4 \odot 90 \text{ }/\text{ } 13578$ & $4 \odot 50 \text{ }/\text{ } 3025$ \\ 
   $f_{opt} (\eta_{o}, m, \tau)$ & $\text{SGD} (1e^{-2}, 0.9, 0.5)$ & $\text{Adam} (3e^{-4}, 0.5, \times)$ \\ 
  \Xhline{2\arrayrulewidth}
\end{tabular}
\label{svam_tab_train}
\end{table}%

Subsequently, the pre-trained weights are exported into the SVAM-Net model for its end-to-end training on underwater imagery. The loss components applied at the output layers of SAM\textsuperscript{aux}, SAM\textsuperscript{bu}, SAM\textsuperscript{td}, and SAM\textsuperscript{tdr} are  
\begin{align}\small
    \mathcal{L}^{aux}_{E2E} &\equiv \mathcal{L}_{BCE}(Y^{aux}, Y), \\
    \mathcal{L}^{bu}_{E2E} &\equiv \lambda_w \mathcal{L}_{WCE}(Y^{bu}, Y) + \lambda_b \mathcal{L}_{BLE}(Y^{bu}, Y), \\
    \mathcal{L}^{td}_{E2E} &\equiv \lambda_w \mathcal{L}_{WCE}(Y^{td}, Y) + \lambda_b \mathcal{L}_{BLE}(Y^{td}, Y), \text{ and} \\
    \mathcal{L}^{tdr}_{E2E} &\equiv \mathcal{L}_{BCE}(Y^{tdr}, Y).
\label{svam_net_eq_aux}
\end{align}
We formulate the combined objective function as a linear combination of these loss terms as follows
\begin{equation}\small
    \mathcal{L}_{E2E} = \lambda_{aux} \mathcal{L}^{aux}_{E2E} + \lambda_{bu} \mathcal{L}^{bu}_{E2E} + \lambda_{td} \mathcal{L}^{td}_{E2E} + \lambda_{tdr} \mathcal{L}^{tdr}_{E2E}.
\label{svam_net_eq_e2e}
\end{equation}
Here, $\lambda_{\Box}$ symbols are scaling factors that represent the contributions of respective loss components; their values are empirically tuned as hyper-parameters. In our evaluation, the selected values of $\lambda_w$, $\lambda_b$, $\lambda_{aux}$, $\lambda_{bu}$,  $\lambda_{td}$, and   
$\lambda_{tdr}$ are $0.7$, $0.3$, $0.5$, $1.0$, $2.0$, and $4.0$, respectively. As shown in Table~\ref{svam_tab_train}, we use the Adam optimizer~\cite{kingma2014adam} for the global optimization of $\mathcal{L}_{E2E}$ with a learning rate of $3e^{-4}$ and a momentum of $0.5$. 
%the network converges within $100$ epochs of training in this setup with a batch-size of $4$.

\begin{figure}[t]
    \centering
    \includegraphics[width=0.9\linewidth]{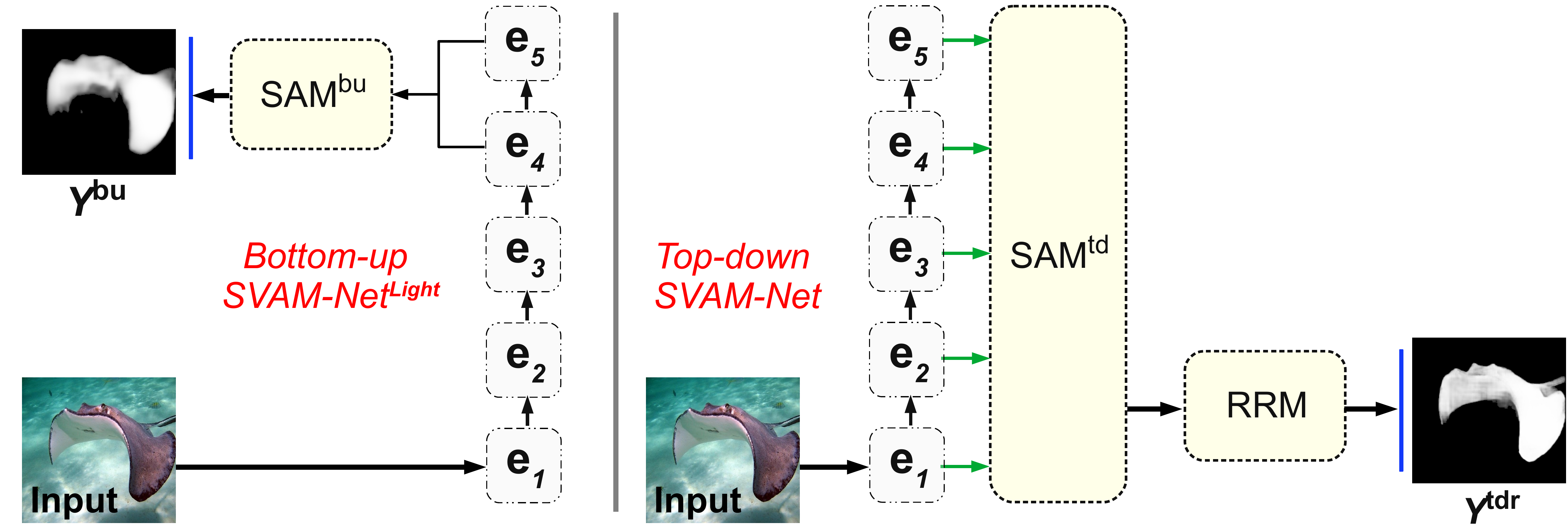}%
\vspace{-2mm}
\caption{The decoupled pipelines for bottom-up and top-down inference: SVAM-Net\textsuperscript{Light} and SVAM-Net (default), respectively.}
\label{svam_fig_inference}  
\end{figure}

\subsection{SVAM-Net Inference}\label{svam_sec_inference}
Once the end-to-end training is completed, we decouple a bottom-up and a top-down branch of SVAM-Net for fast inference. As illustrated in Figure~\ref{svam_fig_inference}, the $\{\mathbf{e_{1:5}}\rightarrow \text{SAM\textsuperscript{td}} \rightarrow \text{RRM}\}$ branch is the default SVAM-Net top-down pipeline that generates fine-grained saliency maps; here, we discard the SAM\textsuperscript{aux} and SAM\textsuperscript{bu} modules to avoid unnecessary computation. On the other hand, we exploit the shallow bottom-up branch, \ie, the $\{\mathbf{e_{1:5}}\rightarrow \text{SAM\textsuperscript{bu}}\}$ pipeline to generate rough yet reasonably accurate saliency maps at a significantly faster rate. Here, we discard SAM\textsuperscript{aux} and both the top-down modules (SAM\textsuperscript{td} and RRM); we denote this computationally light pipeline as SVAM-Net\textsuperscript{Light}. Next, we analyze the SOD performance of SVAM-Net and SVAM-Net\textsuperscript{Light}, demonstrate potential use cases, and discuss various operational considerations.

\begin{figure}[h]
\centering
    \subfigure[Spatial saliency learning over $e=100$ epochs of backbone pre-training; outputs of $Y^{td}$ are shown after $5$, $30$, $60$, and $90$  epochs.]{
        \includegraphics[width=0.9\textwidth]{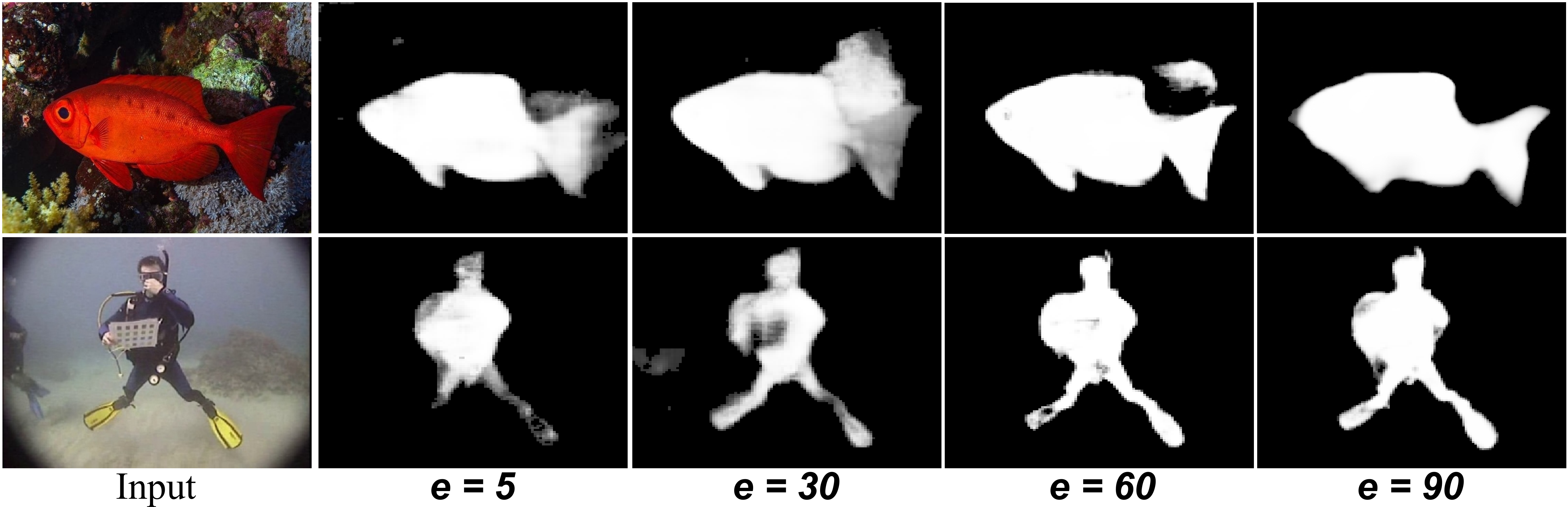} \label{svam_fig_ablation_a}
    }
    \vspace{1mm}

    \subfigure[Snapshots of SVAM-Net output after $40$ epochs of subsequent end-to-end training; notice the spatial attention of early encoding layers (in $Y^{aux}$) and the gradual progression and refinement by the deeper layers (through $Y^{bu} \rightarrow Y^{td} \rightarrow Y^{tdr}$).]{
        \includegraphics[width=0.9\textwidth]{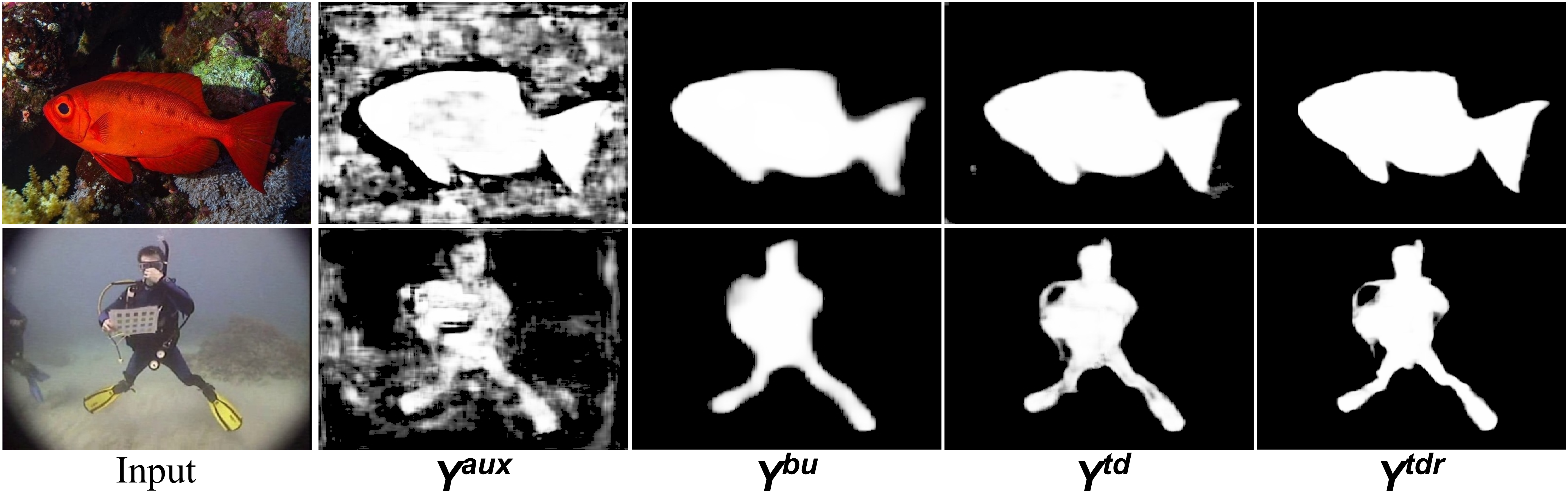} \label{svam_fig_ablation_b}
    }
    \vspace{-1mm}
    \caption{Demonstrations of progressive learning behavior of SVAM-Net and effectiveness of its learning components.}
    \label{svam_fig_ablation}
\end{figure} 

\section{Experimental Evaluation}\label{svam_beval}
\subsection{Implementation Details and Ablation Studies}
As mentioned in~\ref{svam_training}, SVAM-Net training is supervised by paired data $(\{X\}, \{Y\})$ to learn a pixel-wise predictive function $f : X \rightarrow Y^{aux}, \text{ } Y^{bu}, \text{ } Y^{td}, \text{ } Y^{tdr}$. TensorFlow and Keras libraries~\cite{abadi2016tensorflow} are used to implement its network architecture and optimization pipelines (Equation~\ref{svam_net_eq_bce}-\ref{svam_net_eq_e2e}). A Linux machine with two NVIDIA\texttrademark~GTX 1080 graphics cards is used for its backbone pre-training and end-to-end training with the learning parameters provided in Table~\ref{svam_tab_train}.

We demonstrate the progression of SOD learning by SVAM-Net and visualize the contributions of its learning components in Figure~\ref{svam_fig_ablation}. The first stage of learning is guided by supervised pre-training with over $13.5$K instances including both terrestrial and underwater images. This large-scale training facilitates effective feature learning in the backbone encoding layers and by SAM\textsuperscript{td}. As Figure~\ref{svam_fig_ablation_a} shows, the $\{\mathbf{e_{1:5}}\rightarrow \text{SAM\textsuperscript{td}}\}$ pipeline learns spatial attention with a reasonable precision within $90$ epochs. We found that it is crucial to not over-train the backbone for ensuring a smooth and effective end-to-end learning with the integration of SAM\textsuperscript{aux}, SAM\textsuperscript{bu}, and RRM. As illustrated in Figure~\ref{svam_fig_ablation_b}, the subsequent end-to-end training on underwater imagery enables more accurate and fine-grained saliency estimation by SVAM-Net. 

\begin{figure}[t]
    \centering
    \includegraphics[width=\linewidth]{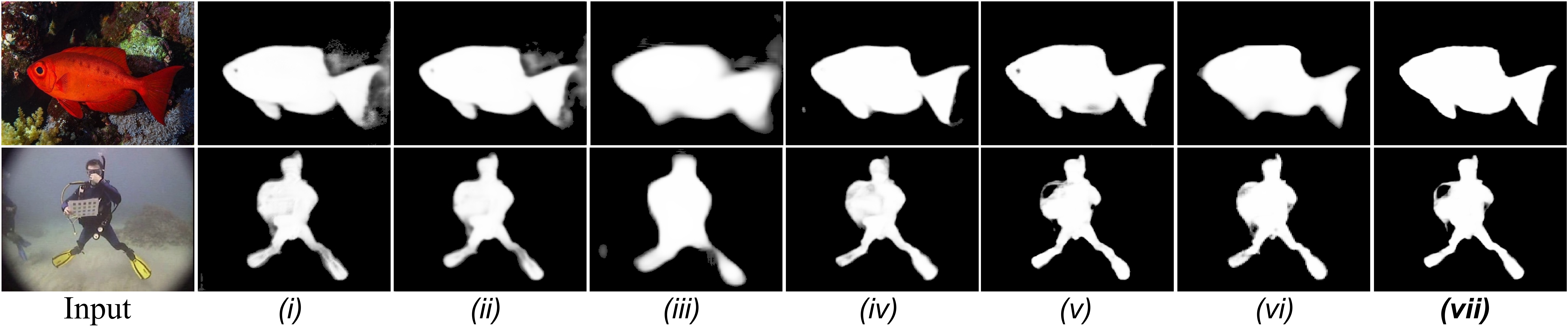}%
    \vspace{-8mm}
\caption{Results of ablation experiments (for the same input images) showing contributions of various attention modules and loss functions in the SOD learning: $(i)$ without $\mathcal{L}_{BLE}$ ($\lambda_b=0$, $\lambda_w=1$), $(ii)$ without SAM\textsuperscript{aux} and SAM\textsuperscript{bu} ($\lambda_{aux}=\lambda_{bu}=0$), $(iii)$ without SAM\textsuperscript{td} and RRM ($\lambda_{td}=\lambda_{tdr}=0$),  $(iv)$ without RRM ($\lambda_{tdr}=0$), $(v)$ without backbone pre-training, $(vi)$ only backbone pre-training, and $(vii)$ proposed two-step training.}
\label{svam_fig_ablation_c}  
\end{figure}

Moreover, we conduct a series of ablation experiments to visually inspect the effects of various loss functions and attention modules in the learning. As Figure~\ref{svam_fig_ablation_c} demonstrates, the boundary awareness (enforced by $\mathcal{L}_{BLE}$) and bottom-up attention modules (SAM\textsuperscript{aux} and SAM\textsuperscript{bu}) are essential to achieve precise localization and sharp contours of the salient objects. It also shows that important details are missed when we incorporate only bottom-up learning, \ie, without SAM\textsuperscript{td} and subsequent delicate refinements by RRM. Besides, the backbone pre-training step is important to ensure generalizability in the SOD learning and as an effective way to combat the lack of large-scale annotated underwater datasets that are needed for comprehensive training.

\subsection{Evaluation Data Preparation}
We conduct benchmark evaluation on three publicly available datasets: SUIM~\cite{islam2020suim}, UFO-120~\cite{islam2020sesr}, and MUED~\cite{jian2019extended}. As mentioned, SVAM-Net is jointly supervised on $3025$ training instances of SUIM and UFO-120; their test sets contain an additional $110$ and $120$ instances, respectively. These datasets contain a diverse collection of natural underwater images with important object categories such as fish, coral reefs, humans, robots, wrecks/ruins, etc. Besides, MUED dataset contains $8600$ images in $430$ groups of conspicuous objects; although it includes a wide variety of complex backgrounds, the images lack diversity in terms of object categories and water-body types. Moreover, MUED provides bounding-box annotations only. Hence, to maintain consistency in our quantitative evaluation, we select $300$ diverse groups and perform pixel-level annotations on those images to obtain ground truth. 

\begin{figure}[t]
    \centering
    \includegraphics[width=0.98\linewidth]{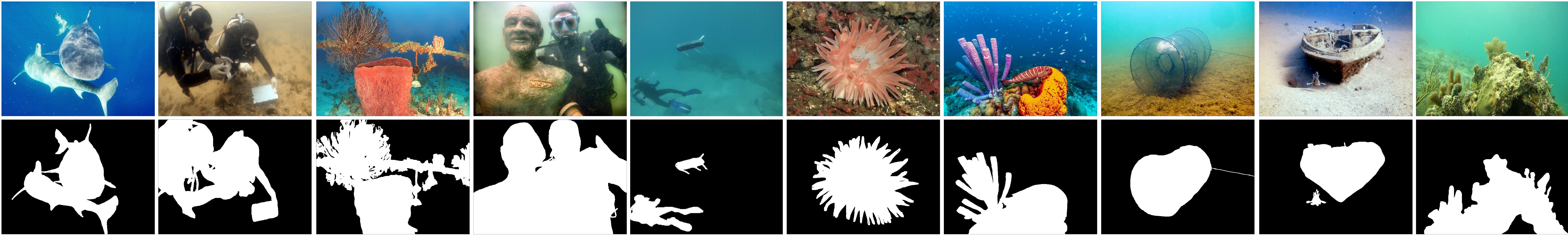}
    \vspace{-3mm}
\caption{There are $300$ test images in the proposed USOD dataset (resolution: $640\times480$); a few sample images and their ground truth saliency maps are shown on the top and bottom row, respectively.}
\label{svam_fig_usod}  
\end{figure}

In addition to the existing datasets, we prepare a challenging test set named \textbf{USOD} to evaluate underwater SOD methods. It contains $300$ natural underwater images which we exhaustively compiled to ensure diversity in the objects categories, water-body, optical distortions, and aspect ratio of the salient objects. We collect these images from two major sources: $(i)$ \textbf{Existing unlabeled datasets}: we utilize benchmark datasets that are generally used for underwater image enhancement and super-resolution tasks; specifically, we select subsets of images from datasets named USR-248~\cite{islam2019underwater}, UIEB~\cite{li2019underwater}, and EUVP~\cite{islam2019fast}. $(ii)$ \textbf{Field trials}: we have collected data from several oceanic trials and explorations in the Caribbean sea at Barbados. The selected images include diverse underwater scenes and setups for human-robot cooperative experiments (see Section~\ref{svam_deploy}).
Once the images are compiled, we annotate the salient pixels to generate ground truth labels; a few sample images are illustrated in Figure~\ref{svam_fig_usod}.

\begin{table}[H]
\centering
\caption{Quantitative performance comparison of SVAM-Net and SVAM-Net\textsuperscript{Light} with existing solutions and SOTA methods for both underwater (first six) and terrestrial (last four) domains are shown. All scores of F-measure ($\mathbf{F}_{\beta}^{max}$), S-measure ($\mathbf{S}_m$), and mean absolute error ($\mathbf{MAE}$) are in $[0, 1]$; two (column-wise) top scores are bold-faced.
}
\vspace{2mm}
\scriptsize
\begin{tabular}{l||ccc|ccc}
\multicolumn{7}{c}{(a) Comparison for SUIM~\cite{islam2020suim} and UFO-120~\cite{islam2020sesr} dataset.} \\ 
  \Xhline{2\arrayrulewidth}
   & \multicolumn{3}{c|}{SUIM} & \multicolumn{3}{c}{UFO-120} \\  
  %\cline{2-10} 
  Method & $\mathbf{F}_{\beta}^{max}$ & $\mathbf{S}_m$ & $\mathbf{MAE}$ & $\mathbf{F}_{\beta}^{max}$ & $\mathbf{S}_m$ & $\mathbf{MAE}$ \\
  & ($\uparrow$) & ($\uparrow$) & ($\downarrow$) & ($\uparrow$) & ($\uparrow$) & ($\downarrow$) \\
  \Xhline{2\arrayrulewidth}
  SAOE~\cite{wang2013saliency} & $0.2698$ & $0.3965$ & $0.4015$ & $0.4011$  & $0.4420$ & $0.3752$    \\
  SSRC~\cite{li2016saliency} & $0.3015$ & $0.4226$ & $0.3028$ & $0.3836$  & $0.4534$ & $0.4125$    \\
  Deep SESR~\cite{islam2020sesr} & $0.3838$ & $0.4769$ & $0.2619$ & $0.4631$  & $0.5146$ & $0.3437$    \\
  LSM~\cite{chen2019underwater} & $0.5443$ & $0.5873$ & $0.1504$ & $0.6908$  & $0.6770$ & $0.1396$   \\
  SUIM-Net~\cite{islam2020suim} & {$\mathbf{0.8413}$} & $0.8296$ & {$\mathbf{0.0787}$} & $0.6628$  & $0.6790$ & $0.1427$    \\
  QDWD~\cite{jian2018integrating} & $0.7328$ & $0.6978$ & $0.1129$ & $0.7074$  & $0.7044$ & $0.1368$    \\ \hline 
  \textbf{SVAM-Net\textsuperscript{Light}} & $0.8254$ & {$\mathbf{0.8356}$} & $0.0805$ & {$\mathbf{0.8428}$} & {$\mathbf{0.8613}$} & {$\mathbf{0.0663}$}  \\
  \textbf{SVAM-Net} & {$\mathbf{0.8830}$} & {$\mathbf{0.8607}$} & {$\mathbf{0.0593}$} & {$\mathbf{0.8919}$} & {$\mathbf{0.8808}$} & {$\mathbf{0.0475}$}    \\ \hline
  BASNet~\cite{qin2019basnet} & $0.7212$ & $0.6873$ & $0.1142$ & $0.7609$  & $0.7302$ & $0.1108$    \\
  PAGE-Net~\cite{wang2019salient} & $0.7481$ & $0.7207$ & $0.1028$ & $0.7518$  & $0.7522$ & $0.1062$    \\
  ASNet~\cite{wang2018salient} & $0.7344$ & $0.6740$ & $0.1168$ & $0.7540$  & $0.7272$ & $0.1153$   \\
  CPD~\cite{wu2019cascaded} & $0.6679$ & $0.6254$ & $0.1387$ & $0.6947$  & $0.6880$ & $0.3752$   \\
  \Xhline{2\arrayrulewidth}
  \end{tabular}
  \vspace{3mm}
  
\begin{tabular}{l||ccc|ccc}
\multicolumn{7}{c}{(b) Comparison for MUED~\cite{jian2019extended} and USOD dataset.} \\
  \Xhline{2\arrayrulewidth}
   & \multicolumn{3}{c|}{MUED~\cite{jian2019extended}} & \multicolumn{3}{c}{\textbf{USOD}} \\  
  %\cline{2-10} 
  Method &  $\mathbf{F}_{\beta}^{max}$ & $\mathbf{S}_m$ & $\mathbf{MAE}$ & $\mathbf{F}_{\beta}^{max}$ & $\mathbf{S}_m$ & $\mathbf{MAE}$ \\
  & ($\uparrow$) & ($\uparrow$) & ($\downarrow$) & ($\uparrow$) & ($\uparrow$) & ($\downarrow$) \\
  \Xhline{2\arrayrulewidth}
  SAOE~\cite{wang2013saliency} & $0.2978$  & $0.3045$ & $0.3849$ & $0.2520$  & $0.2418$  & $0.4678$    \\
  SSRC~\cite{li2016saliency} & $0.4040$  & $0.3946$ & $0.2295$ & $0.2143$  & $0.2846$  & $0.3872$    \\
  Deep SESR~\cite{islam2020sesr} & $0.3895$  & $0.3565$ & $0.2118$ & $0.3914$  & $0.4868$  & $0.3030$    \\
  LSM~\cite{chen2019underwater} & $0.4174$  & $0.4025$ & $0.1934$ & $0.6775$  & $0.6768$  & $0.1186$    \\
  SUIM-Net~\cite{islam2020suim} & $0.5686$  & $0.5070$ & $0.1227$ & $0.6818$  & $0.6754$  & $0.1386$    \\
  QDWD~\cite{jian2018integrating} & $0.6248$  & $0.5975$ & $0.0771$ & $0.7750$  & $0.7245$  & $0.0989$    \\ \hline 
  \textbf{SVAM-Net\textsuperscript{Light}} & $0.8492$  & $0.8588$ & $0.0184$ & {$\mathbf{0.8703}$}  & {$\mathbf{0.8723}$}  & {$\mathbf{0.0619}$}    \\
  \textbf{SVAM-Net} & {$\mathbf{0.9013}$} & {$\mathbf{0.8692}$} & {$\mathbf{0.0137}$} & {$\mathbf{0.9162}$}  & {$\mathbf{0.8832}$}  & {$\mathbf{0.0450}$}    \\ \hline
  BASNet~\cite{qin2019basnet} & {$\mathbf{0.8556}$}  & {$\mathbf{0.8820}$} & {$\mathbf{0.0145}$} & $0.8425$  & $0.7919$  & $0.0745$    \\
  PAGE-Net~\cite{wang2019salient} & $0.6849$  & $0.7136$ & $0.0442$ & $0.8430$  & $0.8017$  & $0.0713$    \\
  ASNet~\cite{wang2018salient} & $0.6413$  & $0.7476$ & $0.0370$ & $0.8310$  & $0.7732$  & $0.0798$    \\
  CPD~\cite{wu2019cascaded} & $0.7624$  & $0.7311$ & $0.0330$ & $0.7877$  & $0.7436$  & $0.0917$    \\
  \Xhline{2\arrayrulewidth}
  \end{tabular}  
\label{svam_tab_quan_uw}
\end{table}

%\begin{figure}[h]
%    \centering
%    \includegraphics[width=0.82\linewidth]{figs/usod_10.png}%
%\caption{Comparison of PR curves on USOD dataset is shown for the top ten SOD models based on the results shown in Table~\ref{svam_tab_quan_uw}.}
%\label{svam_fig_quant_usod}  
%\end{figure}

\begin{figure*}[h]
    \centering
    \includegraphics[width=0.49\linewidth]{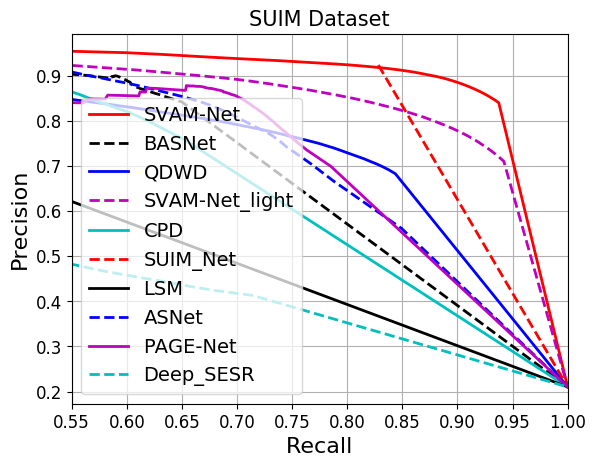}~\hspace{1mm}
    \includegraphics[width=0.49\linewidth]{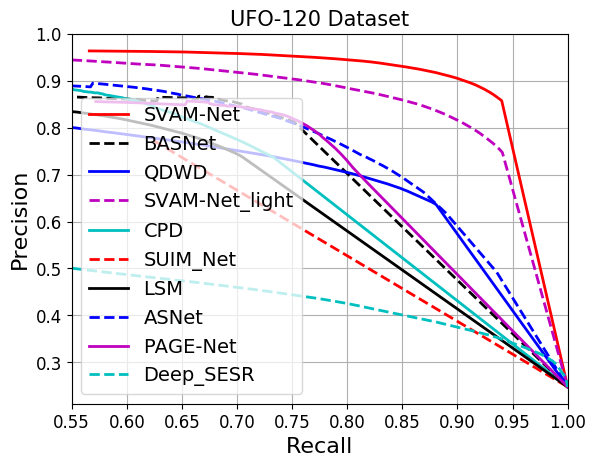} \\
    \vspace{3mm}
    
    \includegraphics[width=0.49\linewidth]{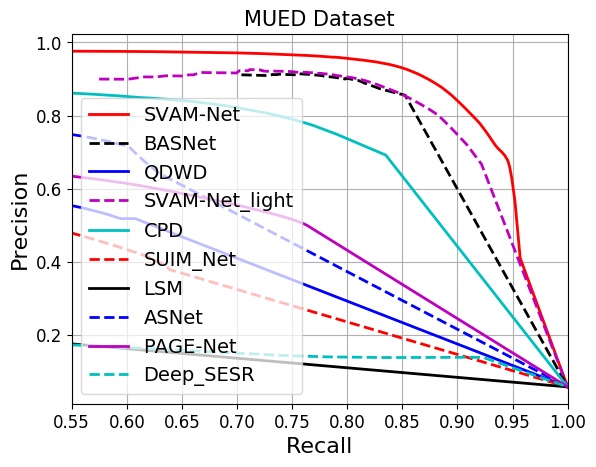}~\hspace{1mm}
    \includegraphics[width=0.49\linewidth]{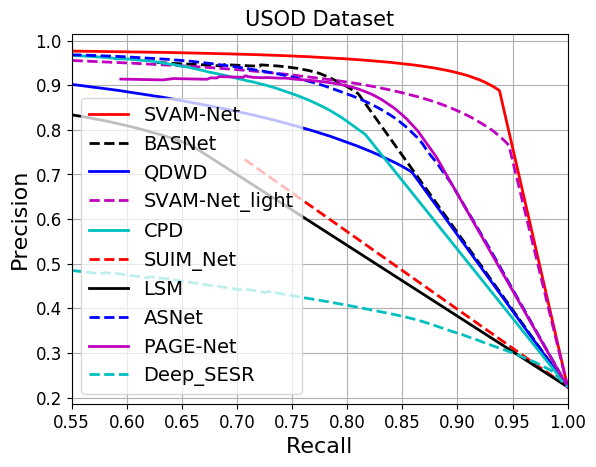}%
\vspace{-2mm}
\caption{Comparisons of PR curves on three benchmark datasets (SUIM, UFO-120, MUED) and the proposed USOD dataset are shown; to maintain clarity, we consider the top ten SOD models based on the results shown in Table~\ref{svam_tab_quan_uw}.}
\label{svam_fig_quant_uw}  
\end{figure*}

\begin{figure*}[ht]
    \centering
    \includegraphics[width=\linewidth]{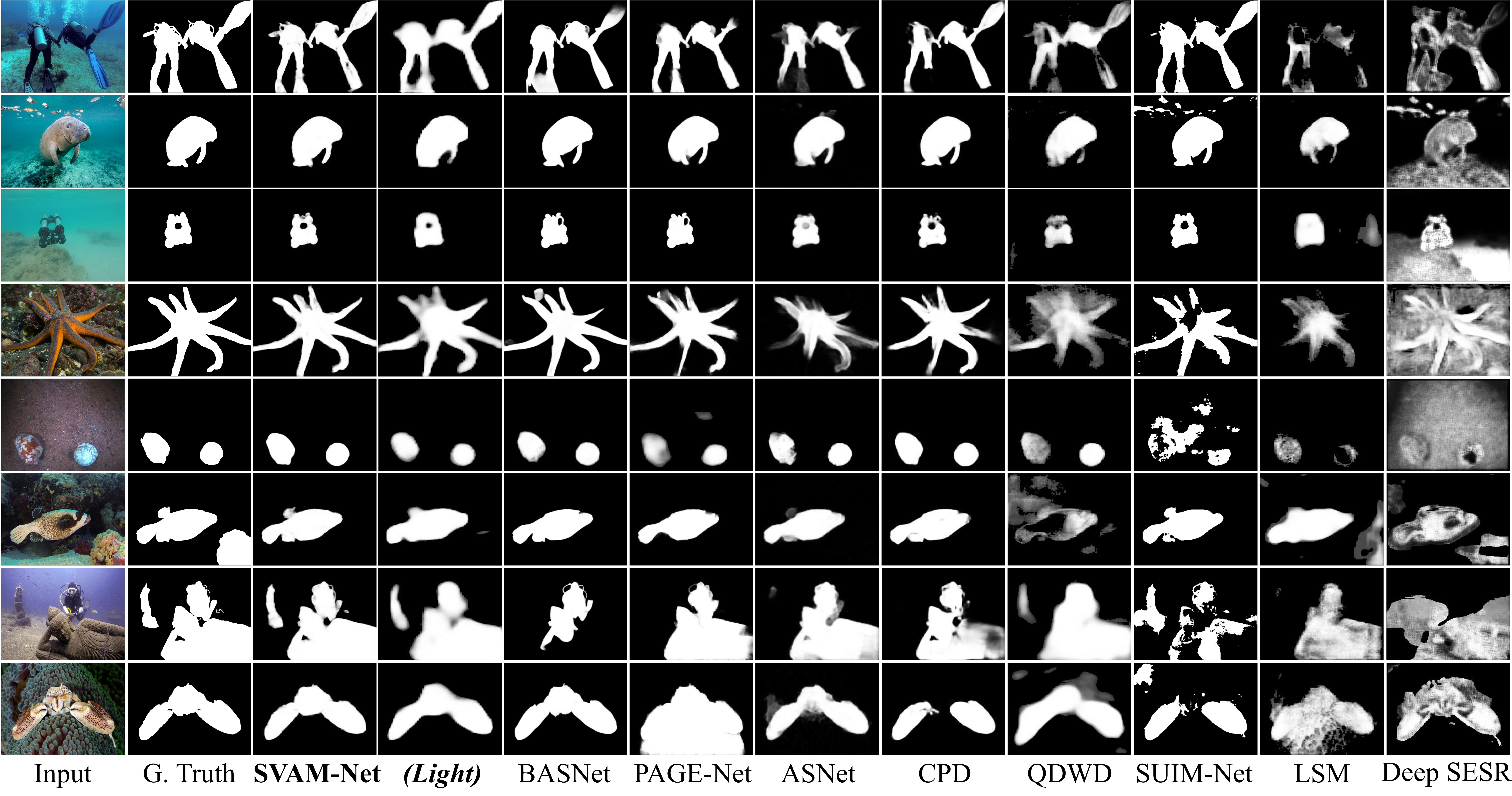}%
\vspace{-7mm}
\caption{A few qualitative comparisons of saliency maps generated by the top ten SOD models (based on the results of Table~\ref{svam_tab_quan_uw}). From the top: first four images belong to the test sets of SUIM~\cite{islam2020suim} and UFO-120~\cite{islam2020sesr}, the next one to MUED~\cite{jian2019extended}, whereas the last three images belong to the proposed USOD dataset. (Best viewed at 300\% zoom)}%
\label{svam_fig_comp_uw}  
\end{figure*}

\subsection{Quantitative and Qualitative Analysis}
We evaluate the performance of SVAM-Net and other existing SOD methods based on four widely-used evaluation criteria~\cite{borji2019salient,feng2019attentive,qin2019basnet,wang2018detect}: F-measure ($\mathbf{F}_{\beta}^{max}$), S-measure ($\mathbf{S}_{m}$), Mean  Absolute Error ($\mathbf{MAE}$), and Precision-recall (PR) curve; their definitions and relevant details are provided in Appendix~\ref{ApenSOD}. For performance comparison, we consider the following six methods that are widely used for underwater SOD and/or saliency estimation: ($i$) SOD by Quaternionic
Distance-based Weber Descriptor (QDWD)~\cite{jian2018integrating}, ($ii$) saliency estimation by the Segmentation of Underwater IMagery Network (SUIM-Net)~\cite{islam2020suim}, ($iii$) saliency prediction by the Deep Simultaneous Enhancement and Super-Resolution (Deep SESR) model~\cite{islam2020sesr}, ($iv$) SOD by a Level Set-guided Method (LSM)~\cite{chen2019underwater}, ($v$) Saliency Segmentation by evaluating Region Contrast (SSRC)~\cite{li2016saliency}, and ($vi$) SOD by Saliency-based Adaptive Object Extraction (SAOE)~\cite{wang2013saliency}. We also include the performance margins of four SOTA SOD models: ($i$) Boundary-Aware Saliency Network (BASNet)~\cite{qin2019basnet}, ($ii$) Pyramid Attentive and
salient edGE-aware Network (PAGE-Net)~\cite{wang2019salient}, ($iii$) Attentive Saliency Network (ASNet)~\cite{wang2018salient}, and ($iv$) Cascaded Partial Decoder (CPD)~\cite{wu2019cascaded}. We use their publicly released weights (pre-trained on terrestrial imagery) and further train them on combined SUIM and UFO-120 data by following the same setup as SVAM-Net (see Table~\ref{svam_tab_train}). We present detailed results for this comprehensive performance analysis in Table~\ref{svam_tab_quan_uw}. 

As the results in the first part of Table~\ref{svam_tab_quan_uw} suggest, SVAM-Net outperforms all the underwater SOD models in comparison with significant margins. Although QDWD and SUIM-Net perform reasonably well on particular datasets (\eg, SUIM, and MUED, respectively), their $\mathbf{F}_{\beta}^{max}$, $\mathbf{S}_{m}$, and $\mathbf{MAE}$ scores are much lower; in fact, their scores are comparable to and often lower than those of SVAM-Net\textsuperscript{Light}. The LSM, Deep SESR, SSRC, and SAOE models offer even lower scores than SVAM-Net\textsuperscript{Light}. The respective comparisons of PR curves shown in Figure~\ref{svam_fig_quant_uw} further validate the superior performance of SVAM-Net and SVAM-Net\textsuperscript{Light} by an area-under-the-curve (AUC)-based analysis. Moreover, Figure~\ref{svam_fig_comp_uw} demonstrates that SVAM-Net-generated saliency maps are accurate with precisely segmented boundary pixels. Although not as fine-grained, SVAM-Net\textsuperscript{Light} also generates reasonably well-localized saliency maps that are still more accurate and consistent compared to the existing models. These results corroborate our discussion on the lack of advancements of underwater SOD literature (see Section~\ref{related_work}).

For a comprehensive validation of SVAM-Net, we compare the performance margins of SOTA SOD models trained through the same learning pipeline. As shown in Figure~\ref{svam_fig_comp_uw}, the saliency maps of BASNet, PAGE-Net, ASNet, and CPD are mostly accurate and often comparable to  SVAM-Net-generated maps. 
The quantitative results of Table~\ref{svam_tab_quan_uw} and Figure~\ref{svam_fig_quant_uw} also confirm their competitive performance over all datasets. Given the substantial learning capacities of these models, one may exhaustively find a better choice of hyper-parameters that further improves their baseline performances. Nevertheless, unlike these standard models, SVAM-Net incorporates a considerably shallow computational pipeline and offers an even lighter bottom-up sub-network (SVAM-Net\textsuperscript{Light}) that ensures fast inference on single-board devices. 
In the following section, we demonstrate SVAM-Net's generalization performance and discuss its utility in various practical applications.

\subsection{Generalization Performance}\label{svam_gen}
\begin{figure}[t]
\centering
    \subfigure[Lack of contrast and/or color distortions.]{
      \includegraphics[width=0.85\textwidth]{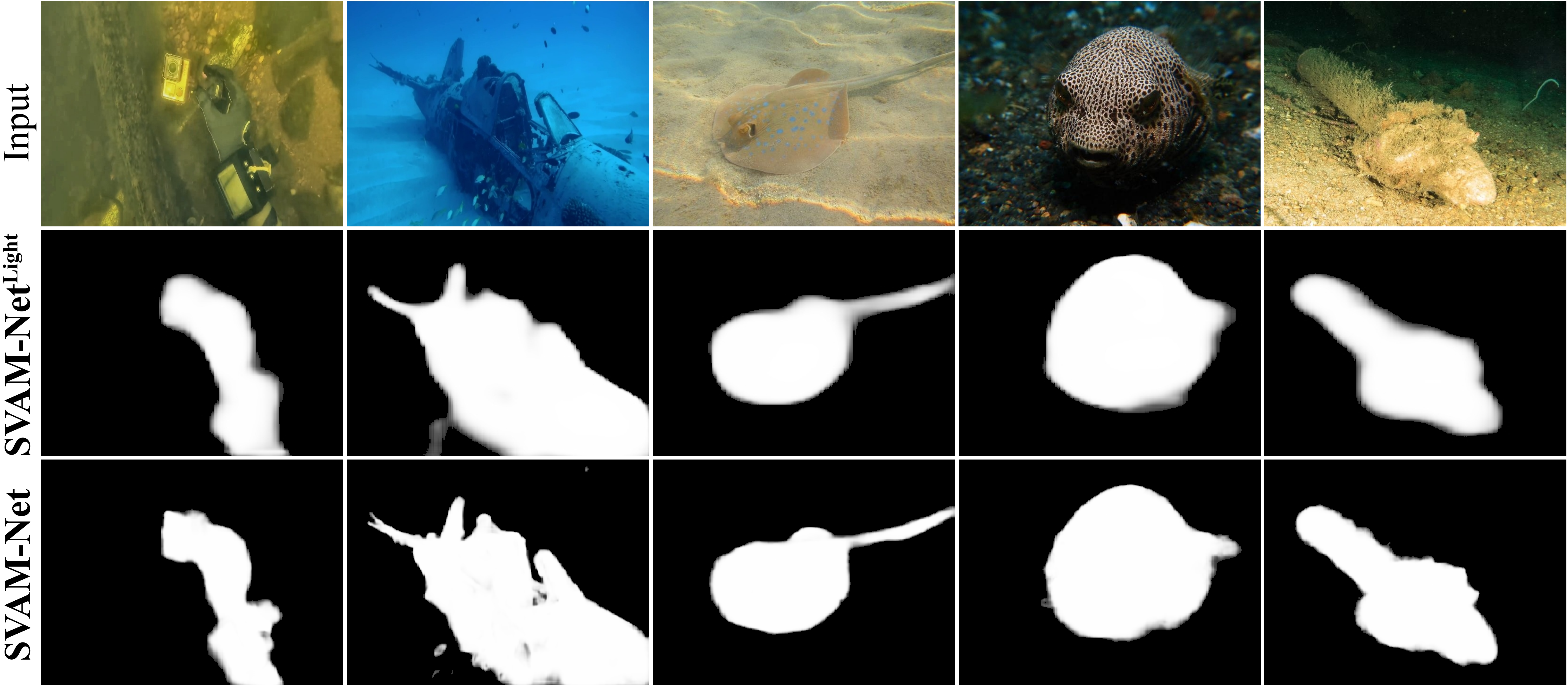} \label{svam_fig_gen_a}
    }
    
    \subfigure[Cluttered background and/or confusing textures.]{
        \includegraphics[width=0.85\textwidth]{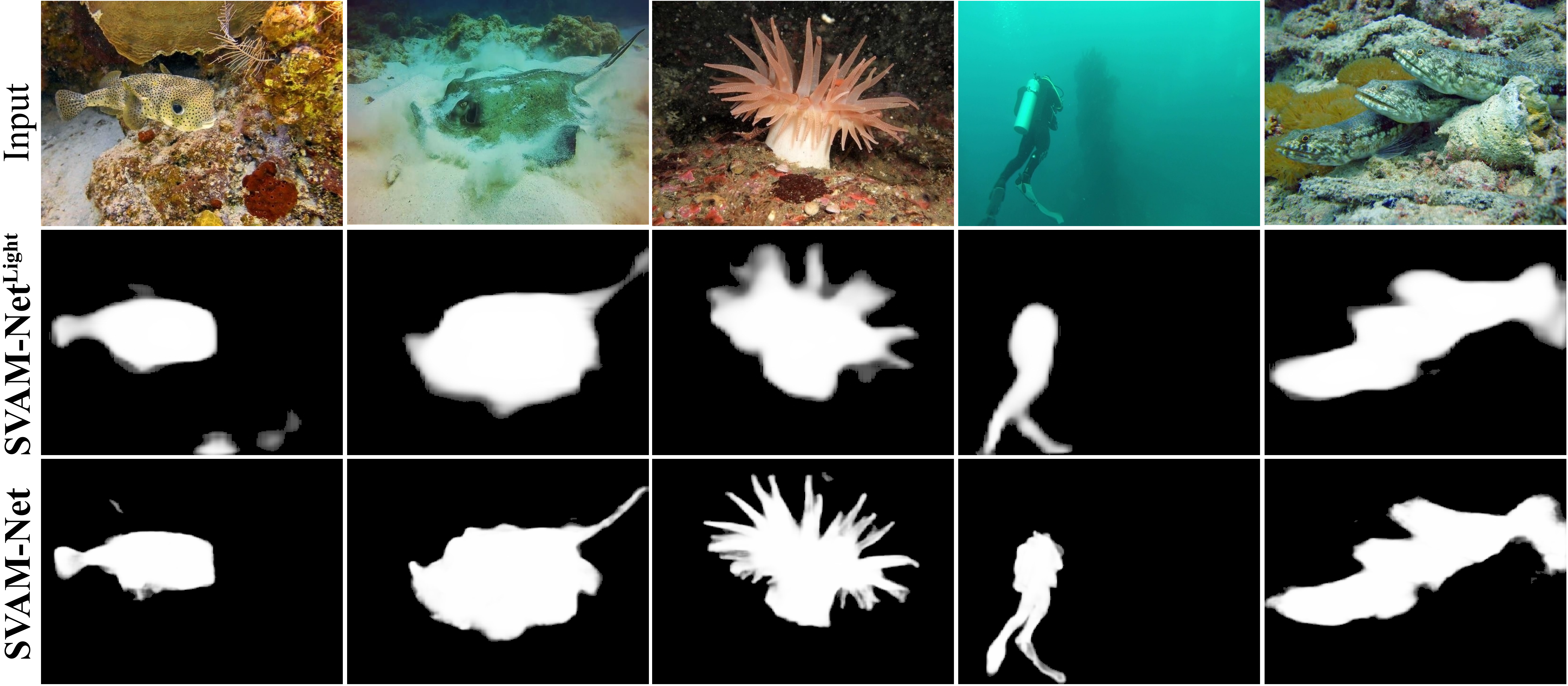} \label{svam_fig_gen_b}
    }
    \vspace{-1mm}
    \caption{Demonstrations of generalization performance of SVAM-Net over various categories of challenging test cases.}
    \label{svam_fig_general}
\end{figure}

Underwater imagery suffers from a wide range of non-linear distortions caused by the waterbody-specific properties of light propagation~\cite{islam2019fast,islam2020sesr}. The image quality and statistics also vary depending on visibility conditions, background patterns, and the presence of artificial light sources and unknown objects  in a scene. Consequently, learning-based SOD solutions oftentimes fail to generalize beyond supervised data. To address this issue,  SVAM-Net adopts a two-step training pipeline (see Section~\ref{svam_training}) that includes supervision by ($i$) a large collection of samples with diverse scenes and object categories to learn a generalizable SOD function, and ($ii$) a wide variety of natural underwater images to learn to capture the inherent optical distortions. In Figure~\ref{svam_fig_general}, we demonstrate the robustness of SVAM-Net with a series of challenging test cases.        

As shown in Figure~\ref{svam_fig_gen_a}, underwater images tend to have a dominating green
or blue hue because red wavelengths get absorbed in
deep water (as light travels further)~\cite{akkaynak2018revised}. Such wavelength dependent attenuation, scattering, and other optical properties of the waterbodies cause irregular and non-linear distortions which result in low-contrast, often blurred,
and color-degraded images~\cite{islam2019fast,torres2005color}. We notice that both SVAM-Net and SVAM-Net\textsuperscript{Light} can overcome the noise and image distortions and successfully localize the salient objects. 
They are also robust to other pervasive issues such as occlusion and cluttered backgrounds with confusing textures. As Figure~\ref{svam_fig_gen_b} demonstrates, the salient objects are mostly well-segmented from the confusing background pixels having similar colors and textures. Here, we observe that although SVAM-Net\textsuperscript{Light} introduces a few false-positive pixels, SVAM-Net's predictions are consistently accurate and more fine-grained.

\begin{figure}[t]
\centering
    \subfigure[Unseen objects/shapes and variations in scale.]{
      \includegraphics[width=0.85\textwidth]{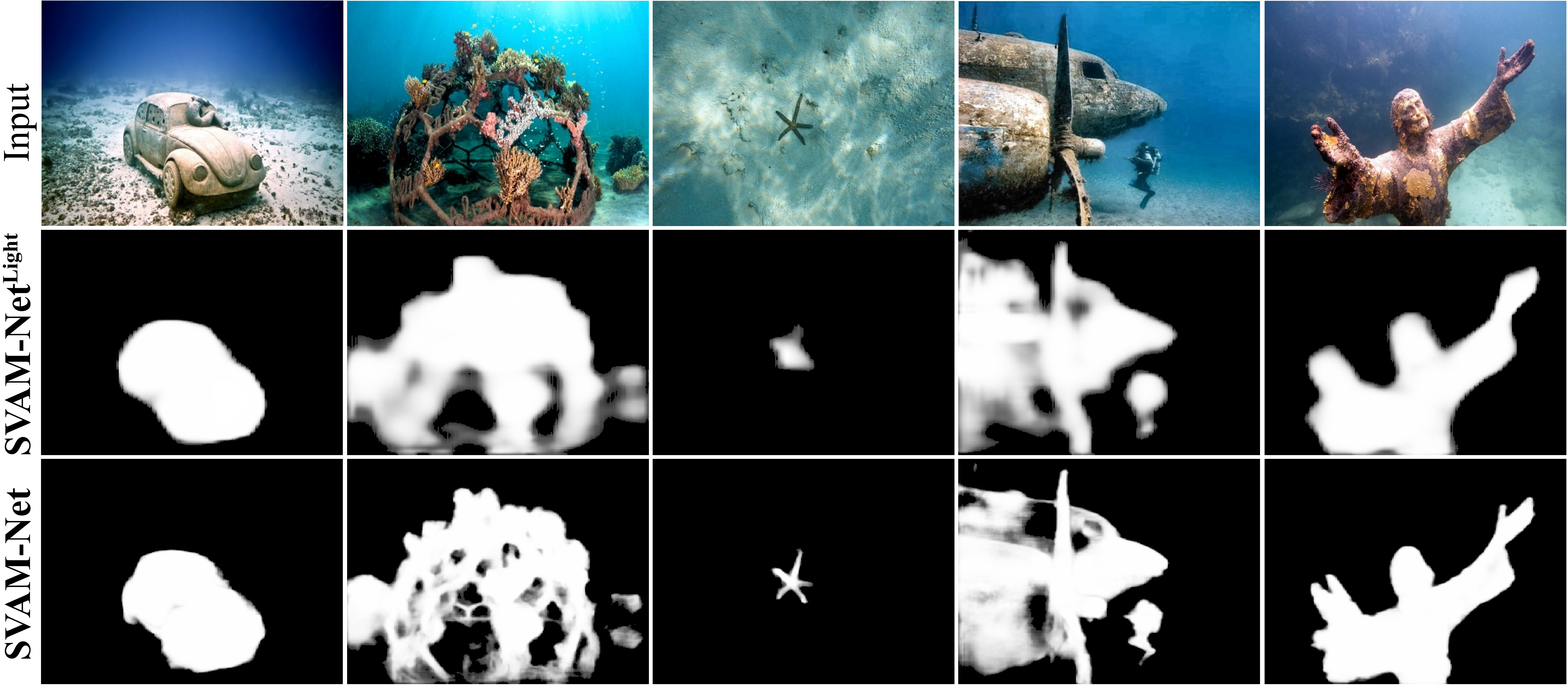} \label{svam_fig_gen_c}
    }
    
    \subfigure[Unseen terrestrial images with arbitrary objects.]{
      \includegraphics[width=0.85\textwidth]{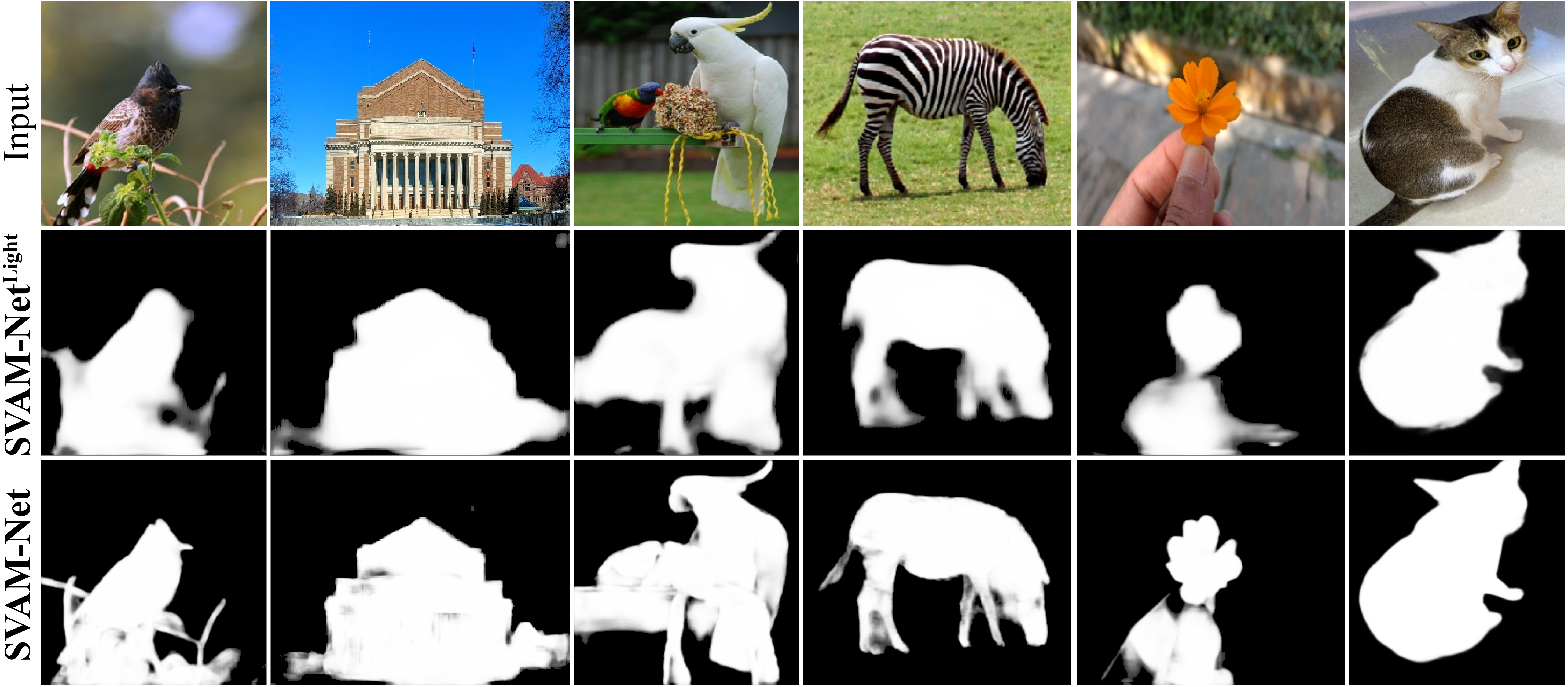} \label{svam_fig_gen_d}
    }
    \vspace{-1mm}
    \caption{Demonstrations of generalization performance of SVAM-Net on arbitrary images with unseen natural objects.}
    \label{svam_fig_general2}
\end{figure} 
 
Another important feature of general-purpose SOD models is the ability to identify novel salient objects, particularly with complicated shapes. As shown in Figure~\ref{svam_fig_gen_c}, objects such as wrecked/submerged cars, planes, statues, and cages are accurately segmented by both SVAM-Net and SVAM-Net\textsuperscript{Light}. Their SOD performance is also invariant to the scale and orientation of salient objects. We postulate that the large-scale supervised pre-training step contributes to this robustness as the terrestrial datasets include a variety of object categories. In fact, we find that they also perform reasonably well on arbitrary terrestrial images (see Figure~\ref{svam_fig_gen_d}), which suggest that with domain-specific end-to-end training, SVAM-Net could be effectively used in terrestrial applications as well.

\subsection{Operational Feasibility Analysis}\label{svam_deploy}
SVAM-Net offers an end-to-end run-time of $49.82$ milliseconds (ms) per-frame, \ie, $20.07$ frames-per-second (FPS) on a single NVIDIA\texttrademark~GTX 1080 GPU. As Table~\ref{svam_tab_time} shows, SVAM-Net\textsuperscript{Light} operates at a much faster rate of $11.60$ ms per-frame ($86.15$ FPS). These inference rates surpass the reported speeds of SOTA SOD models~\cite{wang2019salientsurvey,borji2019salient} and are adequate for GPU-based use in real-time applications. More importantly, SVAM-Net\textsuperscript{Light} runs at $21.77$ FPS rate on a single-board computer named NVIDIA\texttrademark~Jetson AGX Xavier with an on-board memory requirement of only $65$ MB. 
These computational aspects make SVAM-Net\textsuperscript{Light} ideally suited for
single-board robotic deployments, and justify our design intuition of decoupling the bottom-up pipeline $\{\mathbf{e_{1:5}}\rightarrow \text{SAM\textsuperscript{bu}}\}$
from the SVAM-Net architecture (see Section~\ref{svam_sec_inference}).

\begin{table}[H]
\centering
\caption{Run-time comparison for SVAM-Net and SVAM-Net\textsuperscript{Light} on a GTX 1080 GPU and on a single-board NVIDIA\texttrademark~Jetson AGX Xavier device.}
\footnotesize
\vspace{3mm}
\begin{tabular}{l||l|l}
  \Xhline{2\arrayrulewidth}
   & SVAM-Net  & \textbf{SVAM-Net\textsuperscript{Light}}   \\ \Xhline{2\arrayrulewidth} 
  GTX 1080 &  $49.82$ ms ($20.07$ FPS)  &  $11.60$ ms ($86.15$ FPS)   \\
  %Jetson TX2 &  $925.92$ ms ($1.08$ FPS)  &  $172.41$ ms ($5.8$ FPS)   \\ 
  AGX Xavier &  $222.2$ ms ($4.50$ FPS)  &  $\mathbf{45.93}$ \textbf{ms} ($\mathbf{21.77}$ \textbf{FPS})   \\ \Xhline{2\arrayrulewidth}
\end{tabular}
\label{svam_tab_time}
\end{table}

\begin{figure*}[t]
    \centering
    \includegraphics[width=\linewidth]{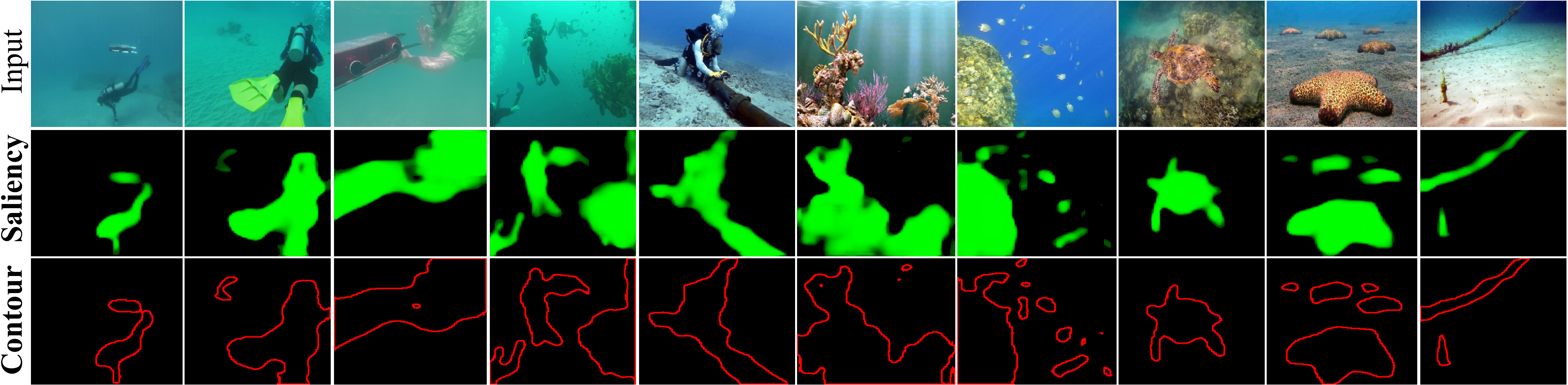}%
    \vspace{-7mm}
    \caption{SVAM-Net\textsuperscript{Light}-generated saliency maps and  respective object contours are shown for a variety of snapshots taken during human-robot cooperative experiments and oceanic explorations. A video demonstration can be seen here: {\tt \url{https://youtu.be/SxJcsoQw7KI}}.
    }
    \label{svam_fig_vam}
\end{figure*}

In the last two sections, we discussed the practicalities involved in designing a generalized underwater SOD model and identified several drawbacks of existing solutions such as QDWD, SUIM-Net, LSM, and Deep SESR. Specifically, we showed that their predicted saliency maps lack important details, exhibit improperly segmented object boundaries, and incur plenty of false-positive pixels (see Figure~\ref{svam_fig_comp_uw}). Although such sparse detection of salient pixels can be useful in specific higher-level tasks (\eg, contrast enhancement~\cite{islam2020sesr}, rough foreground extraction~\cite{li2016saliency}), these models are not as effective for general-purpose SOD. It is evident from our experimental results that the proposed SVAM-Net model overcomes these limitations and offers a robust SOD solution for underwater imagery. 
For underwater robot vision, in particular, robust identification of salient pixels is the most essential first step in these approaches irrespective of the high-level application-specific tasks, \eg, enhanced object detection~\cite{zhu2020saliency,rizzini2015investigation,ravanbakhsh2015automated}, place recognition~\cite{maldonado2019learning}, coral reef monitoring~\cite{modasshir2020enhancing,ModasshirFSR2019}, autonomous exploration~\cite{girdhar2016modeling,rekleitis2001multi}, etc. SVAM-Net\textsuperscript{Light} offers a general-purpose solution to this, while ensuring fast inference rates on single-board devices. As shown in Figure~\ref{svam_fig_vam}, SVAM-Net\textsuperscript{Light} reliably detects humans, robots, wrecks/ruins, instruments, and other salient objects in a scene. Additionally, it accurately discards all background (waterbody) pixels and focuses on salient foreground pixels only. Such precise segmentation of salient image regions enables fast and effective spatial attention modeling, which is key to the operational success of visually-guided underwater robots. 

\section{Concluding Remarks}
\textit{``Where to look''} is a challenging and open problem in underwater robot vision. An essential capability of visually-guided AUVs is to identify interesting and salient objects in the scene to accurately make important operational decisions. 
In this chapter, we present a novel deep visual model named SVAM-Net, which combines the power of bottom-up and top-down SOD learning in a holistic encoder-decoder architecture. We design dedicated spatial attention modules to effectively exploit the coarse-level and fine-level semantic features along the two learning pathways. In particular, we configure the bottom-up pipeline to extract semantically rich hierarchical features from early encoding layers, which facilitates an abstract yet accurate saliency prediction at a significantly faster rate; we denote this decoupled bottom-up pipeline as SVAM-Net\textsuperscript{Light}. %On the other hand, we design a residual refinement module that ensures fine-grained saliency estimation through the deeper top-down pipeline. 

In the implementation, we incorporate comprehensive end-to-end supervision of SVAM-Net by large-scale diverse training data consisting of both terrestrial and underwater imagery. Subsequently, we validate the effectiveness of its learning components and various loss functions by extensive ablation experiments. In addition to using existing datasets, we release a new challenging test set named USOD for the benchmark evaluation of SVAM-Net and other underwater SOD models. By a series of qualitative and quantitative analyses, we show that SVAM-Net provides SOTA performance for SOD on underwater imagery, exhibits significantly better generalization performance on challenging test cases than existing solutions, and achieves fast end-to-end inference on single-board devices. Moreover, we demonstrate that a delicate balance between robust performance and computational efficiency makes SVAM-Net\textsuperscript{Light} suitable for real-time use by visually-guided underwater robots. %In the near future, we plan to optimize the end-to-end SVAM-Net architecture further to achieve a faster run-time. %The subsequent pursuit will be to analyze its feasibility in online learning pipelines for task-specific model adaptation. 

\chapter{Platform-specific Adaptations and Further Discussions}\label{do_better}
In Chapter~\ref{diver_following} through Chapter~\ref{svam}, we presented our proposed methodologies for autonomous diver following~\cite{islam2017mixed,islam2018towards}, robot-to-robot pose estimation~\cite{islam2019robot}, human-to-robot communication~\cite{islam2018dynamic,islam2018understanding}, underwater image enhancement and super-resolution~\cite{islam2019fast,islam2020sesr}, and visual attention modeling~\cite{islam2020svam}. We also discussed their underlying algorithms and implementation details for single-board embedded platforms. For the deep visual models, in particular, we provided various platform-specific design choices to ensure their computational feasibility for real-time robotic deployments. In this chapter, we further extend our discussion on these aspects from the practical \emph{usability} perspective. First, we specify the hardware and software platforms used in our implementations and field deployments in Section~\ref{hw_sw_platforms}. Subsequently, in Section~\ref{uc_srdrm}-\ref{uc_suim}, we explore a few intriguing research questions and elaborately discuss our research findings for the relevant use cases. In particular, we focus on their platform-specific expected performance margins and operational feasibility for use by underwater robots in real-time applications.         

\section{Hardware and Software Platforms}\label{hw_sw_platforms}
\begin{figure}[t]
\centering
    \subfigure[Aqua-8 AUV~\cite{dudek2007aqua} (MinneBot).]{
      \includegraphics[width=0.35\textwidth]{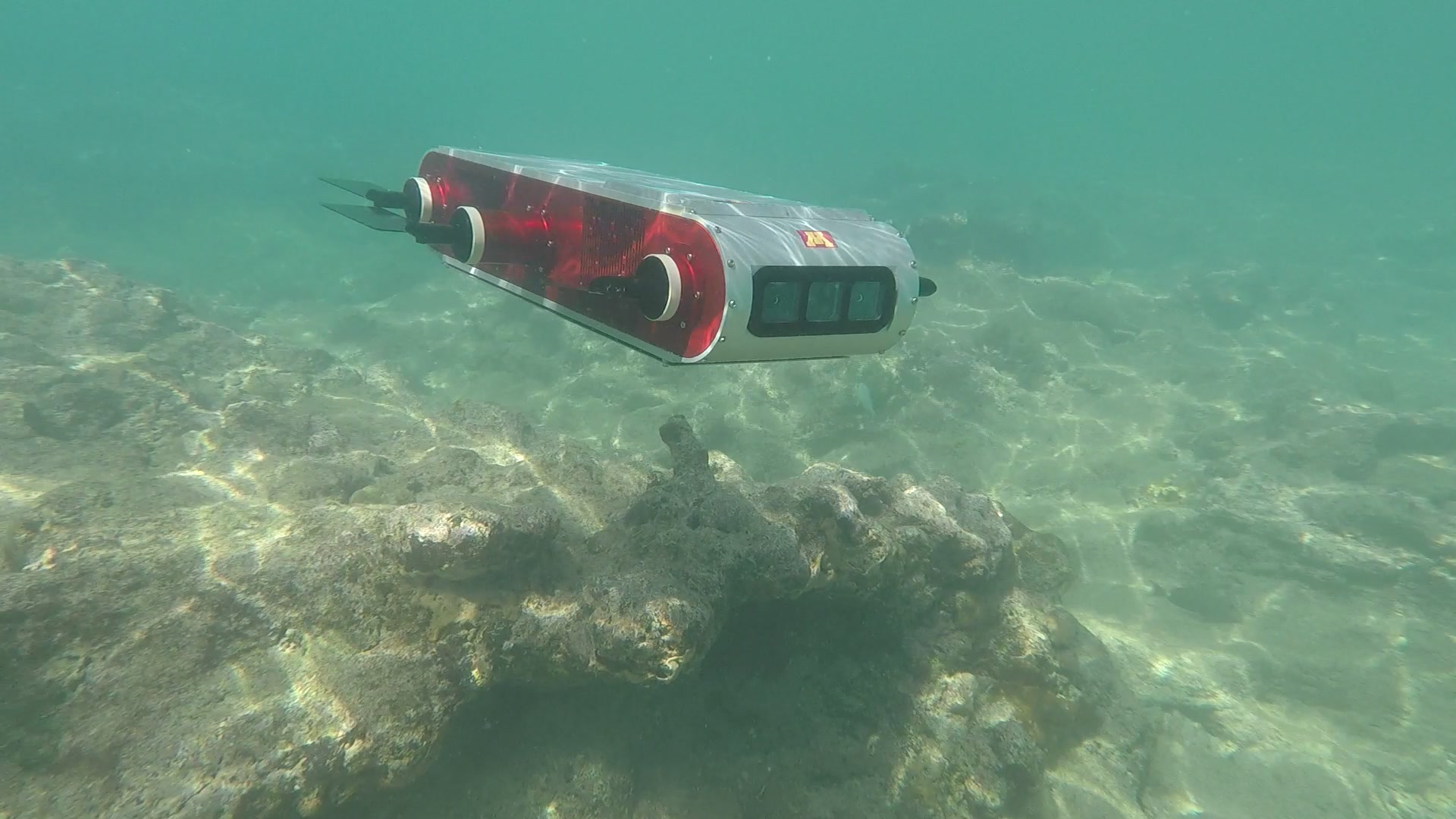} \label{fig_all_robots_a}
    }\subfigure[LoCO AUV~\cite{LoCOAUV}.]{
      \includegraphics[width=0.262\textwidth]{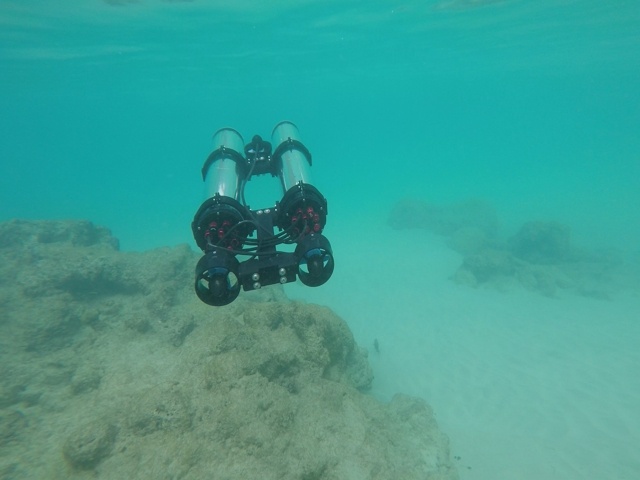} \label{fig_all_robots_b}
    }\subfigure[Trident ROV~\cite{trident}.]{
      \includegraphics[width=0.35\textwidth]{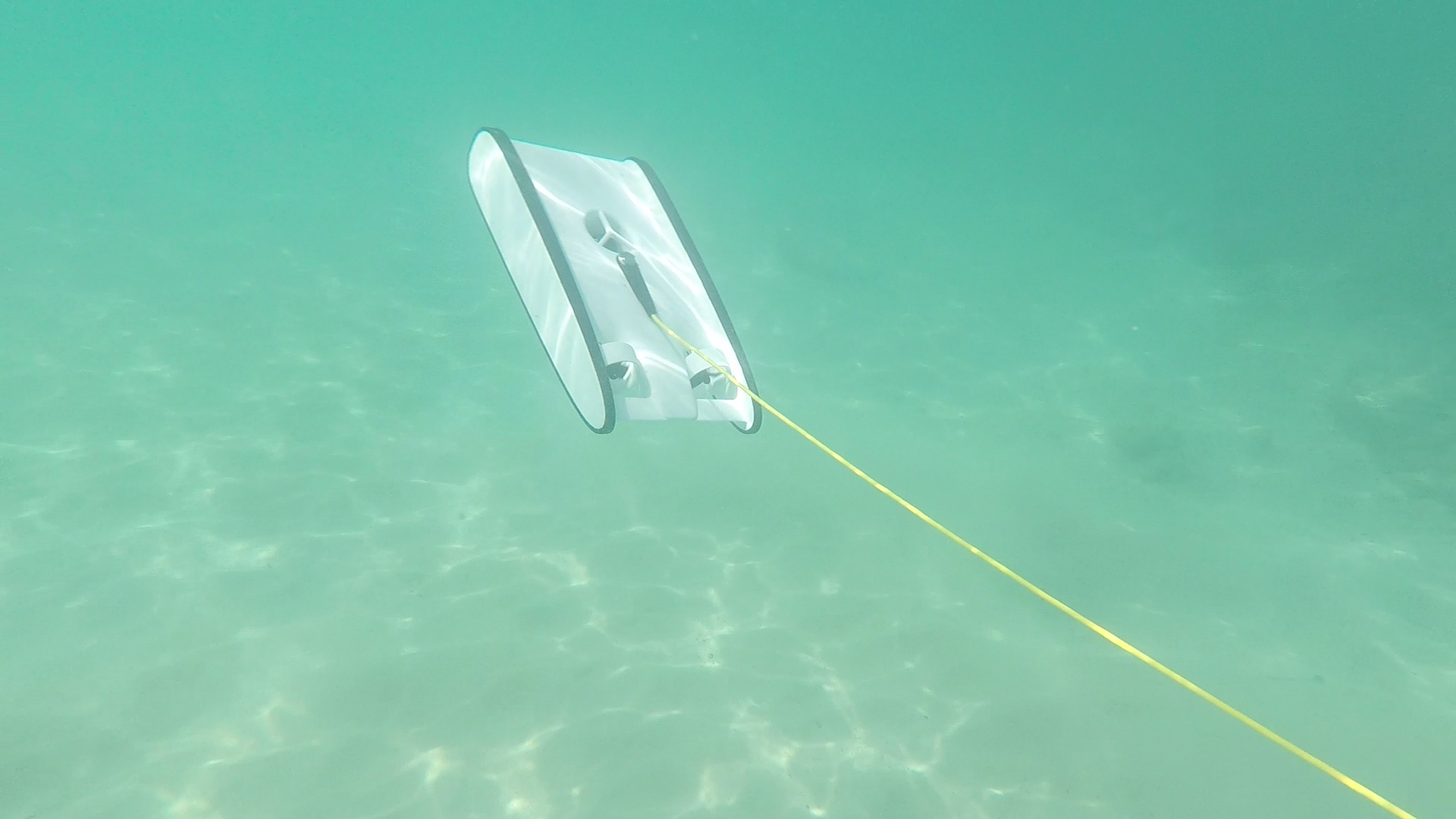} \label{fig_all_robots_c}
    }
    
    \subfigure[Stereo rig (built in-house).]{
      \includegraphics[width=0.285\textwidth]{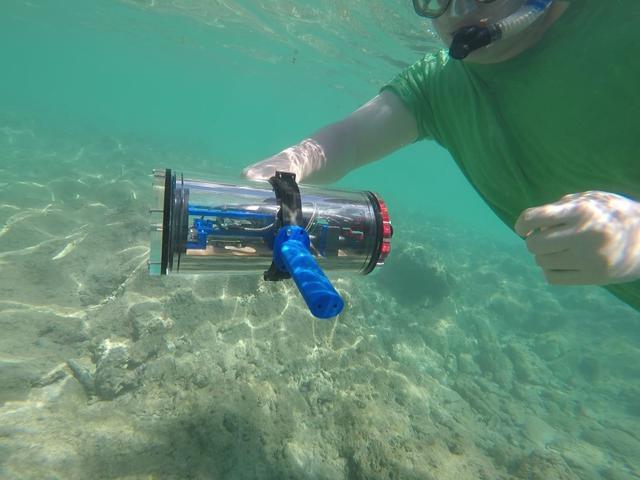} \label{fig_all_robots_d}
    }\subfigure[OpenROV~\cite{OpenROV}.]{
      \includegraphics[width=0.365\textwidth]{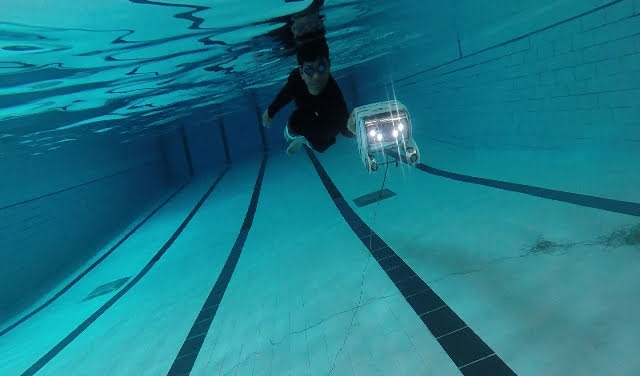} \label{fig_all_robots_e}
    }\subfigure[Single-board devices~\cite{Jetson}.]{
      \includegraphics[width=0.315\textwidth]{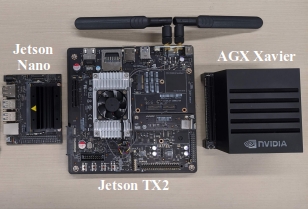} \label{fig_all_robots_f}
    }
    \vspace{-2mm}
    \caption{AUVs, ROVs, camera rigs, and embedded devices that are used in our field experiments, during data collection, and for benchmark performance analyses.}
    \label{fig_all_hws}
\end{figure}

As shown in Figure~\ref{fig_all_hws}, we use several underwater robots and embedded platforms for real-world field experimental validations and data collection. In most of our oceanic experiments, we use an Autonomous Underwater Vehicle (AUV) named MinneBot, an eighth generation Aqua robot~\cite{dudek2007aqua}. It uses six flippers for underwater propulsion with five degrees of freedom: surge, heave, pitch, roll, and yaw. It has three on-board cameras for perception: stereo vision on the front, and monocular vision on the back. Information passing through these cameras is handled by one of the two on-board computers, namely the vision stack. The other on-board computer, the control stack, is responsible for handling motor commands controlling the robot either teleoperated or autonomously. There is an inertial measurement unit (IMU) and a depth sensor as well.

A recent modification to Aqua led to the addition of NVIDIA\texttrademark{}~Jetson TX2, a single-board device, to facilitate fast visual computing. The LoCO AUV~\cite{LoCOAUV}, which is built in-house, also has one Jetson TX2 on-board. We use another single-board super-computer named NVIDIA\texttrademark{}~AGX Xavier for benchmark performance analyses. Besides, we use multiple NVIDIA\texttrademark{}~Jetson Nano devices in our stereo camera rigs that are also built in-house for data collection. Additionally, we use multiple Remotely Operated Vehicles (ROVs) named Trident~\cite{trident} and OpenROV~\cite{OpenROV} for data collection in human-robot cooperative experiments. We collect video footage of underwater scenes (\eg, reefs, subsea structures) using GoPros~\cite{gopro} and low-light cameras~\cite{lowlight} as well.

As mentioned earlier, our proposed methodologies are deployed on underwater robots (\eg, Aqua AUV, LoCO AUV) and validated through field experiments at the Caribbean sea~\cite{Bellairs} by mimicking real application scenarios such as subsea inspection, environmental monitoring, and autonomous explorations. For on-board implementation, we used standard libraries for OpenCV~\cite{opencv}, TensorFlow, and Keras~\cite{abadi2016tensorflow} integrated within a Robot Operating System (ROS) platform~\cite{ros}. To ensure reproducibility and rapid innovation, we released these modules, software packages, and datasets to the broader academic community at \url{http://irvlab.cs.umn.edu/resources}. We refer interested readers to the respective code repositories for specific implementation details of each algorithm; we also included the specifics in their respective discussions (in Chapter~\ref{diver_following}-\ref{svam}). In the following sections, we explore the usability and computational aspects specific to single-board platforms for a few particular case studies.        

\section{Is a Standalone Super-resolution Module Useful?}\label{uc_srdrm}
In Chapter~\ref{sesr}, we demonstrated how the single image super-resolution (SISR) capability~\cite{islam2019underwater} can be useful in detailed scene understanding and image synthesis by underwater robots. We also showed that for noisy and distorted visual data, simultaneous enhancement and super-resolution (SESR)~\cite{islam2020sesr} is computationally more feasible for real-time use. Nevertheless, 
\emph{in favorable visual conditions and when more computational resources are available, are standalone SISR modules useful?}. We now investigate this research question by a benchmark analysis of the state-of-the-art (SOTA) SISR models. 

\begin{figure}[t]
\centering
    \subfigure[A few instances sampled from the HR set; the HR images are of size $640\times480$.]{
        \includegraphics[width=0.98\textwidth]{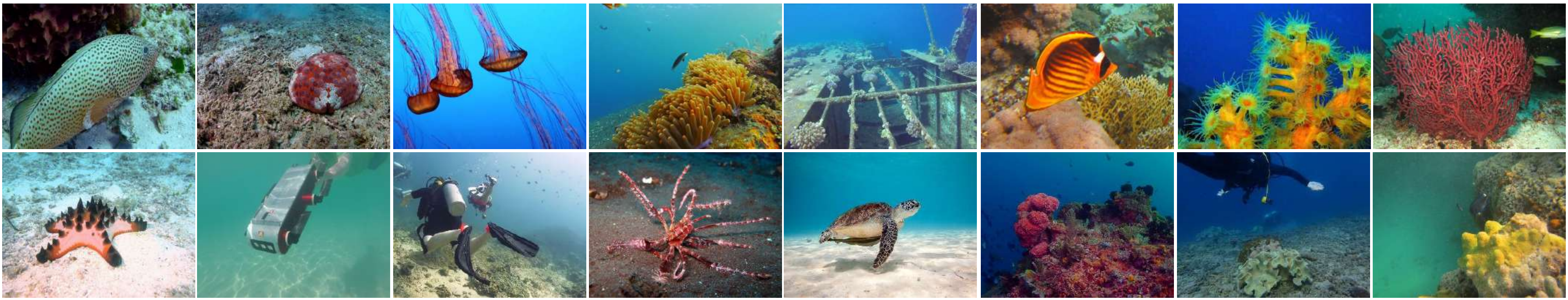}
        \label{fig:data_srdrm_a}
    }
    
    \subfigure[A particular HR ground truth image and its corresponding LR images are shown.]{
        \includegraphics[width=0.95\textwidth]{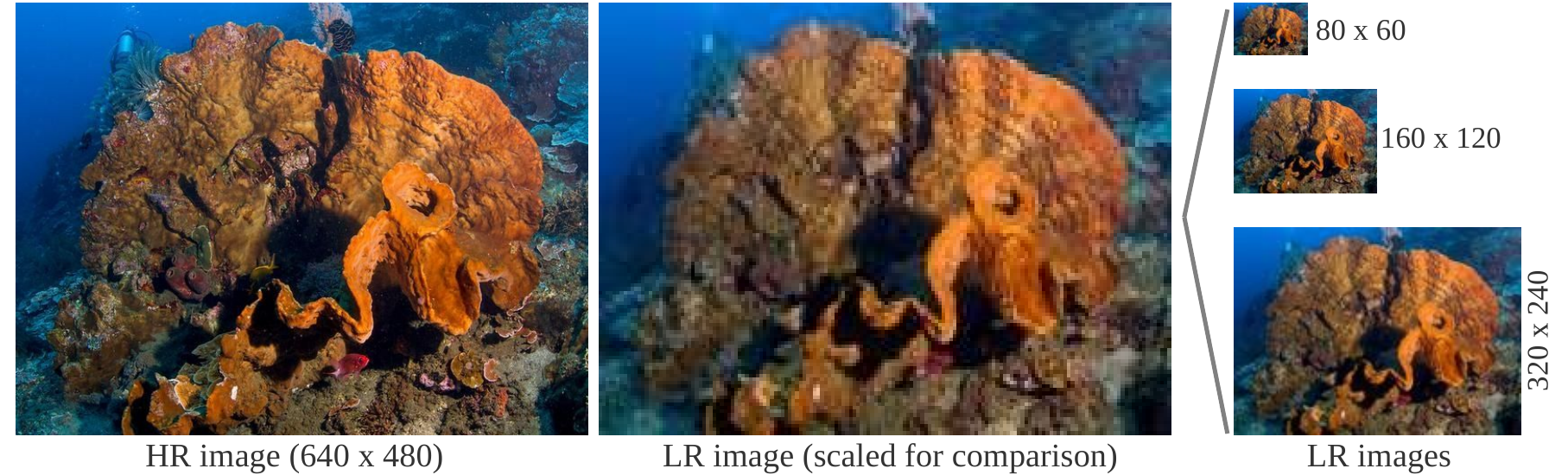}
        \label{fig:data_srdrm_b}
    }
    
    \caption{A few sample images from the USR-248 dataset are shown; it facilitates supervised training of $2\times$, $4\times$ and $8\times$ SISR models.}
    \label{fig:data_srdrm}
\end{figure}

\subsection{Data Preparation and SOTA Models}
First, we select a large collection of high-quality images of natural underwater scenes from our field experimental footage, and also from a few publicly available online resources such as YouTube\texttrademark{} and  Flickr\texttrademark{}. The compile dataset, named \textbf{USR-248}, contains $1060$ high-resolution (HR) images of size $640\times480$. We prepare three sets of low-resolution (LR) pairs of sizes $80\times60$, $160\times120$, and $320\times240$ by using standard Gaussian blurring and bicubic down-sampling with a $7\times7$ kernel. A separate set of $248$ images are provided for testing. A few samples from the USR-248 dataset are shown in Figure~\ref{fig:data_srdrm}. For benchmark evaluation, we consider three SOTA generative models named SRCNN~\cite{dong2015image}, SRResNet~\cite{ledig2017photo,yang2019deep}, and DSRCNN~\cite{mao2016image}; and three SOTA adversarial models named SRGAN~\cite{ledig2017photo}, ESRGAN~\cite{wang2018esrgan}, and EDSRGAN~\cite{lim2017enhanced}; these are originally proposed and benchmarked for terrestrial imagery.

\begin{figure}[t]
\centering
    \subfigure[Architecture of a Deep Residual Multiplier (DRM) block.]{
        \includegraphics[width=0.57\textwidth]{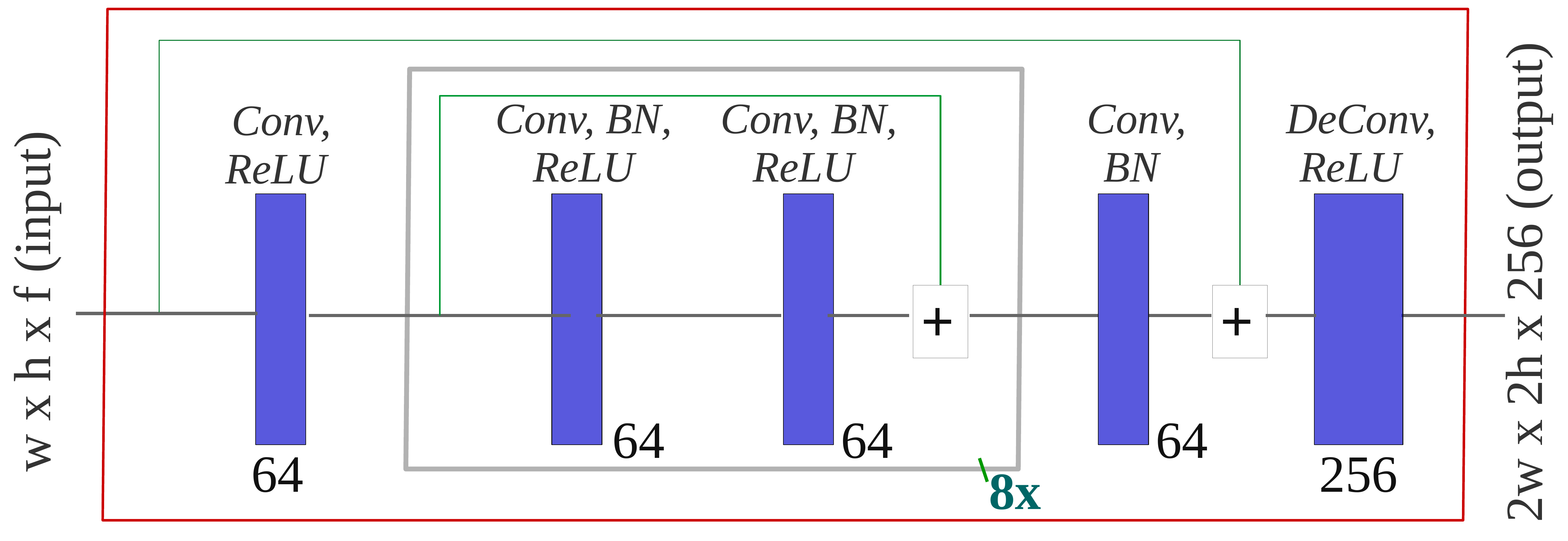}
        \label{fig:model_drm}
    }
    \vspace{1mm}
    
    \subfigure[Generator: one, two, or (up to) three DRM blocks in sequence.]{
    \includegraphics[width=0.67\textwidth]{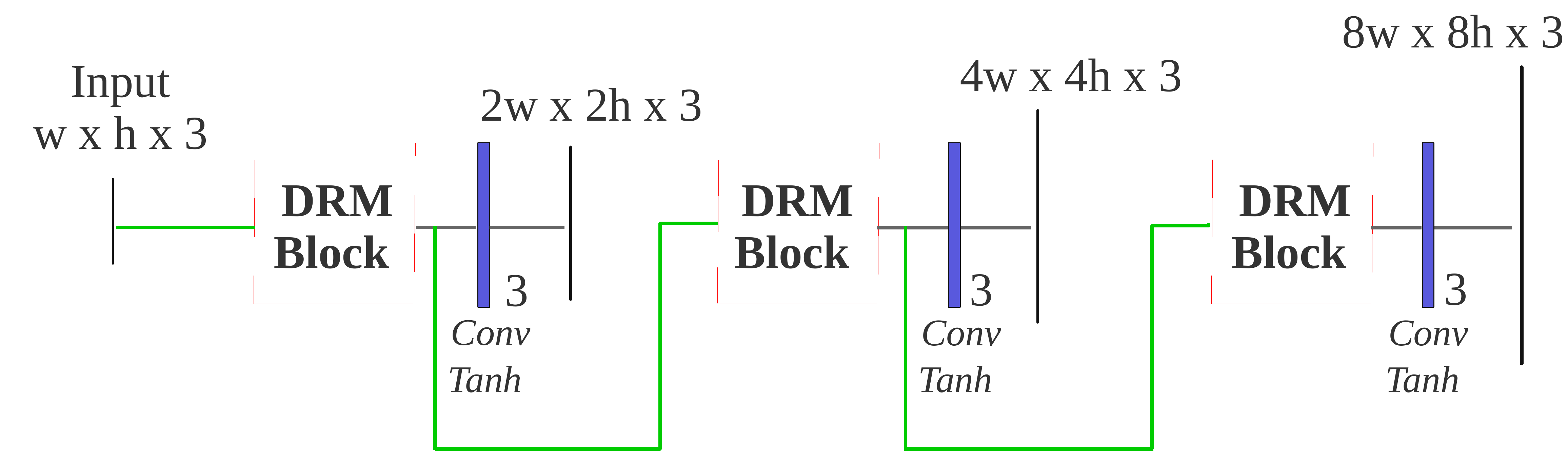}
    \label{fig:model_srdrm}
    }
    
    \subfigure[Discriminator: a Markovian PatchGAN~\cite{isola2017image} with nine layers and a patch-size of $40\times30$.]{
    \includegraphics[width=0.98\textwidth]{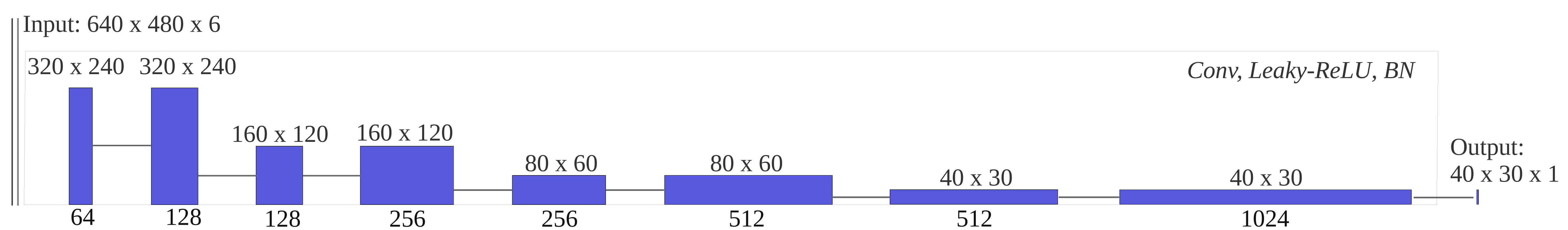}
    \label{fig:model_srdrmgan}
    }
    
\caption{Network architecture and parameter specifications of the proposed SRDRM model for underwater SISR~\cite{islam2019underwater}.}
\label{fig:model_sr}
\end{figure}

\subsection{SRDRM and SRDRM-GAN Architecture}
We also introduce a deep residual multiplier (\textbf{DRM}) module that can be utilized for learning $2\times$, $4\times$, or $8\times$ SISR by generative (\textbf{SRDRM}) or adversarial (\textbf{SRDM-GAN}) training. DRM consists of a convolutional ({\tt Conv}) layer, followed by $8$ repeated residual layers, then another {\tt Conv} layer, and finally a de-convolutional ({\tt DeConv}) layer for up-scaling.
%Each of the repeated residual layers (consisting of two {\tt conv} layers) is designed by following the principles outlined in the EDSR model~\cite{lim2017enhanced}. 
%Several choices of hyper-parameters, \eg, 
The number of filters in each layer, the use of {\tt ReLU} non-linearity~\cite{nair2010rectified}, and/or Batch Normalization (BN)~\cite{ioffe2015batch} are annotated in Figure~\ref{fig:model_drm}. The purpose of such computationally \textit{shallow} design is to gain fast inference on single-board platforms.

As a whole, DRM is only a $10$-layer residual network that learns $2\times$ spatial super-resolution of its input features. It uses a series of 2D convolutions of size $3\times3$ (in repeated residual block) and $4\times4$ (in the rest of the network) to learn this spatial interpolation from paired training data. For generative learning, SRDRM makes use of $n \in \{1, 2, 3\}$ DRM blocks in order to learn to generate $2^n\times$ HR outputs, as shown in Figure~\ref{fig:model_srdrm}. An additional {\tt Conv} layer with {\tt tanh} non-linearity~\cite{raiko2012deep} is added after the final DRM block to project the learned features to desired output shape. 
%Specifically, it generates a $2^nw \times 2^nh \times3$ output for an input of size $w\times h \times3$. 
For adversarial training, we use the same SRDRM model as a {generator} and employ a standard Markovian PatchGAN~\cite{isola2017image} for the {discriminator}, as illustrated in Figure~\ref{fig:model_srdrmgan}.
%nine {\tt conv} layers are used to transform a $640\times480\times6$ input (real and generated image) to a $40\times30\times1$ output that represents averaged \textit{validity} responses of the discriminator. At each layer, $3\times3$ convolutional filters are used with a strides of $2$, followed by a {\tt Leaky-ReLU} non-linearity~\cite{maas2013rectifier} and BN.

\subsection{Supervised SISR Learning}
For the paired supervision of SRDRM, we use the global similarity loss, perceptual loss, and content loss terms, which we already introduced in the SESR learning pipeline (see Section~\ref{loss_fun}). The standard adversarial objective is applied for SRDRM-GAN training with the additional min-max formulation~\cite{ledig2017photo}. We use TensorFlow libraries~\cite{abadi2016tensorflow} to implement the optimization pipelines of SRDRM and SRDRM-GAN. Both models are trained on the USR-248 dataset for up to $200$ epochs with a batch size of $4$, using two NVIDIA\textsuperscript{TM} GeForce GTX 1080 graphics cards. We also use TensorFlow implementations of all the SOTA models with default configurations and train them on the USR-248 dataset using the same hardware setup. We refer interested readers to the source repository~\cite{islam2019underwater} for more details about the SISR learning pipelines.  

\subsection{Underwater SISR Performance Analyses}
We now conduct quantitative evaluation of SRDRM, SRDRM-GAN, and the SOTA SISR models named SRCNN~\cite{dong2015image}, SRResNet~\cite{ledig2017photo,yang2019deep}, DSRCNN~\cite{mao2016image}, SRGAN~\cite{ledig2017photo}, ESRGAN~\cite{wang2018esrgan}, and EDSRGAN~\cite{lim2017enhanced}. We consider the standard metrics: Peak Signal-to-Noise Ratio (PSNR), Structural Similarity (SSIM), and Underwater Image Quality Measure (UIQM); their definitions and relevant details are provided in Appendix~\ref{ApenD}.

\begin{table}[H]

\centering
\caption{Comparison of averaged PSNR, SSIM, and UIQM scores for $2\times$/$4\times$/$8\times$ SISR on the USR-248 test set.}
\footnotesize
\vspace{3mm}
\begin{tabular}{l|c|c|c}
  \Xhline{2\arrayrulewidth}
   & $PSNR$ & $SSIM$ & $UIQM$ \\
  \Xhline{2\arrayrulewidth}
  SRResNet & $25.98$/$24.15$/$19.26$ & $0.72$/$0.66$/$0.55$ & $2.68$/$2.23$/$1.95$ \\ \hline
  SRCNN & $26.81$/$23.38$/$19.97$ & $0.76$/$0.67$/$0.57$ & $2.74$/$2.38$/$2.01$ \\ \hline
  DSRCNN & $27.14$/$23.61$/$20.14$ & $0.77$/$0.67$/$0.56$ & $2.71$/$2.36$/$2.04$ \\ \hline
  \textbf{SRDRM} & $28.36$/$24.64$/$21.20$ & $0.80$/$0.68$/$0.60$ & $2.78$/$2.46$/$2.18$ \\ \hline
  \textbf{SRDRM-GAN} & $28.55$/$24.62$/$20.25$ & $0.81$/$0.69$/$0.61$ & $2.77$/$2.48$/$2.17$ \\ \hline
  ESRGAN & $26.66$/$23.79$/$19.75$ & $0.75$/$0.66$/$0.58$ & $2.70$/$2.38$/$2.05$ \\ \hline
  EDSRGAN & $27.12$/$21.65$/$19.87$ & $0.77$/$0.65$/$0.58$ & $2.67$/$2.40$/$2.12$ \\ \hline
  SRGAN & $28.05$/$24.76$/$20.14$ & $0.78$/$0.69$/$0.60$ & $2.74$/$2.42$/$2.10$ \\ \Xhline{2\arrayrulewidth}
\end{tabular}
\vspace{-1mm}
\label{tab:psnr_ssim_srdrm}
\end{table}

\begin{figure}[t]
    \centering
    \includegraphics[width=0.99\linewidth]{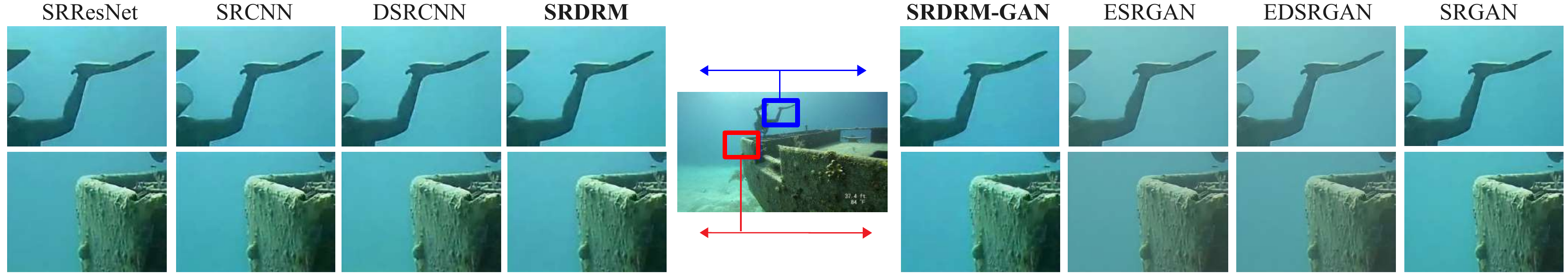}  \\
    \vspace{1mm}
    \includegraphics[width=0.99\linewidth]{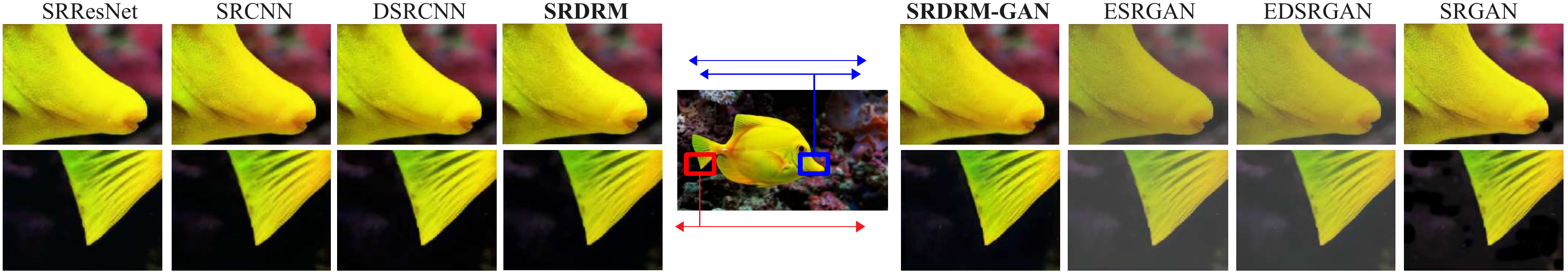}\\
    \vspace{1mm}
    \includegraphics[width=0.99\linewidth]{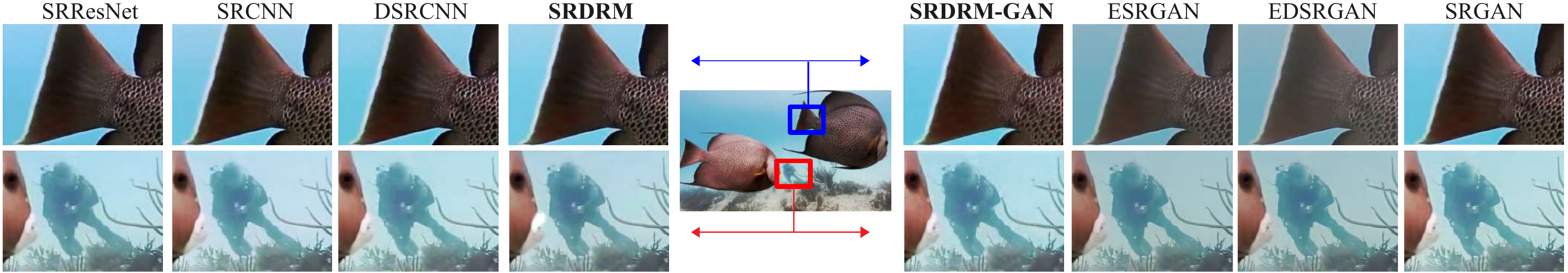}\\
    \vspace{-2mm}
    \caption{Qualitative performance comparison of SRDRM and SRDRM-GAN with SRCNN~\cite{dong2015image}, SRResNet~\cite{ledig2017photo,yang2019deep},  DSRCNN~\cite{mao2016image}, 
    SRGAN~\cite{ledig2017photo}, 
    ESRGAN~\cite{wang2018esrgan}, and EDSRGAN~\cite{lim2017enhanced}. (Best viewed at $400\%$ zoom)   
    }
    \label{fig:comp_srdrm}
\end{figure}

We evaluate all the SISR models on USR-248 test images and show the performance comparison in Table~\ref{tab:psnr_ssim_srdrm}. The results indicate that SRDRM-GAN, SRDRM, SRGAN, and SRResNet produce comparable values for PSNR and SSIM, and perform better than other models. SRDRM and SRDRM-GAN also produce higher UIQM scores than other models in comparison. 
In Figure~\ref{fig:comp_srdrm}, we provide a qualitative performance comparison with the SOTA models for the particular case of $4\times$ SISR. Here, we select multiple $160\times120$ patches on the test images containing interesting textures and objects of contrasting backgrounds. Then, we apply all the SISR models (trained on $4\times$ USR-248 data) to generate respective HR images of size $640\times480$. 

In the evaluation, we observe that SRDRM performs at least as well and often better compared to the generative models, \ie, SRResNet, SRCNN, and DSRCNN. Moreover, SRResNet and SRGAN are prone to inconsistent coloring and over-saturation in bright regions. 
On the other hand, ESRGAN and EDSRGAN often fail to restore the sharpness and global contrast. 
Furthermore, SRDRM-GAN generates sharper images and does a better texture recovery than SRDRM (and other generative models) in general. We postulate that the PatchGAN-based discriminator contributes to this, as it forces the generator to learn high-frequency local texture and style information~\cite{isola2017image}. 
%These statistics are consistent with our qualitative analysis. 

\subsection{Practical Feasibility}
The qualitative and quantitative results suggest that SRDRM, SRDRM-GAN, and SOTA models such as SRGAN and EDSRGAN provide good quality HR visualizations, particularly for $2\times$ and $4\times$ SISR. Moreover, the architectural simplicity of SRDRM facilitates significantly faster inference rates compared to the SOTA models, which generally provide less than $2$ FPS run-time on single-board devices such as NVIDIA\textsuperscript{TM} Jetson TX2. Hence, we particularly focus on the computational aspects (\ie, run-times and memory requirements) of SRDRM/SRDRM-GAN, and analyze their feasibility for near real-time use by underwater robots.

\begin{figure}[t]
    \centering
    \includegraphics[width=0.98\linewidth]{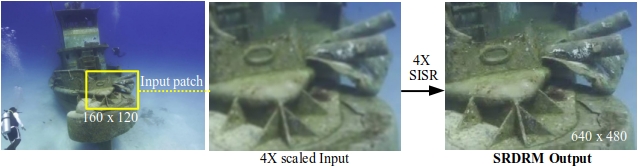}  
    \vspace{-2mm}
    \caption{A demonstration of using $4\times$ SRDRM on an image RoI specified by a human operator; such zoom-in capability is particularly useful for guiding ROVs in distant surveying, inspection, and mapping applications.  
    }
    \label{fig:srdrm_final}
\end{figure}

As we demonstrate in Table~\ref{tab:time_srdrm}, the memory requirement of SRDRM/SRDRM-GAN is only $3.5$--$8$\,MB and it runs at $6.86$--$7.11$\,FPS on NVIDIA\textsuperscript{TM} Jetson TX2. Essentially, a robot would require $140$--$146$ milliseconds to take a closer look at a LR image patch, which is admissible in near real-time applications. Even on FUnIE-GAN-enhanced images, we get $2\times$-$4\times$ SISR capability at over $5.4$-$5.54$ FPS rates, which is reasonable. Therefore, a combination of FUnIE-GAN and SRDRM can be used on ROVs, where a human operator (on the surface) would be able to zoom in a particular region of interest (RoI), as shown in Figure~\ref{fig:srdrm_final}. This is particularly useful for operating ROVs in applications such as target following, coral reef monitoring, seabed surveying, and mapping~\cite{hoegh2007coral,shkurti2012multi}. For autonomous exploration and scene understanding by AUVs, however, visual attention modeling is needed for using the SISR functionalities effectively.

\begin{table}[t]
\centering
\caption{Run-times and memory requirements of $2\times$ and $4\times$ SRDRM/SRDRM-GAN as a standalone model (\textbf{\textit{R}}) or in combination with the FUnIE-GAN model (\textbf{\textit{F}} + \textbf{\textit{R}}) are shown; all evaluations are performed on an NVIDIA\textsuperscript{TM} Jetson TX2 device.}
\vspace{3mm}
\footnotesize
\begin{tabular}{l||r|r|r|r}
  \Xhline{2\arrayrulewidth}
  \textbf{Model} &  \textbf{\textit{R}} ($2\times$)  & \textbf{\textit{R}} ($4\times$)  & \textbf{\textit{F}} + \textbf{\textit{R}} ($2\times$) & \textbf{\textit{F}} + \textbf{\textit{R}} ($4\times$) \\ \Xhline{2\arrayrulewidth}
  Inference time &  $140.6$\,ms  &  $145.7$\,ms & $180.2$\,ms & $185.1$\,ms \\
  Frame rate  &  $7.11$\,FPS &  $6.86$\,FPS & $5.54$\,FPS & $5.4$\,FPS \\
  %\hline
  Memory requirement &  $3.5$\,MB &  $8$\,MB & $21$\,MB & $25$\,MB \\ \Xhline{2\arrayrulewidth}
\end{tabular}
\vspace{1mm}
\label{tab:time_srdrm}
\end{table}

\section{How Does Visual Attention Modeling Help?}\label{uc_svam}
In Chapter~\ref{svam}, we presented a holistic approach to saliency-guided visual attention modeling (SVAM) for underwater robots. We also showed that the proposed SVAM-Net\textsuperscript{Light} model can be used as a general-purpose solution, irrespective of any high-level application-specific tasks of the robot. We now demonstrate
its effectiveness for two important use cases: salient RoI enhancement and effective SISR/SESR.

%robust identification of salient pixels 

\begin{figure*}[t]
    \centering
    \includegraphics[width=\linewidth]{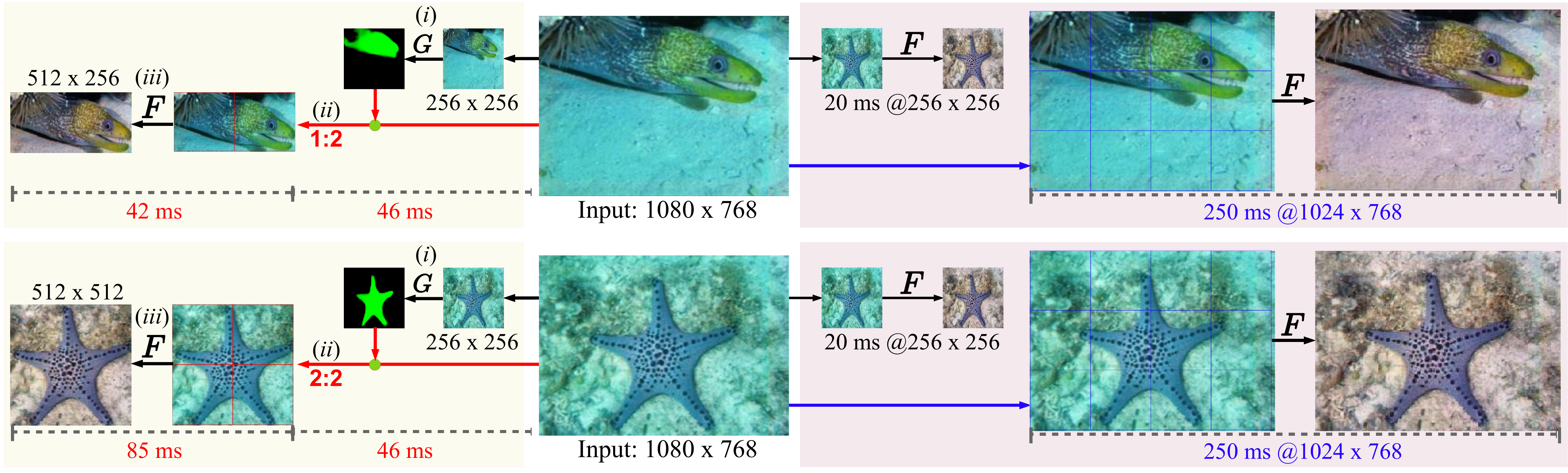}%
    \vspace{-8mm}
    \caption{Benefits of salient RoI enhancement are shown for two high-resolution input images. On the left: ($i$) SVAM-Net\textsuperscript{Light} (\textbf{\textit{G}})-generated saliency maps are used for RoI pooling, ($ii$) the salient RoIs are reshaped based on their area, and then ($iii$) FUnIE-GAN (\textbf{\textit{F}}) is applied on all $256\times256$ patches; the total processing time is $88$ ms for a $512\times256$ RoI (top image) and $131$ ms for a $512\times512$ RoI (bottom image). In comparison, as shown on the right, it takes $250$ ms to enhance the entire image.
    }
    \label{svam_enhancement}
\end{figure*}

\subsection{Salient RoI Enhancement}
AUVs and ROVs operating in noisy visual conditions frequently use various image enhancement models to restore the perceptual image qualities for improved visual perception~\cite{torres2005color,fabbri2018enhancing,roznere2019real}. However, these models typically have a low-resolution input reception, \eg, $224\times224$, $256\times256$, or $320\times240$. Hence, despite the robustness of SOTA underwater image enhancement models~\cite{islam2019fast,li2019underwater,fabbri2018enhancing}, their applicability to high-resolution robotic visual data is limited. For instance, the fastest available model, FUnIE-GAN~\cite{islam2019fast}, has an input resolution of $256\times256$, and it takes $20$ ms processing time to generate $256\times256$ outputs (on NVIDIA\textsuperscript{TM}~AGX Xavier). As a result, it eventually requires $250$ ms to enhance and combine all patches of a $1080\times768$ input image, which is too slow to be useful in near real-time applications. 

An effective alternative is to adopt a salient RoI enhancement mechanism to intelligently enhance useful image regions only. As shown in Figure~\ref{svam_enhancement}, SVAM-Net\textsuperscript{Light}-generated saliency maps are used to \textit{pool} salient image RoIs, which are then reshaped to convenient image patches for subsequent enhancement. Although this process requires an additional $46$ ms of processing time (by SVAM-Net\textsuperscript{Light}), it is still considerably faster than enhancing the entire image. As demonstrated in Figure~\ref{svam_enhancement}, we can save over $45\%$ processing time even when the salient RoI occupies more than half the input image. Overall, we found $45\%$-$75\%$ faster computation for RoI enhancement over the combined test images of EUVP~\cite{islam2019fast} and UFO-120~\cite{islam2020sesr} datasets.   
%In our experiments, we found up to $75\%$ and $67\%$ faster computation for RoI enhancement and RoI super-resolution tasks, respectively.

\begin{figure*}[t]
    \centering
    \includegraphics[width=\linewidth]{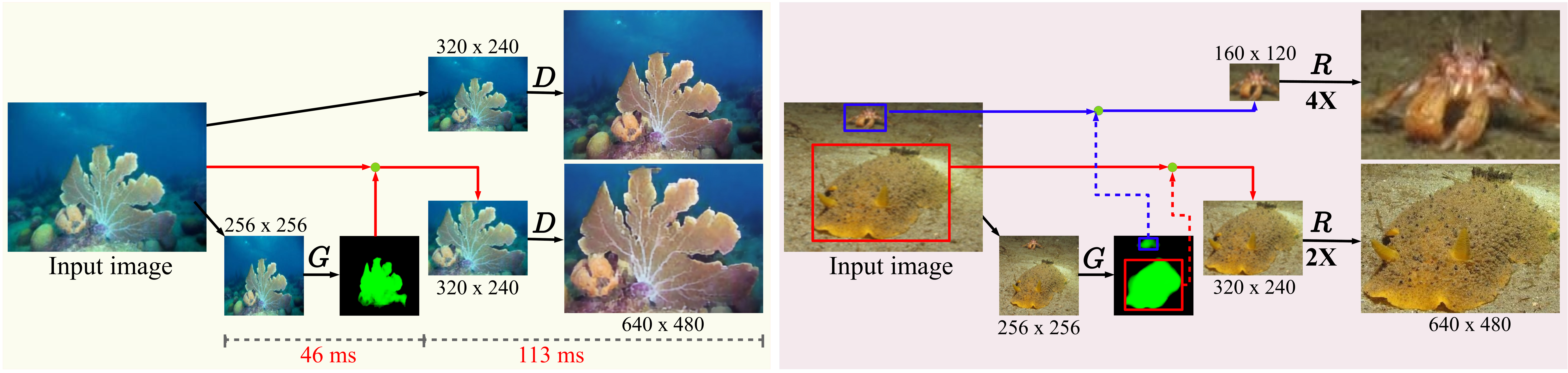}%
    \vspace{-8mm}
    \caption{Utility of visual attention modeling for effective image super-resolution is illustrated by two examples. As shown on the left, Deep SESR (\textbf{\textit{D}}) on the salient image RoI is potentially more useful for detailed perception rather than SESR on the entire image. Moreover, as seen on the right, SVAM-Net\textsuperscript{Light} (\textbf{\textit{G}})-generated saliency maps can also be used to determine the scale for super-resolution; here, we use $2\times$ and $4\times$ SRDRM (\textbf{\textit{R}}) on two salient RoIs based on their respective resolutions.
    }
    \label{svam_sesr}
\end{figure*}

\subsection{Effective SISR/SESR}
Since performing SISR/SESR on the entire input image is not computationally feasible, it is important to determine the salient image regions while discarding the background (waterbody) pixels. As shown in Figure~\ref{svam_sesr}, 
SVAM-Net\textsuperscript{Light} can be used to find the salient image RoIs for effective SISR or SESR. Moreover, the super-resolution scale (\eg, $2\times$, $3\times$, or $4\times$) can be readily determined based on the shape/pixel-area of a salient RoI. Hence, a class-agnostic SOD module is of paramount importance to gain the operational benefits of image super-resolution, especially in vision-based tasks such as tracking fast-moving targets~\cite{zhu2020saliency,shkurti2017underwater,chuang2014tracking} and autonomous surveying~\cite{modasshir2020enhancing,manderson2018vision,ModasshirFSR2019}. For its computational efficiency and robustness, SVAM-Net\textsuperscript{Light} is an ideal choice to be used alongside a SISR/SESR module in practical applications.

\section{Is Class-aware Saliency Estimation Feasible?}\label{uc_suim}
We demonstrated that SVAM-Net~\cite{islam2020svam} and other class-agnostic SOD models are useful for general-purpose visual attention modeling by AUVs. On the other hand, pixel-level detection of specific objects of interest is important for detailed scene understanding in many application-specific tasks~\cite{islam2020suim}. Such \textit{class-aware} saliency estimation relates most closely to the research problem of semantic image segmentation~\cite{garcia2017review,chen2017deeplab,badrinarayanan2015segnet}. Despite potential use cases, there are no underwater datasets to facilitate large-scale training and benchmark evaluation of semantic segmentation models. The existing large-scale annotated data and relevant methodologies are tied to specific applications such as coral-reef classification and coverage estimation~\cite{beijbom2012automated,alonso2019coralseg,VAIME}, fish detection and segmentation~\cite{ravanbakhsh2015automated,chuang2011automatic}, etc. Other datasets contain either binary annotations for salient foreground pixels~\cite{islam2020sesr} or semantic labels for very few object categories (\eg, sea-grass, rocks/sand, etc.)~\cite{zhou2019underwater}. Therefore, the large-scale learning-based semantic segmentation methodologies for underwater imagery are not explored in depth in the literature.  
%We now explore feasible solutions to this problem for real-time underwater robot vision. 

\begin{figure}[h]
\centering
    \subfigure[Four comparisons are shown for \textit{class-agnostic} SOD (by SVAM-Net~\cite{islam2020svam}) and \textit{class-aware} saliency estimation by semantic segmentation of underwater imagery; the object categories and labels (from the SUIM dataset~\cite{islam2020suim}) are provided on the right.]{
        \includegraphics[width=0.8\textwidth]{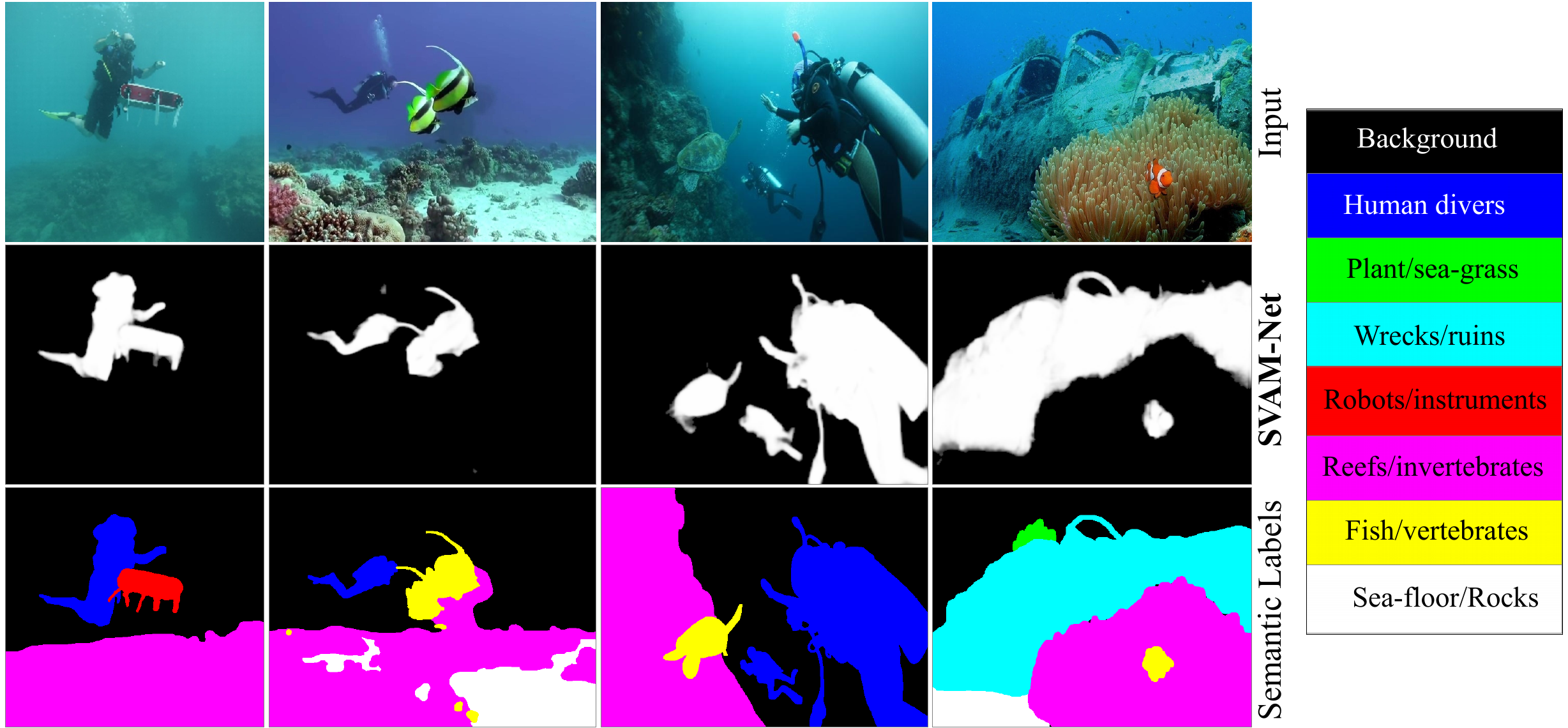}
        \label{fig:suim_intro_a}
    }%
    
    \subfigure[Pixel-level detection of specific object categories: humans, robots, fish, and wrecks/ruins are shown; the semantic masks are overlayed on the bottom row images.]{
    \includegraphics[width=0.8\textwidth]{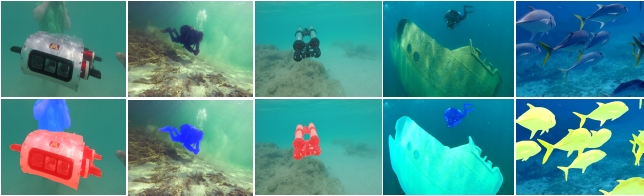}
    \label{fig:suim_intro_b}
    }%
    
\caption{A few demonstrations for class-aware saliency estimation by semantic segmentation of underwater scenes~\cite{islam2020suim}.}
\label{fig:suim_intro}
\end{figure}%

We attempt to address these limitations by compiling a large-scale annotated dataset for semantic Segmentation of Underwater Imagery (\textbf{SUIM})~\cite{islam2020suim} by considering object categories for fish, reefs, aquatic plants, and wrecks/ruins, which are of primary interest in underwater exploration and surveying tasks~\cite{girdhar2014autonomous,shkurti2012multi,bingham2010robotic}. As shown in Figure~\ref{fig:suim_intro}, it also contains pixel annotations for human divers, robots/instruments, and sea-floor/rocks; these are major objects of interest in human-robot cooperative applications~\cite{islam2018understanding,sattar2009vision}. Subsequently, we conduct a benchmark evaluation of several SOTA semantic segmentation models and compare their performance on the proposed SUIM dataset. We also design a novel encoder-decoder model named \textbf{SUIM-Net}, which offers much faster run-time than SOTA models while achieving competitive performance margins. We further analyze its computational aspects and practical utility in robotic applications over a series of quantitative and qualitative experiments.

\subsection{SUIM Dataset}
\begin{figure*}[t]
    \centering
    \vspace{-1mm}
    \includegraphics[width=0.98\linewidth]{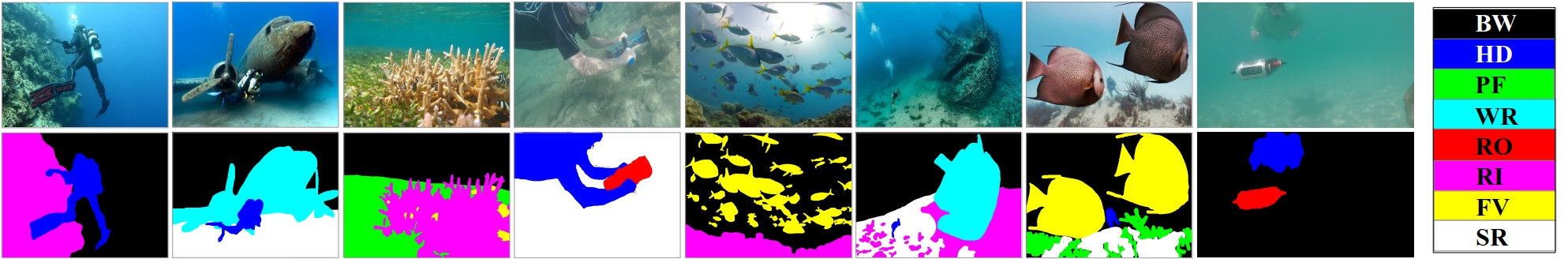} 
    \vspace{-3mm}
    \caption{A few sample images and corresponding pixel-annotations from the SUIM dataset are shown on the top, and bottom row, respectively.}
    \label{fig:data_suim}
\end{figure*}

We consider the following object categories for semantic labeling in the SUIM dataset: {$i$)} waterbody background ({\tt BW}), {$ii$)} human divers ({\tt HD}), {$iii$)} aquatic plants/flora ({\tt PF}), {$iv$)} wrecks/ruins ({\tt WR}), {$v$)} robots and instruments ({\tt RO}), {$vi$)} reefs and other invertebrates ({\tt RI}), {$vii$)} fish and other vertebrates ({\tt FV}), and {$viii$)} sea-floor and rocks ({\tt SR}). 
As depicted in Table~\ref{tab:my_label}, we use 3-bit binary RGB colors to represent these eight object categories in the image space; a few annotated samples are shown in Figure~\ref{fig:data_suim}.

\begin{table}[t]
    \centering
    \caption{The object categories and corresponding color codes in the SUIM dataset.}
    \vspace{2mm}
    \footnotesize
    \begin{tabular}{l||l|c}
      \Xhline{2\arrayrulewidth}
      \textbf{Object category} & {\textbf{RGB color}} & {\textbf{Code (value)}} \\ \Xhline{2\arrayrulewidth}
      Background (waterbody) & Black & {\tt BW} ($000$) \\ \hline
      Human divers & Blue & {\tt HD} ($001$) \\ \hline
      Aquatic plants and sea-grass & Green  & {\tt PF} ($010$) \\ \hline
      Wrecks or ruins & Sky  & {\tt WR} ($011$) \\ \hline
      Robots (AUVs/ROVs/instruments) & Red & {\tt RO} ($100$) \\ \hline
      Reefs and invertebrates & Pink  & {\tt RI} ($101$) \\ \hline
      Fish and vertebrates & Yellow & {\tt FV} ($110$) \\ \hline
      Sea-floor and rocks  & White & {\tt SR} ($111$) \\
      \Xhline{2\arrayrulewidth}
    \end{tabular}
    \label{tab:my_label}
\end{table}

The SUIM dataset has $1525$ RGB images for training and validation; another $110$ test images are provided for benchmark evaluation of semantic segmentation models. The images 
%are of various resolutions, \eg, $1906 \times 1080$, $1280 \times 720$, $640 \times 480$, and $256 \times 256$. They 
are carefully chosen from a large pool of samples collected during oceanic explorations and human-robot cooperative experiments in several locations of various water types. We also utilize a few images from EUVP~\cite{islam2019fast}, USR-248~\cite{islam2019srdrm}, and UFO-120~\cite{islam2020sesr} datasets, which we previously proposed for image enhancement and super-resolution problems. The images are chosen to accommodate a diverse set of natural underwater scenes and various setups for human-robot collaborative experiments. All images are pixel-annotated by seven human participants independently; the guidelines discussed in~\cite{ConfRef1} and~\cite{ConfRef2} are followed for classifying potentially confusing objects such as plants/reefs, vertebrates/invertebrates, etc.

\begin{figure}[t]
\centering
    \subfigure[Architecture of an RSB: the skip-connection can be either fed from an intermediate {\tt Conv} layer (by setting \textit{skip=0}) or from the input (by setting \textit{skip=1}) for local residual learning.]{
        \includegraphics[width=0.7\textwidth]{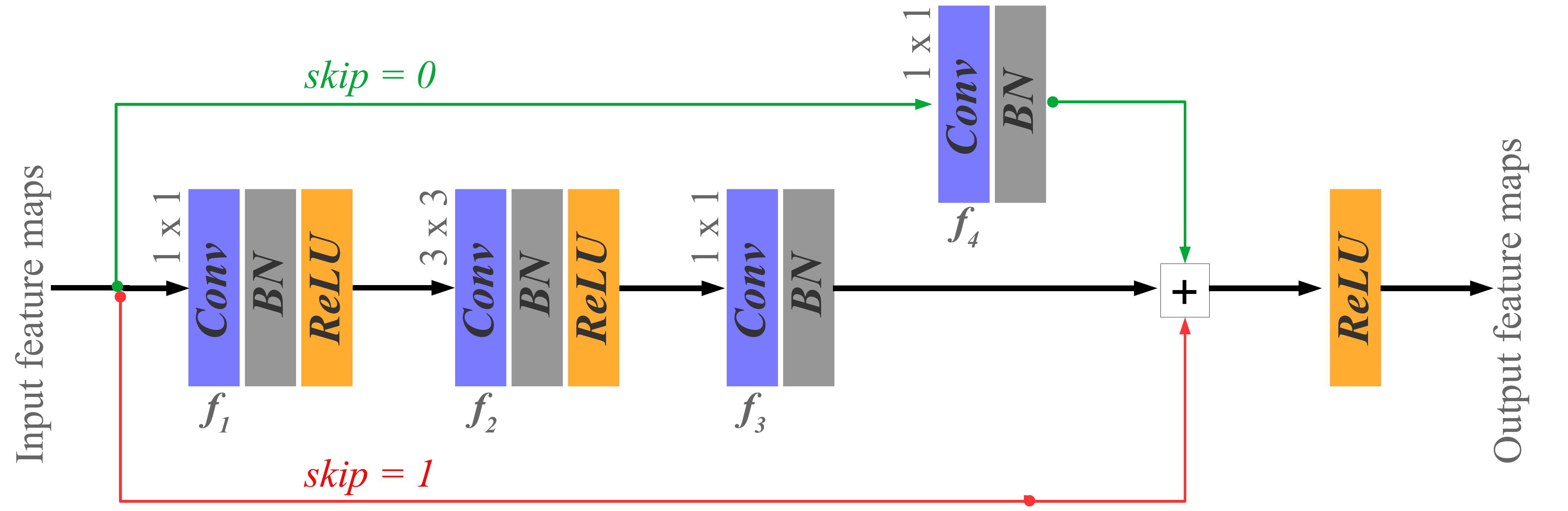}
        \label{fig:model_suim_a}
    }%
    
    \subfigure[The end-to-end architecture of SUIM-Net$_{RSB}$: three composite layers of encoding is performed by a total of seven RSBs, followed by three decoder blocks with mirrored skip-connections.]{
    \includegraphics[width=0.75\textwidth]{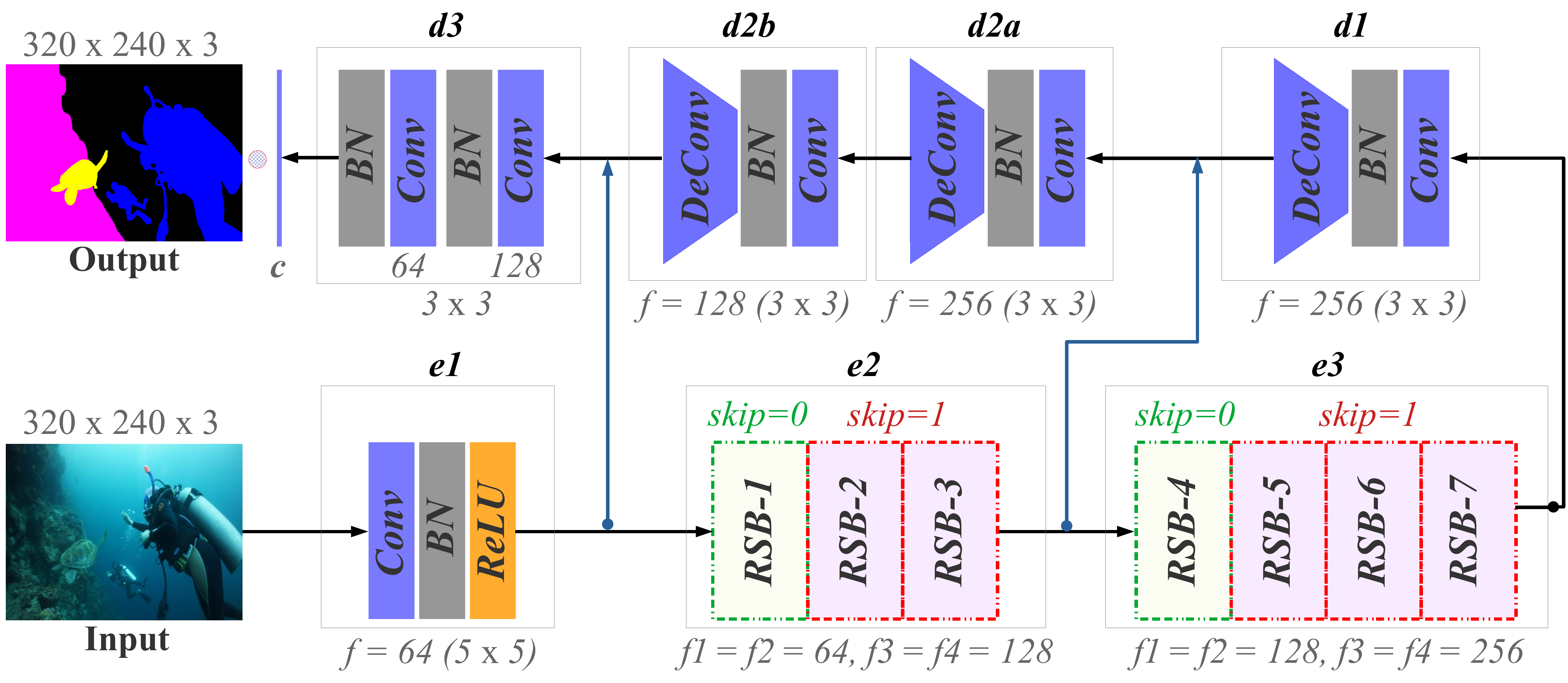}
    \label{fig:model_suim_b}
    }%
    
\caption{Detailed specification of the SUIM-Net$_{RSB}$ model~\cite{islam2020suim}.}
\label{fig:model_suim}
\end{figure}%

\subsection{SUIM-Net Model}
The proposed SUIM-Net model incorporates a fully convolutional encoder-decoder architecture with skip connections between mirrored composite layers. As shown in Figure~\ref{fig:model_suim_a}, the base model embodies residual learning~\cite{he2016deep} with an optional skip layer named RSB (residual skip block). Each RSB consists of three convolutional ({\tt Conv}) layers, each followed by Batch Normalization ({\tt BN})~\cite{ioffe2015batch} and {\tt ReLU} non-linearity~\cite{nair2010rectified}. As Figure~\ref{fig:model_suim_b} shows, two sets of RSBs are used in the second and third encoder layers; the number of filters, feature dimensions, and other parameters are marked in the figure. The encoder network extracts $256$ feature maps from input RGB images; the encoded features are then exploited by three sequential decoder layers. Each decoder layer consists of a {\tt Conv} layer that receives skip-connections from respective conjugate encoder layer; it is followed by {\tt BN} and a de-convolutional ({\tt DeConv}) layer~\cite{zeiler2010deconvolutional} for spatial up-sampling. The final {\tt Conv} layer generates the per-channel binary pixel labels for each object category, which is then post-processed for visualizing in the RGB space.

\begin{figure*}[h]
    \centering
    \includegraphics[width=0.75\linewidth]{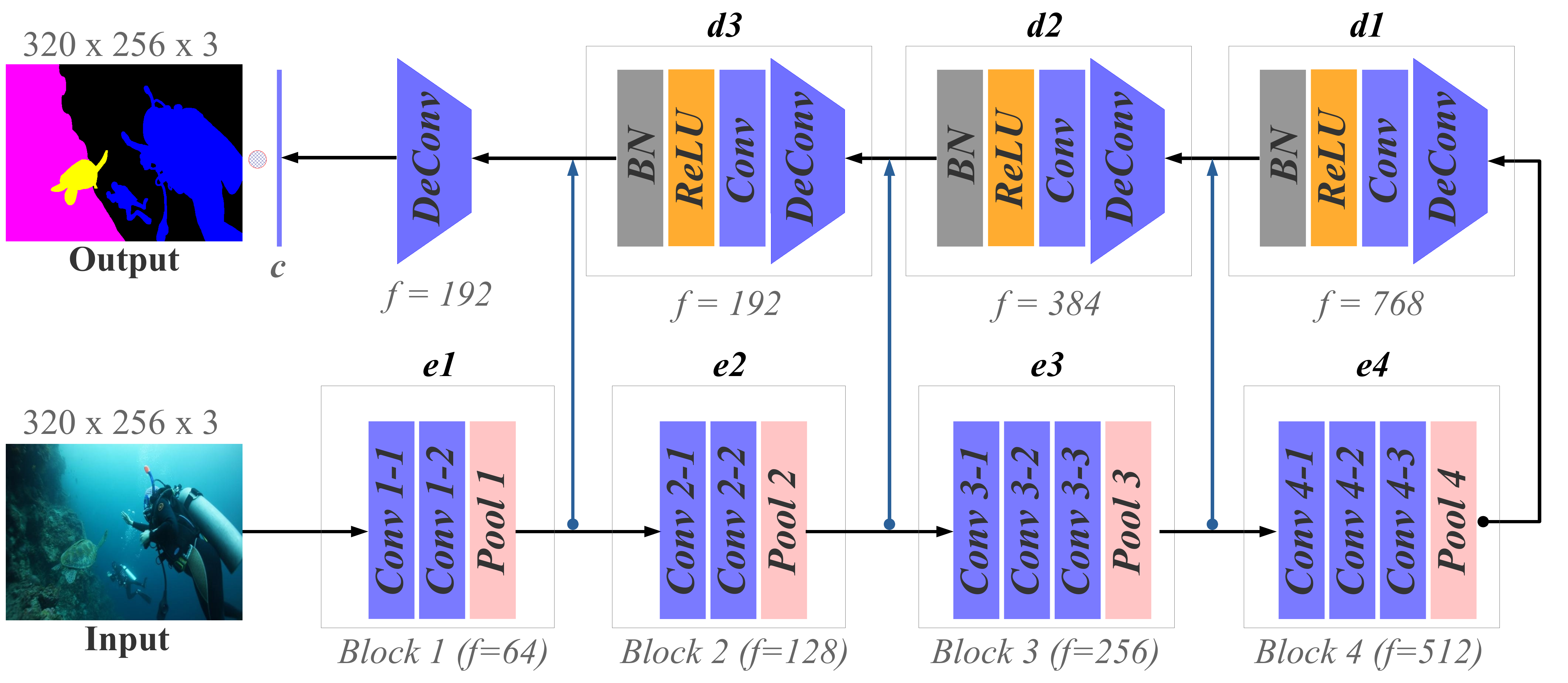} 
    \vspace{-3mm}
    \caption{The end-to-end architecture of SUIM-Net$_{VGG}$: first four blocks of a pre-trained VGG-16~\cite{simonyan2014very} model are used for encoding, followed by three mirrored decoder blocks and a {\tt DeConv} layer.}
    \label{fig:model_suim_c}
\end{figure*}

As evident from Figure~\ref{fig:model_suim_b}, we try to combine the benefits of skip-connections~\cite{ronneberger2015u} and residual learning~\cite{he2016deep,badrinarayanan2015segnet} in the \textbf{SUIM-Net$_{RSB}$} model. Our motive for such design is to ensure real-time inference while achieving a reasonable segmentation performance. 
On the other hand, we focus more on improved performance in the \textbf{SUIM-Net$_{VGG}$} model, which utilizes $12$ encoding layers of a pre-trained VGG-16 network~\cite{simonyan2014very}; detailed architecture is illustrated in Figure~\ref{fig:model_suim_c}. 
We present the experimental analysis of SUIM-Net and other SOTA models in the following sections.

\subsection{Training Pipeline and SOTA Models}
We formulate the problem as learning a mapping from input domain of natural underwater images to its target semantic labeling. The end-to-end training is supervised by the standard cross-entropy loss~\cite{zhang2018generalized,badrinarayanan2015segnet} which evaluates the discrepancy between predicted and ground truth pixel labels. We use TensorFlow libraries~\cite{abadi2016tensorflow} to implement the optimization pipeline; a Linux host with one Nvidia\texttrademark{ }GTX 1080 graphics card is used for training. Adam optimizer~\cite{kingma2014adam} is used for the global iterative learning with a rate of $0.0001$ and a momentum of $0.5$. Moreover, we used the standard Keras libraries for data augmentation; the specific parameters are: rotation range of $0.2$; width shift, height shift, shear, and zoom range of $0.05$; horizontal flip is enabled; and the rest of the parameters are left as default. We consider the following SOTA models for performance evaluation on the SUIM dataset: 
$i$) FCN8~\cite{long2015fully} with two variants of base model: vanilla CNN (FCN8$_{CNN}$) and VGG-16 (FCN8$_{VGG}$), 
$ii$)  SegNet~\cite{badrinarayanan2015segnet} with two variants of base model: vanilla CNN (SegNet$_{CNN}$) and ResNet-50 (SegNet$_{ResNet}$), $iii$)  UNet~\cite{ronneberger2015u} with two variants of input: grayscale images (UNet$_{GRAY}$) and RGB images (UNet$_{RGB}$), $iv$) pyramid scene parsing network~\cite{zhao2017pyramid} with MobileNet~\cite{howard2017mobilenets} as the base model (PSPNet$_{MobileNet}$), and $v$) DeepLab$_{V3}$~\cite{chen2017rethinking}. We use TensorFlow implementations of all these models and train them on the SUIM datasest using the same hardware setup; further information can be found in their source repositories.

\subsection{Benchmark Evaluation and Performance Analyses}
We conduct experimental evaluation of SUIM-Net and other SOTA models for the following two training configurations: 
\begin{itemize}
    \item Semantic segmentation with five major object categories (see Table~\ref{tab:my_label}): {\tt HD}, {\tt WR}, {\tt RO}, {\tt RI}, and {\tt FV}; the rest are considered as background, \ie, {\tt BW}={\tt PF}={\tt SR}={\tt (000)\textsubscript{RGB}}. Each model is configured for five channels of output, one for each category. The predicted separate pixel masks are combined to RGB masks for visualization.

    \item Single-channel saliency prediction: the ground truth intensities of {\tt HD}, {\tt RO}, {\tt FV}, and {\tt WR} pixels are set to $1.0$, and the rest are set to $0.0$. The output is thresholded and visualized as binary images.    
\end{itemize}
We compare the performance of all the models based on standard metrics that evaluate region similarity and contour accuracy~\cite{garcia2017review,perazzi2016benchmark}. The region similarity metric quantifies the correctness of predicted pixel labels by using the notion of `dice coefficient' aka $\mathcal{F}$ score. Besides, contour accuracy represents the object boundary localization performance by the mean intersection over union ($mIOU$). Their definitions and relevant details are provided in Appendix~\ref{ApenObj}.

\begin{table*}[ht]
\centering
\caption{Quantitative performance comparison for semantic segmentation and saliency prediction on the SUIM dataset.}
\footnotesize
%\scriptsize
\vspace{2mm}
\begin{tabular}{c|l||c|c|c|c|c|c||c}
  \Xhline{2\arrayrulewidth}
  & \textbf{Model} & {\tt HD} & {\tt WR} & {\tt RO} & {\tt RI} & {\tt FV} & \textbf{Combined} & Saliency  \\ \Xhline{2\arrayrulewidth}
  % change this number 7 when adding/deleting rows
  \parbox[t]{2mm}{\multirow{8}{*}{\rot{$\mathcal{F}$ ($\rightarrow$)}}} & FCN8$_{CNN}$ & $76.34$ & $70.24$  & $39.83$  & $61.65$ & $76.24$ & $64.86$ & $75.62$ \\ %\hline
  
  & FCN8$_{VGG}$ & {$89.10$} & {$82.03$} & {$74.01$} & {$79.19$} & {$90.46$}
 & {$82.96$} &  {$89.63$} \\ %\cline{2-9} 
  
  & SegNet$_{CNN}$ & $59.60$ & $41.60$  & $31.77$ & $41.88$ & $60.08$ & $46.97$ & $56.96$  \\ %\hline
  
  & SegNet$_{ResNet}$ & $80.52$  & {$77.65$}  & $62.45$ & {$82.30$} & {$91.47$} & $76.88$ & {$86.88$}  \\ %\hline

  & UNet$_{GRAY}$ &  $85.47$ & {$79.77$} & $60.95$ & $69.95$ & $84.47$ & $75.12$ & $83.96$ \\ %\hline  \\ 
  
  & UNet$_{RGB}$ & {$89.60$} & {$86.17$}  & {$68.87$}  & {$79.24$} & {$91.35$} & {$83.05$} & {$89.99$}  \\ %\hline

  & PSPNet$_{MobileNet}$ &  {$80.21$} & $70.94$  & {$72.04$} & $72.65$ & $79.19$ & $76.01$ & $78.42$ \\ %\hline

  & DeepLab$_{V3}$ &   {$89.68$} & {$77.73$}  & {$72.72$}   & {$78.28$}  & {$87.95$}  & {$81.27$} & {$85.94$} \\ %\hline
  
  & \textbf{SUIM-Net$_{RSB}$} & {$89.04$} & $65.37$ & {$74.18$} & $71.92$ & $84.36$ & {$78.86$} & $81.36$ \\ %\hline  \\ 
  
  & \textbf{SUIM-Net$_{VGG}$} & {$93.56$} & {$86.02$} & {$78.06$} & 
  {$83.49$} & {$93.73$} & {$86.97$} & 
  {$91.91$} \\ %\hline  \\ 
  \hline
  
    % change this number 7 when adding/deleting rows
  \parbox[t]{2mm}{\multirow{8}{*}{\rot{$mIOU$ ($\rightarrow$)}}} 
  & FCN8$_{CNN}$ & $67.27$  &$81.64$   & $36.44$ &$78.72$  & $70.25$  
  & $66.86$ & $75.63$ \\ %\hline

  & FCN8$_{VGG}$ & {$79.86$} & {$85.77$} & {$65.05$} & {$85.23$} & {$81.18$} & {$79.42$} & {$85.22$} \\ %\cline{2-9} 
 
  & SegNet$_{CNN}$ & $62.76$ & $66.75$  & $36.63$ & $63.46$ & $62.48$ & $58.42$ & $65.90$ \\ %\hline
  
  & SegNet$_{ResNet}$ & $74.00$ & $82.68$ & $58.63$ & {$89.61$} & {$82.96$} & $77.58$ & {$83.09$} \\ %\hline
  
  & UNet$_{GRAY}$ & {$78.33$} & {$85.14$} & $57.25$ & $79.96$ & $78.00$ &  $75.74$ & $82.77$ \\ %\hline 

  & UNet$_{RGB}$ & {$81.17$} &  {$87.54$} & $62.07$  & {$83.69$} & {$83.83$} & {$79.66$} & {$85.85$}  \\ %\hline
  
  & PSPNet$_{MobileNet}$ & $75.76$ & {$86.82$} & {$72.66$}  & {$85.16$} & $74.67$ & $77.41$ & $80.87$ \\ %\hline

  & DeepLab$_{V3}$ & {$80.78$}  & {$85.17$}  & {$66.03$} & {$83.96$} & {$79.62$}  & {$79.10$} & {$83.55$} \\ %\hline
  
  & \textbf{SUIM-Net$_{RSB}$} & {$81.12$}  & $80.68$ & {$65.79$} & {$84.90$} & $76.81$ & {$77.77$}& $80.86$  \\ %\hline  \\
  
  & \textbf{SUIM-Net$_{VGG}$} & {$85.09$}  & {$89.90$} & {$72.49$} & {$89.51$} & 
  {$83.78$} & {$84.14$} & 
  {$87.67$} \\ %\hline  \\
  \Xhline{2\arrayrulewidth}
  
\end{tabular}
\label{tab:quant_suim}
%\vspace{-1mm}
\end{table*}

We present the quantitative results in Table~\ref{tab:quant_suim}. It compares the $\mathcal{F}$ and \textit{mIOU} scores for semantic segmentation of each object category; it also compares the respective scores for saliency prediction. The results suggest that UNet$_{RGB}$, FCN8$_{VGG}$, and DeepLab$_{V3}$ generally perform better than other models. In particular, they achieve the top three $\mathcal{F}$ and \textit{mIOU} scores for both semantic segmentation and saliency prediction. SegNet$_{ResNet}$ and PSPNet$_{MobileNet}$ also provide competitive results; however, their performances are slightly inconsistent over various object categories. Moreover, significantly better scores of SegNet$_{ResNet}$ (FCN8$_{VGG}$) over SegNet$_{CNN}$ (FCN8$_{CNN}$) validate the benefits of using a powerful feature extractor. As Table~\ref{tab:usability_suim} shows, SegNet$_{ResNet}$ (FCN8$_{VGG}$) has about twice (five times) the number of network parameters than SegNet$_{CNN}$ (FCN8$_{CNN}$). On the other hand, consistently better performance of UNet$_{RGB}$ over UNet$_{GRAY}$ validates the utility of learning on RGB image space (rather than using grayscale images as input).

\begin{figure*}[t]
    \centering
    \includegraphics[width=0.98\linewidth]{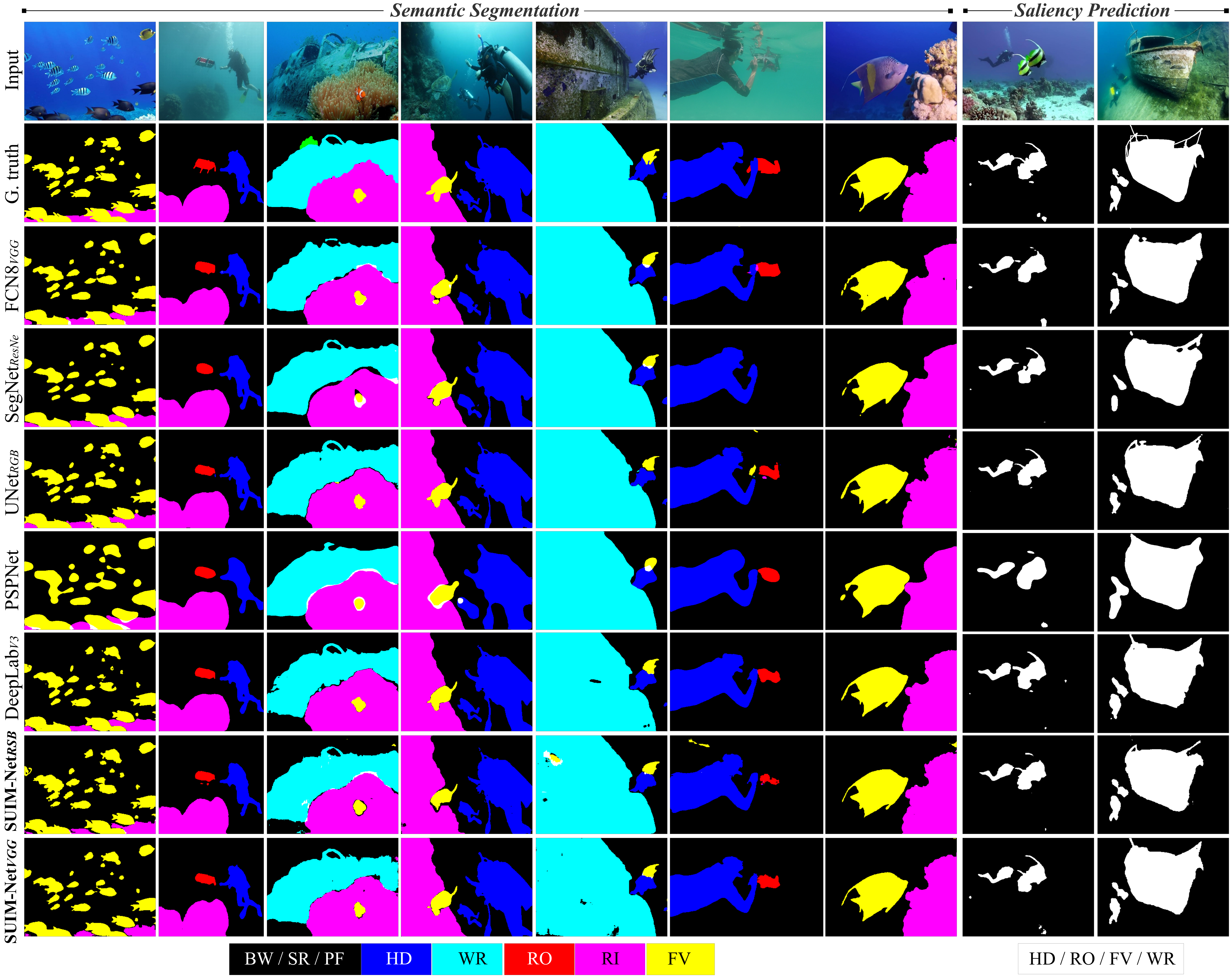} 
    \vspace{-3mm}
    \caption{A few qualitative comparison for the experiment of Table~\ref{tab:quant_suim}: (left) semantic segmentation with {{\tt HD}}, {{\tt WR}}, {{\tt RO}}, {{\tt RI}}, and {{\tt FV}} as object categories; (right) saliency prediction with {\tt HD}={\tt RO}={\tt FV}={\tt WR}={\tt 1} and {\tt RI}={\tt PF}={\tt SR}={\tt BW}={\tt 0}. Results for the top  performing models are shown; best viewed digitally at $300\%$ zoom.}
    \label{fig:qual_suim}
\end{figure*}

SUIM-Net$_{RSB}$ and SUIM-Net$_{VGG}$ provides consistent and competitive performance for both region similarity and object localization. As Table~\ref{tab:quant_suim} suggests, their $\mathcal{F}$ and \textit{mIOU} scores are within $5$\% margins of the respective top scores. The accuracy of semantic labeling and object localization can be further visualized in Figure~\ref{fig:qual_suim}, which shows that the SUIM-Net-generated masks are qualitatively comparable to the ground truth labels. Although UNet$_{RGB}$, FCN8$_{VGG}$, and DeepLab$_{V3}$ achieve much fine-grained object contours, the loss is not perceptually significant. 
Moreover, as shown in Table~\ref{tab:usability_suim}, SUIM-Net$_{RSB}$ operates at a rate of $28.65$ frames-per-second (FPS) on a Nvidia\texttrademark~GTX 1080 GPU, which is much faster than other SOTA models in comparison. Also, it is over $10$ times more memory efficient than UNet$_{RGB}$, FCN8$_{VGG}$, and DeepLab$_{V3}$. These computational aspects are ideal for its use in near real-time applications. 
However, its computational efficiency comes at a cost of lower performance margins and slightly poor generalization performance. 
To this end, with more learning capacity, SUIM-Net$_{VGG}$ achieves better object localization performance in general; it also offers an inference rate of $22.46$ FPS, which is still considerably faster than FCN8$_{VGG}$.

\begin{table}[ht]
\centering
\caption{Comparison for the input resolution, number of model parameters, and inference rates (averaged end-to-end processing times on a single image).}
%\footnotesize
\scriptsize
\vspace{2mm}
\begin{tabular}{l||c|r|r|r}
  \Xhline{2\arrayrulewidth}
  \textbf{Model} & Resolution & Parameters & Run-time (GTX 1080) & Run-time (AGX Xavier)  \\ \Xhline{2\arrayrulewidth}
  FCN8$_{CNN}$ & $320\times240$ & $69.744$ M & $17.11$ FPS & $3.42$ FPS \\
  FCN8$_{VGG}$& $320\times240$ & $134.286$ M & $8.79$ FPS  & $1.85$ FPS \\
  SegNet$_{CNN}$& $320\times256$ & $2.845$ M & $17.52$ FPS  & $4.38$ FPS \\
  SegNet$_{ResNet}$& $320\times256$ & $15.012$ M & $10.86$ FPS  & $2.39$ FPS \\
  UNet$_{GRAY}$& $320\times240$ &$31.032$ M & $20.13$ FPS & $4.62$ FPS \\
  UNet$_{RGB}$& $320\times240$ & $31.033$ M & $19.98$ FPS & $4.18$ FPS \\
  PSPNet$_{MobileNet}$& $384\times384$ & $63.967$ M & $6.65$ FPS  & $1.20$ FPS \\
  DeepLab$_{V3}$& $320\times320$ &$41.254$ M  & $16.00$ FPS  & $3.55$ FPS \\
  \textbf{SUIM-Net$_{RSB}$}& $320\times240$ & ${3.864}$ {M} & {${28.65}$} {FPS}  & $7.16$ FPS \\
  \textbf{SUIM-Net$_{VGG}$}& $320\times256$ & ${12.219}$ {M} & {${22.46}$} {FPS} & $5.84$ FPS \\
  \Xhline{2\arrayrulewidth}
\end{tabular}
\label{tab:usability_suim}
\end{table}%

%& Frame rate (Nvidia\texttrademark~GTX 1080 GPU)

\begin{figure*}[t]
    \centering
    \includegraphics[width=0.98\linewidth]{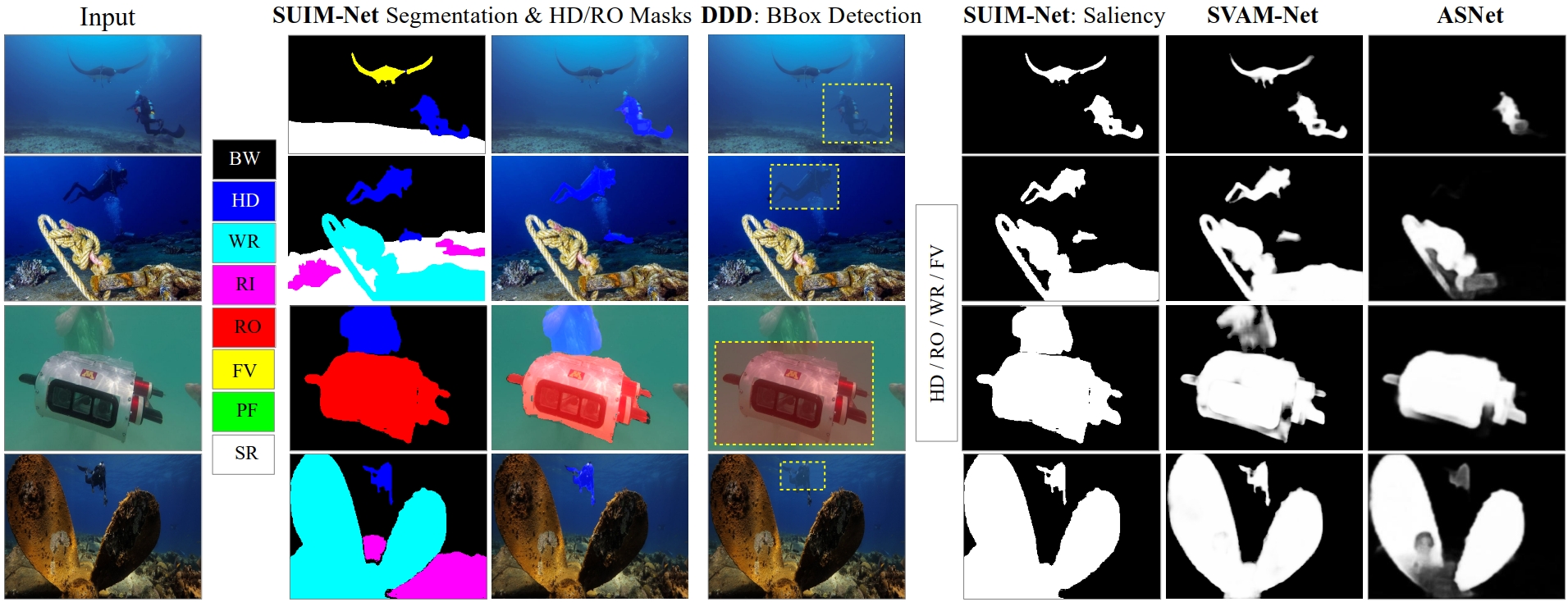}
    \vspace{-3mm}
    \caption{SUIM-Net$_{VGG}$-generated segmentation masks are shown alongside the bounding box outputs of DDD~\cite{islam2018towards} for the detection of human divers and robots; also, their corresponding saliency masks are compared with the class-agnostic predictions of SVAM-Net~\cite{islam2020svam} and ASNet~\cite{wang2018salient}.}
    \label{fig:use_suim}
\end{figure*}

Figure~\ref{fig:use_suim} further demonstrates the effectiveness of SUIM-Net$_{VGG}$-generated segmentation masks for fine-grained object localization in the image space, comparing its pixel-level detection of human divers and robots with the standard object detectors such as DDD~\cite{islam2018towards}. In addition to providing more precise object localization, SUIM-Net$_{VGG}$ incurs considerably fewer cases of missed detection, particularly in occluded or low-contrast regions in the image. Moreover, one advantage over class-agnostic models (\eg, SVAM-Net~\cite{islam2020svam} ASNet~\cite{wang2018salient}) is that SUIM-Net$_{VGG}$ offers additional semantic information about a scene, which can be useful in visual tracking, obstacle avoidance, and other application-specific attention modeling tasks.

\subsection{Challenges of Full-scale Underwater Scene Parsing}
\begin{figure*}[t]
    \centering
    \includegraphics[width=0.8\linewidth]{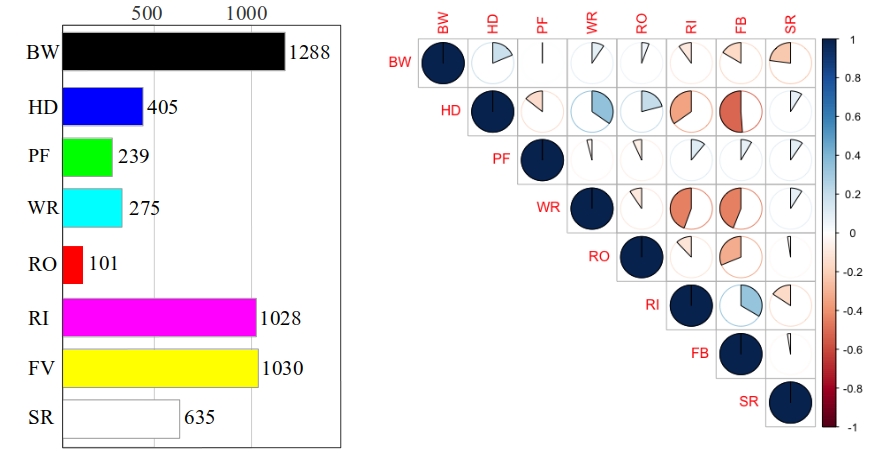}
    \vspace{-2mm}
    \caption{Statistics concerning the object categories of SUIM dataset: the number of images in each category (left), and their pairwise correlations of occurrences (right).}
    \label{fig:gran_suim}
\end{figure*}

SUIM-Net and other SOTA models demonstrate inspiring results for semantic segmentation and class-agnostic saliency estimation in underwater imagery. However, there are a few limitations and challenges involved in achieving full-scale underwater scene parsing. First, we would require a lot more training samples, possibly with an additional set of useful object classes. A few object categories in the SUIM dataset appear in very few images, resulting in a \textit{class imbalance} problem. As shown in Figure~\ref{fig:gran_suim}, the object categories such as {\tt RO} (AUVs/ROVs/instruments), {\tt PF} (aquatic plants and sea-grass), and {\tt WR} (wrecks/ruins) appear very rarely in the dataset. Despite our best efforts, it remains a challenging undertaking to ensure that all object categories appear in different waterbody and scenarios. Additional object categories for various species of reefs/fish and sub-categories for wrecks/ruin/trash are also going to be useful as a potential extension in achieving more granular underwater scene parsing.

Secondly, ensuring computational efficiency of deep visual models for underwater scene parsing remains challenging, particularly on single-board platforms. We showed that even for five-class image segmentation, SUIM-Net models offer only $5$-$7$ FPS inference rates on an Nvidia\texttrademark~AGX Xavier device (see Table~\ref{tab:usability_suim}). Hence, for an increasing number of object categories and more granular semantic prediction, it is considerably more challenging to develop a single-board deployable solution without compromising the performance. Further research efforts and more expansive dataset formulations are required to gain the benefits of full-scale underwater scene parsing in general-purpose robotics applications. Along that line, faster mobile supercomputers and embedded parallel computing
solutions~\cite{Jetson} will be useful as well. 

\section{How to Enable On-demand Image Enhancement?}
\begin{figure}[t]
\centering
    \subfigure[Probability density distribution of UICM scores.]{
        \includegraphics[width=0.98\textwidth]{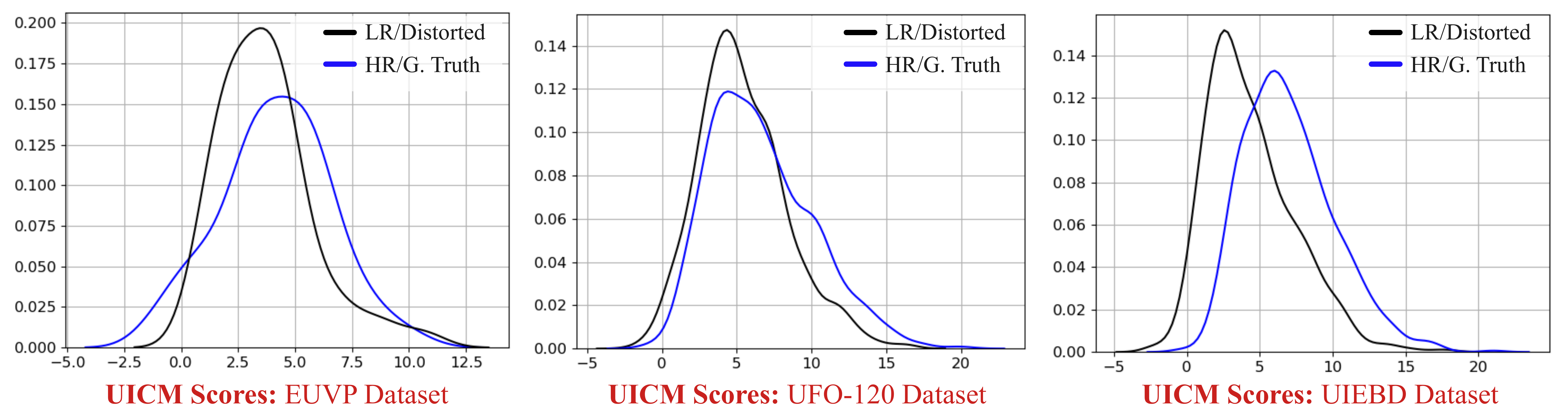}
        \label{fig:uicm_uiqm_dist_a}
    }%
    
    \subfigure[Probability density distribution of UIQM scores.]{
    \includegraphics[width=0.98\textwidth]{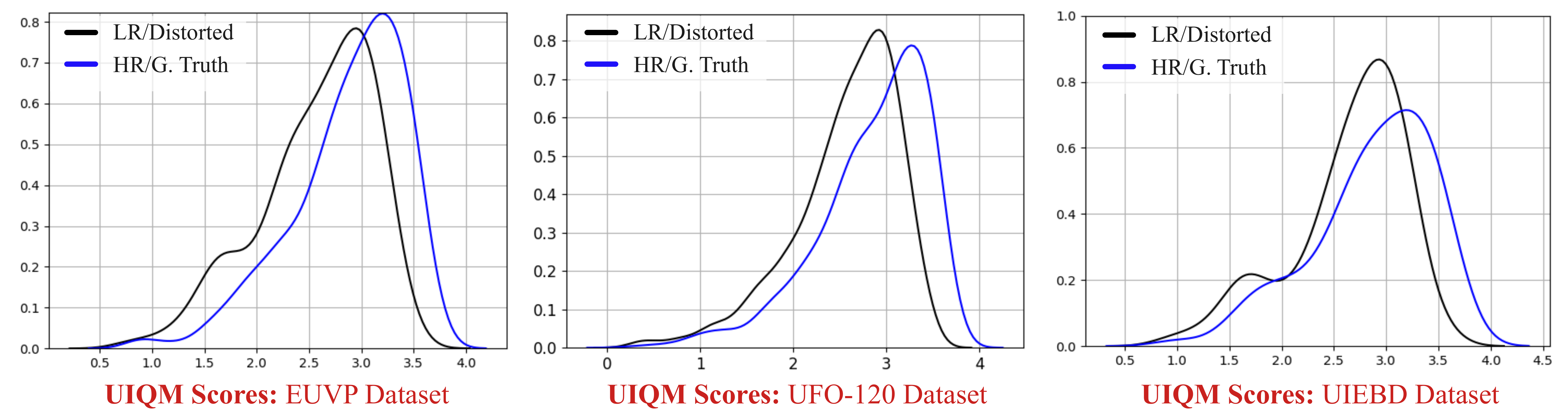}
    \label{fig:uicm_uiqm_dist_b}
    }%
    
\caption{The distributions of UICM and UIQM scores~\cite{panetta2016human,liu2019real} are illustrated for LR/distorted images versus HR ground truth images on benchmark datasets: EUVP~\cite{islam2019fast}, UFO-120~\cite{islam2020sesr}, and UIEBD~\cite{li2019underwater}. Significant overlaps between the distributions in every case reveal their ambiguity in underwater image quality assessment.}
\label{fig:uicm_uiqm_dist}
\end{figure}

One important dilemma in real-time underwater robot vision is to determine when to use image enhancement filters. Ideally, robots would sense visual conditions and trigger enhancement modules only when the image quality is poor. This saves unnecessary computation and also avoids the risk of introducing noise to undistorted images in good visual conditions. From a practical standpoint, the trivial choice is to use classical underwater image quality metrics such as UICM or UIQM~\cite{panetta2016human,liu2019real} (see Appendix~\ref{ApenD}). However, these metrics are not reliable predictors of image quality; as shown in Figure~\ref{fig:uicm_uiqm_dist},  their distributions of projected values for LR/distorted images and HR undistorted images of benchmark datasets have substantial overlapping regions. Addressing such inconsistencies of the existing image quality metrics is an active research problem in both terrestrial and underwater domains~\cite{prashnani2018pieapp,yang2015underwater}. 

\begin{figure*}[t]
    \centering
    \includegraphics[width=0.98\linewidth]{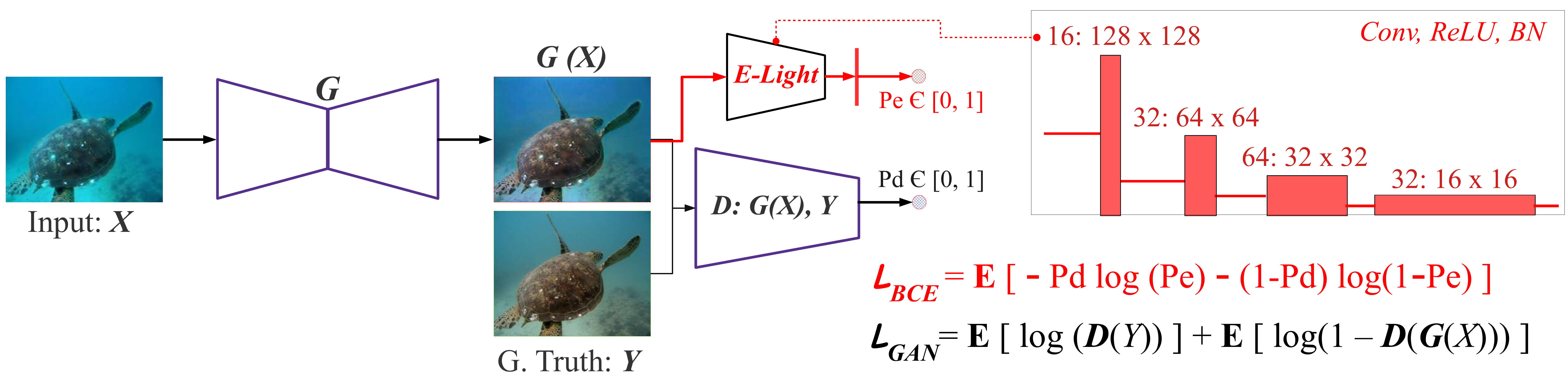}
    \vspace{-2mm}
    \caption{An outline of the E-light discriminator training inside an adversarial learning pipeline of underwater image enhancement. In addition to the adversarial loss $\mathcal{L}_{GAN}$, a binary cross-entropy function $\mathcal{L}_{BCE}$ is applied between E-light and the primary discriminator \textbf{\textit{D}} for learning image quality assessment.}
    \label{fig:delight_v0}
\end{figure*}

Another option is to leverage the discriminator models of GAN-based image enhancement pipelines; this is plausible because, in a loose sense, a discriminator learns to predict image quality which in turn forces the generator to learn to produce better-quality images. Evidently, a discriminator needs to be significantly lighter than its generator model for this to be a computationally feasible alternative. However, using a substantially shallow discriminator is not ideal because it generally leads to instability in the GAN training~\cite{mao2017least,arjovsky2017wasserstein}. To address this issue, we introduce the idea of an entangled light (\textbf{E-Light}) discriminator in the adversarial learning pipelines of GAN-based underwater image enhancement models. As outlined in Figure~\ref{fig:delight_v0}, E-light module is entangled with the primary discriminator (\textbf{\textit{D}}) via a binary-cross entropy loss ($\mathcal{L}_{BCE}$)~\cite{zhang2018generalized}, while being independent from the adversarial min-max game. That is, E-Light learns image quality assessment guided by \textbf{\textit{D}}, which independently evolves with a generator (\textbf{\textit{G}}) for learning image enhancement via $\mathcal{L}_{GAN}$ (and possibly other perceptual loss terms~\cite{islam2019fast}). The intuition here is to decouple the E-light module after training and take advantage of its shallow network architecture for fast on-device use.

\begin{table*}[t]
	\centering
	\footnotesize
	\caption{Run-time (on NVIDIA\texttrademark~Jetson AGX Xavier) and memory efficiency of the proposed E-light discriminator is demonstrated in comparison with several \textit{entangled} GAN-based underwater image enhancement models.}
	\vspace{2mm}
	\begin{tabular}{l||r|c||r|c}
		\Xhline{2\arrayrulewidth}
		 Models & \multicolumn{2}{c||}{{ Memory requirement (MB)}} & \multicolumn{2}{c}{{ Frame rate (FPS)}} \\ \cline{2-5}
		 &  Primary model & \textbf{E-Light} & Primary model & \textbf{E-Light} \\
		\Xhline{2\arrayrulewidth}
		FUnIE-GAN~\cite{islam2019fast}  & $11.90$ &   &  $48.20$ &  \\
		%\hline
		Pix2Pix~\cite{isola2017image}  & $28.74$ & $\mathbf{2.1}$ & $18.91$ & $\mathbf{134.85}$  \\
		%\hline
		UGAN-P~\cite{fabbri2018enhancing}  & $217.55$ &  & $5.715$ &   \\
		
		\Xhline{2\arrayrulewidth} 
	\end{tabular}
	\label{tab:elight_fps}
\end{table*}

\begin{figure*}[t]
    \centering
    \includegraphics[width=0.98\linewidth]{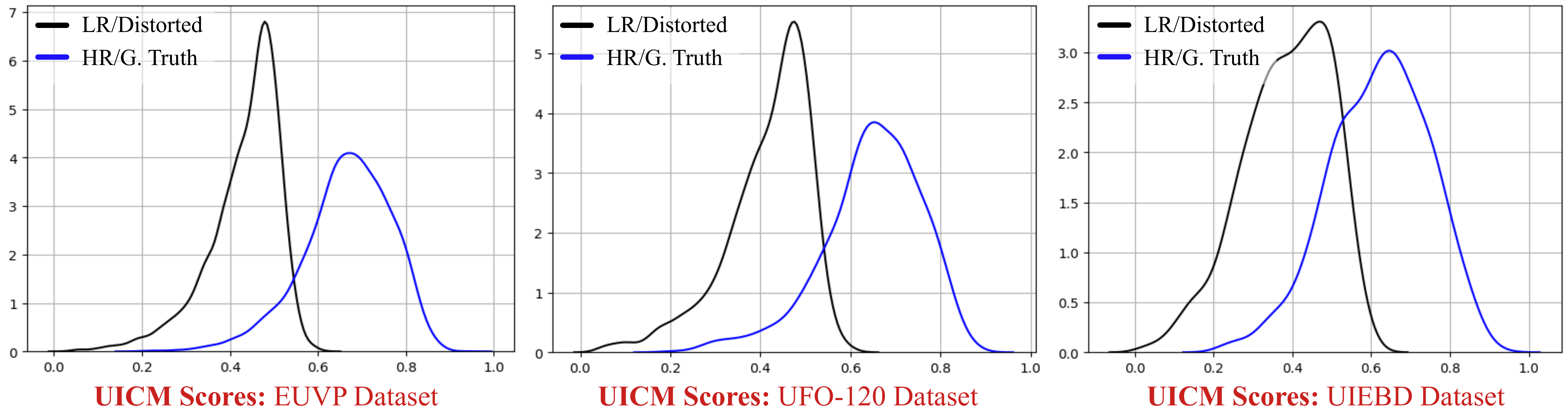}
    \vspace{-2mm}
    \caption{E-light discriminator's predictions demonstrate much better separation between the distributions of LR/distorted images and HR ground truth images than UICM/UIQM scores (refer to Figure~\ref{fig:uicm_uiqm_dist} for pairwise comparisons).}
    \label{fig:elight_all}
\end{figure*}

As demonstrated in Table~\ref{tab:elight_fps}, the E-light discriminator trained 
alongside FUnIE-GAN~\cite{islam2019fast}, Pix2Pix~\cite{isola2017image}, and UGAN-P~\cite{fabbri2018enhancing} offers significantly faster inference on a single-board NVIDIA\texttrademark~Jetson AGX Xavier device. It is at least $2.8$ times faster and over $5.6$ times memory efficient than the best available alternative, the FUnIE-GAN model. For its $134.85$ FPS frame rate and a memory requirement of only $2.1$ MB, AUVs can run E-light module on-board all the time and invoke the relatively heavier image enhancement filters (\eg, FUnIE-GAN) only when necessary. Our analyses showed in Figure~\ref{fig:elight_all} suggest that a threshold value of $0.40$-$0.55$ can be chosen for this purpose.

\section{Summary}
In this chapter, we discussed several platform-specific practices to gain the benefits of our proposed methodologies for real-time underwater machine vision. We also explored several intriguing research questions and empirically analyzed the underlying practicalities. First, we identified the challenges involved in using a standalone super-resolution module on single-board platforms. Then, we demonstrated how visual attention modeling can be applied to deal with the computational constraints of HR image enhancement and super-resolution problems. Subsequently, we investigated the utility of class-aware saliency predictors by exploring the research problem of semantic segmentation of underwater imagery. Finally, we introduced the idea of entangled light discriminator training for underwater image quality assessment, which facilitates the design of an on-demand image enhancement filtering pipeline. These features and model adaptation techniques are vital for the operational success of visually-guided AUVs in real-time applications.
\chapter{Conclusion and Future Work}\label{con}
The primary objective of this dissertation is to address the sensing and operational challenges of real-time machine vision for underwater human-robot cooperation in shallow-water and coastal-water applications. We developed robust and efficient perception capabilities for visually-guided AUVs to intelligently deal with adverse sensing conditions and other domain-specific practicalities. Our research on autonomous diver following~\cite{islam2017mixed,islam2018towards,islam2018person}, diver-robot interaction~\cite{islam2018understanding,islam2018dynamic,islam2019robot}, underwater image enhancement~\cite{islam2019fast,fabbri2018enhancing}, super-resolution~\cite{islam2020sesr,islam2019srdrm}, and visual attention modeling~\cite{islam2020svam,islam2020suim} provide novel and improved solutions for underwater visual perception. They have been deployed on underwater robots (\eg, Aqua AUV~\cite{dudek2007aqua}, LoCO AUV~\cite{LoCOAUV}) and validated through real application scenarios by extensive field experiments at the Caribbean sea~\cite{Bellairs}.
In this chapter, we summarize the contributions and discuss several follow-up research directions.
%, and highlight several open problems.

\section{Summary of Contributions}

\subsection{Enabling the Key Capabilities of a Companion AUV}
In Chapter~\ref{diver_following}, we presented two novel methods for autonomous diver following by underwater robots. The first method, Mixed-Domain Periodic Motion (MDPM) tracker~\cite{islam2017mixed}, introduces an intuitive representation of human swimming motion in the spatio-temporal domain. It combines spatial-domain features and frequency-domain motion signatures into a Hidden Markov Model (HMM)-based efficient tracking pipeline to relax the directional constraint of existing frequency-domain trackers~\cite{Sattar09RSS}. In the second method, we designed an architecturally simple Convolutional Neural Network (CNN)-based diver detection model~\cite{islam2018towards}, which is much faster than the state-of-the-art (SOTA) deep models yet provides comparable detection performances. The proposed model is robust to noise and invariant to the color, scale, and appearance of divers. In addition to the algorithmic design and implementation details, we experimentally analyzed its robustness-efficiency trade-off by numerous closed-water and open-water experiments.

Then, in Chapter~\ref{r2r_pose}, we introduced a robot-to-robot relative pose estimation method that only uses mutually visible humans' body-poses as correspondences~\cite{islam2019robot}. In the proposed method, we
designed an efficient person re-identification technique and an iterative key-point refinement algorithm to establish geometric validity of the associated humans across viewpoints. We demonstrated the utility of such pose estimation method as an \textit{implicit} way to maintain spatial coordination by cooperative robots in feature-deprived underwater scenarios. Subsequently in Chapter~\ref{hrc}, we presented RoboChatGest~\cite{islam2018dynamic}, a hand gesture-based \textit{explicit} diver-to-robot communication system that incorporates $i$) a syntactically simple visual language, $ii$) various hand gesture recognizer models, and $iii$) a robust instruction decoder. The motive for its simplistic design was to reduce cognitive load on divers during interaction by eliminating the need for carrying and manipulating tags, as done in the predecessor RoboChat language~\cite{dudek2007visual}. Over a series of oceanic field trails and a user interaction study~\cite{islam2018understanding}, we demonstrated its operational feasibility and usability benefits over existing systems.

\subsection{Conquering Adverse Underwater Visual Conditions}
In Chapter~\ref{en_sr}, we presented the design and implementation details of the FUnIE-GAN~\cite{islam2019fast}, a conditional Generative Adversarial Network (GAN)-based model for fast underwater image enhancement. We supervised its \textit{paired} and \textit{unpaired} adversarial training by evaluating the perceptual image quality based on color, texture, content, and style information. We also released the EUVP dataset used for FUnIE-GAN training; it has paired and unpaired collections of underwater images (of poor and good quality) for large-scale deep visual learning. By a series of qualitative and quantitative experiments, we demonstrated that FUnIE-GAN-enhanced images significantly boost the performance of standard visual perception tasks such as object detection, human pose estimation, and saliency prediction. Moreover, we validated its real-time utility in the autonomy pipeline of visually-guided AUVs as an image enhancement filter for improved visual perception. FUnIE-GAN is the current SOTA for real-time underwater image enhancement; it overcomes the computational aspects of our previously proposed UGAN model~\cite{fabbri2018enhancing} and offers $25.4$ FPS inference rates on an NVIDIA\texttrademark{} Jetson TX2.  

Subsequently, we introduced the new research problem of Simultaneous Enhancement and Super-Resolution (SESR) for underwater robot vision in Chapter~\ref{sesr}. Our proposed solution, {Deep SESR}~\cite{islam2020sesr}, is a residual-in-residual model that performs perceptual image enhancement at $2\times$, $3\times$, or $4\times$ higher spatial resolutions. We supervised its training by addressing the chrominance-specific underwater image distortions and designing novel loss functions to learn contrast enhancement in salient image regions. We also released {UFO-120}, the first dataset to facilitate large-scale SESR learning. Deep SESR individually outperforms existing underwater image enhancement and super-resolution models and exhibits significantly better generalization performance. Additionally, it offers several application-specific design choices and offers fast inference on single-board platforms, \eg, $7.75$-$10.02$ FPS rates on an Nvidia\texttrademark AGX Xavier.

\subsection{Meeting On-board Computational Constraints}
In Chapter~\ref{svam}, we developed a general-purpose solution of Saliency-guided Visual Attention Modeling (SVAM)
for underwater robots by combining the power of bottom-up and top-down deep visual learning. The proposed model, {SVAM-Net}~\cite{islam2020svam}, incorporates dedicated Spatial Attention Modules (SAMs) to jointly learn coarse-level and fine-level semantic features for accurate salient object detection in underwater imagery. The end-to-end SVAM-Net model provides SOTA scores on benchmark datasets, exhibits better generalization performance on challenging test cases than existing approaches, and offers over $20$ FPS inference rate on an NVIDIA\texttrademark~GTX 1080 GPU. Moreover, its \textit{light}  variant, SVAM-Net\textsuperscript{Light} runs as fast as $86.15$ FPS on a GPU, and offers $21.77$ FPS inference on an NVIDIA\texttrademark{}~AGX Xavier device. More importantly, SVAM-Net\textsuperscript{Light} facilitates effective visual attention modeling in various perception tasks such as image enhancement, super-resolution, scene parsing, and uninformed visual search. In particular, we found $45\%$-$67\%$ faster processing on single-board platforms for image enhancement and super-resolution depending on the location and scales of salient objects in a given scene.

We discussed several other platform-specific practices and expected performance margins of our proposed solutions in Chapter~\ref{do_better}. We empirically analyzed various problem formulations and model adaptation techniques to meet the real-time operating constraints for AUVs. 
In particular, we discussed the operational feasibility of a standalone super-resolution module for underwater imagery~\cite{islam2019srdrm}. 
Then, we explored the research problem of class-aware saliency prediction and semantic segmentation of underwater imagery~\cite{islam2020suim}. Lastly, we investigated the design of an \textit{on-demand} image enhancement filter by altering the adversarial training pipeline of existing GAN-based models. For all these intriguing studies, we presented our research findings, identified the underlying challenges and limitations, and highlighted various open areas for future research.         

%which are important for the operational success of visually-guided AUVs in human-robot cooperative applications.

\section{Future Research Directions}
\noindent
\hspace{2mm}~$\bullet$ \emph{\textbf{A universal API for degraded machine vision.}} 
Despite recent advancements of interactive vision APIs and AutoML technologies, there are no universal platforms or criteria to measure the \textit{goodness} of visual sensing conditions to extrapolate the performance bounds of a perception algorithm. Hence, the design of adaptable solutions to combat degraded vision (\textit{e.g.}, via online learning or reinforcement learning) has not been explored in the literature. One immediate future direction is to address these limitations by devising a universal API for visually-guided underwater robots. Moreover, extending these ideas to thermal imaging and sonar imaging modalities can be useful for generating augmented visuals in autonomous exploration, manned/unmanned rescue operations, bathymetric surveys, and other remote sensing applications. 

\hspace{2mm}~$\bullet$ \emph{\textbf{Following a team of divers by a team of robots.}} 	
Many underwater missions consist of multiple divers who work as a team during the deployment. Following the team as a whole is operationally more efficient in general. The perception problem can be solved by a simple extension (\ie, multi-object detection); however, motion planning and control modules are not straightforward. Additionally, the rules for spatial conduct and interaction need to be identified and quantified. Tracking a team of independently moving objects is a challenging problem in general~\cite{islam2018person}; it gets even more challenging in a 3D environment while dealing with real-time constraints. 
%Despite the challenges, it is potentially invaluable in numerous applications of underwater human-robot teams~\cite{islam2018person}.
On the other hand, multi-robot convoying led by a human is useful in cooperative estimation problems~\cite{rekleitis2001multi,shkurti2017underwater}.  
A simple approach to this problem is to assign leader-follower pairs; that is, one of the robots is assigned to follow the person, and every other robot is individually assigned another robot as its leader. Each robot follows its leader and together they move as a team. Another approach is to let the robots communicate with each other and cooperatively plan their motions. 
%The underlying planning pipeline is similar to that of a multi-robot convoying problem, which is particularly challenging in underwater and aerial scenarios~\cite{shkurti2017underwater,minaeian2016vision}. 
Moreover, this can be further generalized into the problem of {following a team of divers by a team of robots} in a cooperative setting. Our proposed robot-to-robot relative pose estimation method can be useful in these scenarios for ensuring spatial coordination of the follower robots. However, it requires complex cooperative planning pipelines to achieve the optimal positioning and motion trajectories for each robot, which is a challenging open problem.

\noindent
\hspace{2mm}~$\bullet$ \emph{\textbf{A vision-based long-term monitoring system.}} Long-term monitoring of underwater habitats provides critical data that help us understand their health and interactions within the marine ecosystems~\cite{alonso2019coralseg,beijbom2012automated,shkurti2012multi}. Our proposed solutions for object detection~\cite{islam2020suim}, visual attention modeling~\cite{islam2020svam}, and image enhancement~\cite{islam2019fast,islam2020sesr} can be used and extended to develop an on-device vision-based monitoring system. Such platforms will be potentially useful in applications such as long-term monitoring of coral reefs, tracking invasive species of fish, measuring/detecting changes in seasonal water-quality or waterborne pathogens, etc. Along that line, a research study of existing systems and solutions for long-term biological and climatic observations~\cite{NCRMP} can lead to subsequent developments of the vision-based system for certain coastal-water applications.

\noindent
\hspace{2mm}~$\bullet$ \emph{\textbf{Harnessing the power of robot learning from demonstration.}} 
Not all desired behavior of an intelligent agent can be modeled as tractable optimization problems or scripted by traditional robot programming paradigms. In recent years, the end-to-end learning of autonomous robot behaviors from demonstration has been an active research topic. Researchers have reported exciting results in the domains of 2D robot navigation in cluttered environments~\cite{pfeiffer2017perception}, simple autonomous driving~\cite{codevilla2018end}, imitating driving styles~\cite{kuderer2015learning}, etc. These results indicate that the end-to-end learning models, particularly the idea of {Learning from Demonstration} (LfD) can be very effective for companion robots of the next generation. 
There are a few research efforts already in this regard in simulation environments~\cite{pierre2018end,dewantara2016generation}; however, more extensive research and real-world experiments are necessary. For underwater human-robot cooperative applications, in particular, LfD can provide a natural and expressive way to program artificially intelligent behavior under complex environmental constraints.

\noindent
\hspace{2mm}~$\bullet$ \emph{\textbf{Addressing privacy and safety concerns for human-robot cooperation.}} 	
There has been an increasing amount of concerns about the privacy and safety issues of companion robots, particularly in the terrestrial domain. For instance, a recent study by Pew Research Center has found that about $54\%$ of the US population thinks that drones should not be allowed to fly in residential areas as it undermines the ability to assess context and measure trust. Such growing concerns are pervasive across cyberspace toward numerous intelligent products and services. Our research~\cite{islam2018person} identified various forms of \textit{implicit} and \textit{explicit} human-robot interaction for companion robots in manufacturing, health care, the entertainment industry, and social interactions. We discussed some of these aspects in our diver-robot interaction modules as well~\cite{islam2018understanding,islam2019robot,islam2018towards}. However, there are significant rooms for improvement in the existing human-robot interaction systems across all domains, \ie, for AGVs, AUVs, or UAVs.    
%I intend to continue pursuing research on these aspects with a broader goal of ensuring safe and effective human-robot cooperation in practical applications. 

\section{Concluding Remarks}
We are in the midst of a technology revolution of autonomous systems, robot vision, and deep visual computing. New hardware and software platforms with improved solutions are emerging every day which taking us closer toward complete autonomy. As several technical challenges remain to be solved, we are still far away from being able to ensure near-perfect robot perception capabilities in adverse underwater visual conditions. In this dissertation, we developed practical solutions to address some of these challenges to advance the current state-of-the-art. We hope that our proposed methodologies and systems contribute toward closing the gap between theory and practice for fast and effective underwater human-robot cooperation.

\bibliographystyle{hunsrt} 
\bibliography{Refs}

\appendix
\chapter{Appendices for Chapter 2}
\section{Recursive Formulation of $\mu ^*(T)$}\label{ApenA1}
In Section~\ref{mdpm}, we defined $\mu ^*(T)$, \ie, the most likely sequence of states  ($G_{0}, \dots G_{T-1}$) that leads to a given state $G_{T}$, as:
\begin{equation}
\footnotesize
\begin{aligned}
\mu ^*(T) &= \argmax_{w^{(0)}_{i}, \dots, w^{(T-1)}_{i}}{ P \Big\{ G_{0}=w^{(0)}_{i}, \dots, G_{T}=w^{(T)}_{i} \Big| e_{0}, \dots, e_{T}  \Big\}} \\ &= \argmax_{ w^{(0:T-1)}_{i} }{ P \Big\{ G_{0:T}=w^{(0:T)}_{i} \Big| e_{0:T}  \Big\}}.
\end{aligned}
\end{equation} 
By using the properties of the Bayesian chain rule and Markovian transition, the recursive formulation of $\mu ^*(T)$ is derived as follows:
\begin{equation*}
\scriptsize
\begin{aligned}
\mu ^*(T) &= \argmax_{ w^{(0:T-1)}_{i} }{ P \Big\{ G_{0:T}=w^{(0:T)}_{i} \Big| e_{0:T}  \Big\}} \\
 &= \argmax_{w^{(0:T-1)}_{i}}{ \Big( P \Big\{ e_T \Big| G_{0:T}=w^{(0:T)}_{i} \Big\} } \times   {P \Big\{ G_{0:T}=w^{(0:T)}_{i} \Big| e_{0:T-1}  \Big\} \Big) } \\
 &=  \argmax_{w^{(0:T-1)}_{i}}{ \Big( P \Big\{ e_T \Big| G_{T}=w^{(T)}_{i} \Big\} \times } { P\Big\{ G_{T}=w^{(T)}_{i} \Big| G_{T-1}=w^{(T-1)}_{i}  \Big\}} \\
 & \qquad \qquad \qquad  \qquad \qquad  \qquad \qquad  { \times P \Big\{ G_{0:T-1}=w^{(0:T-1)}_{i} \Big| e_{0:T-1}  \Big\} \Big) }    \\
     &=  P \Big\{ e_T \Big| G_{T}=w^{(T)}_{i} \Big\} \times  { \argmax_{w^{(T-1)}_{i}}{ \Big( P\Big\{ G_{T}=w^{(T)}_{i} \Big| G_{T-1}=w^{(T-1)}_{i}  \Big\} } } \\
     & \qquad \qquad \qquad  \qquad \qquad   {  \times \argmax_{w^{(0:T-2)}_{i}} {P \Big\{ G_{0:T-1}=w^{(0:T-1)}_{i} \Big| e_{0:T-1}  \Big\} } \Big)} \\
   &= P \Big\{ e_T \Big| G_{T}=w^{(T)}_{i} \Big\} \times { \argmax_{w^{(T-1)}_{i}}{ \Big( P\Big\{ G_{T}=w^{(T)}_{i} \Big| G_{T-1}=w^{(T-1)}_{i}  \Big\} } } \times   {\mu ^*(T-1)  \Big).  }
\end{aligned}
\label{Recur_old}
\end{equation*}

\section{Standard Deep Visual Object Detection Models}\label{ApenB}
\begin{figure}[ht]
\centering
    \subfigure[\textbf{Faster R-CNN with Inception V2}: Faster R-CNN~\cite{renNIPS15fasterrcnn} is an improvement of R-CNN~\cite{Girshick2014RCNN_CVPR} that introduces a Region Proposal Network (RPN) to make the whole object detection network end-to-end trainable. The RPN uses the last convolutional feature-maps to produce region proposals which are then fed to the fully connected layers for the final detection. The original implementation of Faster R-CNN uses VGG-16~\cite{simonyan2014very} model for feature extraction. We use Inception V2~\cite{szegedy2016rethinking} model for feature extraction instead, as it is known to provide better object detection performances on standard datasets~\cite{tfzoo}.]{\includegraphics [width=0.95\linewidth]{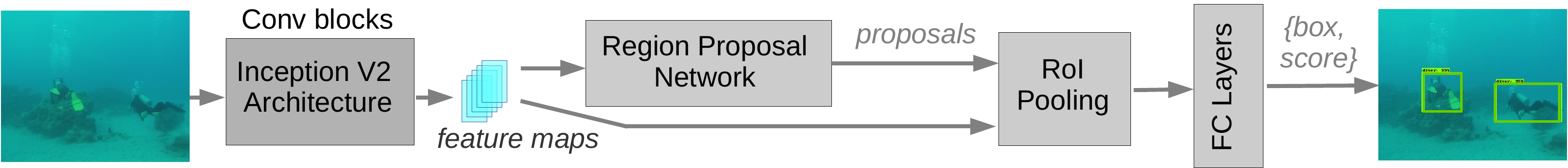}
    }
    
    \vspace{3mm}
    
    \subfigure[\textbf{YOLO V2 and Tiny TOLO}: YOLO models~\cite{redmon2016you,redmon2016yolo9000} formulate object detection as a regression problem in order to avoid using computationally expensive RPNs. They divide the image space into rectangular grids and predict a fixed number of bounding boxes, their corresponding confidence scores, and class probabilities. Although there are restrictions on the maximum number of object categories, they perform faster than the standard RPN-based object detectors. Tiny YOLO~\cite{tinyYOLO} is a scaled down version of the original model having sparser layers that runs much faster compared to the original model; however, it sacrifices detection accuracy in the process.]{\includegraphics [width=0.95\linewidth]{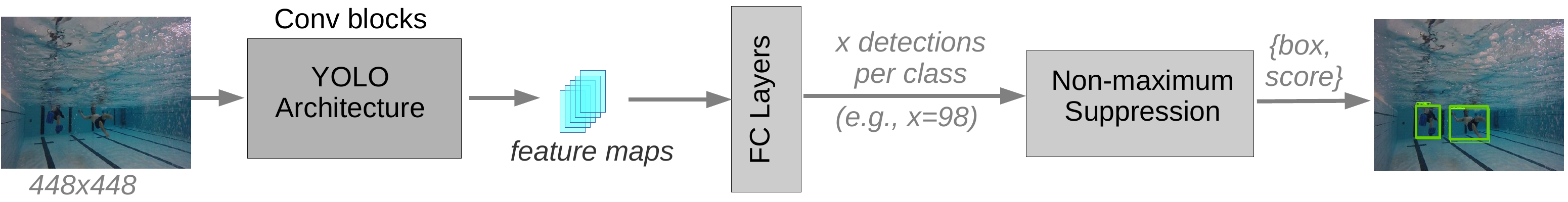}
    }

    \vspace{3mm}
    
    \subfigure[\textbf{SSD with MobileNet V2}: SSD (Single-Shot Detector)~\cite{liu2016ssd} also performs object localization and classification in a single pass of the network using the regression trick as in the YOLO~\cite{redmon2016you} model. The architectural difference of SSD with YOLO is that it introduces additional convolutional layers to the end of a base network, which results in improved performances. In our implementation, we use MobileNet V2 \cite{sandler2018inverted} as the base network to ensure faster running time.]{\includegraphics [width=0.95\linewidth]{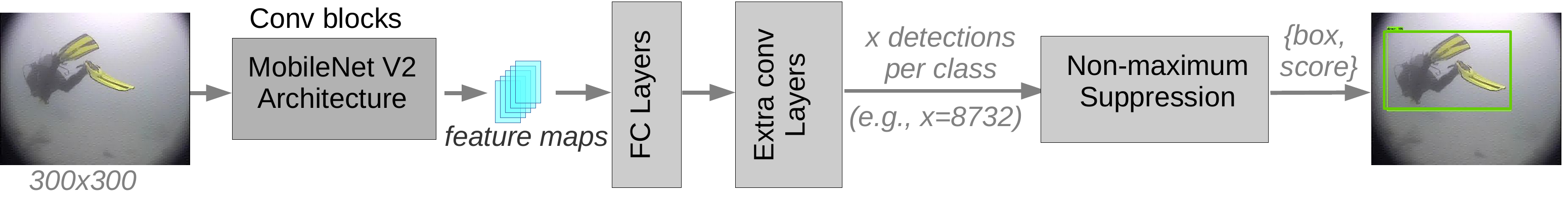}
    }
    
\caption{Methodological overview of a few SOTA object detection models.}
\label{dr_detect_sota}
\end{figure}

\chapter{Standard Performance Metrics}
\section{Object Detection Performance}\label{ApenObj}

\begin{enumerate}[i.]
\item \textbf{mAP (mean Average Precision)}: The mAP measures object detection accuracy as the average of the maximum precision at different recall values. The precision and recall are defined as $precision = \frac{TP}{TP+FP}$ and $recall = \frac{TP}{TP+FN}$; here, the standard terms TP, FP, and FN are short forms for true positive, false positive, and false negative, respectively. 

\item \textbf{IoU (Intersection over Union)}: The IoU measures object localization performance, \ie, how well a model predicts the locations of the objects. It is calculated using the area of overlapping regions of the predicted and ground truth labels, defined as $IoU = \frac{Area\text{ }of\text{ }overlap}{Area\text{ }of\text{ }union}$.

\item \textbf{Dice coefficient ($\mathcal{F}$ score)}: The $\mathcal{F}$ score is a {region similarity} metric that quantifies the correctness of predicted pixel labels compared to ground truth. It is calculated by the normalized precision ($\mathcal{P}$) and recall ($\mathcal{R}$) scores as $\mathcal{F}=\frac{2 \times \mathcal{P} \times \mathcal{R}}{\mathcal{P}+\mathcal{R}}$.
\end{enumerate}

\section{Image-quality Analysis}\label{ApenD}
\begin{enumerate}[i.]
    %\vspace{5mm}
    \item \textbf{Peak Signal-to-Noise Ratio (PSNR)}: The PSNR approximates the reconstruction quality of a generated image $\mathbf{x}^{m \times n}$ compared to its ground truth $\mathbf{y}^{m \times n}$ based on their Mean Squared Error (MSE) as follows:
\begin{equation}
\begin{aligned}
\footnotesize
    MSE(\mathbf{x},\mathbf{y})&=\frac{1}{mn} \sum_{i=1}^{m} \sum_{j=1}^{n} |\mathbf{x}_{i,j}-\mathbf{y}_{i,j}|^2 \\
    PSNR(\mathbf{x}, \mathbf{y}) &= 20 \log_{10} \big[255/MSE(\mathbf{x},\mathbf{y})\big]. 
\end{aligned}
\label{eq:psnr}
\end{equation}

    \vspace{5mm}
    \item \textbf{Structural Similarity (SSIM)}: The SSIM~\cite{wang2004image} compares the image patches based on three properties: luminance, contrast, and structure. It is defined as:  
\begin{equation}
\footnotesize
        SSIM(\mathbf{x},\mathbf{y}) =  \Big(\frac{2 \mathbf{\mu}_\mathbf{x} \mathbf{\mu}_\mathbf{y} + c_1}{\mathbf{\mu}_\mathbf{x}^2+\mathbf{\mu}_\mathbf{y}^2 + c_1}\Big) \Big(\frac{2 \mathbf{\sigma}_{\mathbf{xy}} + c_2}{\mathbf{\sigma}_\mathbf{x}^2+\mathbf{\sigma}_\mathbf{y}^2 + c_2}\Big). 
\label{eq:ssm}
\end{equation}
Here, $\mathbf{\mu}_\mathbf{x}$ ($\mathbf{\mu}_\mathbf{y}$) denotes the mean, and $\mathbf{\sigma}_\mathbf{x}^2$ ($\mathbf{\sigma}_\mathbf{y}^2$) denotes the variance of $\mathbf{x}$ ($\mathbf{y}$); whereas $\mathbf{\sigma}_{\mathbf{xy}}$ denotes the cross-correlation between $\mathbf{x}$ and $\mathbf{y}$. Additionally, $c_1 = (255\times0.01)^2$ and $c_2 = (255\times0.03)^2$ are constants that ensure numeric stability.

    \vspace{5mm}
    \item \textbf{Underwater Image Quality Measure (UIQM)}: The UIQM~\cite{panetta2016human,liu2019real} is a linear combination of three metrics: image colorfulness, sharpness, and contrast. The UIQM is expressed as follows:  
\begin{equation}
\begin{aligned}
\footnotesize
    UIQM(\mathbf{x}) = c_c &\times UICM(\mathbf{x}) +  c_s \times  UISM(\mathbf{x}) + c_m \times UIConM(\mathbf{x})
\end{aligned}
\label{eq:uiqm}
\end{equation}
Here, the constant values are $c_c=0.0282$, $c_s=0.2953$, and $c_m=3.5753$; We follow the standard definition of Eq.~\ref{eq:uiqm} and relevant procedures for computing UICM, UISM, and UIConM that are described in~\cite{islam2019fast}.
\end{enumerate}

\subsection{SOD Performance Evaluation}\label{ApenSOD}
\begin{enumerate}[i.]
    \item \textbf{F-measure} ($\mathbf{F}_{\beta}$) is an overall performance measurement that is computed by the weighted harmonic mean of the precision and recall as: 
    \begin{equation}
    \footnotesize
        \mathbf{F}_{\beta} = \frac{(1 + \beta^2) \times Precision \times Recall}{\beta^2 \times Precision + Recall}.
    \end{equation}
    Here, $\beta^2$ is set to $0.3$ as per the SOD literature to weight precision more than recall~\cite{borji2019salient,feng2019attentive}. Also, the maximum scores ($\mathbf{F}_{\beta}^{max}$) are reported for the benchmark quantitative comparison.
    \vspace{3mm}

    \item \textbf{Mean absolute error (MAE)} is a generic metric that measures the average pixel-wise differences between the predicted ($\hat{s}^{m \times n}$) and ground truth ($s^{m \times n}$) saliency maps as follows: 
    \begin{equation}
    \footnotesize
    MAE(\mathbf{s},\mathbf{\hat{s}})=\frac{1}{mn} \sum_{i=1}^{m} \sum_{j=1}^{n} |\mathbf{s}_{i,j}-\mathbf{\hat{s}}_{i,j}|. 
    \label{eq:mae}
    \end{equation}
    \vspace{3mm}
    
    \item \textbf{S-measure}  ($\mathbf{S}_{m}$) is a recently proposed metric~\cite{fan2017structure} that simultaneously evaluates region-aware ($\mathbf{S}_{o}$) and object-aware ($\mathbf{S}_{r}$) structural similarities between the predicted and ground truth saliency maps as follows: 
    \begin{equation}
    \footnotesize
        \mathbf{S}_{m} = \alpha \times \mathbf{S}_{o} + (1-\alpha) \times \mathbf{S}_{r}.
    \end{equation}
    Here, we set $\alpha=0.5$ as suggested in~\cite{fan2017structure}; detailed implementations can be found here: \url{https://github.com/xahidbuffon/SOD-Evaluation-Tool-Python}. 
    \vspace{3mm}
    
    \item \textbf{Precision-recall (PR) curve} is a standard performance metric and is complementary to MAE. It is evaluated by \textit{binarizing} the predicted saliency maps with a threshold sliding from $0$ to $255$ and then performing bin-wise comparison with the ground truth values. Then, the relationship between \textit{precision} and \textit{recall} is plotted for every possible cut-off values. 
\end{enumerate}

\end{document}